\documentclass[pdflatex,sn-basic]{sn-jnl}

\usepackage{times,amsmath,epsfig}
\usepackage{paralist}
\usepackage{amsmath}
\usepackage{multirow}
\usepackage{colortbl}
\usepackage{url}
\usepackage{subfigure}
\usepackage{colortbl}
\usepackage{paralist}
\usepackage{hyperref}
\usepackage{ragged2e,tabularx, makecell}
\usepackage{times,amsmath,epsfig}
\usepackage{paralist}
\usepackage{float}
\usepackage{amsmath}
\usepackage{multirow}
\usepackage{colortbl}
\usepackage{url}
\usepackage{subfigure}
\usepackage{colortbl}
\usepackage{paralist}
\usepackage{hyperref}
\usepackage{graphicx}
\usepackage{adjustbox}
\usepackage{amsmath}
\usepackage{cleveref}
\usepackage{longtable}
\usepackage{tabu}

\mathchardef\mhyphen="2D

\newcolumntype{H}{>{\columncolor{black}\color{white}}c}

\graphicspath {{./} {Figures/}}
\AtBeginDocument{%
  \providecommand\BibTeX{{%
    \normalfont B\kern-0.5em{\scshape i\kern-0.25em b}\kern-0.8em\TeX}}}

\theoremstyle{thmstyleone}%
\theoremstyle{thmstyletwo}%

\theoremstyle{thmstylethree}%

\begin{document}

\title[AutoMLBench]{AutoMLBench: A Comprehensive Experimental Evaluation of  Automated Machine Learning Frameworks}

\author[1]{\fnm{Hassan} \sur{Eldeeb}}\email{hassan.eldeeb@ut.ee}

\author[1]{\fnm{Mohamed} \sur{Maher}}\email{mohamed.maher@ut.ee}
\equalcont{These authors contributed equally to this work.}

\author[1]{\fnm{Oleh} \sur{Matsuk}}\email{oleh.matsuk@ut.ee} \equalcont{These authors contributed equally to this work.} 

\author[1]{\fnm{Abdelrhman} \sur{Eldallal}}\email{abdelrhman.eldallal@ut.ee}

\author*[1]{\fnm{Radwa} \sur{Elshawi}}\email{radwa.elshawi@ut.ee}

\author*[1]{\fnm{Sherif} \sur{Sakr}}\email{sherif.sakr@ut.ee}

\affil[1]{\orgdiv{Institute of Computer Science}, \orgname{University of Tartu}, \orgaddress{\city{Tartu}, \country{Estonia}}}



\abstract{
Nowadays, machine learning is playing a crucial role in harnessing the power of the massive amounts of data that we are currently producing every day in our digital world. With the booming demand for machine learning applications, it has been recognized that the number of knowledgeable data scientists can not scale with the growing data volumes and application needs in our digital world. In response to this demand, several automated machine learning (AutoML) techniques and frameworks have been developed to fill the gap of human expertise by automating the process of building machine learning pipelines. In this study, we present a comprehensive evaluation and comparison of the performance characteristics of six popular AutoML frameworks, namely, \texttt{AutoWeka}, \texttt{AutoSKlearn}, \texttt{TPOT}, \texttt{Recipe}, \texttt{ATM} and \texttt{SmartML} across 100 data sets from established AutoML benchmark suites. Our experimental evaluation considers different aspects for its comparison including the performance impact of several design decisions including \emph{time budget}, \emph{size of search space}, \emph{meta-learning} and \emph{ensemble construction}. The results of our study reveal various interesting insights that can significantly guide and impact the design of AutoML frameworks.
}




\maketitle

\section{Introduction}
We are witnessing tremendous interest in artificial intelligence applications across governments, industries and research communities with a yearly cost of around 12.5 billion US dollars~\citep{Worldwidesemiannual2017}. The driver for this interest is the advent and increasing popularity of machine learning (ML) and deep learning (DL) techniques. The rise of generated data from different sources, processing capabilities, and ML algorithms opened the way for adopting ML in a wide range of real-world applications~\citep{zomaya2017handbook}. This situation is increasingly contributing towards a potential \emph{data science crisis}, similar to the software crisis~\citep{fitzgerald2012software}, due to the crucial need of having an increasing number of data scientists with solid knowledge and good experience so that they can keep up with harnessing the power of the massive amounts of data produced daily. Thus, we are witnessing a growing interest in automating the process of building ML pipelines where the presence of a human in the loop can be dramatically reduced. Research in Automated machine learning (AutoML) aims to alleviate both the computational cost and human expertise required for developing ML pipelines through automation with efficient algorithms. In particular, AutoML techniques enable the widespread use of ML techniques by domain experts and non-technical users.

\begin{figure}
\centering
	\includegraphics[width=0.7\linewidth]{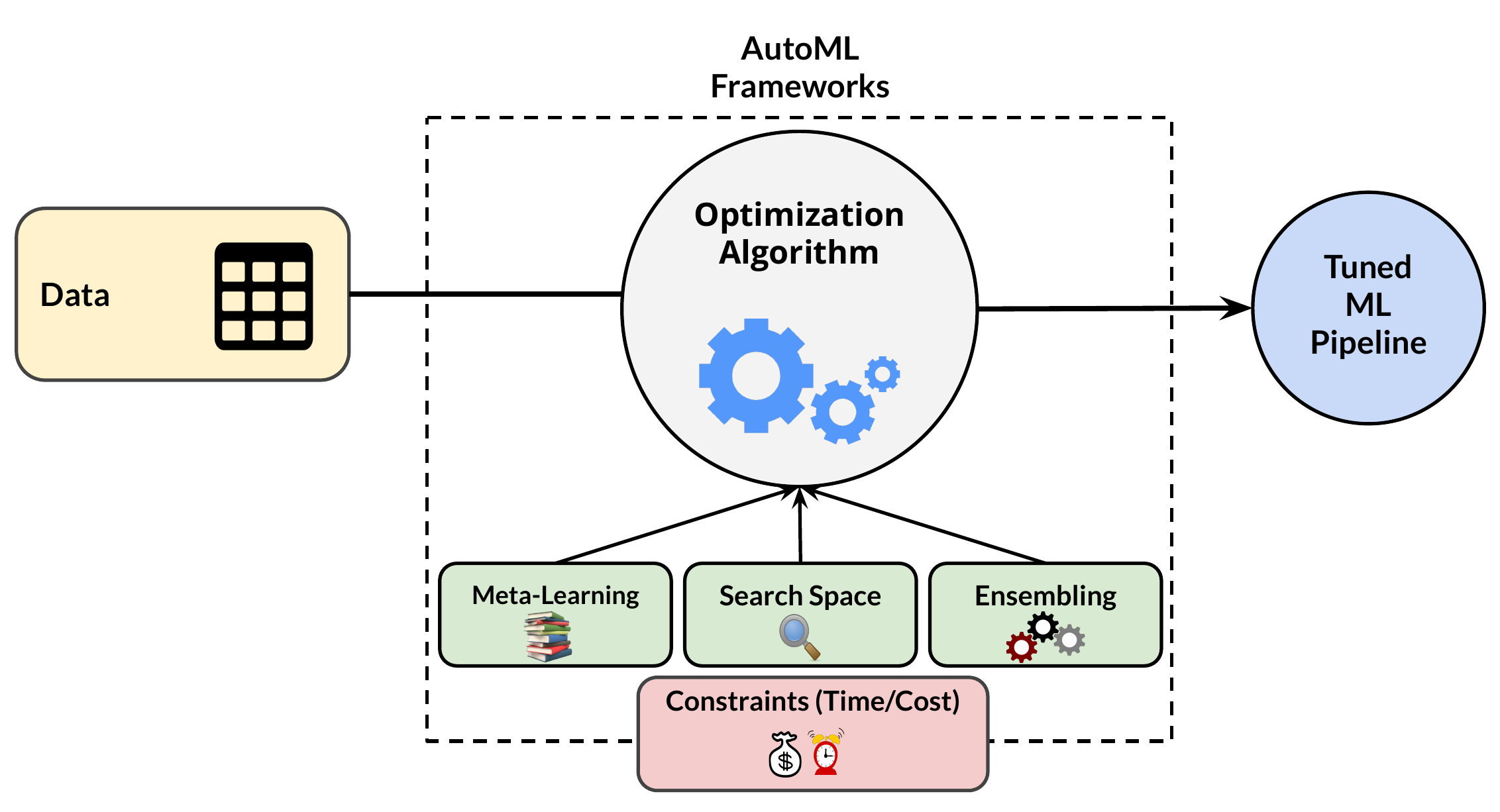}
	\caption{The general Workflow of the AutoML process.}
	\label{fig:Cash}
\end{figure}

Applying ML to real-world problems is a multi-stage process and highly iterative exploratory process. \textcolor{red}{AutoML aims to automatically produce test set predictions for a new dataset within a fixed computational budget using different optimization techniques (See Figure~\ref{fig:Cash}).}
\textcolor{red}{The problem of AutoML can formally stated as follows~\citep{feurer2019auto}: For $i=1,..., n'+m'$, let $x_{i}\in \mathbb{R}$ denote a feature vector and $y_{i}\in Y$ the corresponding target value. Given a training dataset $D_{train}=\{(x_{1},y_{1}),..., (x_{n'},y_{n'})\}$ and the feature vectors $x_{n'+1},..., x_{n'+m'}$ of a test dataset $D_{test}=\{(x_{n'+1},y_{n'+1}),..., (x_{n'+m'},y_{n'+m'})\}$ drawn from the same underlying data distribution, as well as a resource budget $b$ and a loss metric $L(.,.)$, the AutoML problem is to automatically produce test set predictions $\hat{y}_{n'+1}, ..., \hat{y}_{n'+m'}$. The loss of a solution $\hat{y}_{n'+m'}$ to the AutoML problem is given by $\frac{1}{m'} \sum_{n'=1}^{m'}$ $L(\hat{y}_{n'+j},y_{n'+m'}$).}

\textcolor{red}{The budget $b$ would comprise computational resources (e.g. CPU and/or wallclock time, memory usage).} \textcolor{red}{In particular, solving the AutoML problem aims to select and tune a ML algorithm from a defined search space that achieves (near)-optimal performance in terms of the user-defined evaluation metric (e.g., accuracy, sensitivity, specificity, F1-score) within the user-defined budget for the search process, as shown in Figure~\ref{fig:Cash}.} Additionally, different AutoML frameworks consider various \textcolor{red}{search spaces} and design decisions. For example, \texttt{SmartML}~\citep{maher2019smartml} adopts a \emph{meta-learning} based mechanism to improve the performance of the automated search process by starting with the most promising classifiers that performed well with similar datasets in the past. Another example, \texttt{AutoSKlearn}~\citep{feurer2015efficient} employs an option to take a weighted average of the predictions of an ensemble composed of the top trained models during the optimization process.
Auto-Tuned Models \textcolor{red}{(\texttt{ATM})~\citep{swearingen2017atm}} restricts the default search space into only three classifiers, namely, decision tree, K-nearest neighbours and logistic regression. In general, there is no good understanding of the impact of various design decisions of the different AutoML frameworks on the performance of the output pipeline. In this work, we aim to answer the following four questions:

(1) What is the impact of the time budget on the performance of different AutoML frameworks? Given more time budget, can AutoML frameworks guarantee consistent performance improvement?

(2) What is the impact of the search space size of the AutoML framework on the performance? \textcolor{red}{ How does limiting the search space into a predefined portfolio affect the predictive performance?}

(3) Does meta-learning always yield a consistent performance improvement across different time budgets? Is there a relationship between the characteristics of the
datasets and the improvement caused by employing the meta-learning version of the AutoML framework?

(4) Does ensemble construction always yield better performance compared to single learners across different time budgets? Is there a relationship between the characteristics of the
datasets and the improvement caused by employing the ensembling version of the AutoML framework?

This work is an extension of our
initial work~\citep{Eldeeb2021TheIO} that mainly focused on studying the impact of different design decisions on the performance of \texttt{AutoSKlearn}. In
particular, in this work, we follow a holistic approach to
design and conduct a comparative study of six AutoML frameworks, namely \texttt{AutoWeka}~\citep{Kotthoff:2017:AAM:3122009.3122034}, \texttt{AutoSKlearn}, \texttt{TPOT}~\citep{olson2016tpot}, \texttt{Recipe}~\citep{DBLP:conf/eurogp/SaPOP17}, \texttt{ATM} and \texttt{SmartML}, focusing on comparing their general performance and their performance under various design decsions including \emph{time budget}, \emph{size of search space}, \emph{meta-learning} and \emph{ensembling}. For ensuring reproducibility as one of the main targets of this work, we provide access to the source codes and the detailed results for the experiments of our studies\footnote{\url{https://datasystemsgrouput.github.io/AutoMLBench/}}. 

The remainder of this paper is organized as follows.
The related work is reviewed in Section~\ref{RelatedWork}.
Section~\ref{SEC:tools} provides an overview of the evaluated frameworks included in our study.
Section~\ref{SEC:expeimental_setup} describes our benchmark design. The evaluation of the general performance of the benchmark frameworks and the evaluation of the different design decisions on the performance of the benchmark frameworks are presented in Section~\ref{Sec:ExperimentalEvaluation}. We discuss the results and future direction in Section~\ref{Sec:DiscussionFutureDirection} before we finally conclude the paper in  Section~\ref{conclusion}.

\section{Related Work} \label{RelatedWork}
\textcolor{red}{\texttt{Auto-Weka}~\citep{autoweka} is the first and pioneer AutoML from the Universities of British Columbia (UBC) and Freiburg, which is implemented on the top of Weka~\citep{Hall:2009:WDM:1656274.1656278}. Other tools followed, including \texttt{Auto-Keras}~\citep{jin2019auto} from the Texas A\&M University running on top of Keras, Tensorflow and Scikit-learn. \texttt{MLjar}~\citep{mljar} also uses Scikit-learn in conjunction with Tensorflow. \texttt{H2O-Automl}~\citep{H2OAutoML20} was introduced by the H2O, using ML models from the H2O platform. \texttt{AutoGluon}~\citep{erickson2020autogluon} automates ML across various data types, including tabular, image, and text. Unlike previously mentioned AutoML frameworks, \texttt{AutoGluon} does not perform a pipeline search or hyperparameter tuning. Instead, it utilizes a predetermined set of models combined through multi-layer stacking and ensembling. The Fast and Lightweight AutoML Library (\texttt{FLAML})~\citep{wang2021flaml} optimizes boosting frameworks~\citep{chen2016xgboost,prokhorenkova2018catboost,ke2017lightgbm} and a few number of Scikit-learn algorithms through a multi-fidelity randomized directed search~\citep{wu2021frugal}. \texttt{Gama}~\citep{gijsbers2020gama} is a genetic-based AutoML framework that optimizes linear ML pipelines with an arbitrary number of preprocessing algorithms. \texttt{Gama} uses a multi-objective optimization technique that maximizes performance while minimizing the complexity of the pipeline. \texttt{Lightautoml} is another AutoML that is mainly designed for financial services~\citep{vakhrushev2021lightautoml}. \texttt{Lightautoml} creates pipelines for quick inference
and interpretability. Only linear models and boosting frameworks are supported. For hyperparameter optimization, \texttt{Lightautoml} opted a Tree-structured Parzen Estimators (TPE)~\citep{bergstra2013hyperopt}, while warm-starting and
early stopping are used to optimize linear models through grid search.}

Recently, few research efforts have attempted to tackle the challenge of benchmarking different AutoML frameworks~\citep{gijsbers2019open,he2019automl,abs-1906-02287,truong2019towards,zoller2021benchmark}.
In general, most experimental evaluation and comparison studies show that there is no clear winner as there are always some trade-offs that need to be considered and optimized according to user-defined objectives. \textcolor{red}{For example, Gijsbers et al.~\citep{gijsbers2022amlb} conducted a study to compare the performance of 9 AutoML frameworks, namely, Autogluon-tabular~\citep{erickson2020autogluon}, \texttt{AutoSKlearn}, \texttt{AutoSKlearn 2}~\citep{feurer2020auto}, \texttt{FLAML}, \texttt{GAMA}, \texttt{H2O AutoML}, \texttt{LightAutoML}, \texttt{MLjar}, \texttt{TPOT}, across 71 classification and 33 regression tasks. The study includes techniques for comparing AutoML frameworks,
including final model accuracy, inference time trade-offs, and failure analysis. Additionally, an interactive visualization tool for further exploration of the results and reproducibility of the analyses performed is supported.} Gijsbers et al.~\citep{gijsbers2019open} have conducted an experimental study to compare the performance of 4 AutoML frameworks, namely, \texttt{AutoWeka}, \texttt{AutoSKlearn}, \texttt{TPOT} and \texttt{H2O} on 39 datasets across two time budgets (60 minutes and 240 minutes). The results showed that no single AutoML framework outperformed others across different time budgets, and on some datasets, none of the frameworks outperformed a Random Forest within 4 hours time budget. Truong et al.~\citep{truong2019towards} compared the performance of 7 AutoML frameworks, namely, \texttt{H2O}, \texttt{Auto-keras},  \texttt{AutoSKlearn}, \texttt{Ludwig}\footnote{\url{https://github.com/uber/ludwig}}, \texttt{Darwin}\footnote{\url{https://www.sparkcognition.com/product/darwin/}}, \texttt{TPOT} and \texttt{Auto-ml}\footnote{\url{https://github.com/ClimbsRocks/auto_ml}} on 300 datasets across different time budgets. The results showed that no single framework outperformed all others
on a plurality of tasks. Across the various evaluations and
benchmarks, \texttt{H2O}, \texttt{Auto-keras} and
\texttt{AutoSKlearn} performed better than the rest of the frameworks. In particular, \texttt{H2O} slightly outperformed other frameworks for binary classification and regression tasks while achieving poor performance on multi-class classification tasks. \texttt{Auto-keras} showed a
stable performance across all tasks and slightly outperformed
other frameworks on multi-class classification tasks while achieving poor performance on binary classification tasks.

Z{\"o}ller and Huber~\citep{zoller2021benchmark} compared the performance of different optimization techniques, namely, \emph{Grid Search}, \emph{Random Search}, \emph{RObust Bayesian Optimization} (ROBO)~\citep{klein2017robo}, \emph{Bayesian Tuning and Bandits} (BTB)~\citep{smith2020machine}, \emph{hyperopt}~\citep{bergstra2013making}, \emph{SMAC}~\citep{hutter2011sequential}, \emph{BOHB}~\citep{falkner2018bohb} and \emph{Optunity}~\citep{smith2020machine}. The results showed that all optimization techniques achieved comparable performance, and a simple search algorithm such as random search did not perform worse than other techniques. Thus, the study suggested that ranking optimization techniques on pure performance measures are not reasonable, and other aspects like scalability should also be considered. The study also compared the performance of 5 AutoML frameworks, namely, \texttt{TPOT}, \texttt{hpsklearn}~\citep{komer2014hyperopt}, \texttt{AutoSKlearn}, \texttt{ATM}, and \texttt{H2O} on 73 real datasets. The study considered \texttt{AutoSKLearn} once with the default optimizer \emph{SMAC} and once replacing \emph{SMAC} with the random search while ensemble building and meta-learning options are disabled. The comparison results showed that, on average, all AutoML frameworks performed quite similar with a maximum performance difference of 2.2\%.

To the best of our knowledge, our study is the first to investigate the impact of different AutoML design decisions on the predictive performance. We benchmark six open-source, centralized and distributed AutoML frameworks, namely, \texttt{AutoWeka}, \texttt{AutoSKlearn}, \texttt{TPOT}, \texttt{Recipe}, \texttt{ATM} and \texttt{SmartML} on 100 datasets from established AutoML benchmark suites. Few recent benchmark studies focused only on comparing the performance of different AutoML frameworks while we take a holistic approach to studying the impact of various design decisions, including the size of the search space, time budget, meta-learning, and ensembling construction on the performance of the AutoML frameworks.

\section{AutoML Frameworks}
\label{SEC:tools}
This section provides an introduction to the evaluated AutoML frameworks used in this study in terms of popularity (measured in terms of the number of stars on GitHub), ML tool-box used, optimization technique, whether they use meta-learning to learn from previous experience, whether they perform post-processing (e.g., ensemble construction), whether they use Graphical User Interface (GUI), or whether they perform pre-processing. Table~\ref{TBL:AutoMLFrameworks} briefly summarizes the comparison across the AutoML frameworks considered in this study. More detailed comparisons follow in the rest of this section.

\begin{table*}
\small
\centering
\begin{adjustbox}{width=\textwidth}
\begin{tabular}{|l|c|c|c|c|c|c|c|c|}
  \hline
  \textbf{} & 
  \parbox[t]{5mm}{\rotatebox[origin=c]{90}{\begin{tabular}[c]{@{}l@{}}\textbf{Release} \\ \textbf{Date}\end{tabular}}} &
  \parbox[t]{9mm}{\rotatebox[origin=c]{90}{\begin{tabular}[c]{@{}l@{}}\textbf{Popularity} \\ \textbf{(\#of stars} \\ \textbf{on GitHub)} \end{tabular}}} & 
  \parbox[t]{3mm}{\rotatebox[origin=c]{90}{\begin{tabular}[c]{@{}l@{}}\textbf{Optimization} \\ \textbf{Technique}\end{tabular}}} & 
  \parbox[t]{3mm}{\rotatebox[origin=c]{90}{\begin{tabular}[c]{@{}l@{}}\textbf{ML} \\ \textbf{Tool Box}\end{tabular}}} & 
  \parbox[t]{5mm}{\rotatebox[origin=c]{90}{\begin{tabular}[c]{@{}l@{}}\textbf{Meta-} \\ \textbf{Learning}\end{tabular}}} & \parbox[t]{5mm}{\rotatebox[origin=c]{90}{\textbf{Post-processing}}} &
  \parbox[t]{5mm}{\rotatebox[origin=c]{90}{\begin{tabular}[c]{@{}l@{}}\textbf{GUI} \\ \textbf{ }\end{tabular}}} & 
  \parbox[t]{5mm}{\rotatebox[origin=c]{90}{\begin{tabular}[c]{@{}l@{}}\textbf{Data} \\ \textbf{Pre-processing}\end{tabular}}} \\
   \hline

  \textbf{AutoWeka} & 2013 & 291 &  Bayesian optimization & Weka & $\times$ & $\times$ & $\checkmark$ & \checkmark\\
  \hline
  
 \textbf{AutoSKlearn} & 2015 & 5.9k &   Bayesian optimization & Scikit-Learn &  \checkmark & \begin{tabular}[c]{@{}l@{}} Ensemble \\ selection \end{tabular}  &  $\times$ & \checkmark \\
  \hline
  
  \textbf{TPOT} & 2016 & 8.4k & \textcolor{red}{Evolutionary optimization} & Scikit-Learn & $\times$ & $\times$ & $\times$ & $\times$ \\
  \hline
  
  \textbf{Recipe} & 2017 & 160 & \begin{tabular}[c]{@{}l@{}l@{}} Grammar- based \\ genetic algorithm \end{tabular} &  Scikit-Learn &  $\times$ & $\times$ & $\times$ & \checkmark \\
  \hline
  
 \textbf{ATM} & 2017 & 512 & \begin{tabular}[c]{@{}l@{}} Distributed  Random search  \\ \& Tree-Parzen estimators  \end{tabular}&  Scikit-Learn & $\times$ & $\times$ & $\times$ & \checkmark \\
  \hline
  
 \textbf{SmartML} & 2019 & 20 &  Bayesian  optimization & \begin{tabular}[c]{@{}l@{}} mlr, RWeka \&\\ other R packages  \end{tabular} & \checkmark & \begin{tabular}[c]{@{}l@{}} Voting \\ ensembles \end{tabular} & $\times$ & \checkmark \\
  \hline
\end{tabular}
\end{adjustbox}
\caption{Comparison table of functionality of the AutoML frameworks considered in this study as of 24/12/2021} \label{TBL:AutoMLFrameworks}
\end{table*}

\texttt{AutoWeka} is implemented in Java on top of \texttt{Weka}, a popular ML library that has a wide range of ML algorithms. \texttt{AutoWeka} employs Bayesian optimization using \texttt{SMAC}~\citep{hutter2011sequential} and
\texttt{TPE}~\citep{bergstra2013hyperopt} for algorithm selection and hyperparameter tuning. In particular, \texttt{SMAC} draws the relationship between algorithm performance and a given set of hyperparameters by estimating the predictive mean and variance of their performance along with the
trees of a random forest model. \texttt{TPE} is a robust technique that discards low-performing parameter configurations quickly after the evaluation of a small number of dataset folds. 

\texttt{AutoSKlearn} is a tool for automating the process of building ML pipelines for classification and regression tasks. \texttt{AutoSKlearn} is implemented on top of \texttt{Scikit-Learn}~\citep{sklearn_api}, a popular Python ML package, and uses \texttt{SMAC} for algorithm selection and hyperparameter tuning. \texttt{AutoSKlearn} uses meta-learning to initialize the optimization procedure. Additionally, ensemble selection is implemented by combining the best pipelines to improve the performance of the output model. \texttt{AutoSKlearn} supports different execution options including the \emph{vanilla} version (\texttt{AutoSKlearn-v}), the meta-learning version (\texttt{AutoSKlearn-m}), the ensembling selection version (\texttt{AutoSKlearn-e}), and the full version (\texttt{AutoSKlearn}), where all options are enabled. 

\textcolor{red}{\texttt{TPOT} is an AutoML framework for building classification and regression pipelines based on genetic algorithms. ML pipelines can be expressed as a computational graph, with different branches representing different preprocessing pipelines. These pipelines are then optimized using a multi-objective optimization technique to minimize pipeline complexity while optimizing for performance to reduce overfitting caused by the large search space~\citep{olson2016evaluation}.}

 \texttt{Recipe} is an AutoML framework for building machine learning pipelines for classification tasks. \texttt{Recipe} follows the same optimization procedure as \texttt{TPOT}, which in turn exploits the advantages of a global search. \texttt{TPOT} suffers from the unconstrained search problem in which resources can be spent on generating and evaluating invalid solutions. \texttt{Recipe} handles this problem by adding a grammar that reduces the generation of invalid pipelines and hence accelerating the optimization process. 

 \texttt{ATM} is a collaborative service for optimizing ML pipelines for classification tasks. In particular, \texttt{ATM} supports parallel execution through multiple nodes/cores with a shared model hub storing the results out of these executions and improving the selection of pipelines that may outperform the current chosen ones. \texttt{ATM} is based on hybrid Bayesian and multi-armed bandit optimization technique to traverse the search space and report the target pipeline.

\texttt{SmartML} is the first AutoML \texttt{R} package for classification tasks. In the algorithm selection phase, \texttt{SmartML} employs a meta-learning approach where the meta-features of the input dataset are extracted and compared to the meta-features of datasets stored in the framework's knowledge base and populated from the results of the previous runs. The similarity search process identifies similar datasets in the knowledge base using the nearest neighbour approach. The retrieved results are used to identify the best performing algorithms on those similar datasets to nominate the candidate algorithms for the dataset at hand. The hyperparameter tuning of \texttt{SmartML} is based on \texttt{SMAC}. \texttt{SmartML} maintains the results of the new runs to continuously enrich its knowledge base to further improve the predictive performance of the similarity search process and thus the performance and robustness of the future
runs. \texttt{SmartML} supports two execution options which are the base version \texttt{SmartML-m} that employs \texttt{SMAC} and meta-leaning for algorithm selection and hyperparameter tuning, and the ensemble version \texttt{SmartML-e} that employs meta-learning for warm-starting and a voting ensemble mechanism, which is adopted by averaging the predicted probability of the top tuned models found during the optimization process based on their validation performance.

\section{Benchmark Design}
\label{SEC:expeimental_setup}

Each benchmark task consists of a dataset, a metric to optimize, and design decisions made by the user, including a specific time budget to use. We will briefly explain our choice for each.

\textbf{Datasets} We used 100 datasets collected from the popular \texttt{OpenML} repository\citep{OpenML2013}, allowing users to query data for different use cases. \textcolor{red}{Detailed descriptions of the datasets used in this study are given in Table~\ref{Tbl:AutoMLDatasets} in Appendix~\ref{secA1}.} To evaluate the AutoML frameworks on a variety of dataset characteristics, we selected multiple datasets according to different criteria, including the number of classes, number of features, number of instances, number of categorical features per sample, number of instances with missing values, and the class entropy. The datasets represent a mix of binary (50\%) and multiclass (50\%) classification tasks, where the size of the largest dataset is 643MB. 

\textbf{Performance metrics} The benchmark can be run with a wide range of measures per user's choice. \textcolor{red}{The reported results in this paper is based on F1-score.} AutoML frameworks are optimized for the same metric they are evaluated on. The measures are estimated with hold-out validation; each dataset is partitioned into two parts, 70\% for training and 30\% for testing. All AutoML frameworks are applied to the same training and testing splits on all datasets. \textcolor{red}{To eliminate the effects of non-determinism, the performance reported in each experiment is based on an average of 10 trials. We report a performance of 0 for any framework if the number of failed trails is more than or equal 5.}

\textbf{Frameworks and design decisions}
The frameworks considered in this paper are selected based on ease of use, variety of underlying optimization techniques and ML toolboxes, popularity measured by the number of stars on GitHub, and citation count. All frameworks considered in this work are open source. \textcolor{red}{A reference to the source code of each framework is given in Table~\ref{Table:AutoMLFrameworksVer} in Appendix~\ref{Appen:FrameworksBaseline}. We do plan to include more frameworks in future work.} \textcolor{red}{For \texttt{AutoSKlearn}, we consider four execution options; \texttt{AutoSKlearn-v}, \texttt{AutoSKlearn-m}, \texttt{AutoSKlearn-e} and \texttt{AutoSKlearn}. For \texttt{SmartML}, we consider two execution options including \texttt{SmartML-m}, and \texttt{SmartML-e}.} We examined different design decisions, including the size of the search space, meta-learning, and ensemble construction as a post-processing step. We study the impact of these design decisions for only AutoML frameworks that support configuring these decisions. It is important to highlight that all the frameworks do not have the same optimization technique. So, no conclusion can be drawn about this point in this benchmark. \textcolor{red}{We consider the following versions of the frameworks: AutoSKLearn 0.11.0, AutoWeka 2.5, TPOT 0.11.6, Recipe 1.0, ATM 0.2.2, and SmartML 0.2.}

\textcolor{red}{\textbf{Baseline method} To asses the effectiveness of the different AutoML frameworks included in this work, we use a baseline method which is a simple pipeline consisting of an imputation of missing values and a random forest model~\citep{scikit-learn}.}

\textcolor{red}{\textbf{Time budget choice} All AutoML frameworks were
used with four different time budgets. Each framework is limited by a soft time budget (10, 30, 60, and 240 minutes) and a hard one (10\% more than soft time budget). If a framework exceeds the hard time budget, the run is terminated and considered failed. Setting a time budget for all experiments is not straightforward. On one hand, we would like to let each tool run as long as it takes to produce the best possible performance. On the other hand, we have 6 AutoML evaluated on 100 datasets across four time-budgets; we have more than 40000 experiments to run for a total of more than 88366-hour EC2 run-time (which includes the overhead of benchmark harness code). To keep the experiment run-time and cost to practical limits, we tested the cutt-off timeouts of 4 and 8 hours on 14 randomly selected datasets. The results are reported in Table \ref{tab:cutoff} in Appendix~\ref{Appen:Cut-offtime}. Additionally, the Wilcoxon signed-rank test was conducted to determine if a statistically significant difference in performance exists between the AutoML frameworks over the two-time budgets (See Table \ref{tab:cutoff}). The results confirm that the difference is not necessarily towards the 8-hours budget, and not statistically significant. Hence, the 8-hours budget is not further considered.}

\textbf{Hardware choice and resource specifications}
Our experiments were conducted on Google Cloud machines; each machine is configured with 2 vCPUs, 7.5 GB RAM and ubuntu-minimal-1804-bionic. \textcolor{red}{Each
virtual machine uses Python 2.7.15, Python 3.6.8, scikit-learn 0.21.3, R 3.4.4, and Java 1.8.} To avoid memory leakage, we have rebooted the machines after each run to ensure that each experiment has the same available memory size.

\section{Experimental Evaluation}\label{Sec:ExperimentalEvaluation}

This section provides empirical evaluations of the different AutoML frameworks. We first compare the general performance of the different AutoML frameworks in Section~\ref{SEC:GeneralPerformanceEval}. Next, we examine the impact of various design decisions on the performance of the different AutoML frameworks in Section~\ref{Sec:PerformancEvaluationDD}.

\subsection{General Performance Evaluation}
\label{SEC:GeneralPerformanceEval}

\begin{figure*}[t!]
\centering \subfigure[Number of successful runs.] {
    \label{FIG:SuccessfulRuns}
    \includegraphics[width=0.87\textwidth]{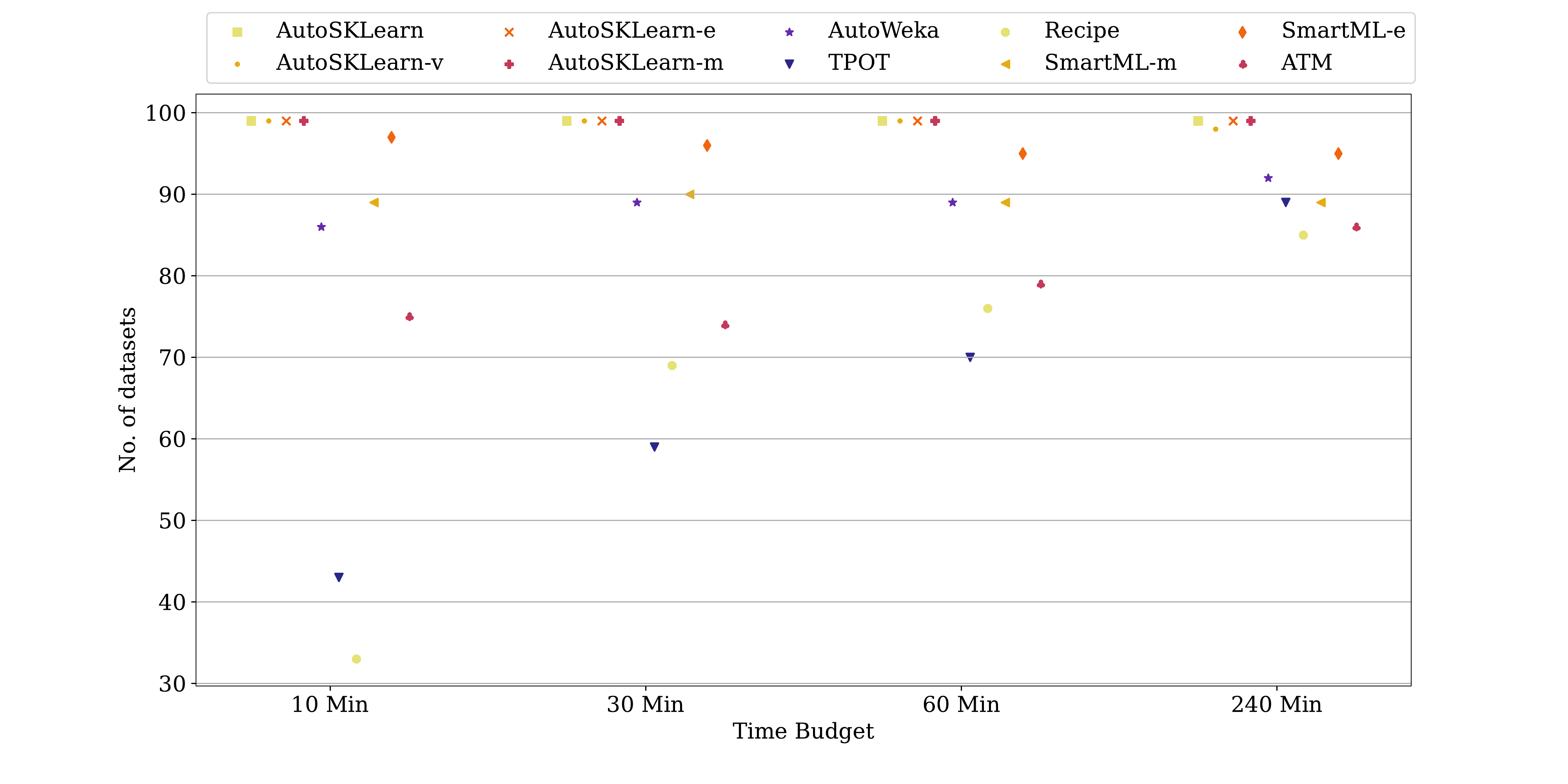}
}

\centering \subfigure[\textcolor{red}{Performance of the final pipeline per AutoML framework for 240 minutes.}] {
    \label{FIG:avg_performance}
    \includegraphics[width=0.73\textwidth]{Figures/box_plot_240 Min.pdf}
}

\caption{General performance trends of the benchmark AutoML frameworks.}
\label{FIG:General}

\end{figure*}

\begin{figure*}[t!]
\centering \subfigure[10 minutes time budget] {
    \label{FIG:Pairwisecomparison10Min}
    \includegraphics[width=0.47\textwidth]{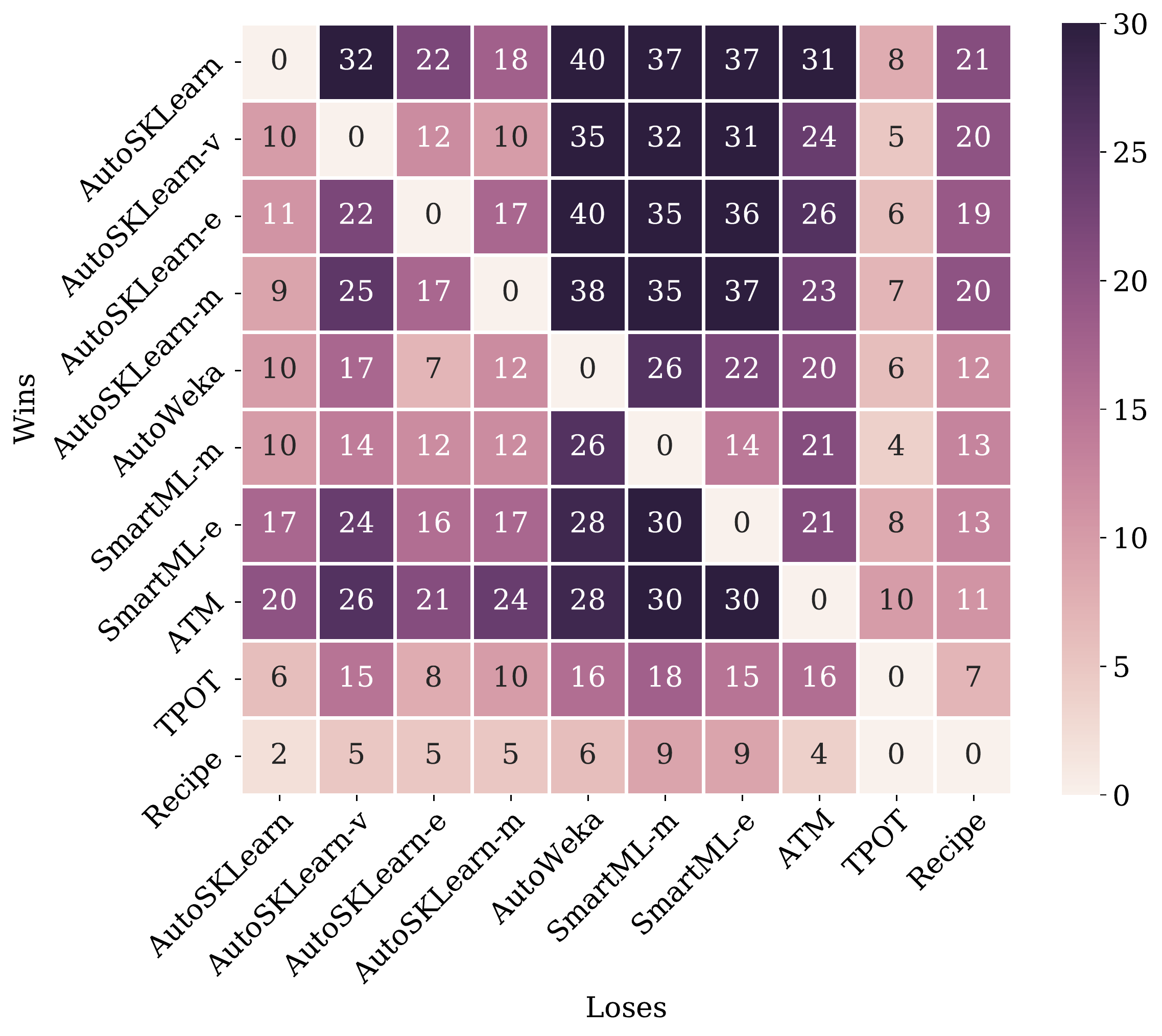}
}
\centering \subfigure[30 minutes time budget] {
    \label{FIG:Pairwisecomparison30Min}
    \includegraphics[width=0.47\textwidth]{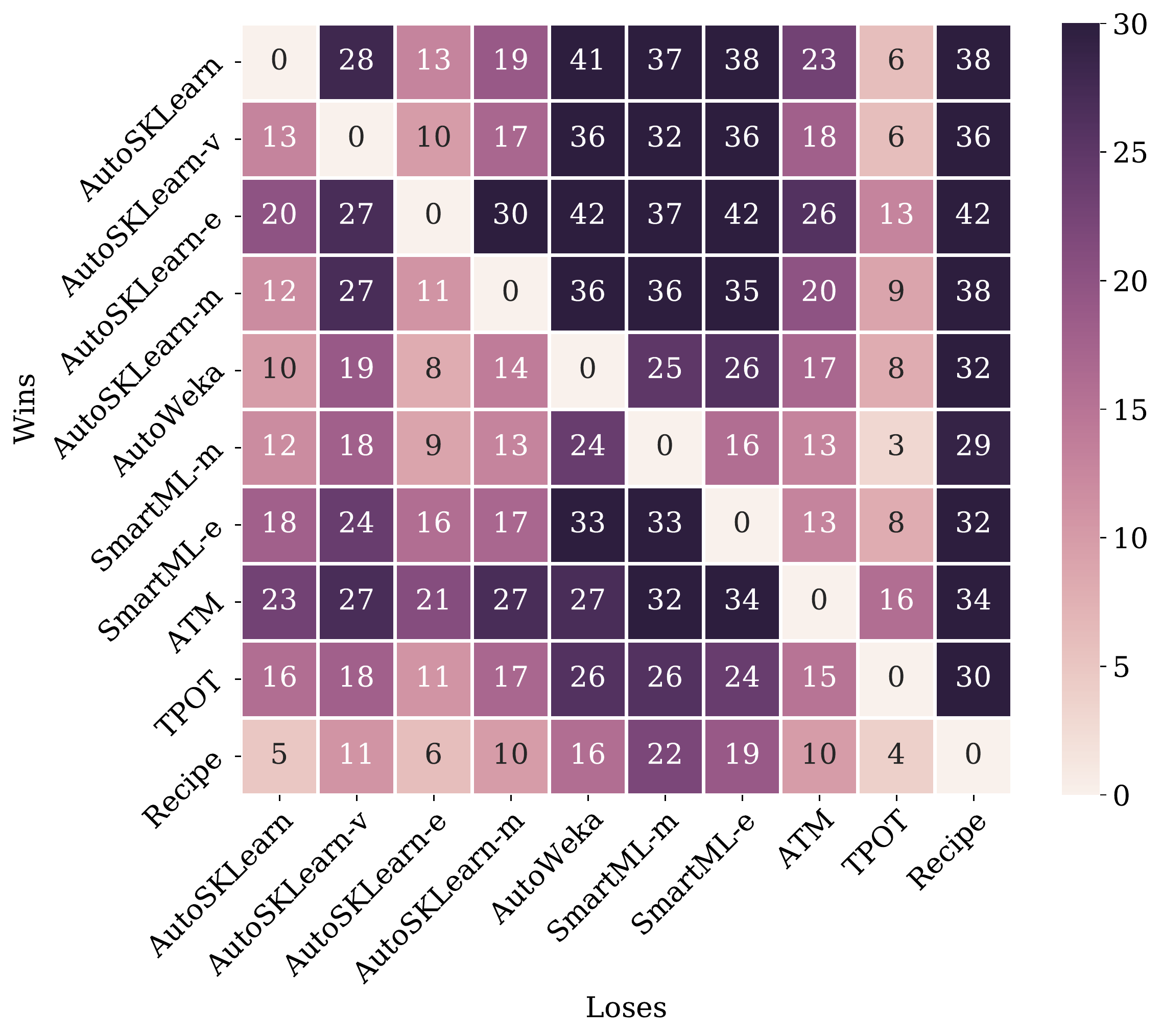}
}
\centering \subfigure[60 minutes time budget] {
    \label{FIG:Pairwisecomparison60Min}
    \includegraphics[width=0.47\textwidth]{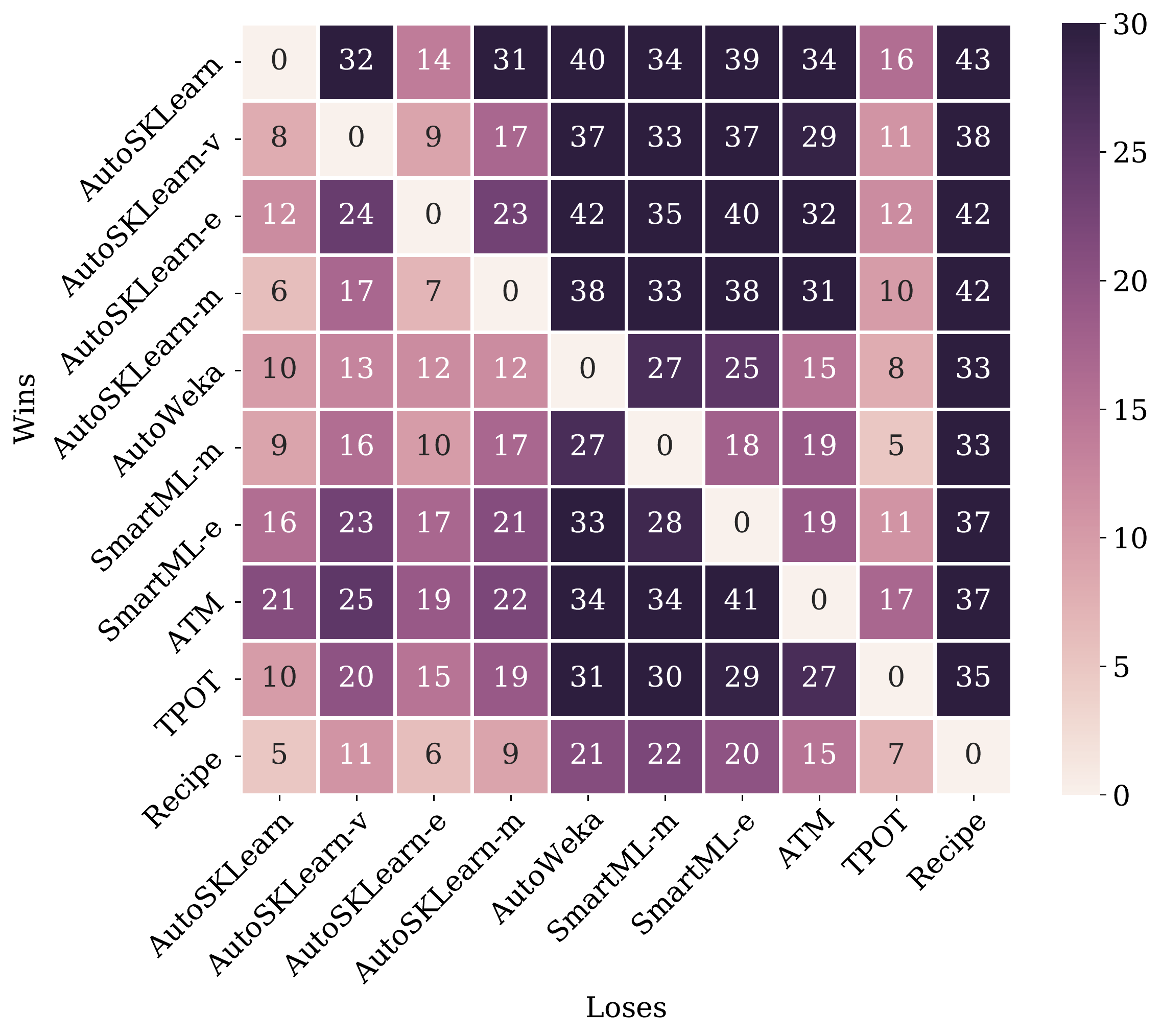}
}
\centering \subfigure[240 minutes time budget] {
    \label{FIG:Pairwisecomparison240Min}
    \includegraphics[width=0.47\textwidth]{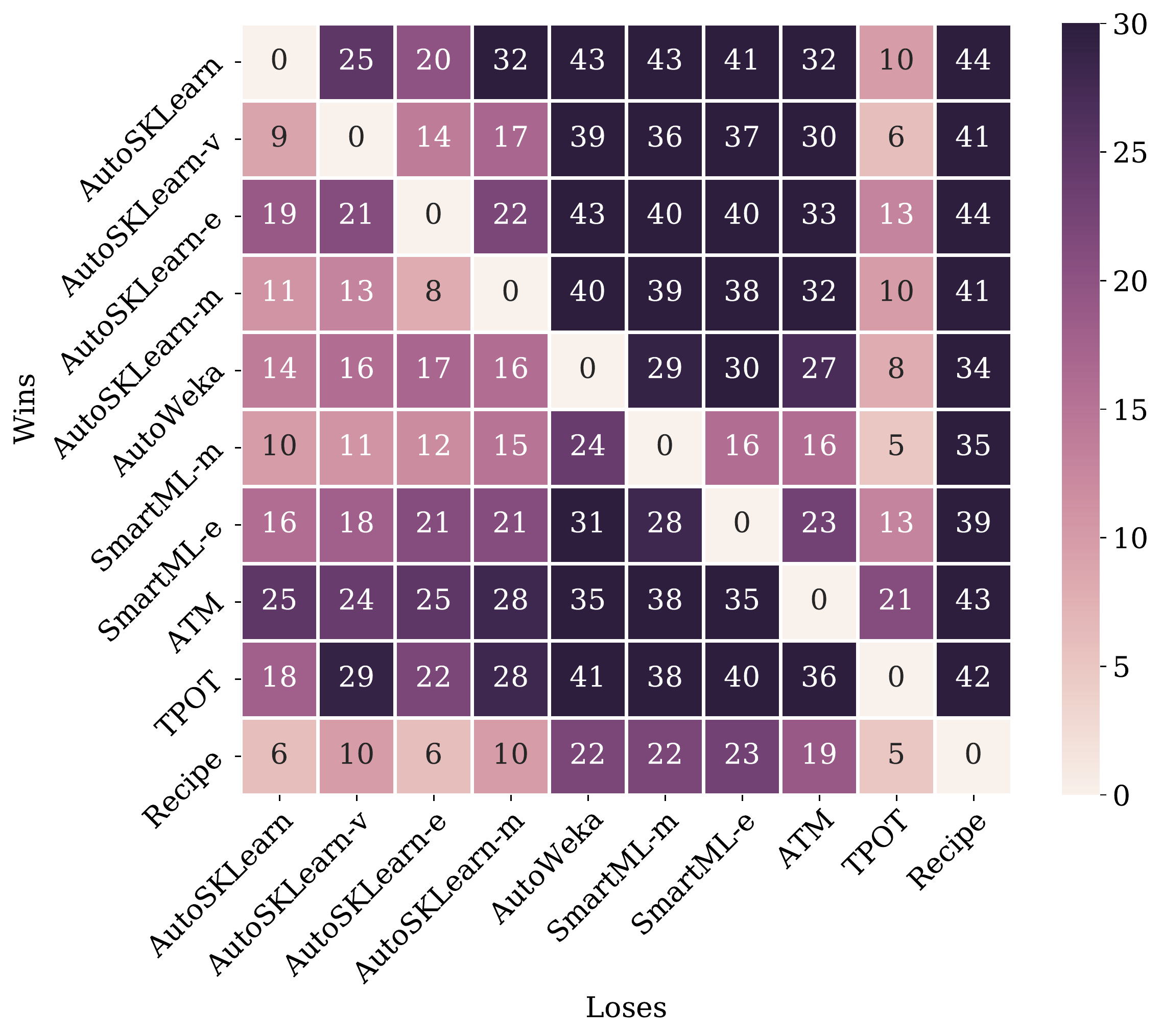}
}
\caption{Heatmaps show the number of datasets a given AutoML framework outperforms another in terms of predictive performance over different time budgets. Two frameworks are considered to have the same performance on a task if they achieve predictive performance within 1\% of each other.}
\label{FIG:heatmap}

\end{figure*}

In this section, we focus on evaluating and comparing the general performance of the benchmark frameworks. Our evaluation considers different aspects for its comparison including (a) number of successful runs, (b) \textcolor{red}{average performance of the final pipeline per AutoML framework across all datasets}, (c) significance of the performance difference between different frameworks across different time budgets, and (d) the robustness of the benchmark frameworks.  

Figure~\ref{FIG:SuccessfulRuns} shows the number of successful runs of each framework on the different time budgets. If an AutoML framework was unable to generate a model in a particular run, then it is considered a failed run. Generally, the results show that increasing the time budget for the AutoML frameworks increases the number of successful runs. \texttt{AutoSKlearn} achieves the largest number of successful runs across all time budgets, as shown in Figure~\ref{FIG:SuccessfulRuns}. Each of the different versions of \texttt{AutoSKlearn} successfully ran on 99 datasets across different time budgets. \texttt{SmartML-e} comes in the second place in terms of the number of successful runs, followed by \texttt{AutoWeka} and \texttt{SmartML}. The genetic-based frameworks, \texttt{TPOT} and \texttt{Recipe} come in the last place, as shown in Figure~\ref{FIG:SuccessfulRuns}. For \texttt{Recipe} and \texttt{TPOT}, the number of successful runs achieved in the longest time budget, 240 minutes, is almost double that achieved for the smallest time budget of 10 minutes. Hence, larger budgets are preferable for \texttt{Recipe} and \texttt{TPOT}.

\begin{figure*}[t!]
\centering \subfigure[\textcolor{red}{Performance of the final pipeline on multi-class classification tasks.}] {
    \label{IG:avg_performance_multi_class}
    \includegraphics[width=0.73\textwidth]{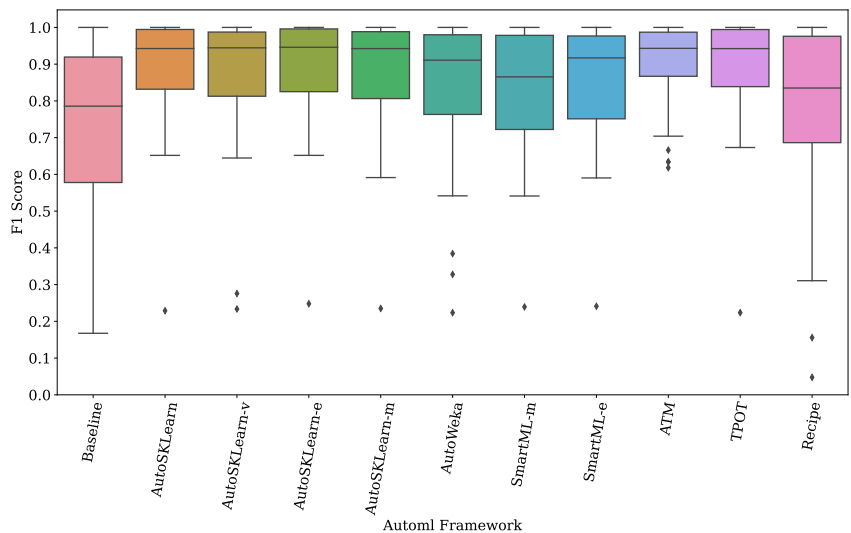}
}

\centering \subfigure[\textcolor{red}{Performance of the final pipeline on datasets with large number of features and small number of instances.}] {
    \label{FIG:avg_performance_large_features_small_instances}
    \includegraphics[width=0.73\textwidth]{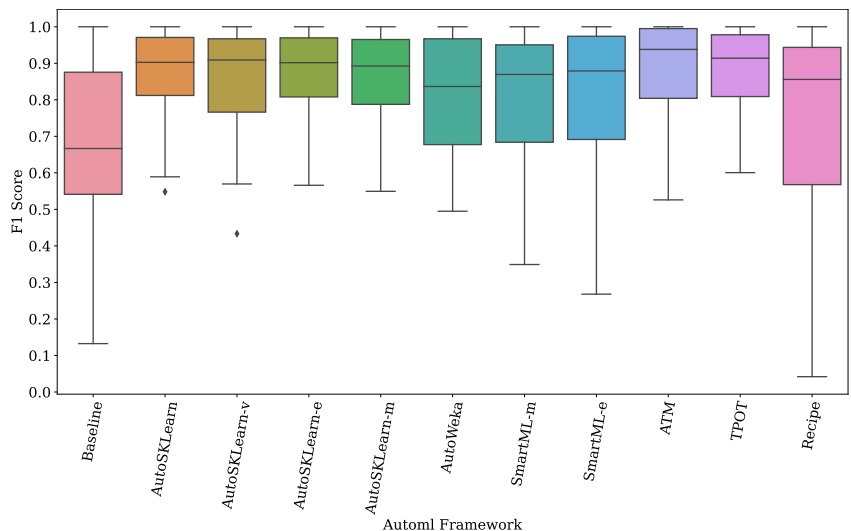}
}

\caption{\textcolor{red}{Performance of the different AutoML frameworks based on
the various characteristics of datasets and tasks over 240 minutes.}}
\label{FIG:avg_performance_main}

\end{figure*}





\begin{table*}[!ht]
\centering\scriptsize
\caption{Wilcoxon pairwise test p-values for AutoML frameworks over different time budgets. Bold entries highlight significant differences ($p\leq 0.05$) and highlighted entries in each row represent a given AutoML framework (row) outperforms another AutoML framework (column).}  

\resizebox{\textwidth}{!}{\begin{tabular}{|c|c|c|c|c|c|c|c|c|c|c|c|}
\toprule
\hline
\multicolumn{10}{c}{\textbf{\LARGE 10 Minutes}}\\
\hline
& Baseline & ATM & AutoWeka & Recipe & AutoSKLearn-e &   AutoSKLearn-m & AutoSKLearn-v &  AutoSKLearn & SmartML-m & SmartML-e &  TPOT \\

\hline
Baseline & 
\cellcolor{black!25}{}&0.0&0.0&0.15&0.0&0.0&0.0&0.0&0.0&0.0&0.0\\
\hline
ATM & 
\cellcolor{blue!25}\textbf{0.0}&\cellcolor{black!25}{}&\cellcolor{blue!25}\textbf{0.008}&0.088&0.768&0.516&0.299&0.587&0.064&0.062&0.879\\
\hline
AutoWeka & 
\cellcolor{blue!25}\textbf{0.0}&0.008&\cellcolor{black!25}{}&0.156&0.0&0.0&0.004&0.0&0.748&0.096&0.06\\
\hline
Recipe & 
0.15&0.088&0.156&\cellcolor{black!25}{}&0.002&0.002&0.004&0.0&0.417&0.248&0.013\\
\hline
AutoSKLearn-e & 
\cellcolor{blue!25}\textbf{0.0}&0.768&\cellcolor{blue!25}\textbf{0.0}&\cellcolor{blue!25}\textbf{0.002}&\cellcolor{black!25}{}&0.569&\cellcolor{blue!25}\textbf{0.014}&0.001&\cellcolor{blue!25}\textbf{0.023}&0.203&0.492\\
\hline
AutoSKLearn-m & 
\cellcolor{blue!25}\textbf{0.0}&0.516&\cellcolor{blue!25}\textbf{0.0}&\cellcolor{blue!25}\textbf{0.002}&0.569&\cellcolor{black!25}{}&\cellcolor{blue!25}\textbf{0.009}&0.009&\cellcolor{blue!25}\textbf{0.004}&0.1&0.33\\
\hline
AutoSKLearn-v & 
\cellcolor{blue!25}\textbf{0.0}&0.299&\cellcolor{blue!25}\textbf{0.004}&\cellcolor{blue!25}\textbf{0.004}&0.014&0.009&\cellcolor{black!25}{}&0.0&\cellcolor{blue!25}\textbf{0.042}&0.663&\cellcolor{blue!25}\textbf{0.026}\\
\hline
AutoSKLearn & 
\cellcolor{blue!25}\textbf{0.0}&0.587&\cellcolor{blue!25}\textbf{0.0}&\cellcolor{blue!25}\textbf{0.0}&\cellcolor{blue!25}\textbf{0.001}&\cellcolor{blue!25}\textbf{0.009}&\cellcolor{blue!25}\textbf{0.0}&\cellcolor{black!25}{}&\cellcolor{blue!25}\textbf{0.001}&\cellcolor{blue!25}\textbf{0.035}&0.258\\
\hline
SmartML-m & 
\cellcolor{blue!25}\textbf{0.0}&0.064&0.748&0.417&0.023&0.004&0.042&0.001&\cellcolor{black!25}{}&0.014&0.022\\
\hline
SmartML-e & 
\cellcolor{blue!25}\textbf{0.0}&0.062&0.096&0.248&0.203&0.1&0.663&0.035&\cellcolor{blue!25}\textbf{0.014}&\cellcolor{black!25}{}&0.452\\
\hline
TPOT & 
\cellcolor{blue!25}\textbf{0.0}&0.879&0.06&\cellcolor{blue!25}\textbf{0.013}&0.492&0.33&0.026&0.258&\cellcolor{blue!25}\textbf{0.022}&0.452&\cellcolor{black!25}{}\\
\hline

\multicolumn{10}{c}{\textbf{\LARGE 30 Minutes}}\\
\hline
& Baseline & ATM & AutoWeka & Recipe & AutoSKLearn-e &   AutoSKLearn-m & AutoSKLearn-v &  AutoSKLearn & SmartML-m & SmartML-e &  TPOT \\

\hline
Baseline & 
\cellcolor{black!25}{}&0.0&0.0&0.346&0.0&0.0&0.0&0.0&0.0&0.0&0.0\\
\hline
ATM & 
\cellcolor{blue!25}\textbf{0.0}&\cellcolor{black!25}{}&\cellcolor{blue!25}\textbf{0.034}&\cellcolor{blue!25}\textbf{0.0}&0.898&0.408&0.159&0.85&\cellcolor{blue!25}\textbf{0.009}&\cellcolor{blue!25}\textbf{0.015}&0.902\\
\hline
AutoWeka & 
\cellcolor{blue!25}\textbf{0.0}&0.034&\cellcolor{black!25}{}&\cellcolor{blue!25}\textbf{0.004}&0.0&0.0&0.003&0.0&0.92&0.195&0.003\\
\hline
Recipe & 
0.346&0.0&0.004&\cellcolor{black!25}{}&0.0&0.0&0.0&0.0&0.134&0.007&0.0\\
\hline
AutoSKLearn-e & 
\cellcolor{blue!25}\textbf{0.0}&0.898&\cellcolor{blue!25}\textbf{0.0}&\cellcolor{blue!25}\textbf{0.0}&\cellcolor{black!25}{}&\cellcolor{blue!25}\textbf{0.015}&\cellcolor{blue!25}\textbf{0.0}&0.694&\cellcolor{blue!25}\textbf{0.001}&\cellcolor{blue!25}\textbf{0.005}&0.94\\
\hline
AutoSKLearn-m & 
\cellcolor{blue!25}\textbf{0.0}&0.408&\cellcolor{blue!25}\textbf{0.0}&\cellcolor{blue!25}\textbf{0.0}&0.015&\cellcolor{black!25}{}&\cellcolor{blue!25}\textbf{0.03}&0.152&\cellcolor{blue!25}\textbf{0.006}&0.075&0.316\\
\hline
AutoSKLearn-v & 
\cellcolor{blue!25}\textbf{0.0}&0.159&\cellcolor{blue!25}\textbf{0.003}&\cellcolor{blue!25}\textbf{0.0}&0.0&0.03&\cellcolor{black!25}{}&0.0&0.064&0.112&0.005\\
\hline
AutoSKLearn & 
\cellcolor{blue!25}\textbf{0.0}&0.85&\cellcolor{blue!25}\textbf{0.0}&\cellcolor{blue!25}\textbf{0.0}&0.694&0.152&\cellcolor{blue!25}\textbf{0.0}&\cellcolor{black!25}{}&\cellcolor{blue!25}\textbf{0.002}&\cellcolor{blue!25}\textbf{0.014}&0.337\\
\hline
SmartML-m & 
\cellcolor{blue!25}\textbf{0.0}&0.009&0.92&0.134&0.001&0.006&0.064&0.002&\cellcolor{black!25}{}&0.015&0.002\\
\hline
SmartML-e & 
\cellcolor{blue!25}\textbf{0.0}&0.015&0.195&\cellcolor{blue!25}\textbf{0.007}&0.005&0.075&0.112&0.014&\cellcolor{blue!25}\textbf{0.015}&\cellcolor{black!25}{}&0.065\\
\hline
TPOT & 
\cellcolor{blue!25}\textbf{0.0}&0.902&\cellcolor{blue!25}\textbf{0.003}&\cellcolor{blue!25}\textbf{0.0}&0.94&0.316&\cellcolor{blue!25}\textbf{0.005}&0.337&\cellcolor{blue!25}\textbf{0.002}&0.065&\cellcolor{black!25}{}\\
\hline

\multicolumn{10}{c}{\textbf{\LARGE 60 Minutes}}\\
\hline
& Baseline & ATM & AutoWeka & Recipe & AutoSKLearn-e &   AutoSKLearn-m & AutoSKLearn-v &  AutoSKLearn & SmartML-m & SmartML-e &  TPOT \\

\hline
Baseline & 
\cellcolor{black!25}{}&0.0&0.0&0.201&0.0&0.0&0.0&0.0&0.0&0.0&0.0\\
\hline
ATM & 
\cellcolor{blue!25}\textbf{0.0}&\cellcolor{black!25}{}&\cellcolor{blue!25}\textbf{0.017}&\cellcolor{blue!25}\textbf{0.0}&0.075&0.358&0.424&0.149&\cellcolor{blue!25}\textbf{0.005}&\cellcolor{blue!25}\textbf{0.004}&0.064\\
\hline
AutoWeka & 
\cellcolor{blue!25}\textbf{0.0}&0.017&\cellcolor{black!25}{}&\cellcolor{blue!25}\textbf{0.015}&0.0&0.0&0.0&0.0&0.59&0.083&0.0\\
\hline
Recipe & 
0.201&0.0&0.015&\cellcolor{black!25}{}&0.0&0.0&0.0&0.0&0.054&0.003&0.0\\
\hline
AutoSKLearn-e & 
\cellcolor{blue!25}\textbf{0.0}&0.075&\cellcolor{blue!25}\textbf{0.0}&\cellcolor{blue!25}\textbf{0.0}&\cellcolor{black!25}{}&\cellcolor{blue!25}\textbf{0.046}&\cellcolor{blue!25}\textbf{0.003}&0.319&\cellcolor{blue!25}\textbf{0.003}&\cellcolor{blue!25}\textbf{0.011}&0.198\\
\hline
AutoSKLearn-m & 
\cellcolor{blue!25}\textbf{0.0}&0.358&\cellcolor{blue!25}\textbf{0.0}&\cellcolor{blue!25}\textbf{0.0}&0.046&\cellcolor{black!25}{}&0.474&0.0&\cellcolor{blue!25}\textbf{0.012}&0.052&0.067\\
\hline
AutoSKLearn-v & 
\cellcolor{blue!25}\textbf{0.0}&0.424&\cellcolor{blue!25}\textbf{0.0}&\cellcolor{blue!25}\textbf{0.0}&0.003&0.474&\cellcolor{black!25}{}&0.0&\cellcolor{blue!25}\textbf{0.039}&0.201&0.01\\
\hline
AutoSKLearn & 
\cellcolor{blue!25}\textbf{0.0}&0.149&\cellcolor{blue!25}\textbf{0.0}&\cellcolor{blue!25}\textbf{0.0}&0.319&\cellcolor{blue!25}\textbf{0.0}&\cellcolor{blue!25}\textbf{0.0}&\cellcolor{black!25}{}&\cellcolor{blue!25}\textbf{0.001}&\cellcolor{blue!25}\textbf{0.015}&0.86\\
\hline
SmartML-m & 
\cellcolor{blue!25}\textbf{0.0}&0.005&0.59&0.054&0.003&0.012&0.039&0.001&\cellcolor{black!25}{}&0.047&0.0\\
\hline
SmartML-e & 
\cellcolor{blue!25}\textbf{0.0}&0.004&0.083&\cellcolor{blue!25}\textbf{0.003}&0.011&0.052&0.201&0.015&\cellcolor{blue!25}\textbf{0.047}&\cellcolor{black!25}{}&0.007\\
\hline
TPOT & 
\cellcolor{blue!25}\textbf{0.0}&0.064&\cellcolor{blue!25}\textbf{0.0}&\cellcolor{blue!25}\textbf{0.0}&0.198&0.067&\cellcolor{blue!25}\textbf{0.01}&0.86&\cellcolor{blue!25}\textbf{0.0}&\cellcolor{blue!25}\textbf{0.007}&\cellcolor{black!25}{}\\
\hline

\multicolumn{10}{c}{\textbf{\LARGE 4 Hours}}\\
\hline
& Baseline & ATM & AutoWeka & Recipe & AutoSKLearn-e &   AutoSKLearn-m & AutoSKLearn-v &  AutoSKLearn & SmartML-m & SmartML-e &  TPOT \\

\hline
Baseline & 
\cellcolor{black!25}{}&0.0&0.0&0.039&0.0&0.0&0.0&0.0&0.0&0.0&0.0\\
\hline
ATM & 
\cellcolor{blue!25}\textbf{0.0}&\cellcolor{black!25}{}&\cellcolor{blue!25}\textbf{0.046}&\cellcolor{blue!25}\textbf{0.0}&0.637&0.943&0.969&0.754&\cellcolor{blue!25}\textbf{0.002}&0.061&0.153\\
\hline
AutoWeka & 
\cellcolor{blue!25}\textbf{0.0}&0.046&\cellcolor{black!25}{}&\cellcolor{blue!25}\textbf{0.027}&0.0&0.001&0.002&0.0&0.773&0.389&0.0\\
\hline
Recipe & 
\cellcolor{blue!25}\textbf{0.039}&0.0&0.027&\cellcolor{black!25}{}&0.0&0.0&0.0&0.0&0.024&0.004&0.0\\
\hline
AutoSKLearn-e & 
\cellcolor{blue!25}\textbf{0.0}&0.637&\cellcolor{blue!25}\textbf{0.0}&\cellcolor{blue!25}\textbf{0.0}&\cellcolor{black!25}{}&\cellcolor{blue!25}\textbf{0.015}&\cellcolor{blue!25}\textbf{0.021}&0.447&\cellcolor{blue!25}\textbf{0.001}&\cellcolor{blue!25}\textbf{0.007}&0.152\\
\hline
AutoSKLearn-m & 
\cellcolor{blue!25}\textbf{0.0}&0.943&\cellcolor{blue!25}\textbf{0.001}&\cellcolor{blue!25}\textbf{0.0}&0.015&\cellcolor{black!25}{}&0.852&0.0&\cellcolor{blue!25}\textbf{0.006}&\cellcolor{blue!25}\textbf{0.043}&0.001\\
\hline
AutoSKLearn-v & 
\cellcolor{blue!25}\textbf{0.0}&0.969&\cellcolor{blue!25}\textbf{0.002}&\cellcolor{blue!25}\textbf{0.0}&0.021&0.852&\cellcolor{black!25}{}&0.001&\cellcolor{blue!25}\textbf{0.004}&0.06&0.0\\
\hline
AutoSKLearn & 
\cellcolor{blue!25}\textbf{0.0}&0.754&\cellcolor{blue!25}\textbf{0.0}&\cellcolor{blue!25}\textbf{0.0}&0.447&\cellcolor{blue!25}\textbf{0.0}&\cellcolor{blue!25}\textbf{0.001}&\cellcolor{black!25}{}&\cellcolor{blue!25}\textbf{0.0}&\cellcolor{blue!25}\textbf{0.002}&0.119\\
\hline
SmartML-m & 
\cellcolor{blue!25}\textbf{0.0}&0.002&0.773&\cellcolor{blue!25}\textbf{0.024}&0.001&0.006&0.004&0.0&\cellcolor{black!25}{}&0.031&0.0\\
\hline
SmartML-e & 
\cellcolor{blue!25}\textbf{0.0}&0.061&0.389&\cellcolor{blue!25}\textbf{0.004}&0.007&0.043&0.06&0.002&\cellcolor{blue!25}\textbf{0.031}&\cellcolor{black!25}{}&0.001\\
\hline
TPOT & 
\cellcolor{blue!25}\textbf{0.0}&0.153&\cellcolor{blue!25}\textbf{0.0}&\cellcolor{blue!25}\textbf{0.0}&0.152&\cellcolor{blue!25}\textbf{0.001}&\cellcolor{blue!25}\textbf{0.0}&0.119&\cellcolor{blue!25}\textbf{0.0}&\cellcolor{blue!25}\textbf{0.001}&\cellcolor{black!25}{}\\
\hline

\bottomrule

\end{tabular}}
\label{table:10-min_wilcoxon_matrix}
\end{table*}

\textcolor{red}{Figure~\ref{FIG:avg_performance} reports the performances of all AutoML frameworks averaged over all datasets over 240 minutes budget. It is apparent that all frameworks are able to outperform the random forest baseline on average. However, single results vary significantly. ~\Cref{FIG:avg_performance_10,FIG:avg_performance_30,FIG:avg_performance_60} in Appendix~\ref{Appen:GeneralPerformanceEval} reports the performance of all AutoML frameworks and the baseline across 10, 30, 60 minutes, respectively}. We investigate pair-wise “outperformance” by calculating the number of datasets for which one framework outperforms another across different time budgets, shown in Figure~\ref{FIG:heatmap}. One framework outperforms another on a dataset if it has at least a 1\% higher predictive performance, representing a minimal threshold for performance improvement. In terms of “outperformance”, it is worth mentioning that no single AutoML framework performs best across all 100 datasets on all-time budgets. For example, for the 10 minutes time budget, there are 2 datasets for which \texttt{Recipe} performs better than \texttt{AutoSKlearn}, despite being the overall worst- and best-ranked algorithms, respectively, as shown in Figure~\ref{FIG:Pairwisecomparison10Min}. On average, the results show that \texttt{AutoSKlearn} framework comes in the first place, outperforming other frameworks on the most significant number of datasets for different time budgets, followed by \texttt{ATM} framework, while \texttt{Recipe} comes in the last place, as shown in Figure~\ref{FIG:heatmap}. The Wilcoxon signed-rank test ~\citep{gehan1965generalized} was conducted to determine if a statistically significant difference in performance exists between the AutoML frameworks including the baseline over different time budgets, the results of which are summarized in Table~\ref{table:10-min_wilcoxon_matrix}. \textcolor{red}{The results show that all AutoML frameworks except \texttt{Recipe} statistically outperform the baseline across all time budgets with a significant difference.}
The results of the Wilcoxon test confirm the fact that there is no clear winner, and the \textcolor{red}{statistical} significance in the performance difference among the AutoML frameworks can vary from one-time budget to another. The ensembling version and the full version of
\texttt{AutoSKlearn} statistically significantly outperform most of the other frameworks across all time budgets. The results show that \texttt{SmartML-m}, \texttt{SmartML-e}, and \texttt{AutoWeka} are \textcolor{red}{statistically} significantly outperformed by the majority of the frameworks, as shown in Table~\ref{table:10-min_wilcoxon_matrix}. For longer time budgets of 60 and 240 minutes, \texttt{TPOT} significantly outperforms \texttt{AutoWeka}, \texttt{Recipe}, \texttt{SmartML-m}, \texttt{SmartML-e}, \texttt{AutoSKlearn-m}, and \texttt{AutoSKlearn-v}.

\begin{figure}[!t]
  \centering
  \includegraphics[width=0.7\linewidth]{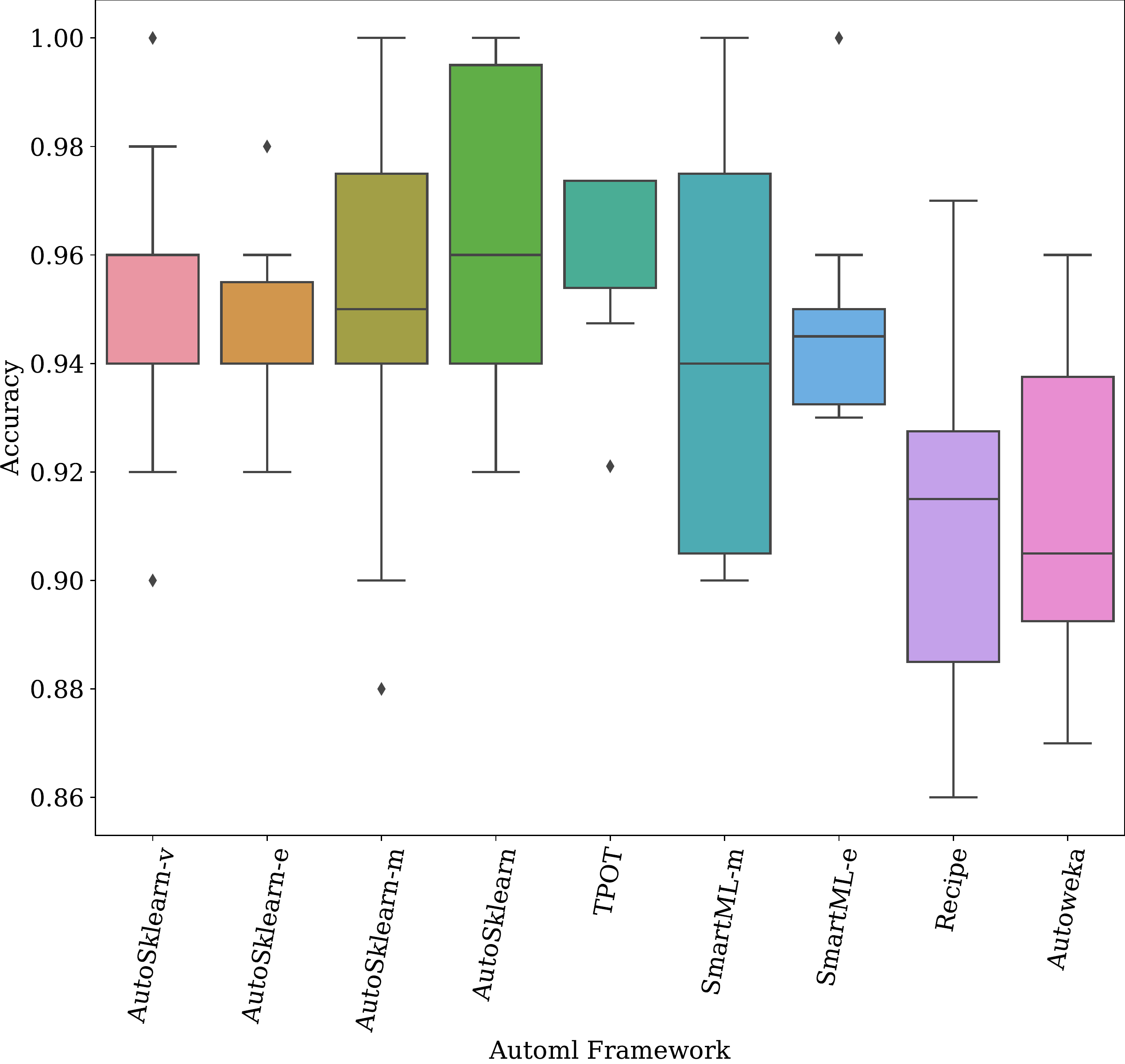}
  \caption{Evaluation of AutoML frameworks on robustness}
  \label{fig:robustness}
\end{figure}

\textcolor{red}{We investigate the performance of the different AutoML frameworks based on the various characteristics of datasets and tasks. Figure~\ref{IG:avg_performance_multi_class} reports the mean performances of the AutoML frameworks on multi-class class classification tasks across 240 minutes budget. Notably, all frameworks outperform the baseline. All the versions of \texttt{AutoSKlearn}, \texttt{ATM}, and \texttt{TPOT} achieve highest mean performance across all the multi-class class classification tasks, while \texttt{SmartML} achieves the lowest mean performance. Figure~\ref{FIG:avg_performance_large_features_small_instances} reports the mean performance of the AutoML frameworks over all datasets with a large number of features and a small number of instances. Notably, the improvement achieved by all AutoML of such datasets, i.e., multi-class or with a large number of features and a small number of instances, is less significant than the average improvement on whole datasets. \texttt{ATM} achieves the highest mean performance, while \texttt{Recipe} achieves a comparable performance to the baseline. ~\Cref{FIG:avg_performance_large_features_large_instances,FIG:avg_performance_small_features_small_instances,FIG:avg_performance_small_features_large_instances} in Appendix~\ref{Appen:GeneralPerformanceEval} report the mean performance of the different AutoML frameworks on datasets with various characteristics including large number of instances and features, small number of features and instances, small number of features and large number of instances. Additionally, we report the mean performance of the all frameworks on binary classification tasks (See Figure~\ref{FIG:avg_performance_binary_class} in Appendix~\ref{Appen:GeneralPerformanceEval}).}

We test the robustness of the AutoML frameworks evaluated by the ability of the framework to achieve the same results across different runs on the same input datasets. \textcolor{red}{For a randomly selected dataset, we run each AutoML framework for 10 different times on 10 minutes time budget.} Figure~\ref{fig:robustness} shows the results of the robustness of the AutoML frameworks. The results show that \texttt{Recipe}, \texttt{AutoWeka}, \texttt{SmartML-m}, and all versions of \texttt{AutoSKlearn} obtain very stable performance, while \texttt{TPOT}, \texttt{ATM} and  \texttt{SmartML-e} get less stability.

\subsection{Performance Evaluation of Different Design Decisions}\label{Sec:PerformancEvaluationDD}
In this section, we study the impact of different design decisions including time budget (Section~\ref{SEC:DDTimeBudget}), size of search space (Section~\ref{SEC:DDSearchSpace}), meta-learning (Section~\ref{Sec:MetaLearSKlearn}), and ensembling (Section~\ref{SEC:DDEnsembling}) on the performance of the different AutoML frameworks across different time budgets. \textcolor{red}{For each framework, the performance reported in each experiment is based on an average of 10 runs.}

\subsubsection{Impact of Time Budget} \label{SEC:DDTimeBudget}

\begin{table}[!t]
\centering
\caption{Mean$_{Succ}$, Mean and standard deviation of the predictive performance of AutoML frameworks per time budget. Bold entries highlight highest Mean$_{Succ}$, mean and lowest standard deviation}
\resizebox{0.9\textwidth}{!}{\begin{tabular}{|c|c|c|c|c|c|c|c|c|c|}
\hline
\rowcolor[HTML]{EFEFEF} 
\cellcolor[HTML]{343434}{\color[HTML]{FFFFFF} \textbf{Time Budget}} & \textbf{Framework} & \textbf{$Mean_{Succ}$} & \textbf{Mean} & \textbf{SD} & \cellcolor[HTML]{333333}{\color[HTML]{FFFFFF} \textbf{Time Budget}} & \textbf{Framework} & \textbf{$Mean_{Succ}$} & \textbf{Mean} & SD \\ \hline
 & \textbf{ATM} & 0.664 &  0.886 &  0.126 &  & \textbf{ATM} & 0.700 &  \textbf{0.886} &  \textbf{0.132} \\ \cline{2-5} \cline{7-10} 
 & \textbf{AutoWeka} & 0.724 & 0.842 & 0.165 &  & \textbf{AutoWeka} & 0.743 &  0.835 &  0.167 \\ \cline{2-5} \cline{7-10} 
 & \textbf{Recipe} & 0.252 & 0.764 & 0.221 &  & \textbf{Recipe} & 0.568 &  0.748 &  0.247 \\ \cline{2-5} \cline{7-10} 
 & \textbf{AutoSKLearn-e} & \textbf{0.859} & 0.868 & 0.145 &  & \textbf{AutoSKLearn-e} & \textbf{0.870} &  0.879 &  0.138 \\ \cline{2-5} \cline{7-10} 
 & \textbf{AutoSKLearn-m} & 0.855 &  0.864 &  0.153 &  & \textbf{AutoSKLearn-m} & 0.861 &  0.870 &  0.144 \\ \cline{2-5} \cline{7-10} 
 & \textbf{AutoSKLearn-v} & 0.853 &  0.862 &  0.151 &  & \textbf{AutoSKLearn-v} & 0.861 &  0.870 &  0.142 \\ \cline{2-5} \cline{7-10} 
 & \textbf{AutoSKLearn} & \textbf{0.859} &  0.868 &  0.152 &  & \textbf{AutoSKLearn} & 0.868 &  0.877 &  0.137 \\ \cline{2-5} \cline{7-10} 
 & \textbf{SmartML-m} & 0.806 &  0.799 &  0.212 &  & \textbf{SmartML-m} & 0.790 & 0.816 & 0.194 \\ \cline{2-5} \cline{7-10}
 & \textbf{SmartML-e} & 0.711 &  0.831 &  0.176 &  & \textbf{SmartML-e} & 0.726 & 0.832 & 0.0172 \\ \cline{2-5} \cline{7-10}
\multirow{-9}{*}{\textbf{10 Min}} & \textbf{TPOT} & 0.383 &  \textbf{0.890} &  \textbf{0.121} & \multirow{-9}{*}{\textbf{60 Min}} & \textbf{TPOT} & 0.620 &  0.885 &  0.137 \\ \hline
 & \textbf{ATM} & 0.665 &  \textbf{0.899} & \textbf{0.121} &  & \textbf{ATM} & 0.768 &  \textbf{0.893} &  \textbf{0.124} \\ \cline{2-5} \cline{7-10} 
 & \textbf{AutoWeka} & 0.747 &  0.839 &  0.166 &  & \textbf{AutoWeka} & 0.771 &  0.838 &  0.166 \\ \cline{2-5} \cline{7-10} 
 & \textbf{Recipe} & 0.516 &  0.748 &  0.254 &  & \textbf{Recipe} & 0.645 &  0.759 &  0.248 \\ \cline{2-5} \cline{7-10} 
 & \textbf{AutoSKLearn-e} & \textbf{0.866} &  0.875 &  0.141 &  & \textbf{AutoSKLearn-e} & 0.874 &  0.883 &  0.132 \\ \cline{2-5} \cline{7-10} 
 & \textbf{AutoSKLearn-m} & 0.859 &  0.868 &  0.152 &  & \textbf{AutoSKLearn-m} & 0.864 &  0.873 &  0.141 \\ \cline{2-5} \cline{7-10} 
 & \textbf{AutoSKLearn-v} & 0.858 &  0.867 &  0.149 &  & \textbf{AutoSKLearn-v} & 0.850 &  0.867 &  0.156 \\ \cline{2-5} \cline{7-10} 
 & \textbf{AutoSKLearn} & 0.862 &  0.871 &  0.148 &  & \textbf{AutoSKLearn} & \textbf{0.875} &  0.884 &  0.132 \\ \cline{2-5} \cline{7-10} 
 & \textbf{SmartML-m} & 0.804 &  0.808 &  0.199 &  & \textbf{SmartML-m} & 0.798 &  0.826 &  0.169 \\ \cline{2-5} \cline{7-10} 
 & \textbf{SmartML-e} & 0.727 &  0.838 &  0.159 &  & \textbf{SmartML-e} & 0.735 &  0.840 &  0.165 \\ \cline{2-5} \cline{7-10} 
\multirow{-9}{*}{\textbf{30 Min}} & \textbf{TPOT} & 0.518 &  0.878 &  0.144 & \multirow{-9}{*}{\textbf{240 Min}} & \textbf{TPOT} & 0.790 &  0.888 &  0.131 \\ \hline \hline

\end{tabular}}
\label{tab:GeneralSummary}
\end{table}

We study the impact of the time budget on the performance of different AutoML frameworks to investigate how quickly the different frameworks can output ML pipelines and whether the different frameworks can guarantee consistent improvement in the performance given more time. We evaluate the performance of the successful runs for each framework bounded by different four-time budgets; 10 minutes, 30 minutes, 60 minutes, and 240 minutes. For each framework, the mean (\texttt{Mean}) and standard deviation (\texttt{SD}) of the performance for all successful runs for each time budget is reported in Table~\ref{tab:GeneralSummary}. \textcolor{red}{Additionally, we report the mean predictive performance weighted by the percentage of successful runs (\texttt{$Mean_{succ}$}).}
\textcolor{red}{\begin{equation}
    Mean_{succ}=Mean \times \frac{N}{T}
\end{equation}}
where $N$ is the number of successful runs and $T$ is the total number of runs.

\begin{table}[!t]
\centering
\caption{Summary of the impact of increasing the time budget. Bold entries highlight highest mean gain, highest maximum gain, smallest mean loss, smallest maximum loss, maximum and minimum number of datasets with gain $>1$ and loss $>$1, respectively.}
\resizebox{0.9\textwidth}{!}
{\begin{tabular}{|c|c|c|c|c|c|c|c|c|}
\hline
\rowcolor[HTML]{EFEFEF}\cellcolor[HTML]{343434}{\color[HTML]{FFFFFF} \textbf{Time Budget}} & & \multicolumn{2}{c}{\textbf{Gain ($g$)}} & \multicolumn{3}{c}{\textbf{\#datasets with}} & \multicolumn{2}{c}{\textbf{Loss ($l$)}} \\
\rowcolor[HTML]{EFEFEF}\cellcolor[HTML]{343434}{\color[HTML]{FFFFFF} in minutes}&  \multirow{-2}{*}{\textbf{Framework}}&  \textbf{Mean} & \textbf{Max} & \textbf{ $g>1\%$} & \textbf{ $g\approx 0 \%$} & \textbf{ $l>1\%$} & \textbf{Mean} & \textbf{Max} \\ \hline
 & \textbf{ATM} &   4.3  &  21.0  &  19  &  33  &  15  &  -4.7  &  \textbf{-21.4} \\ \cline{2-9}  
 & \textbf{AutoWeka} &   5.8  &  20.1  &  16  &  62  &  7  &  -3.8  &  -11.9 \\ \cline{2-9} 
 & \textbf{Recipe} & \textbf{19.3}  &  36.6  &  2  &  17  &  2  &  -4.4  &  -6.2 \\ \cline{2-9} 
 & \textbf{AutoSKLearn-e} &   3.6  &  13.5  &  24  &  67  &  8  &  -2.7  &  -6.8 \\ \cline{2-9} 
 & \textbf{AutoSKLearn-m}&   3.6  &  13.3  &  21  &  65  &  13  &  -2.6  &  -10.7 \\ \cline{2-9}
 & \textbf{AutoSKLearn-v}&   3.9  &  22.1  &  23  &  63  &  13  &  -3.1  &  -7.4 \\ \cline{2-9} 
 & \textbf{AutoSKLearn}&  3.5  &  15.2  &  17  &  72  &  10  &  -2.6  &  -4.8 \\ \cline{2-9} 
 & \textbf{SmartML-m}& 8.0  &  33.3  &  13  &  64  &  11  &  -3.6  &  -8.3 \\ \cline{2-9}
 & \textbf{SmartML-e}& 7.9  &  \textbf{85.2}  &  18  &  67  &  11  &  \textbf{-6.3}  &  -16.9 \\ \cline{2-9}
\multirow{-9}{*}{\textbf{$10 \to 30$}} & \textbf{TPOT}&  6.2  &  17.0  &  7  &  29  &  4  &  -2.2  &  -3.7 \\ \hline
 & \textbf{ATM} &  5.0  &  16.5  &  15  &  34  &  20  &  -6.7  &  -28.6 \\ \cline{2-9} 
 & \textbf{AutoWeka} & 9.1  & \textbf{ 66.6}  &  14  &  61  &  13  &  -9.6  &  -56.7 \\ \cline{2-9} 
 & \textbf{Recipe} & 4.7  &  17.2  &  6  &  58  &  3  &  \textbf{-19.2}  &  -29.1 \\ \cline{2-9} 
 & \textbf{AutoSKLearn-e} &  4.9  &  19.6  &  16  &  66  &  17  &  -2.2  &  -5.7   \\ \cline{2-9}
 & \textbf{AutoSKLearn-m}&  4.9  &  23.4  &  16  &  67  &  16  &  -4.1  &  -13.3 \\ \cline{2-9} 
 & \textbf{AutoSKLearn-v}&   4.1  &  14.2  &  23  &  62  &  14  &  -4.6  &  -13.9 \\ \cline{2-9} 
 & \textbf{AutoSKLearn}& 4.1  &  32.0  &  22  &  65  &  12  &  -2.4  &  -6.8 \\ \cline{2-9}  
 & \textbf{SmartML-m}& \textbf{12.2}  &  40.0  &  10  &  73  &  6  &  -6.2  &  -18.3 \\ \cline{2-9} 
 & \textbf{SmartML-e}& 6.5  &  18.2  &  21  &  54  &  20  &  -9.0  &  \textbf{-84.3} \\ \cline{2-9}
\multirow{-9}{*}{\textbf{$30 \to 60$}} & \textbf{TPOT}&   3.7  &  8.7  &  6  &  42  &  8  &  -2.8  &  -7.7 \\ \hline

& \textbf{ATM} &  5.6  &  31.1  &  21  &  39  &  17  &  -3.4  &  -12.0 \\ \cline{2-9} 
& \textbf{AutoWeka} &  4.1  &  8.7  &  17  &  61  &  8  &  -3.8  &  -11.5 \\ \cline{2-9} 
& \textbf{Recipe} &  13.5  &  38.2  &  4  &  69  &  2  &  \textbf{-20.5}  &  \textbf{-40.0} \\ \cline{2-9} 
& \textbf{AutoSKLearn-e} &  4.3  &  39.0  &  21  &  62  &  16  &  -3.8  &  -12.7 \\ \cline{2-9} 
& \textbf{AutoSKLearn-m}&  4.0  &  13.3  &  20  &  59  &  20  &  -2.7  &  -6.0   \\ \cline{2-9}  
& \textbf{AutoSKLearn-v}&  3.6  &  12.5  &  22  &  63  &  13  &  -8.7  &  -25.3 \\ \cline{2-9}  
& \textbf{AutoSKLearn}&  4.8  &  36.5  &  22  &  63  &  14  &  -3.2  &  -9.9 \\ \cline{2-9} 
& \textbf{SmartML-m}&  \textbf{10.6}  &  \textbf{59.6}  &  19  &  59  &  10  &  -6.6  &  -19.4 \\ \cline{2-9} 
& \textbf{SmartML-e}& 9.1  &  22.3  &  23  &  53  &  19  &  -6.9  &  -18.8 \\ \cline{2-9}
\multirow{-9}{*}{\textbf{$60 \to 240 $}} & \textbf{TPOT}&   2.6  &  5.6  &  18  &  47  &  5  &  -4.1  &  -7.7 \\ \hline

\end{tabular}}
\label{tab:TBSummary}
\end{table}

\textcolor{red}{The results show that for the 10 and 240 minutes budgets, \texttt{AutoSKlearn} and \texttt{AutoSKlearn-e} have comparable \texttt{$Mean_{succ}$}, while \texttt{AutoSKlearn-e} has the highest \texttt{$Mean_{succ}$} over the rest of time budgets. In contrast, \texttt{Recipe} achieves the lowest mean performance and \texttt{Mean$_{succ}$} over all time budgets, as shown in Table~\ref{tab:GeneralSummary}. Notably, the \texttt{$Mean_{succ}$} improves over time for \texttt{Recipe} and \texttt{TPOT}.} For the 10 minutes budget, \texttt{TPOT} and \texttt{ATM} achieve a comparable \textcolor{high} mean performance and \textcolor{red}{low} standard deviation. For the rest of the time budgets, \texttt{ATM} achieves the highest mean performance and lowest standard deviation across all successful runs, as shown in Table~\ref{tab:GeneralSummary}.  ~\Cref{FIG:TimeBudgetSklearn-v,FIG:TimeBudgetSklearn-m,FIG:TimeBudgetSklearn-e,FIG:TimeBudgetSklearn,FIG:TimeBudgetTPOT,FIG:TimeBudgetATM,FIG:TimeBudgetSmartML,FIG:TimeBudgetSmartML-e,FIG:TimeBudgetAutoWeka,FIG:TimeBudgetRecipe} in Appendix~\ref{App:ImpactTimeBudget} show the impact of increasing the time budget for each AutoML framework on 100 datasets.

Table~\ref{tab:TBSummary} reports the gain $(g)$ or loss $(l)$ in the predictive performance of the frameworks when increasing the time budget from 10 to 30 minutes, from 30 to 60 minutes and from 60 to 240 minutes. The gain is measured by the mean predictive performance improvement over all improved datasets and the maximum predictive performance improvement achieved per framework. Similarly, the loss is measured as the mean predictive performance loss over all declined datasets and the maximum predictive performance loss over all declined datasets. When increasing the time budget from 10 to 30 minutes, \texttt{Recipe} achieves the highest mean gain of $19.3$ on 2 datasets, followed by \texttt{SmartML-m}, while \texttt{AutoSKlearn} comes in the last place achieving a mean gain of $3.5$ on 17 datasets, as shown in Table~\ref{tab:TBSummary}. \texttt{SmartML-e} has the highest maximum performance gain and the smallest mean loss when increasing the time budget from 10 to 30 minutes.\texttt{SmartML-m} achieves the highest mean predictive performance gain when increasing the time budget from 30 to 60 minutes and from 60 to 240 minutes. \texttt{AutoWeka} and \texttt{SmartML-m} achieve the highest maximum performance gain when increasing the time budget from 30 to 60 minutes and from 60 to 240 minutes, respectively. \texttt{Recipe} has the smallest mean performance loss when increasing the time budget from 30 to 60 minutes and from 60 to 240 minutes, as shown in Table~\ref{tab:TBSummary}. It is noticeable that \texttt{Recipe} has the smallest number of datasets that witnessed performance improvement and performance degradation when increasing the time budget. \texttt{AutoSKlearn-v} have the largest number of datasets that witnessed performance improvement when increasing the time budget from 30 to 60 minutes and from 60 to 240 minutes, while \texttt{AutoSKlearn-e} witnessed performance improvement across the largest number of datasets when increasing the time budget from 10 to 30 minutes. In contrast, \texttt{ATM} has the most significant number of datasets with performance degradation when increasing the time budget from 10 to 30 minutes and from 30 to 60 minutes.

\begin{table}[!t]
\centering
\caption{Wilcoxon test p-values for all the AutoML frameworks over different time budgets. Bold entries highlight significant difference.}
\resizebox{\textwidth}{!}{\begin{tabular}{|c|c|c|c|c|ccccc}
\hline
\rowcolor[HTML]{656565} 
\cellcolor[HTML]{343434}{\color[HTML]{FFFFFF} \textbf{Framework}} & {\color[HTML]{FFFFFF} \textbf{Time Budget 1}} & {\color[HTML]{FFFFFF} \textbf{Time Budget 2}} & {\color[HTML]{FFFFFF} \textbf{Avg. Acc. Diff}} & {\color[HTML]{FFFFFF} \textbf{$P$ value}} & \multicolumn{1}{c|}{\cellcolor[HTML]{333333}{\color[HTML]{FFFFFF} \textbf{Framework}}} & \multicolumn{1}{c|}{\cellcolor[HTML]{656565}{\color[HTML]{FFFFFF} \textbf{Time Budget 1}}} & \multicolumn{1}{c|}{\cellcolor[HTML]{656565}{\color[HTML]{FFFFFF} \textbf{Time Budget 2}}} & \multicolumn{1}{c|}{\cellcolor[HTML]{656565}{\color[HTML]{FFFFFF} \textbf{Avg. Acc. Diff}}} & \multicolumn{1}{c|}{\cellcolor[HTML]{656565}{\color[HTML]{FFFFFF} \textbf{$P$ value}}} \\ \cline{2-5}\cline{7-10}
\cellcolor[HTML]{EFEFEF} & \cellcolor[HTML]{EFEFEF}30 & \cellcolor[HTML]{EFEFEF}10 & \cellcolor[HTML]{EFEFEF}0.008 & \cellcolor[HTML]{EFEFEF}\textbf{0.016} & \multicolumn{1}{c|}{} & \multicolumn{1}{c|}{30} & \multicolumn{1}{c|}{10} & \multicolumn{1}{c|}{0.003} & \multicolumn{1}{c|}{0.338} \\ \cline{2-5}\cline{7-10}
\cellcolor[HTML]{EFEFEF} & \cellcolor[HTML]{EFEFEF}60 & \cellcolor[HTML]{EFEFEF}10 & \cellcolor[HTML]{EFEFEF}0.003 & \cellcolor[HTML]{EFEFEF}\textbf{0.034} & \multicolumn{1}{c|}{} & \multicolumn{1}{c|}{60} & \multicolumn{1}{c|}{10} & \multicolumn{1}{c|}{0.010} & \multicolumn{1}{c|}{\textbf{0.001}} \\ \cline{2-5}\cline{7-10}
\cellcolor[HTML]{EFEFEF} & \cellcolor[HTML]{EFEFEF}60 & \cellcolor[HTML]{EFEFEF}30 & \cellcolor[HTML]{EFEFEF}0.000 & \cellcolor[HTML]{EFEFEF}0.885 & \multicolumn{1}{c|}{} & \multicolumn{1}{c|}{60} & \multicolumn{1}{c|}{30} & \multicolumn{1}{c|}{0.007} & \multicolumn{1}{c|}{\textbf{0.034}} \\ \cline{2-5}\cline{7-10}
\cellcolor[HTML]{EFEFEF} & \cellcolor[HTML]{EFEFEF}240 & \cellcolor[HTML]{EFEFEF}10 & \cellcolor[HTML]{EFEFEF}0.009 & \cellcolor[HTML]{EFEFEF}\textbf{0.000} & \multicolumn{1}{c|}{} & \multicolumn{1}{c|}{240} & \multicolumn{1}{c|}{10} & \multicolumn{1}{c|}{0.016} & \multicolumn{1}{c|}{\textbf{0.004}} \\ \cline{2-5}\cline{7-10}
\cellcolor[HTML]{EFEFEF} & \cellcolor[HTML]{EFEFEF}240 & \cellcolor[HTML]{EFEFEF}30 & \cellcolor[HTML]{EFEFEF}0.005 & \cellcolor[HTML]{EFEFEF}\textbf{0.039} & \multicolumn{1}{c|}{} & \multicolumn{1}{c|}{240} & \multicolumn{1}{c|}{30} & \multicolumn{1}{c|}{0.013} & \multicolumn{1}{c|}{\textbf{0.018}} \\ \cline{2-5}\cline{7-10}
\multirow{-6}{*}{\cellcolor[HTML]{EFEFEF}\textbf{AutoWeka}} & \cellcolor[HTML]{EFEFEF}240 & \cellcolor[HTML]{EFEFEF}60 & \cellcolor[HTML]{EFEFEF}0.005 & \cellcolor[HTML]{EFEFEF}\textbf{0.042} & \multicolumn{1}{c|}{\multirow{-6}{*}{\textbf{AutoSKLearn}}} & \multicolumn{1}{c|}{240} & \multicolumn{1}{c|}{60} & \multicolumn{1}{c|}{0.006} & \multicolumn{1}{c|}{0.129} \\ \hline
  & 30 & 10 & 0.009 & 0.117 & \multicolumn{1}{c|}{\cellcolor[HTML]{EFEFEF}} & \multicolumn{1}{c|}{\cellcolor[HTML]{EFEFEF}30} & \multicolumn{1}{c|}{\cellcolor[HTML]{EFEFEF}10} & \multicolumn{1}{c|}{\cellcolor[HTML]{EFEFEF}0.005} & \multicolumn{1}{c|}{\cellcolor[HTML]{EFEFEF}0.175} \\ \cline{2-5}\cline{7-10}
  & 60 & 10 & 0.008 & 0.388 & \multicolumn{1}{c|}{\cellcolor[HTML]{EFEFEF}} & \multicolumn{1}{c|}{\cellcolor[HTML]{EFEFEF}60} & \multicolumn{1}{c|}{\cellcolor[HTML]{EFEFEF}10} & \multicolumn{1}{c|}{\cellcolor[HTML]{EFEFEF}0.008} & \multicolumn{1}{c|}{\cellcolor[HTML]{EFEFEF}\textbf{0.001}} \\ \cline{2-5}\cline{7-10}
  & 60 & 30 & 0.001 & 0.428 & \multicolumn{1}{c|}{\cellcolor[HTML]{EFEFEF}} & \multicolumn{1}{c|}{\cellcolor[HTML]{EFEFEF}60} & \multicolumn{1}{c|}{\cellcolor[HTML]{EFEFEF}30} & \multicolumn{1}{c|}{\cellcolor[HTML]{EFEFEF}0.003} & \multicolumn{1}{c|}{\cellcolor[HTML]{EFEFEF}0.088} \\ \cline{2-5}\cline{7-10}
  & 240 & 10 & 0.013 & \textbf{0.016} & \multicolumn{1}{c|}{\cellcolor[HTML]{EFEFEF}} & \multicolumn{1}{c|}{\cellcolor[HTML]{EFEFEF}240} & \multicolumn{1}{c|}{\cellcolor[HTML]{EFEFEF}10} & \multicolumn{1}{c|}{\cellcolor[HTML]{EFEFEF}0.005} & \multicolumn{1}{c|}{\cellcolor[HTML]{EFEFEF}\textbf{0.000}} \\ \cline{2-5}\cline{7-10}
  & 240 & 30 & 0.006 & \textbf{0.035} & \multicolumn{1}{c|}{\cellcolor[HTML]{EFEFEF}} & \multicolumn{1}{c|}{\cellcolor[HTML]{EFEFEF}240} & \multicolumn{1}{c|}{\cellcolor[HTML]{EFEFEF}30} & \multicolumn{1}{c|}{\cellcolor[HTML]{EFEFEF}0.000} & \multicolumn{1}{c|}{\cellcolor[HTML]{EFEFEF}\textbf{0.040}} \\ \cline{2-5}\cline{7-10}
\multirow{-6}{*}{\textbf{TPOT}} & 240 & 60 & 0.004 & \textbf{0.008} & \multicolumn{1}{c|}{\multirow{-6}{*}{\cellcolor[HTML]{EFEFEF}\textbf{AutoSKLearn-v}}} & \multicolumn{1}{c|}{\cellcolor[HTML]{EFEFEF}240} & \multicolumn{1}{c|}{\cellcolor[HTML]{EFEFEF}60} & \multicolumn{1}{c|}{\cellcolor[HTML]{EFEFEF}-0.003} & \multicolumn{1}{c|}{\cellcolor[HTML]{EFEFEF}0.099} \\ \hline
\cellcolor[HTML]{EFEFEF} & \cellcolor[HTML]{EFEFEF}30 & \cellcolor[HTML]{EFEFEF}10 & \cellcolor[HTML]{EFEFEF}0.014 & \cellcolor[HTML]{EFEFEF}0.866 & \multicolumn{1}{c|}{} & \multicolumn{1}{c|}{30} & \multicolumn{1}{c|}{10} & \multicolumn{1}{c|}{0.008} & \multicolumn{1}{c|}{\textbf{0.000}} \\ \cline{2-5}\cline{7-10}
\cellcolor[HTML]{EFEFEF} & \cellcolor[HTML]{EFEFEF}60 & \cellcolor[HTML]{EFEFEF}10 & \cellcolor[HTML]{EFEFEF}-0.002 & \cellcolor[HTML]{EFEFEF}0.955 & \multicolumn{1}{c|}{} & \multicolumn{1}{c|}{60} & \multicolumn{1}{c|}{10} & \multicolumn{1}{c|}{0.012} & \multicolumn{1}{c|}{\textbf{0.000}} \\ \cline{2-5}\cline{7-10}
\cellcolor[HTML]{EFEFEF} & \cellcolor[HTML]{EFEFEF}60 & \cellcolor[HTML]{EFEFEF}30 & \cellcolor[HTML]{EFEFEF}-0.004 & \cellcolor[HTML]{EFEFEF}0.535 & \multicolumn{1}{c|}{} & \multicolumn{1}{c|}{60} & \multicolumn{1}{c|}{30} & \multicolumn{1}{c|}{0.004} & \multicolumn{1}{c|}{0.904} \\ \cline{2-5}\cline{7-10}
\cellcolor[HTML]{EFEFEF} & \cellcolor[HTML]{EFEFEF}240 & \cellcolor[HTML]{EFEFEF}10 & \cellcolor[HTML]{EFEFEF}0.023 & \cellcolor[HTML]{EFEFEF}0.093 & \multicolumn{1}{c|}{} & \multicolumn{1}{c|}{240} & \multicolumn{1}{c|}{10} & \multicolumn{1}{c|}{0.015} & \multicolumn{1}{c|}{\textbf{0.000}} \\ \cline{2-5}\cline{7-10}
\cellcolor[HTML]{EFEFEF} & \cellcolor[HTML]{EFEFEF}240 & \cellcolor[HTML]{EFEFEF}30 & \cellcolor[HTML]{EFEFEF}0.003 & \cellcolor[HTML]{EFEFEF}0.067 & \multicolumn{1}{c|}{} & \multicolumn{1}{c|}{240} & \multicolumn{1}{c|}{30} & \multicolumn{1}{c|}{0.007} & \multicolumn{1}{c|}{\textbf{0.038}} \\ \cline{2-5}\cline{7-10}
\multirow{-6}{*}{\cellcolor[HTML]{EFEFEF}\textbf{Recipe}} & \cellcolor[HTML]{EFEFEF}240 & \cellcolor[HTML]{EFEFEF}60 & \cellcolor[HTML]{EFEFEF}0.002 & \cellcolor[HTML]{EFEFEF}0.345 & \multicolumn{1}{c|}{\multirow{-6}{*}{\textbf{AutoSKLearn-e}}} & \multicolumn{1}{c|}{240} & \multicolumn{1}{c|}{60} & \multicolumn{1}{c|}{0.003} & \multicolumn{1}{c|}{0.291} \\ \hline
  & 30 & 10 & 0.001 & 0.583 & \multicolumn{1}{c|}{\cellcolor[HTML]{EFEFEF}} & \multicolumn{1}{c|}{\cellcolor[HTML]{EFEFEF}30} & \multicolumn{1}{c|}{\cellcolor[HTML]{EFEFEF}10} & \multicolumn{1}{c|}{\cellcolor[HTML]{EFEFEF}0.004} & \multicolumn{1}{c|}{\cellcolor[HTML]{EFEFEF}0.156} \\ \cline{2-5}\cline{7-10}
  & 60 & 10 & -0.007 & 0.254 & \multicolumn{1}{c|}{\cellcolor[HTML]{EFEFEF}} & \multicolumn{1}{c|}{\cellcolor[HTML]{EFEFEF}60} & \multicolumn{1}{c|}{\cellcolor[HTML]{EFEFEF}10} & \multicolumn{1}{c|}{\cellcolor[HTML]{EFEFEF}0.006} & \multicolumn{1}{c|}{\cellcolor[HTML]{EFEFEF}0.105} \\ \cline{2-5}\cline{7-10}
  & 60 & 30 & -0.008 & 0.499 & \multicolumn{1}{c|}{\cellcolor[HTML]{EFEFEF}} & \multicolumn{1}{c|}{\cellcolor[HTML]{EFEFEF}60} & \multicolumn{1}{c|}{\cellcolor[HTML]{EFEFEF}30} & \multicolumn{1}{c|}{\cellcolor[HTML]{EFEFEF}0.002} & \multicolumn{1}{c|}{\cellcolor[HTML]{EFEFEF}0.873} \\ \cline{2-5}\cline{7-10}
  & 240 & 10 & 0.003 & 0.585 & \multicolumn{1}{c|}{\cellcolor[HTML]{EFEFEF}} & \multicolumn{1}{c|}{\cellcolor[HTML]{EFEFEF}240} & \multicolumn{1}{c|}{\cellcolor[HTML]{EFEFEF}10} & \multicolumn{1}{c|}{\cellcolor[HTML]{EFEFEF}0.009} & \multicolumn{1}{c|}{\cellcolor[HTML]{EFEFEF}0.210} \\ \cline{2-5}\cline{7-10}
  & 240 & 30 & -0.001 & 0.799 & \multicolumn{1}{c|}{\cellcolor[HTML]{EFEFEF}} & \multicolumn{1}{c|}{\cellcolor[HTML]{EFEFEF}240} & \multicolumn{1}{c|}{\cellcolor[HTML]{EFEFEF}30} & \multicolumn{1}{c|}{\cellcolor[HTML]{EFEFEF}0.004} & \multicolumn{1}{c|}{\cellcolor[HTML]{EFEFEF}0.920} \\ \cline{2-5}\cline{7-10}
\multirow{-6}{*}{\textbf{ATM}} & 240 & 60 & 0.008 & 0.394 & \multicolumn{1}{c|}{\multirow{-6}{*}{\cellcolor[HTML]{EFEFEF}\textbf{AutoSKLearn-m}}} & \multicolumn{1}{c|}{\cellcolor[HTML]{EFEFEF}240} & \multicolumn{1}{c|}{\cellcolor[HTML]{EFEFEF}60} & \multicolumn{1}{c|}{\cellcolor[HTML]{EFEFEF}0.003} & \multicolumn{1}{c|}{\cellcolor[HTML]{EFEFEF}0.660} \\ \hline
 \cellcolor[HTML]{EFEFEF} & \cellcolor[HTML]{EFEFEF}30 & \cellcolor[HTML]{EFEFEF}10 & \cellcolor[HTML]{EFEFEF}0.007 & \cellcolor[HTML]{EFEFEF}0.636 & \multicolumn{1}{c|}{} & \multicolumn{1}{c|}{30} & \multicolumn{1}{c|}{10} & \multicolumn{1}{c|}{0.008} & \multicolumn{1}{c|}{0.521} \\ \cline{2-5}\cline{7-10}
 \cellcolor[HTML]{EFEFEF} & \cellcolor[HTML]{EFEFEF}60 & \cellcolor[HTML]{EFEFEF}10 & \cellcolor[HTML]{EFEFEF}0.009 & \cellcolor[HTML]{EFEFEF}0.832 & \multicolumn{1}{c|}{} & \multicolumn{1}{c|}{60} & \multicolumn{1}{c|}{10} & \multicolumn{1}{c|}{0.003} & \multicolumn{1}{c|}{0.589} \\ \cline{2-5}\cline{7-10}
 \cellcolor[HTML]{EFEFEF} & \cellcolor[HTML]{EFEFEF}60 & \cellcolor[HTML]{EFEFEF}30 & \cellcolor[HTML]{EFEFEF}0.009 & \cellcolor[HTML]{EFEFEF}0.597 & \multicolumn{1}{c|}{} & \multicolumn{1}{c|}{60} & \multicolumn{1}{c|}{30} & \multicolumn{1}{c|}{-0.004} & \multicolumn{1}{c|}{0.672} \\ \cline{2-5}\cline{7-10}
 \cellcolor[HTML]{EFEFEF} & \cellcolor[HTML]{EFEFEF}240 & \cellcolor[HTML]{EFEFEF}10 & \cellcolor[HTML]{EFEFEF}0.026 & \cellcolor[HTML]{EFEFEF}0.121 & \multicolumn{1}{c|}{} & \multicolumn{1}{c|}{240} & \multicolumn{1}{c|}{10} & \multicolumn{1}{c|}{0.011} & \multicolumn{1}{c|}{0.092} \\ \cline{2-5}\cline{7-10}
 \cellcolor[HTML]{EFEFEF} & \cellcolor[HTML]{EFEFEF}240 & \cellcolor[HTML]{EFEFEF}30 & \cellcolor[HTML]{EFEFEF}0.025 & \cellcolor[HTML]{EFEFEF}\textbf{0.050} & \multicolumn{1}{c|}{} & \multicolumn{1}{c|}{240} & \multicolumn{1}{c|}{30} & \multicolumn{1}{c|}{0.004} & \multicolumn{1}{c|}{0.182} \\ \cline{2-5}\cline{7-10}
\multirow{-6}{*}{\cellcolor[HTML]{EFEFEF}\textbf{SmartML-m}} & \cellcolor[HTML]{EFEFEF}240 & \cellcolor[HTML]{EFEFEF}60 & \cellcolor[HTML]{EFEFEF}0.015 & \cellcolor[HTML]{EFEFEF}0.071 & \multicolumn{1}{c|}{\multirow{-6}{*}{\textbf{SmartML-e}}} & \multicolumn{1}{c|}{240} & \multicolumn{1}{c|}{60} & \multicolumn{1}{c|}{0.008} & \multicolumn{1}{c|}{0.305} \\ \hline
\end{tabular}}

\label{tab:wilcoxon_timebudget}
\end{table}

The Wilcoxon signed-rank test was conducted to determine if a statistically significant difference in terms of the average predictive performance between the AutoML frameworks exists when increasing the time budget, the results of which are summarized in Table~~\ref{tab:wilcoxon_timebudget}. The average performance variations across the specified time budgets are not significantly high. Table~\ref{tab:wilcoxon_timebudget} shows that the significance of the impact of increasing the time budget varies from one framework to another. For example, the results show that for \texttt{AutoSKlearn-m}, 
\texttt{Recipe}, \texttt{ATM}, \texttt{SmartML-m} and \texttt{SmartML-e}, increasing the time budgets do not lead to significant performance impact, while such significant performance impact is achieved in most of the cases for the \texttt{AutoWeka}, \texttt{TPOT} and all versions of \texttt{AutoSKlearn} except \texttt{AutoSKlearn-m}. These results show that end-users should always carefully consider the trade-off between time budget and performance for the benchmark frameworks based on their specific goals.

\begin{figure*}[!htbp]
\centering \subfigure[AutoWeka] {
    \label{FIG:HeatMapWeka}
    \includegraphics[width=0.67\textwidth]{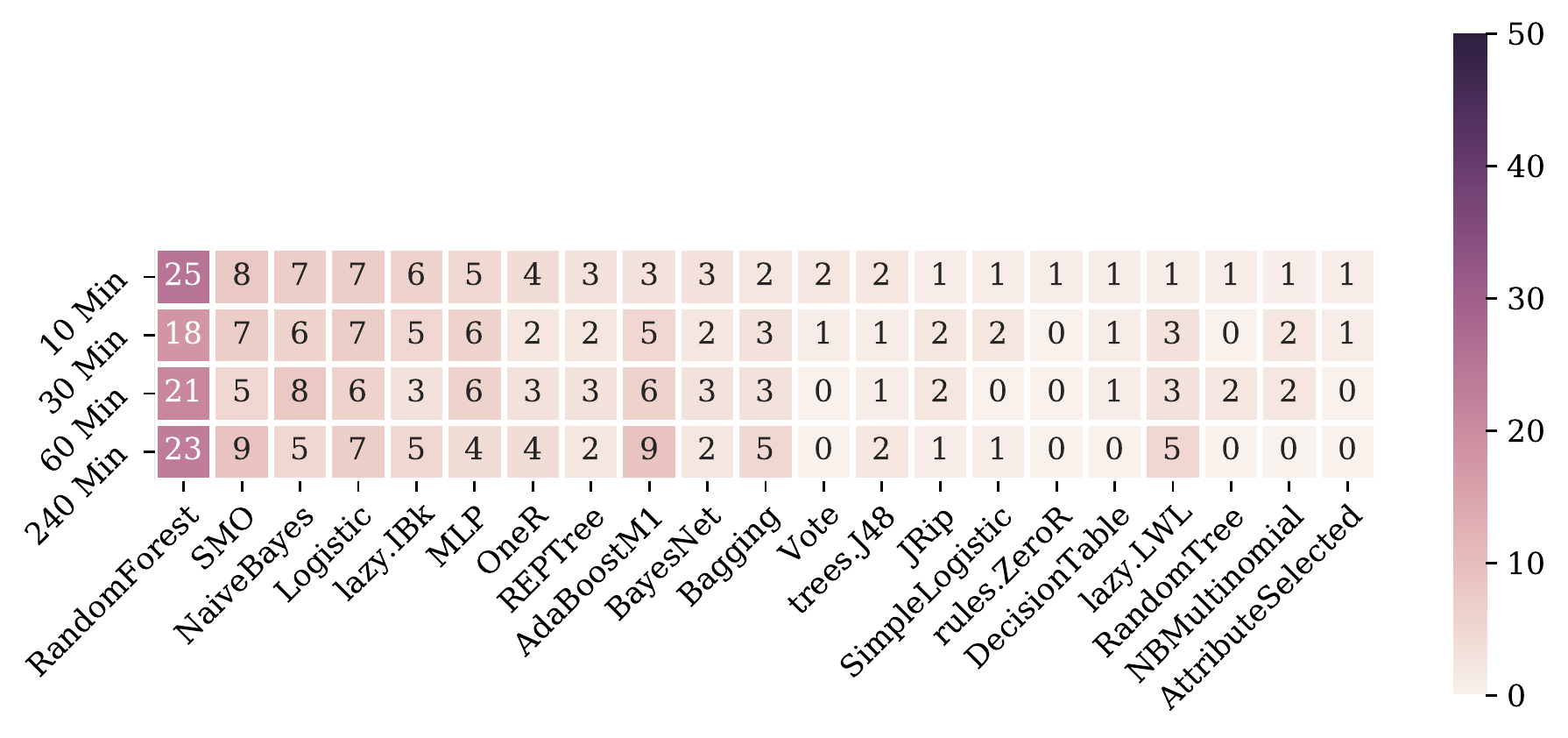}
}
\centering \subfigure[ATM] {
    \label{FIG:HeatMapATM}
    \includegraphics[width=0.27\textwidth]{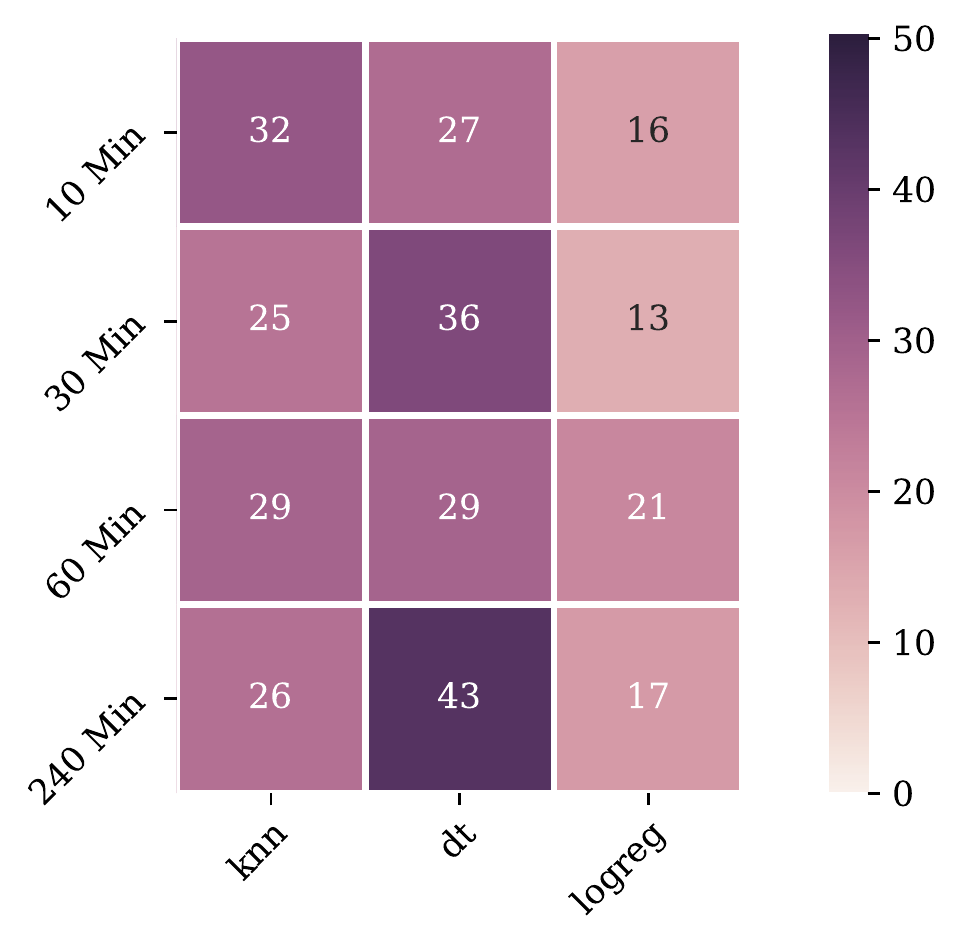}
}
\centering \subfigure[AutoSKlearn-v] {
    \label{FIG:HeatMapAutoSKlearn-v}
    \includegraphics[width=0.47\textwidth]{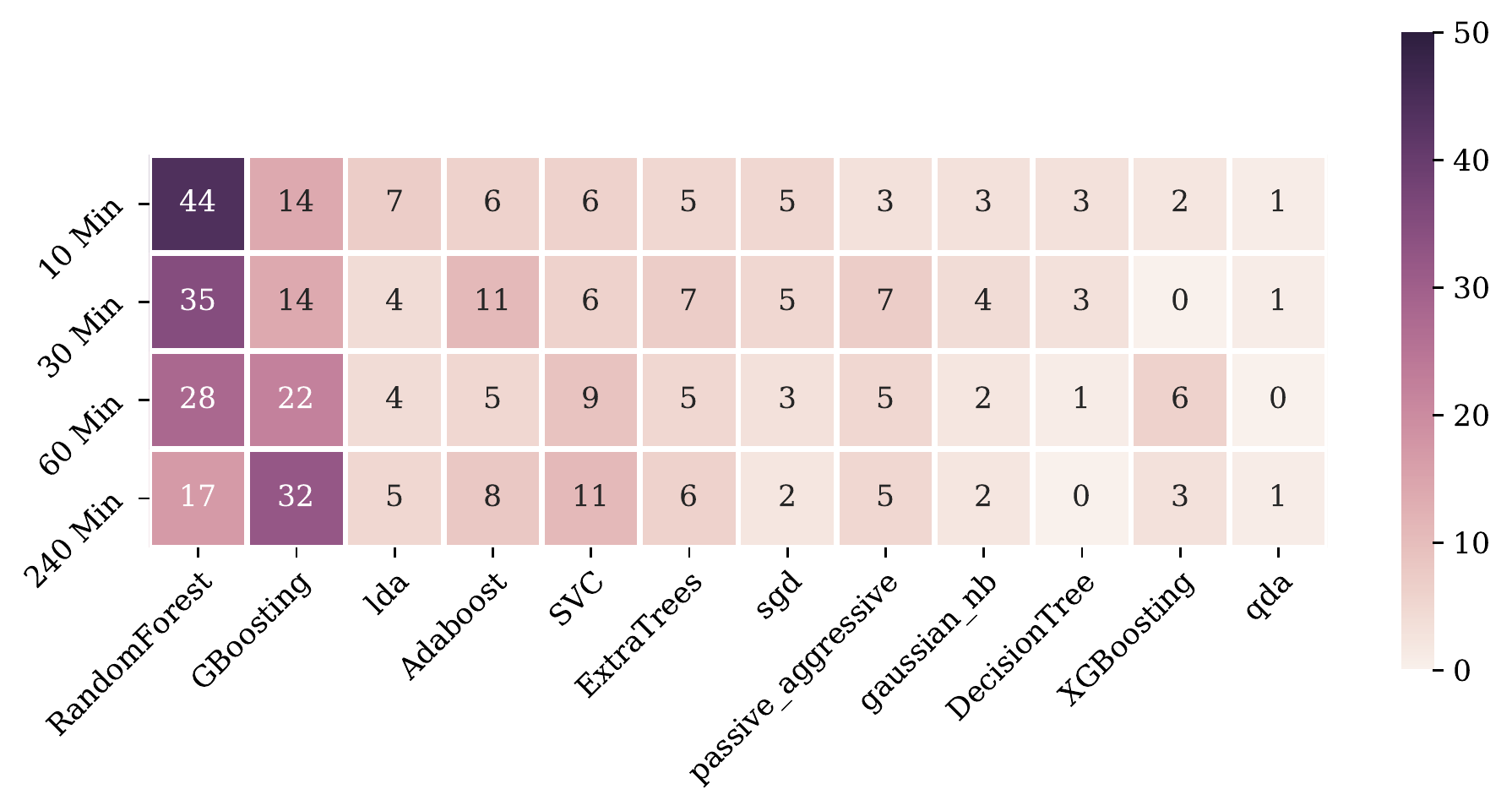}
}
\centering \subfigure[AutoSKlearn-m] {
    \label{FIG:HeatMapAutoSKlearn-m}
    \includegraphics[width=0.47\textwidth]{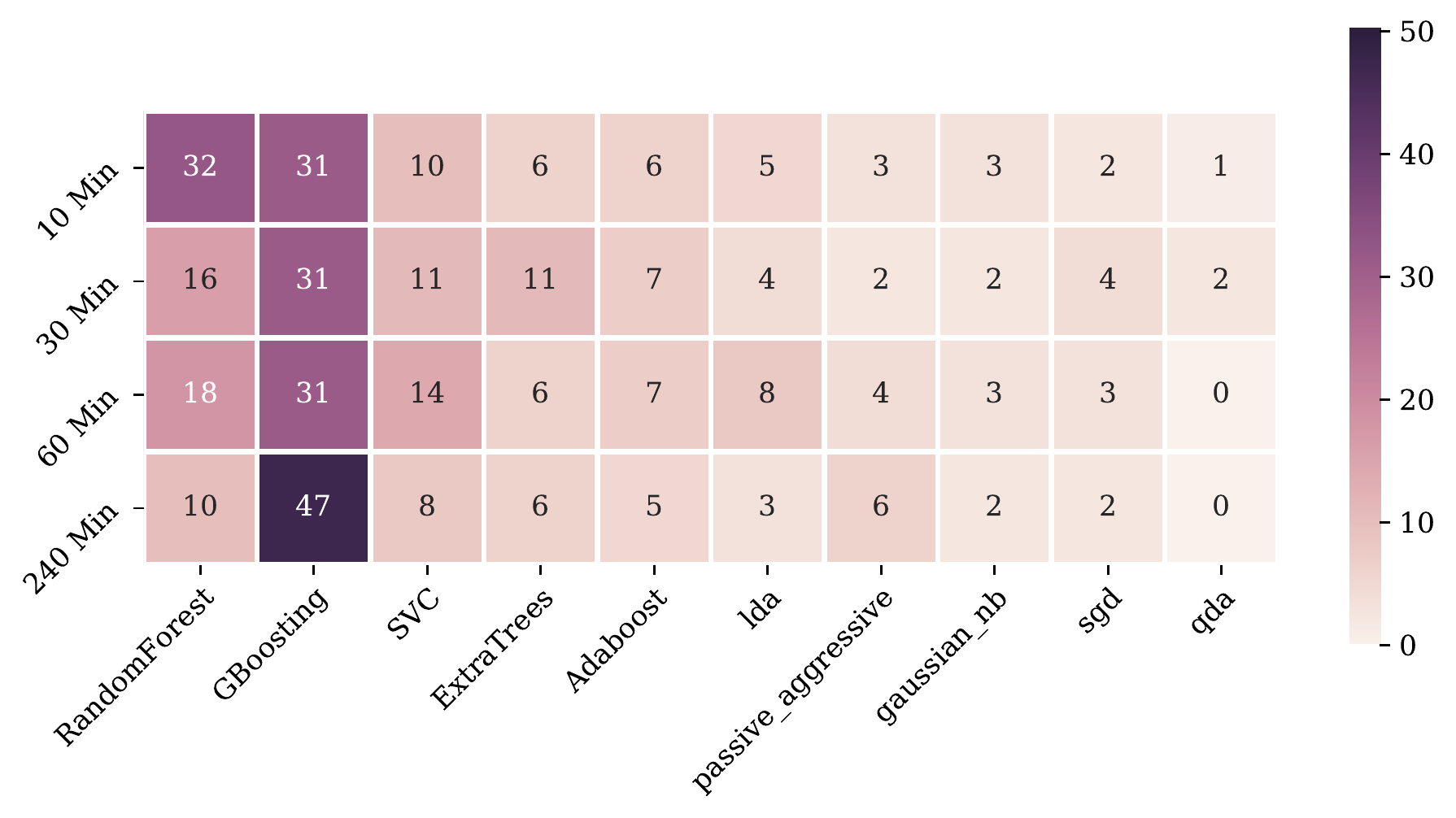}
}
\centering \subfigure[SmartML-m] {
    \label{FIG:HeatMapSmartML}
    \includegraphics[width=0.54\textwidth]{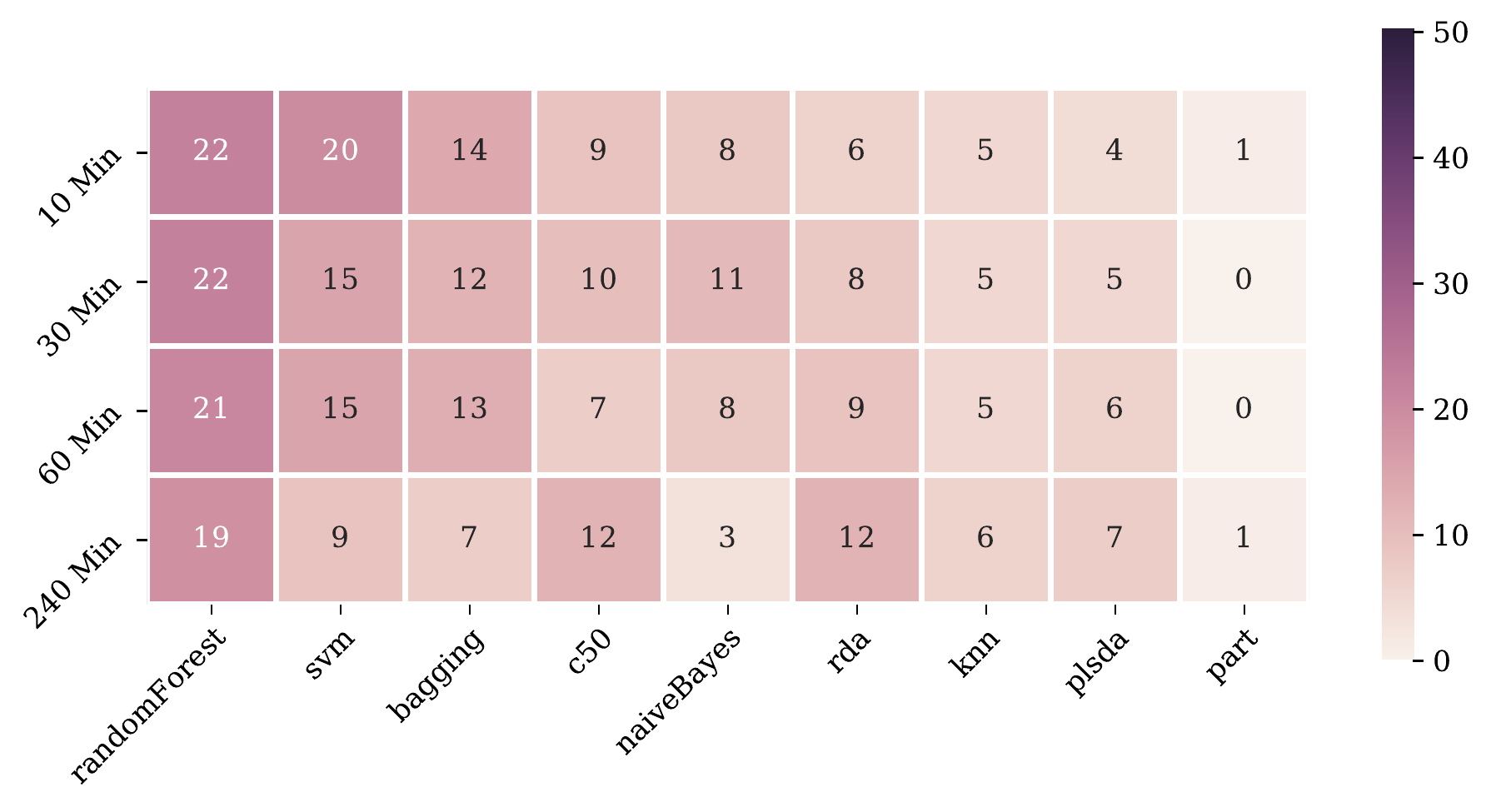}
}
\centering \subfigure[TPOT] {
    \label{FIG:HeatMapTPOT}
    \includegraphics[width=0.40\textwidth]{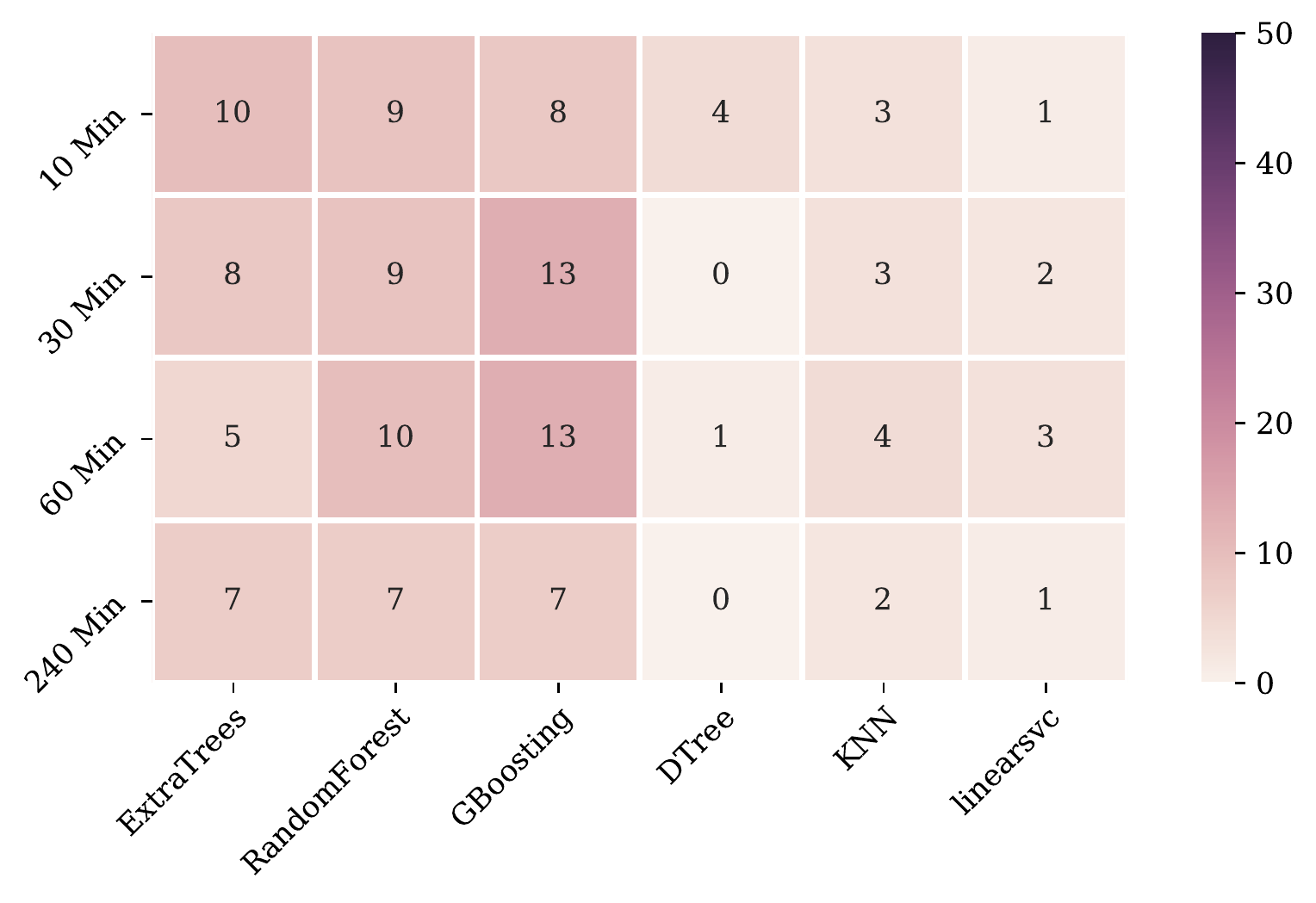}
}

\centering \subfigure[Recipe] {
    \label{FIG:HeatMapRecipe}
    \includegraphics[width=0.70\textwidth]{Figures/Classifiers'_Heat_map_(Recipe).pdf}
}



\caption{The frequency of using different machine learning models by the different AutoML frameworks.}
\label{FIG:FrequentClassifier}
\end{figure*}

\subsubsection{Impact of the Size of Search Space}
\label{SEC:DDSearchSpace}
We study the impact of limiting the search space on the performance across different time budgets. In practice, search space defines the structural paradigm that the different optimization methods can explore; thus, designing a good search space is a vital but challenging problem. Figure~\ref{FIG:FrequentClassifier}, provides an overview on the most frequent ML models commonly used by the different AutoML frameworks. By analyzing the returned best-performing models, it is notable that there is no single ML model that all AutoML frameworks have frequently used; however, it is apparent the tree-based models are the most frequent across all frameworks, except \texttt{Recipe} for all time budgets. For example, the returned pipelines by \texttt{AutoWeka}, \texttt{AutoSKlearn-v}, and \texttt{SmartML-m} show that \emph{random forest} is the most frequently used classifier, as shown in Figures~\ref{FIG:HeatMapWeka}, ~\ref{FIG:HeatMapAutoSKlearn-v}, and ~\ref{FIG:HeatMapSmartML}. The most frequent classifier for \texttt{AutoSKlearn-m}, \texttt{TPOT}, and \texttt{Recipe} is \emph{gradient boosting}, as shown in Figures~\ref{FIG:HeatMapAutoSKlearn-m}, ~\ref{FIG:HeatMapTPOT}, and \ref{FIG:HeatMapRecipe}, respectively. To efficiently utilize the time budget, \texttt{ATM} limits its default search space to only three classifiers, namely, \emph{k-nearest neighbours}, \emph{decision tree}, and \emph{logistic regression}, while \emph{decision tree} is the most frequently used one, as shown in Figure~\ref{FIG:HeatMapATM}.

\begin{figure*}[!t]
\centering \subfigure[AutoSKlearn] {
    \label{FIG:SearchSpaceSklearn}
    \includegraphics[width=0.47\textwidth]{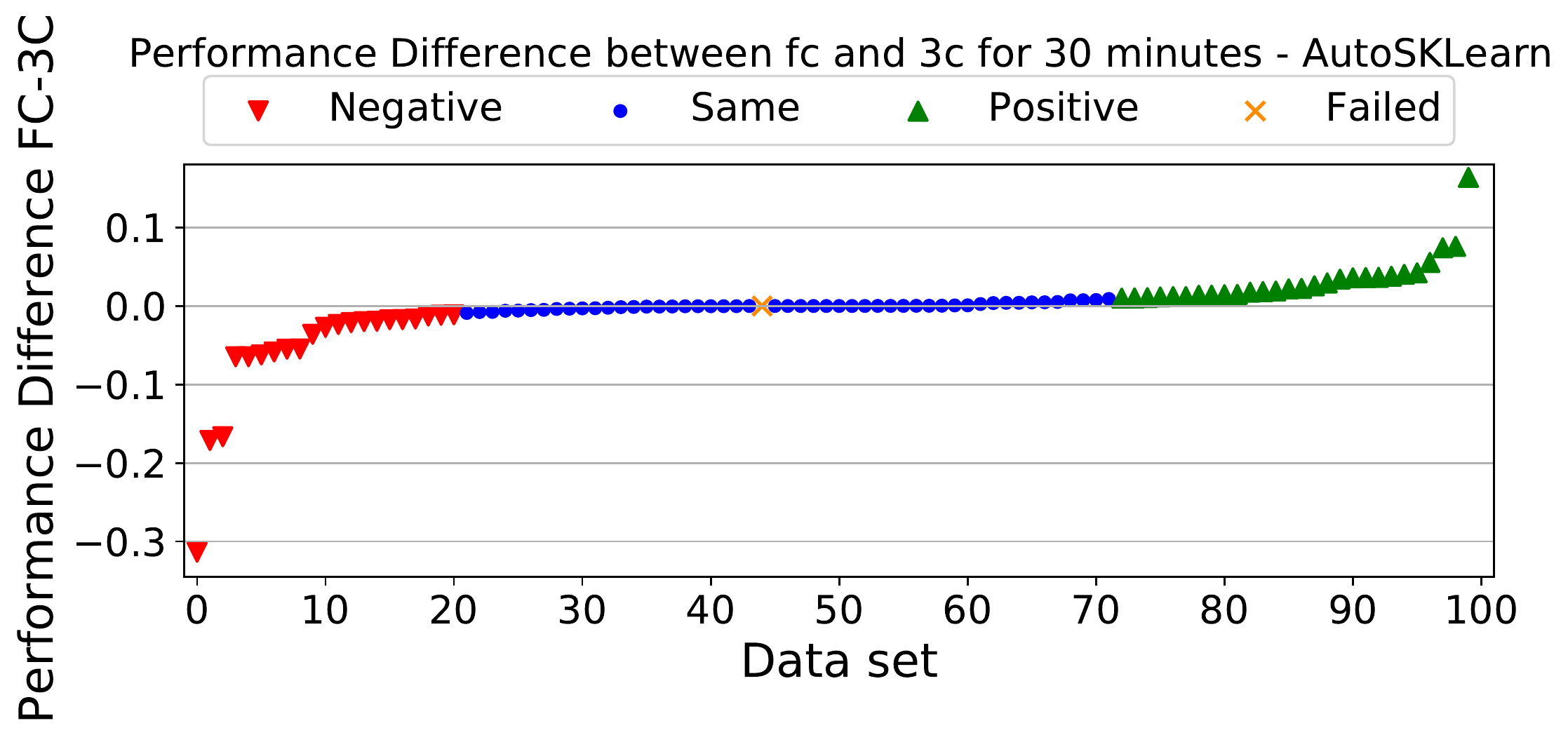}
}
\centering \subfigure[TPOT] {
    \label{FIG:SearchSpaceTPOT}
    \includegraphics[width=0.47\textwidth]{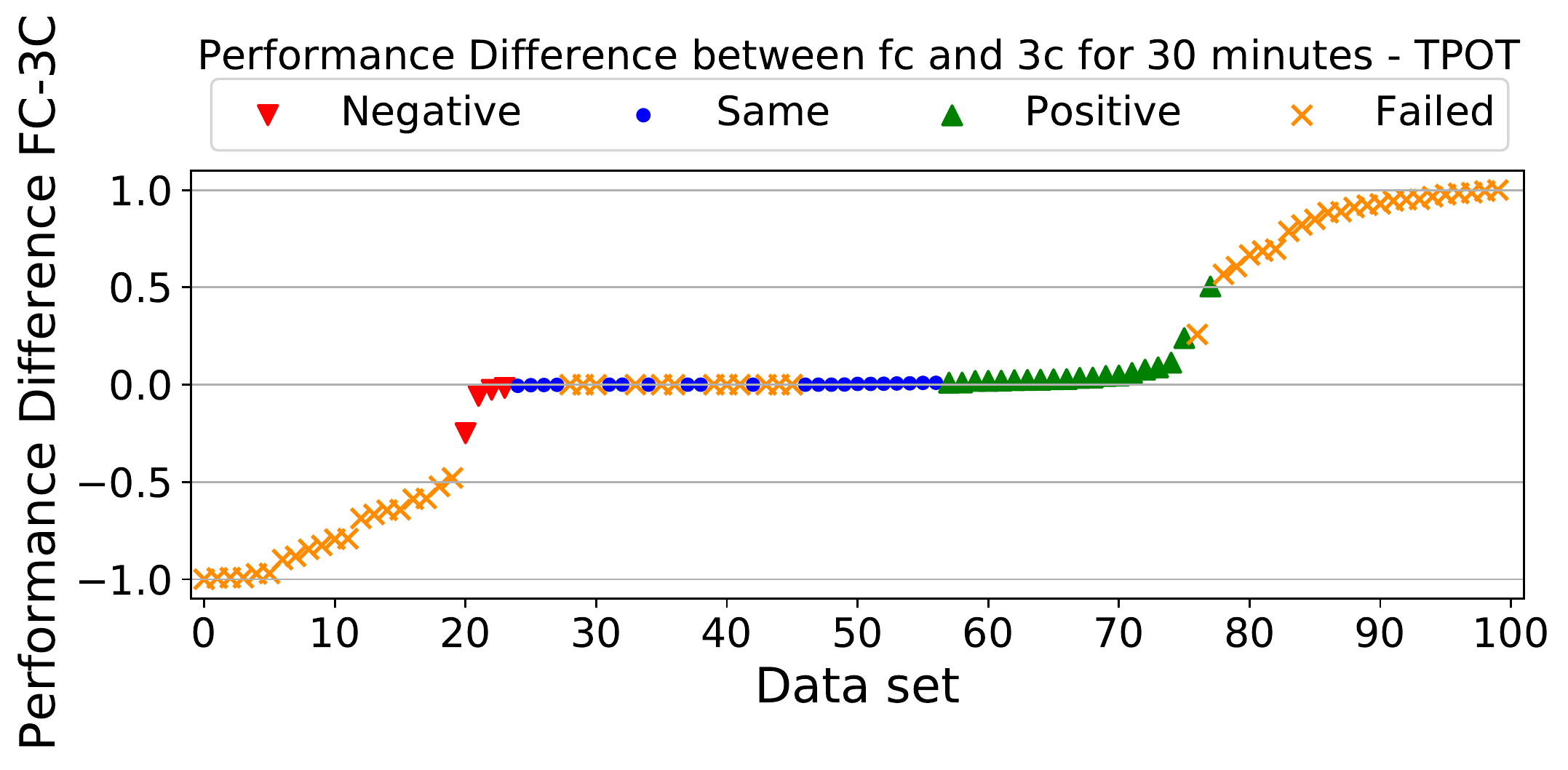}
}
\centering \subfigure[ATM] {
    \label{FIG:SearchSpaceATM}
    \includegraphics[width=0.47\textwidth]{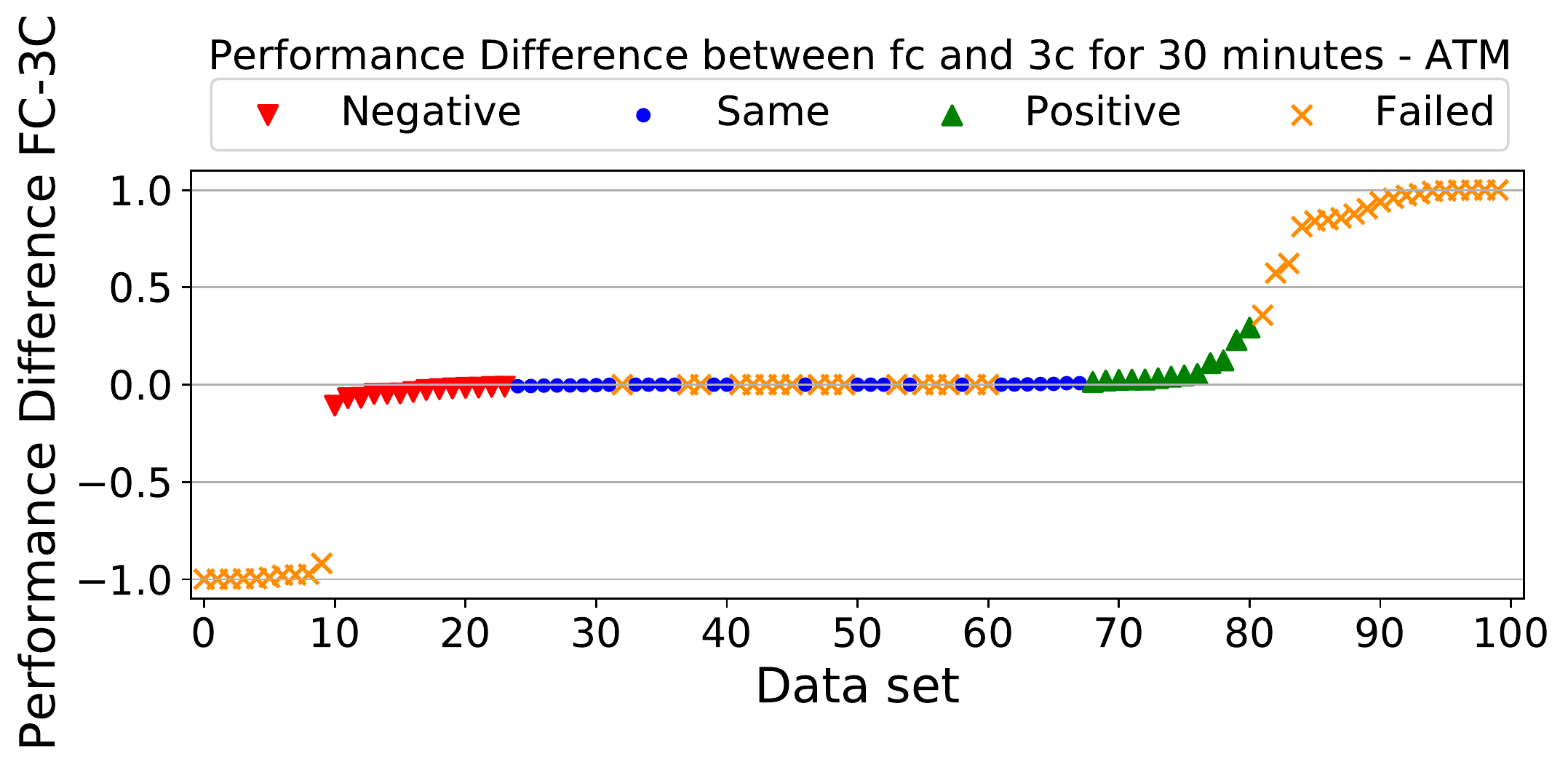}
}
\caption{\textcolor{red}{The impact of using a static portfolio on each AutoML framework. Green markers represent better performance with $FC$ search space, blue markers represent comparable performance with a difference less than 1\%, red markers represent better performance with $3C$ search space, yellow markers on the left represent failed runs with $FC$ but successful with $3C$, yellow markers on the right represent failed runs with $3C$ but successful with $FC$, and yellow markers in the middle represent failed runs with both $FC$ and $3C$.}}
\label{FIG:SearchSpace2}
\end{figure*}

\textcolor{red}{Finding an optimal solution to the time-bounded optimization problem of AutoML requires defining the underlying search space and searching for well-performing ML pipelines as efficiently as possible. Often these search spaces are chosen arbitrarily without any validation, sometimes leading to bloated spaces and the inability to find optimal results~\citep{zoller2021benchmark}. In the following, we examine the impact of a budget allocation strategy as a complementary design decision for AutoML frameworks. The strategy is based on using a static portfolio~\citep{kotthoff2016algorithm} – a set of configurations that covers as many diverse datasets as possible and minimizes the risk of failure
when facing a new task. So, we construct
a portfolio consisting of the top three performing classifiers over the 100 datasets and supported by all AutoML frameworks. These classifiers are \emph{support vector machine}, \emph{random forest}, and \emph{decision tree}. Then for a dataset at hand, all algorithms in this portfolio based on different hyperparameters are evaluated. } For the AutoML frameworks included in this work that allows configuring the search space, \texttt{ATM}, \texttt{AutoSKlearn}, and \texttt{TPOT}, we compare the performance of using the full search space including all available classifiers ($FC$) to the performance when using the static portfolio ($3C$) on 30 minutes time budget, the results of which are summarized in Figure~\ref{FIG:SearchSpace2}.

For \texttt{AutoSKlearn}, the results show that the performance of the $FC$ outperforms the performance of $3C$ on 28 datasets with an average predictive performance gain of 3.3\%. \textcolor{red}{On the other hand, the predictive performance achieved using $3C$ outperforms that achieved using $FC$ on 21 datasets by 5.9\%, as shown in Figure~\ref{FIG:SearchSpaceSklearn}. This difference in performance can be attributed to the focus of the AutoML frameworks on tuning classifiers that tended to yield good performance and hence evaluate more hyperparameters of these classifiers than in $FC$. In other words, AutoML frameworks focus on promising regions in the search space and ignore unimportant ones.} On 50 datasets, both of the $FC$ and the $3C$ achieve comparable performance with predictive performance differences of less than 1\%. For \texttt{TPOT}, the results show that 23 datasets failed to run with $3C$ and 20 datasets failed to run with $FC$, while 12 datasets failed to run using both $FC$ and $3C$, as shown in Figure~\ref{FIG:SearchSpaceTPOT}. For successful runs, \texttt{TPOT} achieved better performance on 21 datasets using $FC$ compared to $3C$ with average predictive performance improvement of 9.6\%, while the performance of using both search spaces is comparable on 18 datasets. For successful runs, \texttt{ATM} achieved better performance using $3C$ over $FC$ on 17 datasets with an average predictive performance improvement of 4\%. On the other hand, the \texttt{ATM} achieved better performance using $FC$ compared to $3C$ on 15 datasets with an average performance improvement of 9.3\%, while the performance of \texttt{ATM} using both search spaces was comparable on 22 datasets, as shown in Figure~\ref{FIG:SearchSpaceATM}. \textcolor{red}{Notably, 19 datasets failed to run with the $FC$ yielded valid results with the $3C$. On the other side, $10$ datasets failed with the $3C$ yielded valid results with $FC$, as shown in Figure~\ref{FIG:SearchSpaceATM}.} The Wilcoxon signed-rank test was conducted to determine if a statistically significant difference in terms of the performance difference on all datasets exists between using $FC$ and $3C$ on \texttt{AutoSKlearn}, \texttt{TPOT}, and \texttt{ATM}. For \texttt{TPOT}, the results showed that the difference in performance between the two search spaces is statistically significant with more than 95\% level of confidence (p-value=0.003). In contrast, for \texttt{AutoSKlearn} and \texttt{ATM}, the results showed no statistically significant difference in performance between the two search spaces.

\subsubsection{Impact of Meta-learning}\label{Sec:MetaLearSKlearn}
One way to define meta-learning is the process of learning from previous experience gained during applying various learning algorithms on different ML tasks, and hence reducing the time needed to learn new tasks~\citep{vanschoren2018meta}. \textcolor{red}{In the following, we study the impact of meta-learning on the performance of AutoML frameworks.} \textcolor{red}{The only framework supporting to configure the meta-learning is \texttt{AutoSKlearn}.} Furthermore, we investigate the relationship between the characteristics of the different datasets and the improvement caused by employing the vanilla version or the meta-learning version of the \texttt{AutoSKlearn}. 

\begin{figure*}[!t]
\centering \subfigure[10 Min.] {
    \label{FIG:Meta10}
    \includegraphics[width=0.47\textwidth]{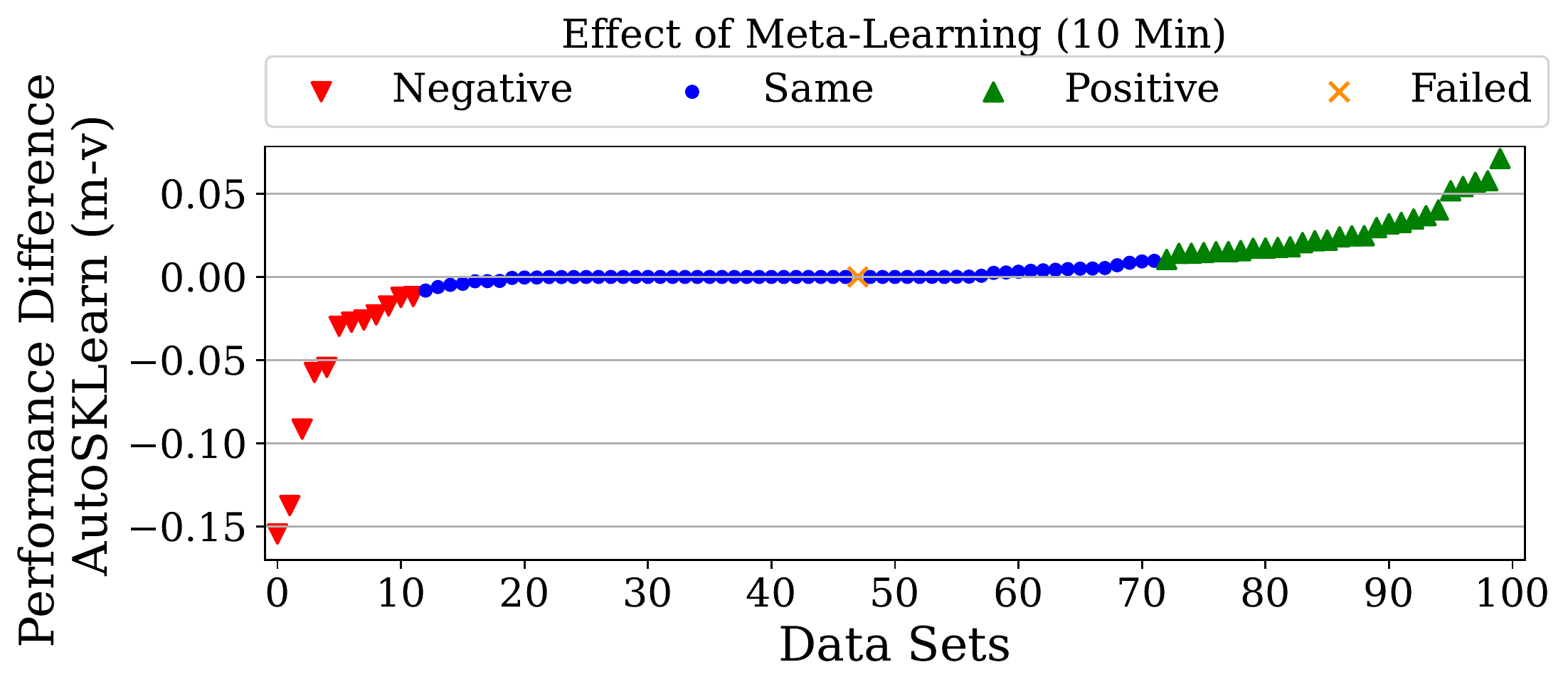}
}
\centering \subfigure[30 Min.] {
    \label{FIG:Meta30}
    \includegraphics[width=0.47\textwidth]{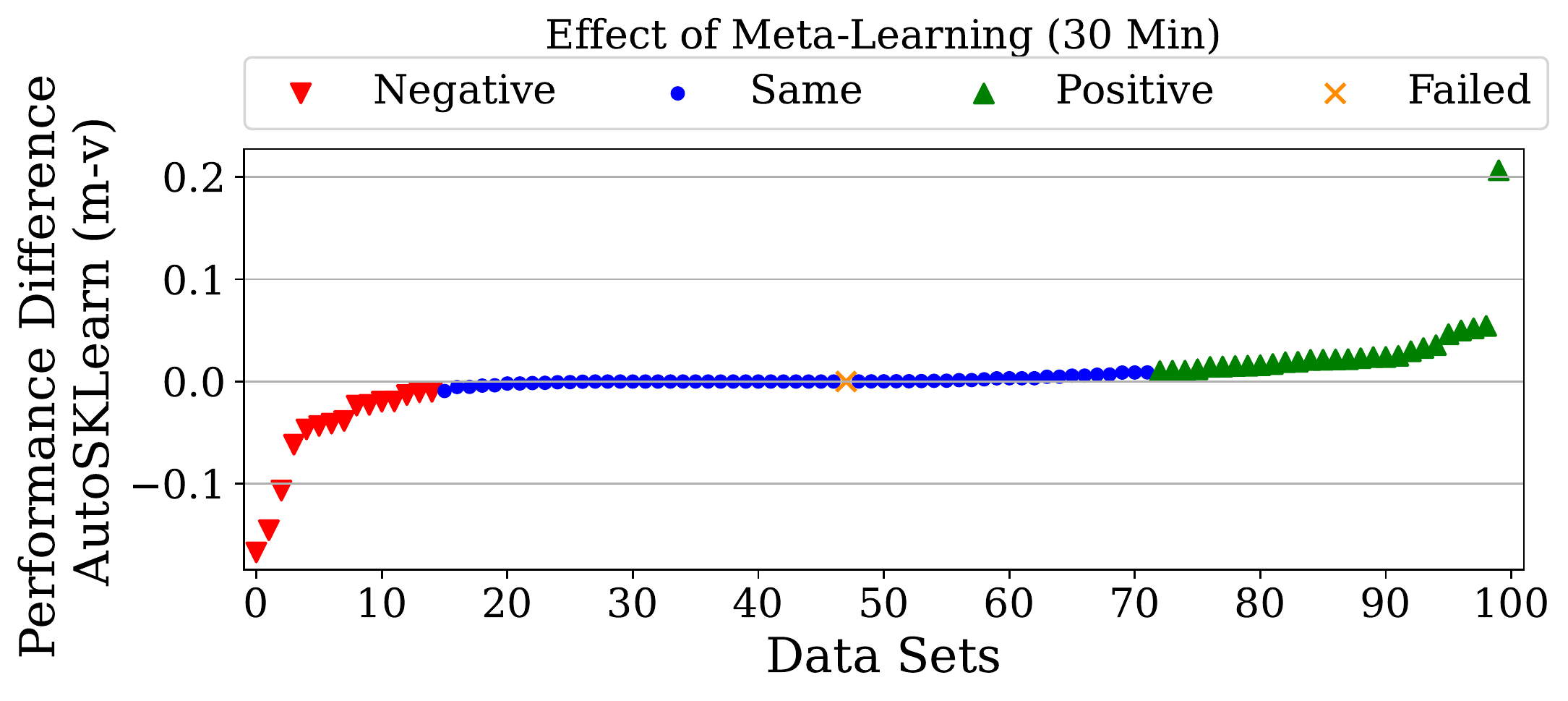}
}
\centering \subfigure[60 Min.] {
    \label{FIG:Meta60}
    \includegraphics[width=0.47\textwidth]{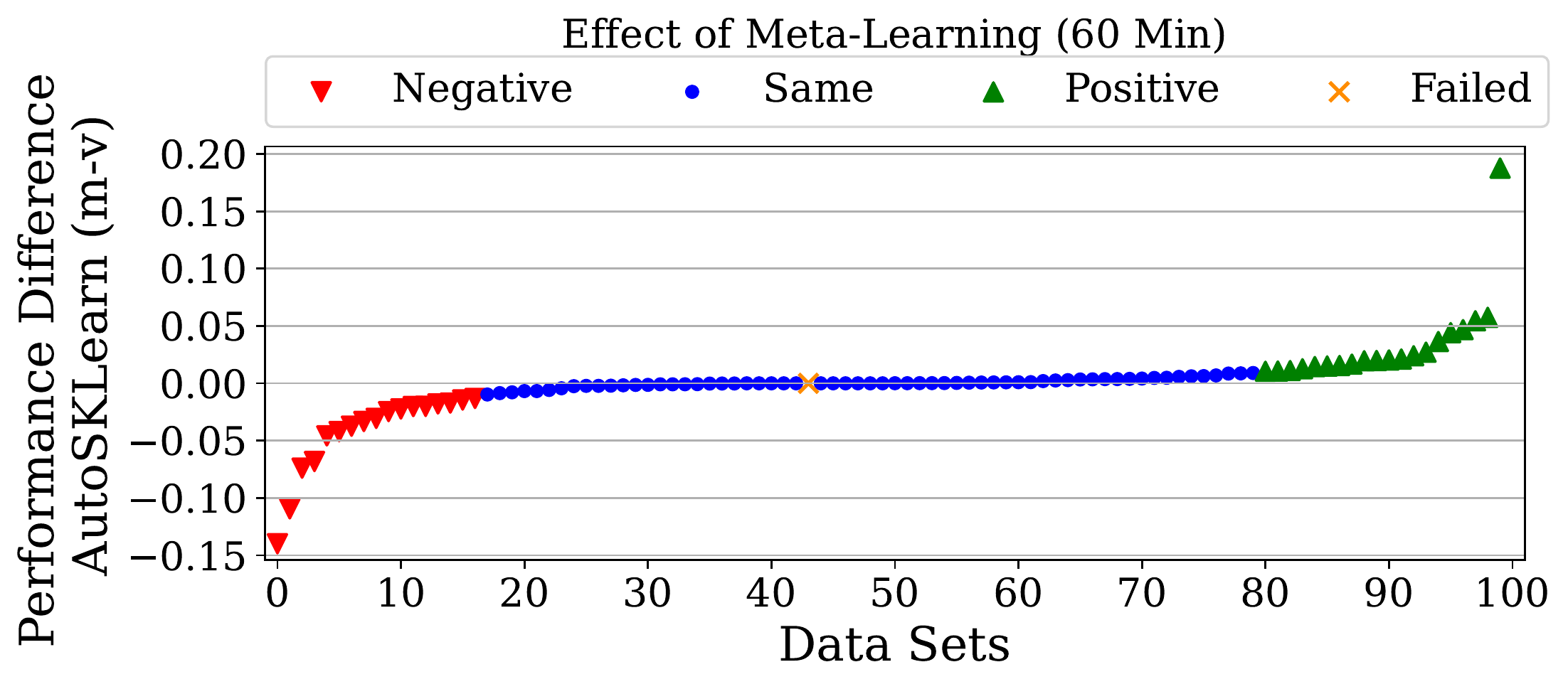}
}
\centering \subfigure[240 Min.] {
    \label{FIG:Meta240}
    \includegraphics[width=0.47\textwidth]{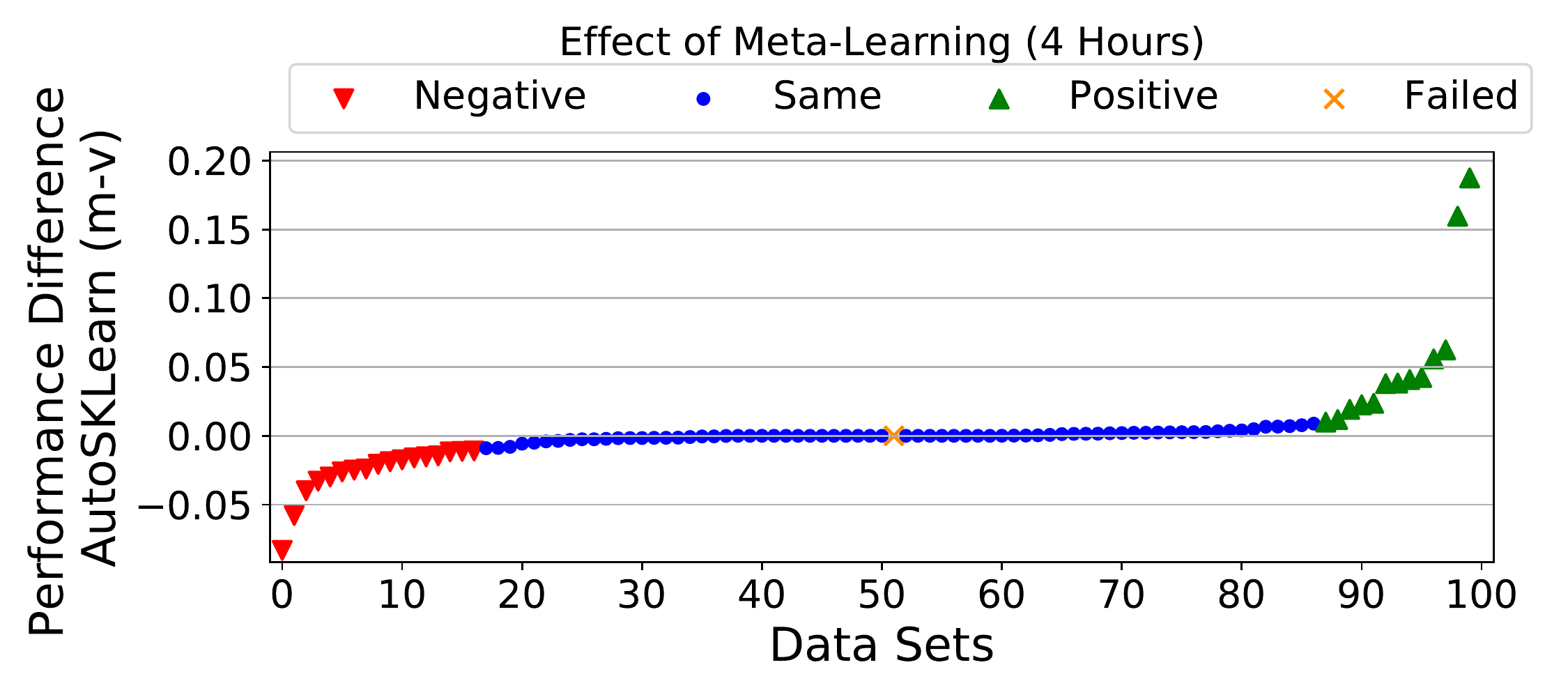}
}
\caption{The impact of meta-learning over all time budgets. Green markers represent better performance with \texttt{AutoSKlearn-m}, blue markers represent comparable performance with a difference
less than 1\%, red markers represent better performance using \texttt{AutoSKlearn-v}, and yellow markers represent failed runs with both runs with both FC and 3C.}
\label{FIG:MetaLearning}
\end{figure*}

\texttt{AutoSKlearn} applies a meta-learning mechanism based on a knowledge base storing the meta-features of datasets as well as the best performing pipelines on these datasets. \texttt{AutoSKlearn} uses 38 meta-features including statistical, information-theoretic and simple meta-features. In an offline phase, the meta-features and the empirically best-performing pipelines are stored for each dataset in their repository (140 datasets from OpenML repository). In an online phase, for any new dataset, the framework extracts the meta-features of the new dataset and searches for the most similar datasets in the knowledge base and return the top $k$ best performing pipelines on these similar datasets. These $k$ pipelines are used as a warm start for the Bayesian optimization algorithm used in the optimization process.
To assess the impact of the meta-learning mechanism, we compare the performance of (\texttt{AutoSKlearn-v}) and (\texttt{AutoSKlearn-m}) on 100 datasets across different time budgets, as shown in Figure~\ref{FIG:MetaLearning}. \textcolor{red}{The results show that using meta-learning is not necessarily associated with performance improvement. \iffalse In this experiment, for presentation purposes, we considered that if the difference in the predictive performance between the two modes, \texttt{AutoSKlearn-m} and \texttt{AutoSKlearn-v}, is within 1\% then it is considered negligible (small circle in the figure). If the performance difference is more 1\% on the side of the meta learning-based version then an upward triangle is used as an indicator and if the performance difference is more than 1\% for the vanilla version then a downward triangle is used. We have conducted the experiment on 4 different time budgets: 10, 30, 60 and 240 minutes.\fi The results show that, on average, the performance of the vanilla and the meta-learning versions is very comparable on the 4 time budgets. In particular, both versions perform similarly on 64, 55, 65, and 69 datasets for 10 minutes, 30 minutes, 60 minutes and 240 minutes, respectively.}
Table~\ref{table:Metasummary} summarizes the performance of both of \texttt{AutoSKlearn-m} and \texttt{AutoSKlearn-v}, in addition to the number of datasets achieved improvement in performance by employing \texttt{AutoSKlearn-m} over \texttt{AutoSKlearn-v} on different time budgets. \textcolor{red}{The results show that the improvement achieved by employing the meta-learning version decreases for more extended time budgets. For example, the number of datasets that achieved performance improvement by using meta-learning dropped from 28 for the 30 minutes budget to 14 for the 240 minutes budget. For the 60 minutes budget, the number of datasets that achieved improvement by employing each of the meta-learning and vanilla versions is 20 and 17, respectively.}
Additionally, employing the meta-learning achieved mean improvement of 2.9\% on  28 datasets over 10 minutes and 30 minutes time budgets, respectively. For the 60 minutes and 240 minutes, the mean performance improvement by employing the meta-learning is higher than that achieved by employing the vanilla version, as shown in Table~\ref{table:Metasummary}. We use \texttt{Wilcoxon} statistical test to assess the significance of the performance difference between the vanilla version and the meta-learning version. The results show that the impact of the meta learning is statistically significant only for the smallest time budget of 10 minutes with more than 95\% level of confidence (pvalue=0.004). 

\begin{table}[!t]
\centering
\caption{The performance of \texttt{Auto\-Sklearn-v} and \texttt{Auto\-Sklearn-m} and the gain in performance achieved by employing the meta-learning on 100 datasets over different time budgets.}
\begin{tabular}{|c|c|c|c|c|c|c|c|}
  \toprule
    \multirow{2}{*}{Time Budget} &    \multirow{2}{*}{Framework}  &    \multicolumn{2}{c}{Predictive Performance}& \multicolumn{3}{c}{Performance Gain} & \#datasets with  \\
    & &    Mean &    SD & Min & Mean & Max &  gain $>$ 1\% \\
     \hline
    \multirow{2}{*}{10}&  AutoSKlearn-m &   0.864 &  0.153 & 1.1\% & 2.9\% & 7.1\% &  28\\
    &  AutoSKlearn-v &   0.862 &  0.151 & 1.1\% & 5.4\% & 15.5\% & 12\\
    \hline
    \multirow{2}{*}{30}    &  AutoSKlearn-m &    0.868 &  0.152 & 1.1\% & 3.1\% & 20.6\% & 28\\
    &  AutoSKlearn-v &    0.8867 &  0.149 & 1.1\% & 5.1\% & 16.7\% & 15 \\
    \hline
    \multirow{2}{*}{60} &  AutoSKlearn-m &    0.870 &  0.144 & 1.1\% & 3.3\% & 18.8\% & 20 \\
    &  AutoSKlearn-v &    0.870 &  0.142 & 1.1\% & 4.3\% & 14.0\% & 17 \\
    \hline
    \multirow{2}{*}{240}&  AutoSKlearn-m &   0.873 &  0.141 & 1.1\% & 7.7\% & 31.6\% & 14 \\
    &  AutoSKlearn-v &   0.867 &  0.156  & 1.1\% & 2.7\% & 8.4\% & 20 \\

 \bottomrule
\end{tabular}
\label{table:Metasummary}
\end{table}

\textcolor{red}{In the following, we explore the relationship between the characteristics of the datasets and the improvement achieved by utilizing the meta-learning version of \texttt{AutoSKlearn} over different time budgets. To this end, we train a model that takes the meta-features of datasets, i.e., characteristics of the datasets as input and provides prediction to whether using meta-learning can improve the performance. To develop this model, we label each dataset as \texttt{Class 1} if utilising the meta-learning improves the performance over the vanilla version and \texttt{Class 0} otherwise. We implemented a total of 42 meta-features from the literature, including
simple, information-theoretic and statistical meta-features~\citep{kalousis2002algorithm,mitchell1990machine}, such as statistics about the number of data points, features, and the number of classes, or data skewness, and the entropy of the targets. All meta-features are listed in Table~\ref{tab:metafeatures} in the Appendix \ref{append:metafeatures}. Using the extracted information from the knowledge base, for our 100 datasets, we fitted a shallow decision tree of depth 4 using the meta-feature variables as the predictors of the model. We considered decision tree classifier due to its interpretable nature that allows rules to be derived from a root-leaf path in the tree. So, given a new dataset, we compute its meta-features and use the decision tree model to recommend whether meta-learning is likely to improve the performance or not. Our model has achieved the following performance: Recall = 0.85 and F1 Score = 0.85. Rules for \texttt{Class 1} and \texttt{Class 0} can be represented as follows:}

\textcolor{red}{
\begin{itemize}
    \item \textbf{R1:} $min(\frac{c}{n}) > 0.5  \implies \texttt{Class 1}$
    \item \textbf{R2:} $min(\frac{c}{n}) < 0.27 \wedge \ noise\mhyphen signal\ ratio > 8.57 \wedge p < 845 \implies \texttt{Class 1}$
    \item \textbf{R3:} $0.1 < min(\frac{c}{n}) < 0.27 \wedge \ noise\mhyphen signal\ ratio < 8.57 \implies \texttt{Class 1}$
    \item \textbf{R4:} $0.27 < min(\frac{c}{n}) < 0.5 \implies \texttt{Class 0}$ 
    \item \textbf{R5:} $min(\frac{c}{n}) < 0.27 \wedge \ noise\mhyphen signal\ ratio > 8.57 \wedge p > 845 \implies \texttt{Class 0}$
    \item \textbf{R6:} $min(\frac{c}{n}) < 0.1 \wedge \ noise\mhyphen signal\ ratio < 8.57 \implies \texttt{Class 0}$
\end{itemize}}

\textcolor{red}{It is clear from the extracted rules that the number of features $p$, the percentage of the minority class to the number of instances ($min(\frac{c}{n})$), and noisiness of data ($noise\mhyphen signal\ ratio$) are quite important features for the prediction.}

\subsubsection{Impact of Ensembling}
\label{SEC:DDEnsembling}

Ensembling~\citep{dietterich2000ensemble} is the process of combining multiple ML base models for the same task to produce a better predictive model. These base models can be combined in different techniques, including simple voting (averaging), weighted voting (averaging), bagging, and boosting~\citep{dietterich2000ensemble}. In the following, we explore the impact of ensembling on the performance of the AutoML frameworks allowing enabling and disabling post-processing ensemble. Such frameworks include \texttt{AutoSKlearn} and \texttt{SmartML-m}. Furthermore, we investigate whether there is a relationship between the characteristics of the different datasets and the improvement caused by employing the vanilla version or the
ensembling version of the AutoML framework. During the optimization process of \texttt{AutoSKlearn} and \texttt{SmartML}, the frameworks store the generated models instead of just keeping the best performing one. These models are used in a post-processing phase to construct an ensemble model. This automatic ensemble construction avoids relying on a single hyperparameter setting which makes the generated model more robust to overfitting. \texttt{AutoSKlearn} uses the ensemble selection methodology introduced by Caruana et al.~\citep{caruana2004ensemble}, while \texttt{SmartML} uses majority voting~\citep{lam1997application}. Ensemble selection is a greedy technique that starts with an empty ensemble and \textcolor{red}{iteratively} adds base models to the ensemble in a way that maximizes the validation performance. The technique uses uniform weights; however, it allows repetitions. The majority voting is considered the simplest and the most effective scheme. The majority voting scheme adheres to democratic principles, i.e., the class with the most votes wins. We kept the default setting of \texttt{AutoSKlearn} and \texttt{SmartML} using 50 and 5 base models in the ensemble, respectively.

To assess the impact of the ensembling, we compare the mean performance of vanilla/base version of each of \texttt{AutoSKlearn} and \texttt{SmartML} to their ensembling versions across different time budgets, as shown in Table~\ref{table:ensemblingsummary}. More detailed performance comparison over all datasets across all time budgets are given in Figures~\ref{FIG:Ensembling} and ~\ref{FIG:Ensembling_smartML} in  Appendix~\ref{Appen:ensembling}.

\textbf{AutoSKlearn}: The results show that ensembling does not always contribute to better performance compared to the vanilla version. The results show that employing ensembling for \texttt{AutoSKlearn} achieved mean improvement of 3.9\%, 3.2\%, 4.7\%, and 8.0\% on 24, 32, 25 and 24 datasets over 10, 30, 60, and 240 minutes budgets, respectively, as shown in Table~\ref{table:ensemblingsummary}. We use Wilcoxon statistical test to assess the significance of the performance difference between \texttt{AutoSKlearn-e} and \texttt{AutoSKlearn-v}. The results show that ensembling enhances the performance with a statistically significant gain of more than 95\% level of confidence ($p$ value $<$ 0.05) on the 4 time budgets. The level of confidence is almost 99\% over all the time budgets combined.

\begin{table}[!t]
\centering
\caption{Performance comparison between vanilla/base version vs ensembling version of \texttt{AutoSKlearn} and \texttt{SmartML} different time budgets.}
\begin{tabular}{|c|c|c|c|c|c|c|c|}
  \toprule

    \multirow{2}{*}{Time Budget} &    \multirow{2}{*}{Framework}  &    \multicolumn{2}{c}{Predictive Performance}& \multicolumn{3}{c}{Performance Gain} & \#datasets with  \\
    & &    Mean &    SD & Min & Mean & Max &  gain $>$ 1\% \\
     \hline

    \multirow{4}{*}{10}&  AutoSKlearn-e &    0.868 &  0.145  & 1.1\% & 3.9\% & 16.7\% & 24 \\
    &  AutoSKlearn-v &    0.868 &  0.151 & 1.1\% & 3.2\% & 8.9\% & 14 \\
    \cline{2-8}
    &  SmartML-e &   0.831 &  0.176 & 1.1\% & 12.6\% & 64.2\% & 30 \\
    &  SmartML &    0.176 &  0.176	 & 1.1\% & 10.2\% & 36.1\% & 14 \\
    \hline
    \multirow{4}{*}{30}    &  AutoSKlearn-e &    0.875     &  0.141 & 1.1\% & 3.2\% & 13.9\% & 32\\
    &  AutoSKlearn-v &    0.867 &  0.149  & 1.1\% & 2.9\% & 11.1\% & 11 \\
    \cline{2-8}
    &  SmartML-e &    0.838 &  0.159 & 1.1\% & 12.5\% & 73.2\% & 33 \\
    &  SmartML &    0.838 &  0.199 & 1.1\% & 10.6\% & 39.4\% & 16 \\
    \hline
    \multirow{4}{*}{60} &  AutoSKlearn-e &    0.879 &  0.138  & 1.1\% & 4.7\% & 12.7\% & 25\\
    &  AutoSKlearn-v &    0.870 &  0.142  & 1.1\% & 2.6\% & 6.1\% & 13\\
    \cline{2-8}
    &  SmartML-e &    0.832 &  0.172 & 1.1\% & 11.2\% & 55.8\% & 28 \\
    &  SmartML &    0.816 &  0.194 & 1.1\% & 10.5\% & 31.1\% & 18 \\
    \hline
    \multirow{4}{*}{240}&  AutoSKlearn-e &    0.883 &  0.132 & 1.1\% & 8.0\% & 69.7\% & 24 \\
    &  AutoSKlearn-v &   0.867 &  0.156 & 1.1\% & 3.1\% & 8.4\% & 14\\
    \cline{2-8}
    &  SmartML-e &    0.842 &  0.165 & 1.1\% & 10.2\% & 34.5\% & 28 \\
    &  SmartML &    0.826 &  0.169 & 1.1\% & 11.9\% & 37.2\% & 16 \\
    
 \bottomrule
\end{tabular}
\label{table:ensemblingsummary}
\end{table}

\textbf{SmartML}: \texttt{SmartML-e} slightly improved the performance over the \texttt{SmartML-m} by average performance of 12.6\%, 12.5\%, 11.2\%, and 10.2\% on 30, 33, 28, and 28 datasets for 10, 30, 60, and 240 minutes time budgets, respectively, as shown in Table~\ref{table:ensemblingsummary}. We use Wilcoxon statistical test to assess the significance of the performance difference between the base (meta-learning) and the ensembling versions of \texttt{SmartML}. The results show that the ensembling version enhance the performance with a statistically significant gain of more than 95\% level of confidence ($p$ value $<$ 0.05) on the 4 time budgets.

\textcolor{red}{In the following, we explore the relationship between the characteristics of the datasets and the improvement achieved by utilizing the ensembling version of \texttt{AutoSKlearn} over different time budgets. To this end, we followed the same approach in Section~\ref{Sec:MetaLearSKlearn} and train a decision tree of depth 3 that takes the meta-features of 100 datasets as input and provides prediction to whether using ensembling can improve the performance (\texttt{Class 1}) or not (\texttt{Class 0}). So, given a new dataset, we compute its meta-features and use the decision tree model to recommend the whether ensembling is likely to improve performance or not. Our model for \texttt{AutoSKlearn} has achieved the following performance: Recall = 0.70 and F1 Score = 0.70. \texttt{AutoSKlearn} rules for \texttt{Class 1} and \texttt{Class 0} can be represented as follows:}

\textcolor{red}{
\begin{itemize}
    \item \textbf{R1:} $\mu(\rho) > 0.13 \ \wedge \ \sigma(\rho) > 0.27 \ \wedge \ max(\rho) > 0.98 \implies \texttt{Class 1} $
    \item \textbf{R2:} $\mu(\rho) > 0.13 \ \wedge \ \sigma(\rho) \leq 0.27 \ \wedge \ min(\rho) > 0.44 \implies \texttt{Class 1}$
    \item \textbf{R3:} $\mu(\rho) \leq 0.13 \ \wedge \ \mu(\pi_{i}) > 3.11 \ \wedge \ \sigma(\frac{n}{p}) > 0.03 \implies \texttt{Class 1}$
    \item \textbf{R4:} $\mu(\rho) \leq 0.13 \ \wedge \ \mu(\pi_{i}) \leq 3.11 \ \wedge \ min(Mutual\ inform.) > 0.03 \implies \texttt{Class 1}$
    \item \textbf{R5:} $\mu(\rho) > 0.13 \ \wedge \ \sigma(\rho) > 0.27 \ \wedge \ max(\rho) \leq 0.98 \implies \texttt{Class 0}$
    \item \textbf{R6:} $\mu(\rho) > 0.13 \ \wedge \ \sigma(\rho) \leq 0.27 \ \wedge \ min(\rho) \leq 0.44 \implies \texttt{Class 0}$
    \item \textbf{R7:} $\mu(\rho) \leq 0.13 \ \wedge \ \mu(\pi_{i}) > 3.11 \ \wedge \ \sigma(\frac{n}{p}) \leq 0.03 \implies \texttt{Class 0}$
    \item \textbf{R8:} $\mu(\rho) \leq 0.13 \ \wedge \ \mu(\pi_{i}) \leq 3.11 \ \wedge \ min(Mutual\ inform.) \leq 0.03 \implies \texttt{Class 0}$
\end{itemize}}
\textcolor{red}{Clearly, the following features are important to the prediction of the model; the mean of the pairwise correlation between features ($\mu(\rho)$), standard deviation of the pairwise correlation between features ($\sigma(\rho)$), maximum of the pairwise correlation between features ($\max(\rho)$), minimum of the pairwise correlation between features ($\min(\rho)$), the standard deviation of the ratio between number of instances and the number of features ($\sigma(\frac{n}{p})$), the minimum of the  mutual information between features and class ($Mutual\ inform.$), and the mean of the unique categorical values of features ($\mu(\pi_{i})$).}

\textcolor{red}{For \texttt{SmartML}, we trained multiple models; however, none of the models could capture the relation of the meta-features and the performance improvement due to employing ensembling.}

\section{Discussion and Future Direction}
\label{Sec:DiscussionFutureDirection}

\textcolor{red}{The global performance average weighted by the percentage of successful runs shows that \texttt{AutoSKlearn-e} and \texttt{AutoSKlearn} achieve the highest performance, while \texttt{Recipe} comes in the last place.} Overall, \texttt{AutoSklearn} achieves the highest number of successful runs across different time budgets and witnessed performance improvement over the most significant number of datasets when increasing the time budget. Our analysis reveals that the impact of meta-learning declines over longer time budgets (i.e., 60 mins, 240 mins). In contrast, ensembling achieves consistent performance improvement across all time budgets. \textcolor{red}{For \texttt{AutoSKlearn}, the analysis reveals a relationship between the characteristics of the datasets (e.g., number of features, noisiness of data, mutual information between features and class) and the improvement achieved by utilizing meta-learning or ensembling.} Generally, AutoML frameworks considered in this work build pipelines with an average length of 2. TPOT yields the shortest pipelines of an average length of 1.5. A possible explanation could be that TPOT generates pipelines that optimize both the pipelines' performance and complexity. \textcolor{red}{Additionally, \texttt{AutoSKlearn}, \texttt{ATM}, and \texttt{TPOT} achieve the highest performance on multi-class classification tasks. For datasets with a large number of instances and a small number of features, \textt{ATM} is a clear winner.}

For some datasets, the performance of the different versions of \texttt{AutoSKlearn} varies significantly across different iterations. These datasets are characterized by having far fewer instances than features. Analyzing the pipelines of the different versions of \texttt{AutoSKlearn} on these datasets across multiple iterations shows that data preprocessing component is responsible for the large performance variance between the different pipelines. For example, the performance difference between \texttt{AutoSKlearn-v} and \texttt{AutoSKlearn-m} on \texttt{phpdo58hj}  varies significantly between $6\%$ to $13\%$ across different iterations. The two generated pipelines for \texttt{AutoSKlearn-v} and \texttt{AutoSKlearn-m} used the same model (lda) with the same set of hyperparameters but different preprocessors. 
For large datasets, meta-learning shows significant performance improvement. For example, \texttt{AutoSKlearn-m} achieves significantly better performance than \texttt{AutoSKlearn-v} on CovPokElec. A possible explanation is that meta-learning warm-starts the optimization process and increases the chances of finding a well-performing configuration in the limited attempts during the defined time budget.

Specifying the time budget needs to be considered carefully as significantly increasing the time budget for the search process (e.g., from 60 minutes to 240 minutes) may not lead to a significant improvement in the predictive performance. This decision varies from one scenario/application to another. For some applications, spending a long budget to achieve an additional predictive performance of 1\% could be crucial while less important for other applications. \textcolor{red}{However, longer time budgets may lead to over-fitting}. Carefully selecting a small search space with few top-performing classifiers can lead to a very comparable performance with a search space that includes a large number of classifiers which is the case for \texttt{AutoSKlearn} and \texttt{ATM} frameworks. 


 \textcolor{red}{Intuitively, an extensive, systematic search for a well-performing machine learning pipeline should bear a high risk of over-fitting, and previous AutoML frameworks have confirmed this intuition~\citep{thornton2013auto}. AutoML tools are on the right extreme of the bias-variance spectrum as they choose among all learners and even construct new and arbitrary large ones using ensemble methods~\citep{mohr2018ml}. Notably, \textt{SmartML} and ~\texttt{AutoWeka} witnessed performance degradation when increasing the time budget from 30 to 60 minutes. One possible explanation is that the data available for the search process is not sufficiently substantial and representative for "real" data. Hence, the danger of over-fitting is higher than for basic learning algorithms. This insight calls for developing novel and more efficient mechanisms to prevent over-fitting.}

\textcolor{red}{While AutoML frameworks optimize predictive performance, many exceed the specified time budget by more than 10\%. This violation of the time constraints caused many runs to be terminated and considered as failed. This problem is observed in all frameworks except for AutoSKLearn, which calls for a robust implementation and careful consideration of the time constraint.}

Most of the current work on AutoML considered automating the preprocessing, algorithm selection and hyperparameter tuning while ignoring the feature engineering part. In practice, the feature engineering part consumes most of the time to build ML pipelines and significantly affects the performance. The right feature engineering phase could turn the feature space into a linearly separable space, so even naive classifiers could achieve relatively high predictive performance. On the other hand, skipping this phase or using the wrong feature engineering preprocessors makes it harder to achieve relatively high predictive performance, even for the most efficient classifiers. Hence, further research in this area can improve the overall performance of the resulting AutoML pipelines.

\section{Conclusion} \label{conclusion}
In this paper, we present a comprehensive evaluation and comparison of the performance characteristics of six AutoML frameworks on 100 datasets from OpenML. Our analysis revealed that there is no single winning framework that outperforms others over all time budgets. Across various evaluations, \texttt{AutoSklearn}, \texttt{ATM}, and \texttt{TPOT} are the top-performing frameworks. The results also show that genetic-based frameworks (\texttt{TPOT} and \texttt{Recipe}) have high frequent failure rates for short time budgets while their success rates are steadily increasing as the time budget increases. Meta-learning has a significant impact on small-time budgets, and such impact declines as the time budget increases. In contrast, ensembling consistently improves performance significantly across all time budgets. Furthermore, carefully selecting a small search space with few top-performing classifiers can lead to a comparable performance with a search space that includes many classifiers. Furthermore, increasing the time budget does not necessarily improve predictive performance. We believe that the results of our analysis are beneficial for guiding and improving the design process of future AutoML techniques.

\section*{Data Availability}
 The datasets generated during and/or analysed during the current study are available in the AutoMLBench repository, \url{https://datasystemsgrouput.github.io/AutoMLBench/datasets}.

\section*{Acknowledgment}
\label{SEC:Acknowledgment}
This work was supported by European Social Fund via "ICT programme“ measure" as well as the European Regional Development Fund and the programme Mobilitas Pluss (2014-2020.4.01.16-0024).

\begin{appendices}





\section{Evaluated Datasets}\label{secA1}
Table~\ref{Tbl:AutoMLDatasets} shows the datasets used in evaluating all the AutoML frameworks included in this work.

\small
\begin{longtable} {ccccccc}


\hline 
\rotatebox[origin=c]{-70}{Dataset Name (openml id)}             & \rotatebox[origin=c]{-70}{Nr features} & \rotatebox[origin=c]{-70}{Nr instances} & \rotatebox[origin=c]{-70}{Nr classes} & \rotatebox[origin=c]{-70}{Nr missing values} & \rotatebox[origin=c]{-70}{Nr categorical features} & \rotatebox[origin=c]{-70}{Class entropy} \\ \hline
AirlinesCodrnaAdult        (1240)  & 30      & 1076790 & 2      & 11896                              & 1                                        & 1,00          \\
Amazon                     (1457)  & 10001   & 1500    & 50     & 0                                  & 0                                        & 0,93          \\
analcatdata\_authorship    (458)   & 71      & 841     & 4      & 0                                  & 0                                        & 0,99          \\
AP\_Breast\_Lung           (1150)  & 10937   & 470     & 2      & 0                                  & 0                                        & 0,99          \\
AP\_Omentum\_Ovary         (1156)  & 10937   & 275     & 2      & 0                                  & 1                                        & 0,93          \\
AP\_Prostate\_Ovary        (1152)  & 10937   & 267     & 2      & 0                                  & 0                                        & 2,18          \\
arrhythmia                 (1017)  & 263     & 452     & 2      & 0                                  & 1                                        & 1,58          \\
audiology                  (999)   & 70      & 226     & 2      & 0                                  & 1                                        & 3,70          \\
avila-tr                   (42932) & 11      & 20867   & 12     & 114                                & 10                                       & 2,27          \\
churn                      (40701) & 21      & 5000    & 2      & 0                                  & 1                                        & 1,00          \\
cifar-10                   (40927) & 3073    & 60000   & 10     & 0                                  & 70                                       & 0,81          \\
connect-4                  (1591)  & 43      & 67557   & 3      & 0                                  & 1                                        & 0,82          \\
CovPokElec                 (149)   & 65      & 1455525 & 10     & 0                                  & 1                                        & 0,86          \\
dataset\_183\_adult        (179)   & 15      & 48842   & 2      & 0                                  & 0                                        & 2,21          \\
dataset\_185\_yeast        (181)   & 9       & 1484    & 10     & 0                                  & 1                                        & 2,19          \\
dataset\_186\_satimage     (182)   & 37      & 6430    & 6      & 1668                               & 3                                        & 1,00          \\
dataset\_187\_abalone      (183)   & 9       & 4177    & 28     & 0                                  & 1                                        & 0,94          \\
dataset\_189\_baseball     (185)   & 18      & 1340    & 3      & 0                                  & 0                                        & 3,32          \\
dataset\_194\_eucalyptus   (188)   & 20      & 736     & 5      & 816                                & 1                                        & 0,99          \\
dataset\_24\_mushroom      (24)    & 22      & 8124    & 2      & 0                                  & 1                                        & 0,84          \\
dataset\_26\_nursery       (26)    & 9       & 12960   & 5      & 0                                  & 0                                        & 4,20          \\
dataset\_28\_optdigits     (28)    & 63      & 5620    & 10     & 0                                  & 0                                        & 4,28          \\
dataset\_31\_credit-g      (31)    & 21      & 1000    & 2      & 0                                  & 1                                        & 2,58          \\
dataset\_36\_segment       (36)    & 19      & 2310    & 7      & 0                                  & 36                                       & 3,84          \\
dataset\_39\_ecoli         (39)    & 8       & 336     & 8      & 32                                 & 1                                        & 0,93          \\
dataset\_40\_sonar         (40)    & 61      & 208     & 2      & 896                                & 6                                        & 2,26          \\
dataset\_42\_soybean       (42)    & 36      & 683     & 19     & 0                                  & 1                                        & 1,79          \\
dataset\_44\_spambase      (44)    & 58      & 4601    & 2      & 0                                  & 1                                        & 2,00          \\
dataset\_54\_vehicle       (54)    & 19      & 846     & 4      & 0                                  & 4                                        & 0,44          \\
dataset\_59\_ionosphere    (59)    & 34      & 351     & 2      & 0                                  & 14                                       & 0,88          \\
dataset\_6\_letter         (6)     & 17      & 20000   & 26     & 0                                  & 0                                        & 4,65          \\
dataset\_60\_waveform-5000 (60)    & 41      & 5000    & 3      & 0                                  & 1                                        & 0,83          \\
dataset\_61\_iris          (61)    & 5       & 150     & 3      & 0                                  & 0                                        & 0,92          \\
dataset\_9\_autos          (9)     & 26      & 205     & 6      & 2792                               & 4                                        & 2,99          \\
devnagari                  (40923) & 785     & 92000   & 46     & 0                                  & 0                                        & 3,17          \\
electricity-normalized     (151)   & 9       & 45312   & 2      & 0                                  & 0                                        & 1,00          \\
eye\_movements             (1044)  & 28      & 10936   & 3      & 40                                 & 2                                        & 0,54          \\
GCM                        (1106)  & 16064   & 190     & 14     & 0                                  & 1                                        & 2,49          \\
gina\_agnostic             (1038)  & 971     & 3468    & 2      & 0                                  & 1                                        & 5,64          \\
hiva\_agnostic             (1039)  & 1618    & 4229    & 2      & 0                                  & 0                                        & 3,32          \\
ipums\_la\_99-small        (378)   & 60      & 8844    & 9      & 0                                  & 0                                        & 6,64          \\
jm1                        (1053)  & 22      & 10885   & 2      & 0                                  & 0                                        & 1,71          \\
jungle\_chess\_2pcs        (40997) & 45      & 4704    & 3      & 0                                  & 0                                        & 6,64          \\
KDDCup99                   (1113)  & 40      & 494020  & 23     & 0                                  & 0                                        & 6,64          \\
kin8nm                     (189)   & 9       & 8192    & 2      & 0                                  & 1                                        & 2,41          \\
leukemia                   (1104)  & 7130    & 72      & 2      & 0                                  & 1                                        & 0,47          \\
lymphoma\_2classes         (1101)  & 4027    & 45      & 2      & 0                                  & 1                                        & 2,81          \\
MagicTelescope             (1120)  & 11      & 19020   & 2      & 0                                  & 0                                        & 0,34          \\
mfeat-pixel                (20)    & 241     & 2000    & 2      & 0                                  & 0                                        & 1,00          \\
mnist\_784                 (554)   & 720     & 70000   & 10     & 0                                  & 0                                        & 1,53          \\
openml\_phpJNxH0q          (15)    & 10      & 699     & 2      & 0                                  & 0                                        & 1,00          \\
page-blocks                (30)    & 11      & 5473    & 2      & 0                                  & 1                                        & 3,60          \\
php0FyS2T                  (1492)  & 65      & 1600    & 100    & 0                                  & 0                                        & 0,22          \\
php3CTpvq                  (1509)  & 5       & 149332  & 22     & 0                                  & 0                                        & 0,97          \\
php5OMDBD                  (40971) & 23      & 1000    & 30     & 0                                  & 6                                        & 1,16          \\
php5s7Ep8                  (40982) & 28      & 1941    & 7      & 0                                  & 0                                        & 1,86          \\
php7KLval                  (1547)  & 21      & 1000    & 2      & 0                                  & 0                                        & 0,59          \\
phpB0xrNj                  (300)   & 618     & 7797    & 26     & 0                                  & 0                                        & 1,58          \\
phpbL6t4U                  (1476)  & 129     & 13910   & 6      & 0                                  & 1                                        & 0,11          \\
phpchCuL5                  (40966) & 81      & 1080    & 8      & 0                                  & 0                                        & 1,71          \\
phpCsX3fx                  (1491)  & 65      & 1600    & 100    & 0                                  & 1                                        & 0,48          \\
phpdo58hj                  (1562)  & 4703    & 64      & 2      & 0                                  & 0                                        & 3,32          \\
phpdReP6S                  (1487)  & 73      & 2534    & 2      & 0                                  & 0                                        & 2,30          \\
phpEZ030X                  (1561)  & 3722    & 64      & 2      & 0                                  & 0                                        & 2,48          \\
phpfLuQE4                  (1485)  & 501     & 2600    & 2      & 0                                  & 0                                        & 1,00          \\
phpfrJpBS                  (1568)  & 9       & 12958   & 4      & 0                                  & 0                                        & 1,00          \\
phpGReJjU                  (40985) & 4       & 45781   & 20     & 0                                  & 1                                        & 4,70          \\
phpGUrE90                  (1494)  & 42      & 1055    & 2      & 0                                  & 22                                       & 1,00          \\
phphQEck0                  (1502)  & 4       & 245057  & 2      & 0                                  & 1                                        & 1,00          \\
phpHyLSNF                  (1515)  & 1083    & 571     & 20     & 0                                  & 26                                       & 0,48          \\
phpkIxskf                  (1461)  & 17      & 45211   & 2      & 0                                  & 1                                        & 2,19          \\
phpmcGu2X                  (1468)  & 857     & 1080    & 9      & 0                                  & 1                                        & 0,94          \\
phpmPOD5A                  (4135)  & 10      & 32769   & 2      & 50                                 & 0                                        & 0,71          \\
phpn1jVwe                  (310)   & 7       & 11183   & 2      & 0                                  & 0                                        & 1,57          \\
phpN4gaxw                  (1477)  & 130     & 13910   & 6      & 0                                  & 0                                        & 0,16          \\
phpNevWWL                  (40477) & 27      & 2800    & 5      & 0                                  & 0                                        & 1,71          \\
phpoOxxNn                  (1493)  & 65      & 1599    & 100    & 0                                  & 9                                        & 1,72          \\
phpoW7Dbi                  (1566)  & 101     & 1212    & 2      & 0                                  & 0                                        & 2,55          \\
phpPbCMyg                  (1475)  & 52      & 6118    & 6      & 0                                  & 0                                        & 2,55          \\
phprAeXmK                  (4535)  & 42      & 299285  & 2      & 0                                  & 1                                        & 0,94          \\
phpSZJq5T                  (1514)  & 1088    & 360     & 10     & 0                                  & 1                                        & 4,70          \\
phptd5jYj                  (1501)  & 37      & 5100    & 2      & 0                                  & 1                                        & 2,64          \\
phpTJRsqa                  (40498) & 257     & 1593    & 10     & 0                                  & 0                                        & 0,32          \\
phpvcoG8S                  (1169)  & 12      & 4898    & 7      & 0                                  & 9                                        & 0,52          \\
phpVeNa5j                  (1497)  & 8       & 539383  & 2      & 0                                  & 1                                        & 0,98          \\
phpvtdNPU                  (1079)  & 25      & 5456    & 4      & 0                                  & 0                                        & 4,25          \\
phpWfYmlu                  (1496)  & 21      & 7400    & 2      & 0                                  & 9                                        & 0,79          \\
phpxijhaP                  (1507)  & 22278   & 95      & 5      & 0                                  & 0                                        & 0,96          \\
phpYLeydd                  (4538)  & 21      & 7400    & 2      & 0                                  & 0                                        & 3,32          \\
phpZrCzJR                  (40900) & 33      & 9873    & 5      & 0                                  & 0                                        & 1,22          \\
pokerhand-normalized       (155)   & 11      & 829201  & 10     & 0                                  & 0                                        & 3,32          \\
schizo                     (466)   & 14      & 340     & 2      & 0                                  & 1                                        & 5,52          \\
shuttle                    (40685) & 10      & 58000   & 7      & 0                                  & 0                                        & 3,99          \\
solar-flare\_1             (40686) & 13      & 315     & 5      & 0                                  & 0                                        & 0,74          \\
synthetic\_control         (377)   & 61      & 600     & 6      & 0                                  & 29                                       & 0,34          \\
tumors\_C                  (1107)  & 7130    & 60      & 2      & 0                                  & 4                                        & 1,56          \\
umistfacescropped          (41084) & 10305   & 575     & 20     & 0                                  & 3                                        & 0,99          \\
vowel                      (307)   & 14      & 990     & 2      & 0                                  & 0                                        & 1,42          \\
wine-quality-red           (40691) & 12      & 1599    & 6      & 0                                  & 11                                       & 0,99          \\
aaaData\_for\_UCI\_named   (43007) & 14      & 10000   & 2      & 0                                  & 0                                        & 1,59          \\
\hline

\caption{\textcolor{red}{List of all tested datasets including information about (abbreviated) name and OpenML id for each data set together with the number of classes, the number of features, the number of instances, how many values are missing in total (Missing values), number of categorical features per sample, and the class entropy.}}
\end{longtable}\label{Tbl:AutoMLDatasets}

\normalfontsize

\section{Framework and Source Code}\label{Appen:FrameworksBaseline}
Table~\ref{Table:AutoMLFrameworksVer} lists the Github repositories of all the open-source AutoML frameworks considered in this work. Some frameworks are still under active development and may differ from the evaluated versions.

\begin{table}[!ht]
\begin{center}
\caption{Source code repositories for all used AutoML frameworks}
 \begin{tabular}{||c c ||} 
 \hline
 AutoML Framework & Source Code   \\ 
 \hline\hline
 AutoSKlearn &   https://automl.github.io/auto-sklearn/ \\ 
 \hline
 TPOT &   https://github.com/EpistasisLab/tpot\\
 \hline
 ATM &   https://github.com/HDI-Project/ATM \\
 \hline
 Recipe &   https://github.com/laic-ufmg/Recipe\\
 \hline
 AutoWeka &  https://github.com/automl/AutoWeka \\  
 \hline
 SmartML & https://github.com/DataSystemsGroupUT/SmartML\\
 \hline
\end{tabular}\label{Table:AutoMLFrameworksVer}
\end{center}
\end{table}

\section{Cut-off time Budget}\label{Appen:Cut-offtime}
We tested the cutt-off timeouts of 4 and 8 hours on 14 randomly selected datasets. Table~\ref{tab:cutoff} shows the mean performance difference between the 8 and 4 hours (Avg. diff) over the 14 datasets. Additionally, we report the results of the Wilcoxon signed-rank test to determine if a statistically significant difference in performance exists between the AutoML frameworks over the two-time budgets.

\begin{table}[!htbp]
\begin{tabular}{ccc}
\hline
\textbf{Framework} & \textbf{P value} & \textbf{Avg. diff} \\ \hline
AutoSKlearn                                  & 0.084                                      & -0.026                                       \\
AutoSKlearn-e                                & \textbf{0.039}            & -0.025                                       \\
AutoSKlearn-m                                & 0.382                                      & -0.031                                       \\
AutoSKlearn-v                                & 0.272                                      & -0.008                                       \\
AutoWeka                                     & 0.133                                      & -0.005                                       \\
Recipe                                       & 0.480                                      & 0.007                                        \\
SmartML                                      & 0.594                                      & -0.003                                       \\
SmartML-e                                    & 0.753                                      & -0.009                                       \\
TPOT                                         & 0.092                                      & -0.050                                       \\ \hline
\end{tabular}
\caption{Performance comparison between the 8 and 4 hours budgets on 14 randomly selected datasets.}
\label{tab:cutoff}
\end{table}

\section{General Performance Evaluation}\label{Appen:GeneralPerformanceEval}
\begin{figure}[!htb]
  \centering
  \includegraphics[width=0.73\linewidth]{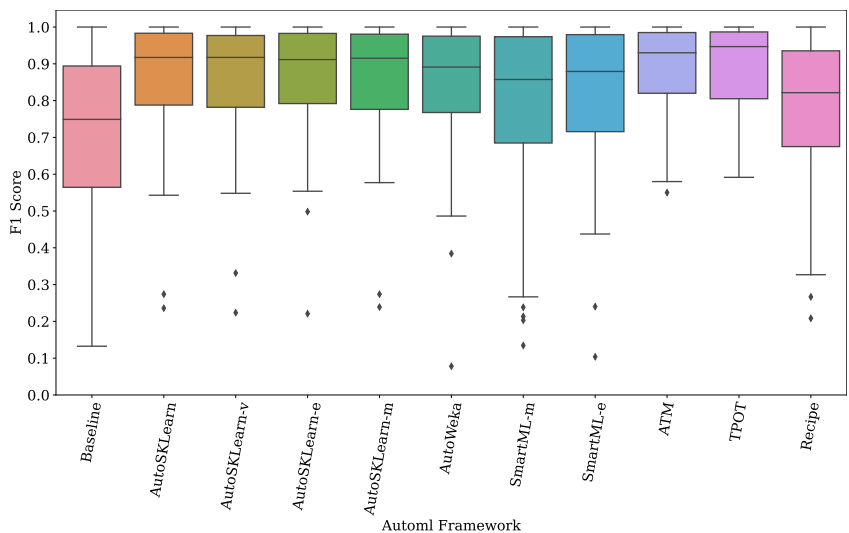}
  \caption{Average performance of all frameworks (10 Min) compared to the baseline.}
  \label{FIG:avg_performance_10}
\end{figure}

\begin{figure}[!t]
  \centering
  \includegraphics[width=0.67\linewidth]{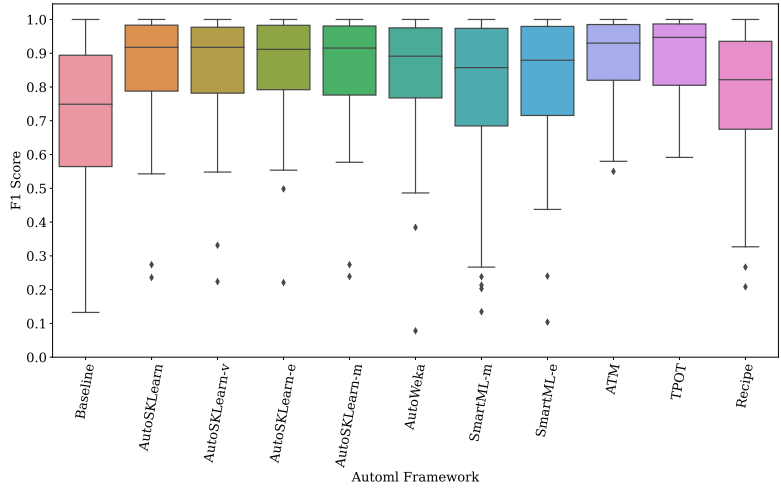}
  \caption{Average performance of all frameworks (30 Min) compared to the baseline.}
  \label{FIG:avg_performance_30}
\end{figure}

\begin{figure}[!t]
  \centering
  \includegraphics[width=0.67\linewidth]{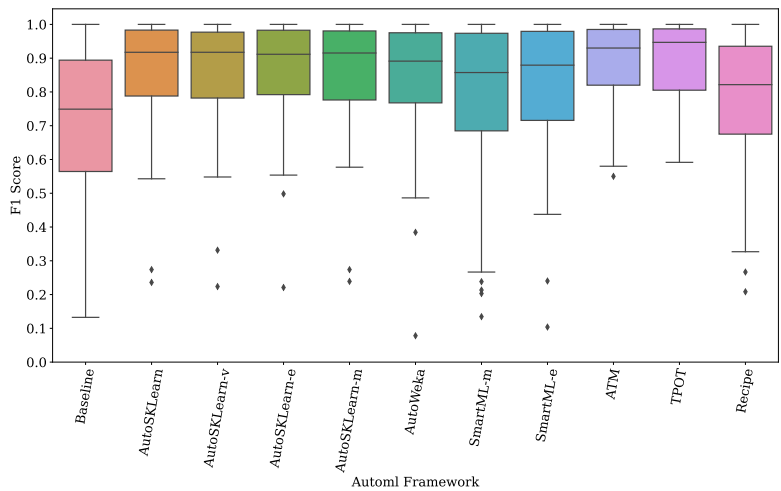}
  \caption{Average performance of all frameworks (60 Min) compared to the baseline.}
  \label{FIG:avg_performance_60}
\end{figure}

\begin{figure}[!h]
  \centering
  \includegraphics[width=0.67\linewidth]{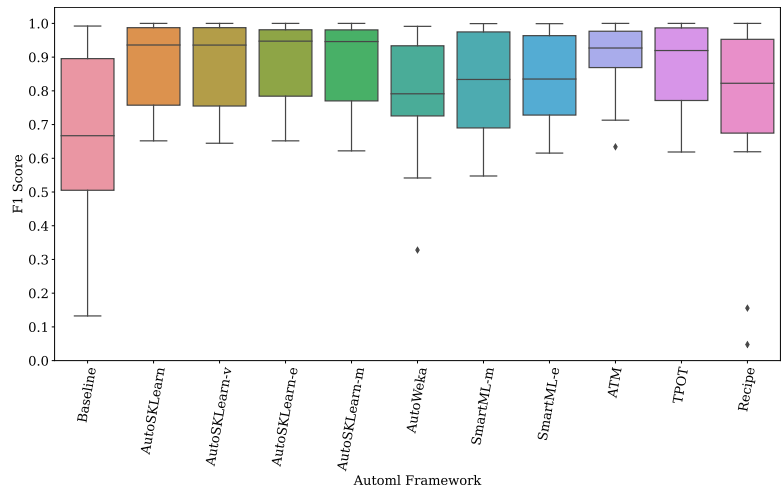}
  \caption{Performance of the final pipeline for datasets with large number of features and large number of instances.}
  \label{FIG:avg_performance_large_features_large_instances}
\end{figure}

\begin{figure}[!h]
  \centering
  \includegraphics[width=0.67\linewidth]{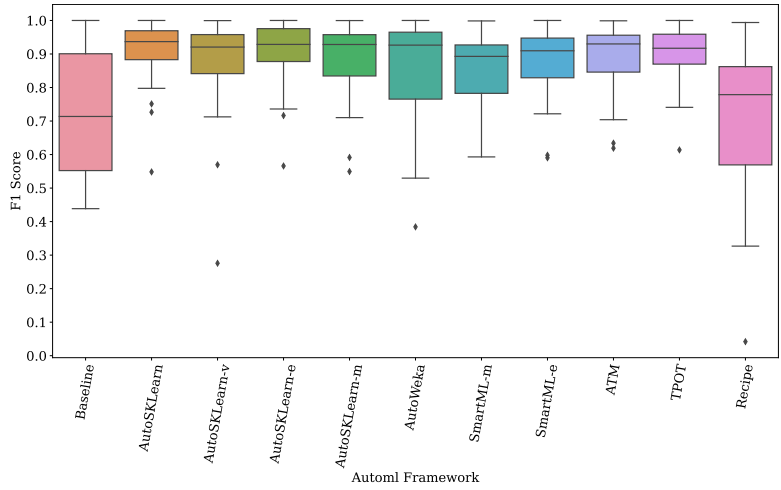}
  \caption{Performance of the final pipeline for datasets with small number of features and small number of instances.}
  \label{FIG:avg_performance_small_features_small_instances}
\end{figure}

\begin{figure}[!b]
  \centering
  \includegraphics[width=0.67\linewidth]{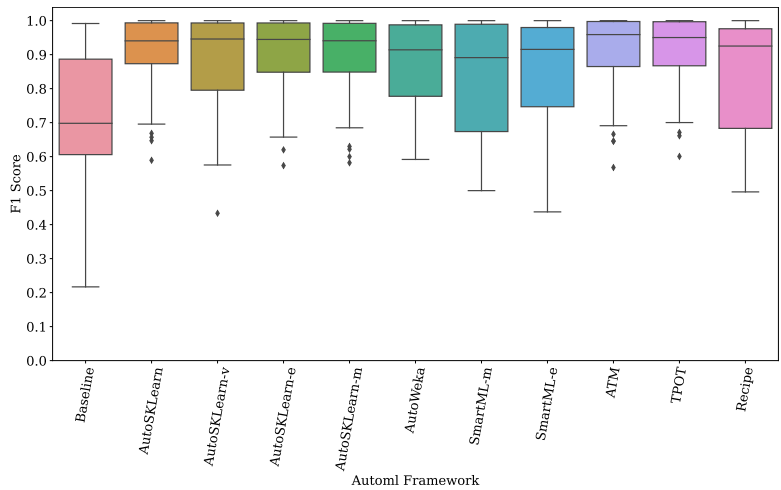}
  \caption{Performance of the final pipeline for datasets with small number of features and large number of instances.}
  \label{FIG:avg_performance_small_features_large_instances}
\end{figure}

~\Cref{FIG:avg_performance_10,FIG:avg_performance_30,FIG:avg_performance_60} shows the average performance of all frameworks for time budgets 10, 30, and 60 minutes compared the average performance of the baseline. ~\Cref{FIG:avg_performance_binary_class,FIG:avg_performance_large_features_large_instances,FIG:avg_performance_small_features_small_instances,FIG:avg_performance_small_features_large_instances} show the AutoML framwworks' average performance on subsets of the datasets with special characteristics, namely binary-class, large number of features and instances, small number of features and instances, and small number of features and large number of instances.

\begin{figure}[!b]
  \centering
  \includegraphics[width=0.67\linewidth]{Figures/box_plot_240 MinBinary Classification.pdf}
  \caption{Performance of the final pipeline for binary-class datasets.}
  \label{FIG:avg_performance_binary_class}
\end{figure}

\section{Impact of Meta Learning}
\label{append:metafeatures}
Table~\ref{tab:metafeatures} lists a total of 42 meta-features including simple, information-theoretic and statistical meta-features.

\begin{table}[!htbp]
\footnotesize
\begin{minipage}{\textwidth}
\begin{center}
\scalebox{1}{
\begin{tabular}{llll}
  \toprule
  Name & Formula & Rationale &  Additional Variants \\
  \midrule
   Nr instances & $n$ & Speed, Scalability \citep{michie1994machine} & $p/n$, $log(n)$  \\ 
   Nr features & $p$ & Curse of dimensionality \citep{michie1994machine} & $log(p)$, Nr/$\mu$/$\sigma$(\pi_{i}) \\ 
   Nr classes & $c$ & Complexity, imbalance \citep{michie1994machine} & min/max/$\sigma(\frac{c}{n})$ \\ 
   Nr missing values & $m$ & Imputation effects \citep{kalousis2002algorithm} &  \\ 
   \midrule
   Skewness & $\frac{E(X-\mu_{X})^{3}}{\sigma_{X}^{3}}$ & Feature normality \citep{michie1994machine} & min,max,$\mu$,$\sigma$,$q_{1},q_{3}$\\
  Kurtosis & $\frac{E(X-\mu_{X})^{4}}{\sigma_{X}^{4}}$ & Feature normality \citep{michie1994machine} & min,max,$\mu$,$\sigma$,$q_{1},q_{3}$\\
   Correlation & $\rho_{X_{1}X_{2}}$ & Feature interdependence \citep{michie1994machine} & min,max,$\mu$,$\sigma$\\
   \midrule
   Class entropy  & $H(\texttt{C})$ & Class imbalance \citep{michie1994machine} & $H(\texttt{C}) / \mu(MI(\texttt{C},\texttt{X}))$\\
   Norm. entropy & $\frac{H(\texttt{X})}{log_{2}n}$ & Feature informativeness \citep{castiello2005meta} & min,max,$\mu$,$\sigma$ \\
   Mutual inform. & $MI(\texttt{C},\texttt{X})$ & Feature importance \citep{michie1994machine} & min,max,$\mu$,$\sigma$ \\
   Noise-signal ratio & $\frac{\overline{H(X)}-\overline{MI(C,X)}}{\overline{MI(C,X)}}$ & Noisiness of data \citep{michie1994machine} & \\
  \bottomrule
\end{tabular}
}
\end{center}
\end{minipage}
\caption{Overview of the used meta-features. Groups from top to bottom: simple, statistical, and information-theoretic. Continuous features $X$ and target $Y$ have mean $\mu_{X}$, stdev $\sigma_{X}$, variance $\sigma^{2}_{X}$. Categorical features $\texttt{X}$ and class $\texttt{C}$ have categorical values $\pi_{i}$, conditional probabilities $\pi_{i\mid j}$, joint probabilities $\pi_{i,j}$, marginal probabilities $\pi_{i+}=\sum_{j}\pi_{ij}$, entropy $H($\texttt{X}$)=-\sum_{i}\pi_{i+}log_{2}(\pi_{i+})$. \citep{vanschoren2018meta}}
\label{tab:metafeatures}
\end{table}

\section{Impact of Ensembling}\label{Appen:ensembling}

\begin{figure*}[!ht]
\centering \subfigure[10 Min.] {
    \label{FIG:Ensembling10}
    \includegraphics[width=0.47\textwidth]{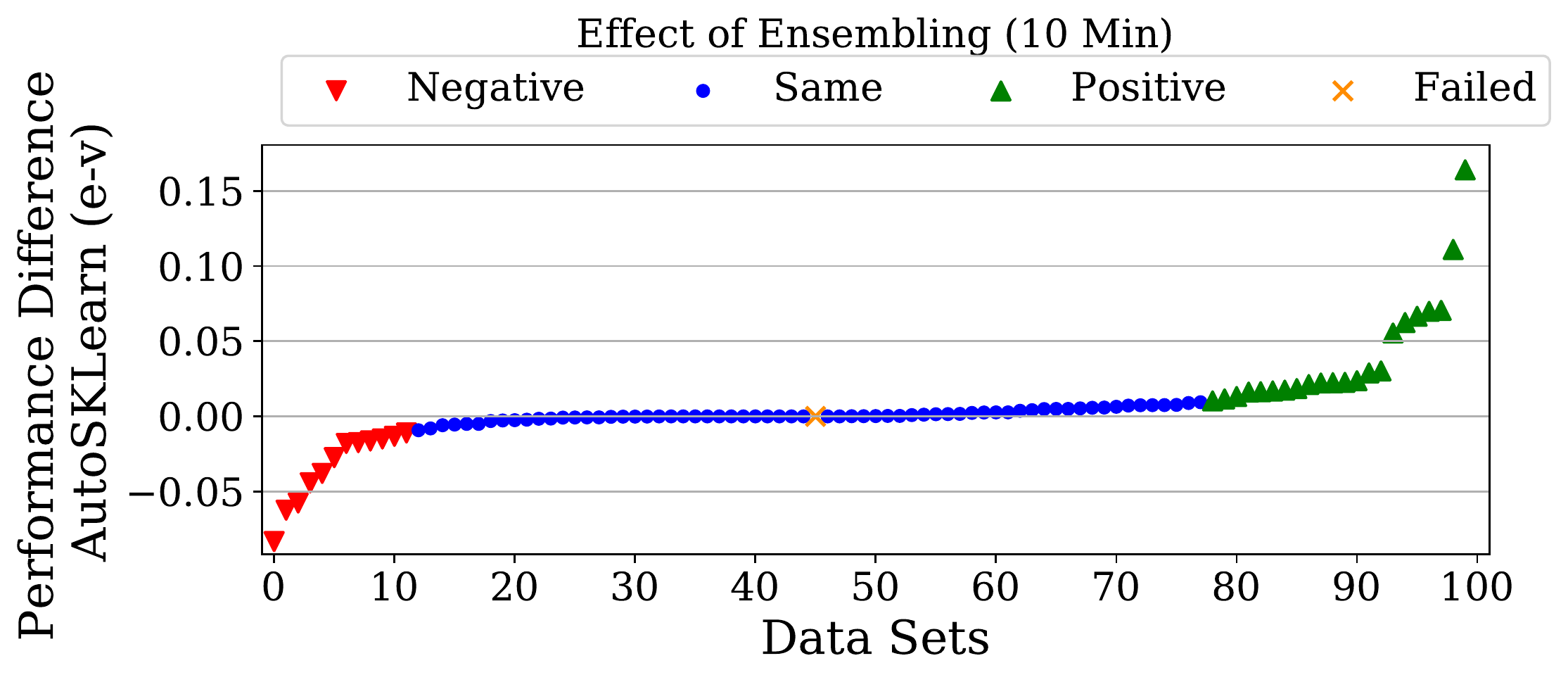}
}
\centering \subfigure[30 Min.] {
    \label{FIG:Ensembling30}
    \includegraphics[width=0.47\textwidth]{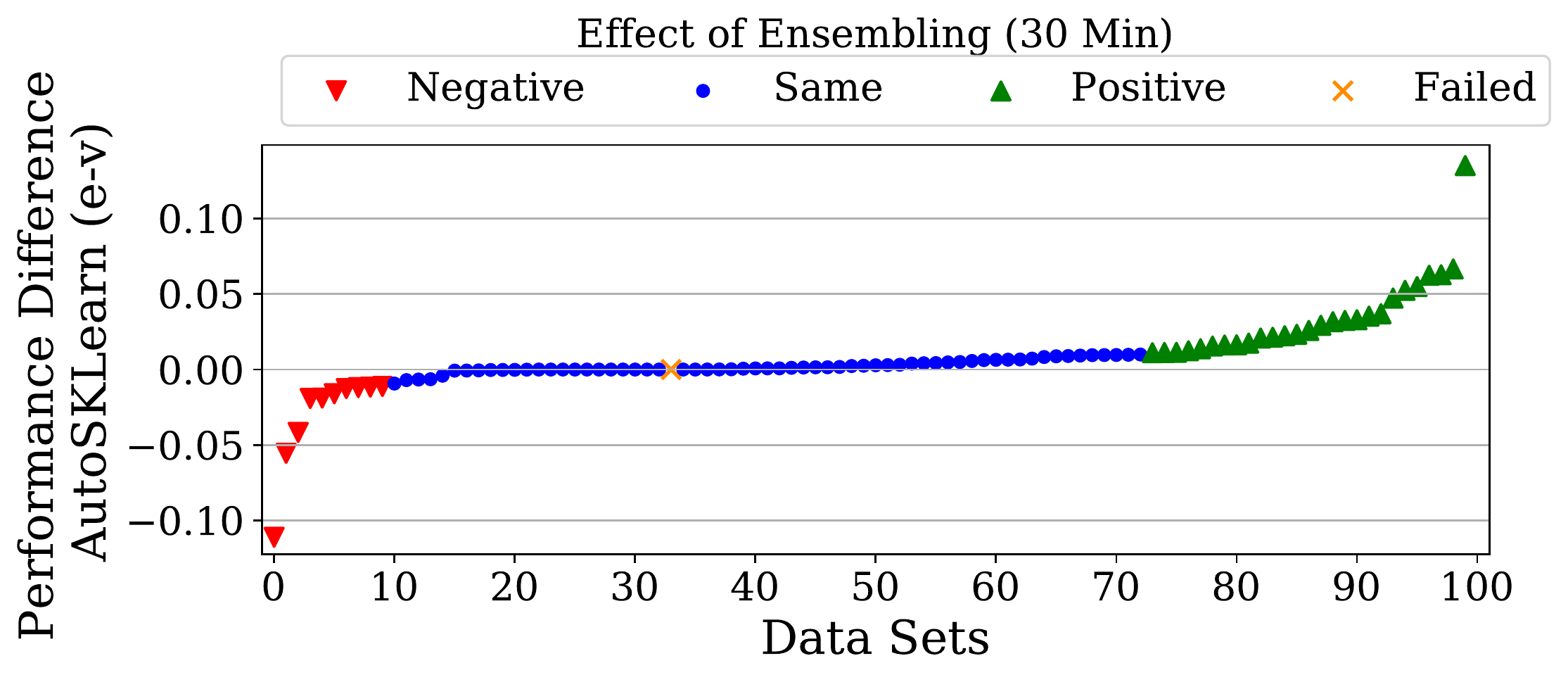}
}
\centering \subfigure[60 Min.] {
    \label{FIG:Ensembling60}
    \includegraphics[width=0.47\textwidth]{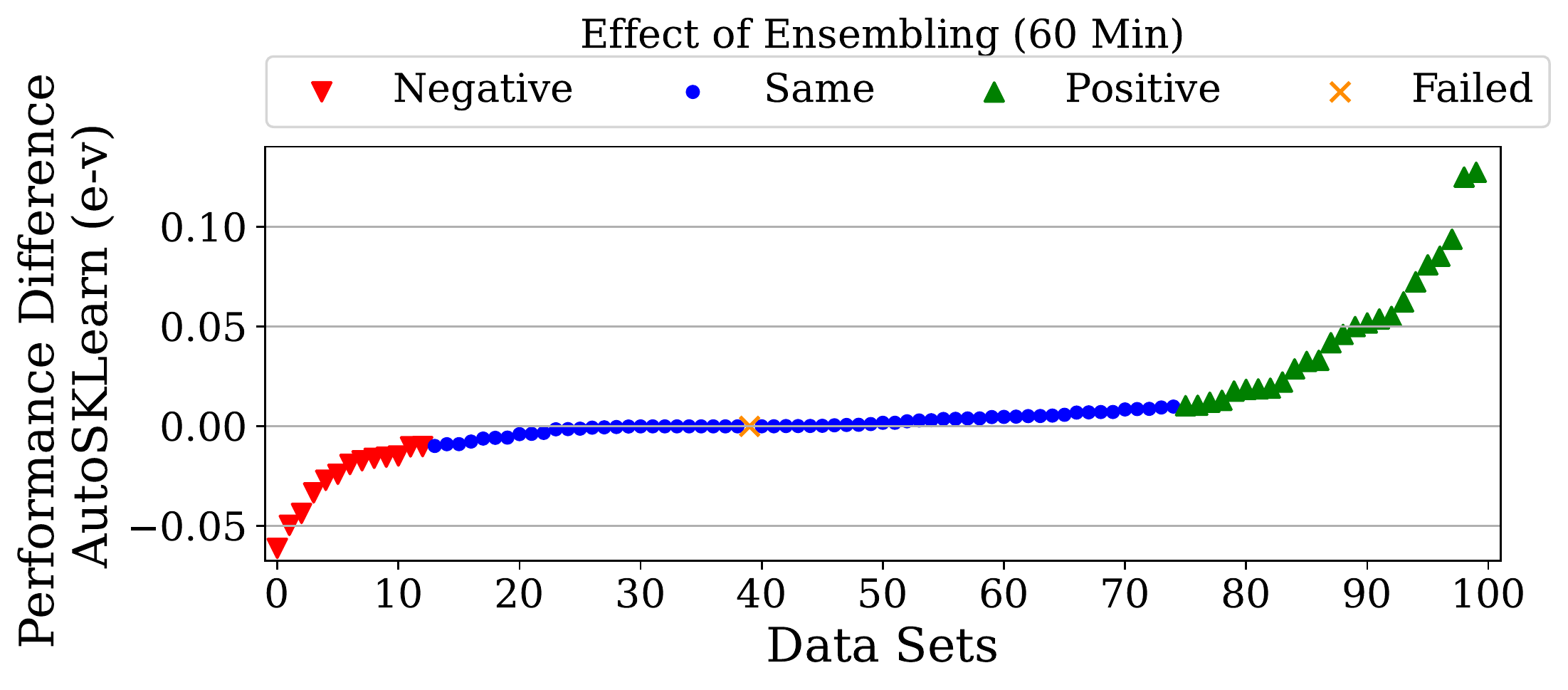}
}
\centering \subfigure[240 Min.] {
    \label{FIG:Ensembling240}
    \includegraphics[width=0.47\textwidth]{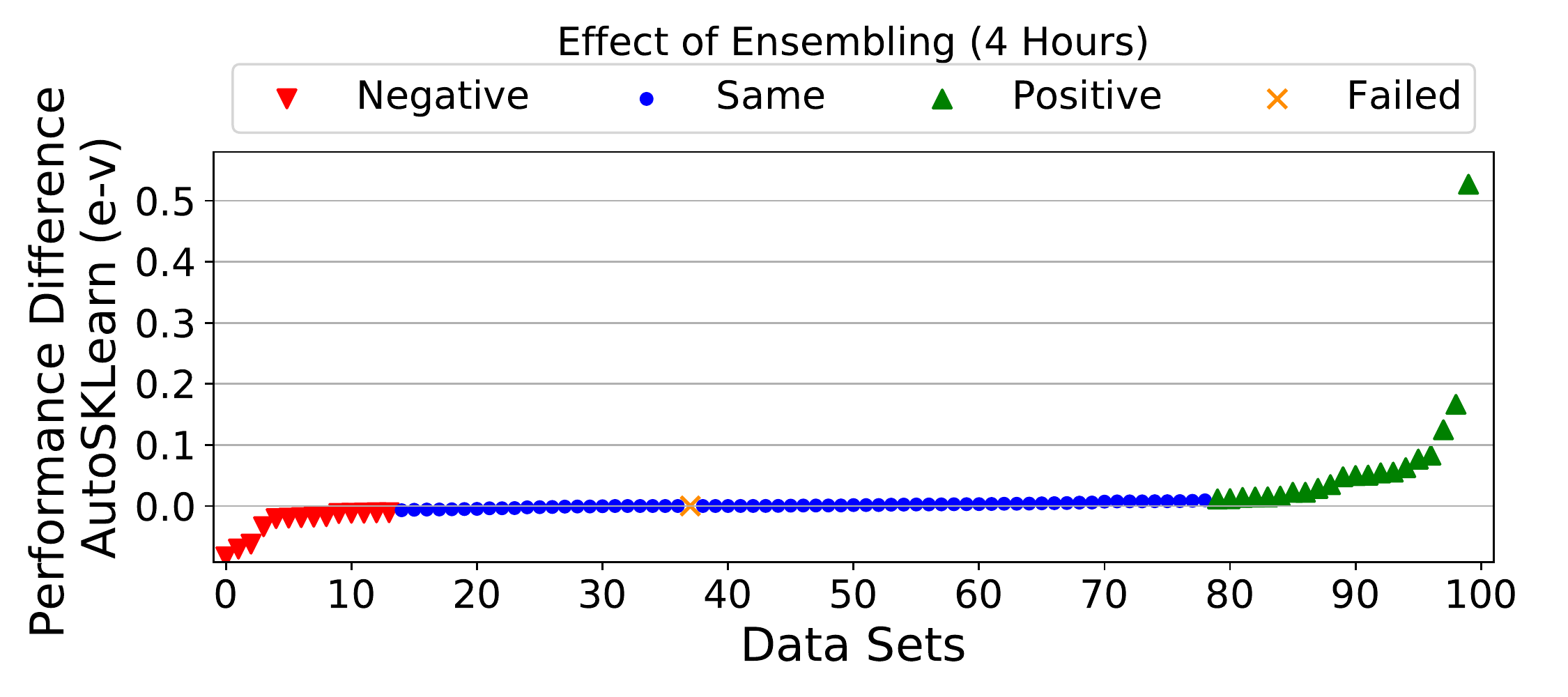}
}
\caption{The performance difference between the \texttt{AutoSKlearn-e} and \texttt{AutoSKlearn-v} over different time budgets. Green markers represent better performance with \texttt{AutoSKlearn-e}, blue markers represent comparable performance with a difference less than 1\%, red markers represent better performance with \texttt{AutoSKlearn-v}, and yellow markers represent failed runs on both versions.}
\label{FIG:Ensembling}
\end{figure*}

\begin{figure*}[th!]
\centering \subfigure[10 Min.] {
    \label{FIG:Ensembling10smart}
    \includegraphics[width=0.47\textwidth]{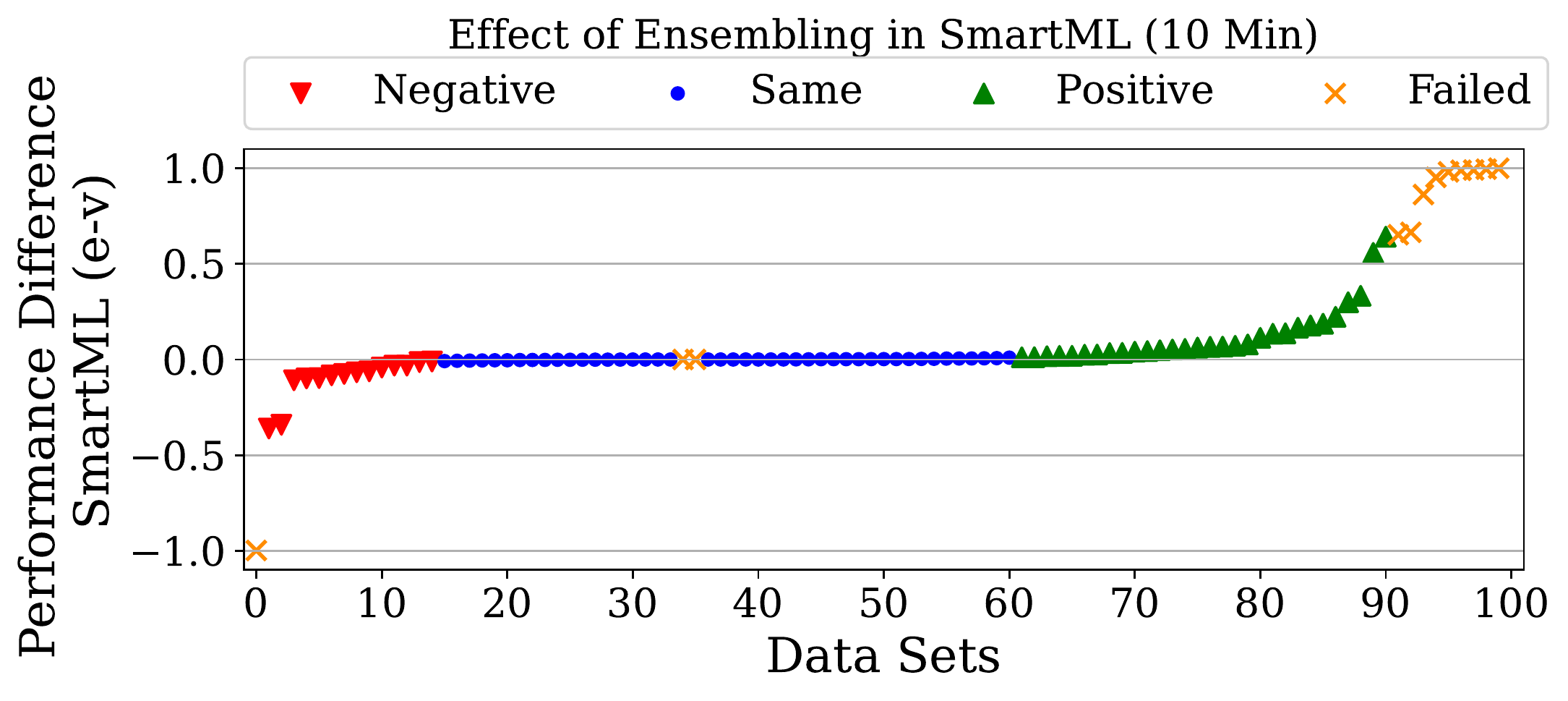}
}
\centering \subfigure[30 Min.] {
    \label{FIG:Ensembling30smart}
    \includegraphics[width=0.47\textwidth]{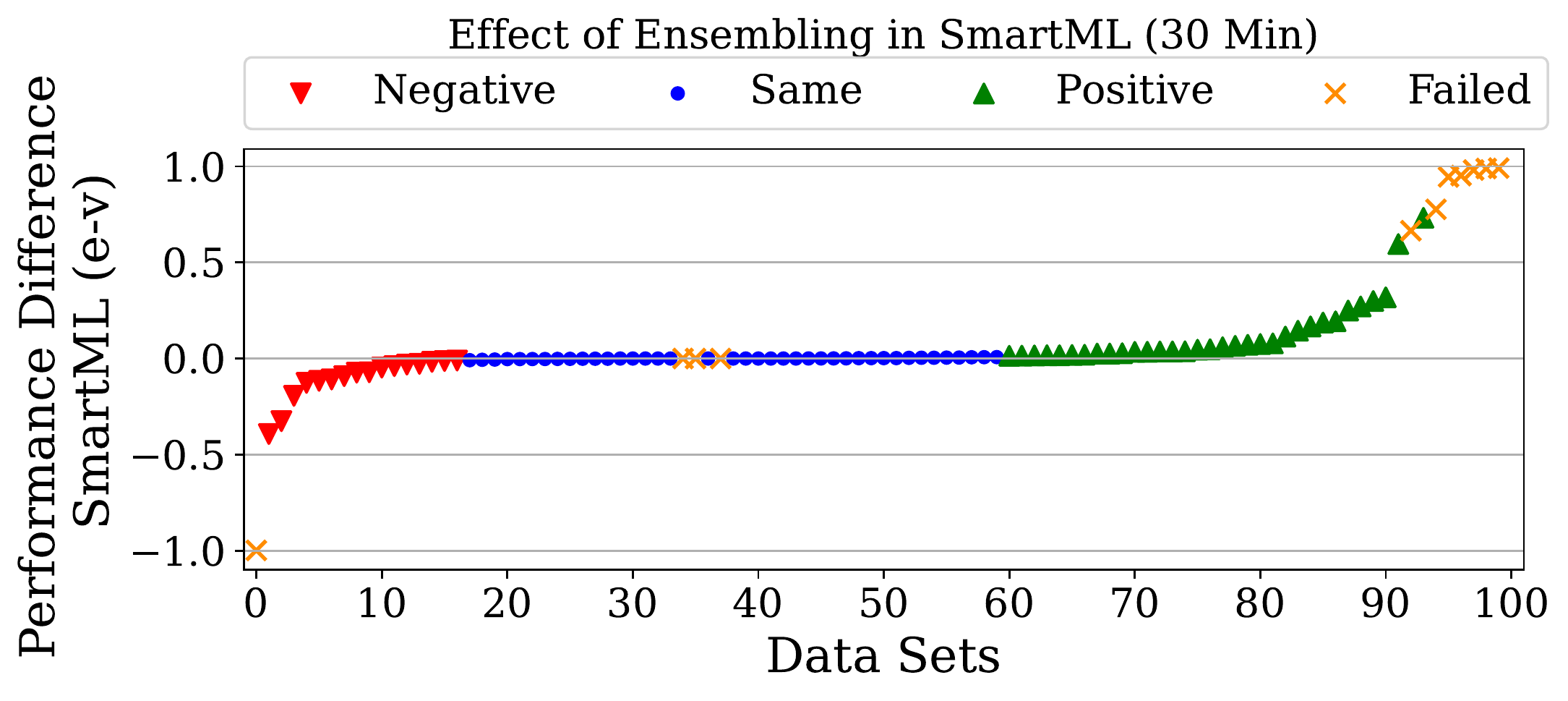}
}
\centering \subfigure[60 Min.] {
    \label{FIG:Ensembling60smart}
    \includegraphics[width=0.47\textwidth]{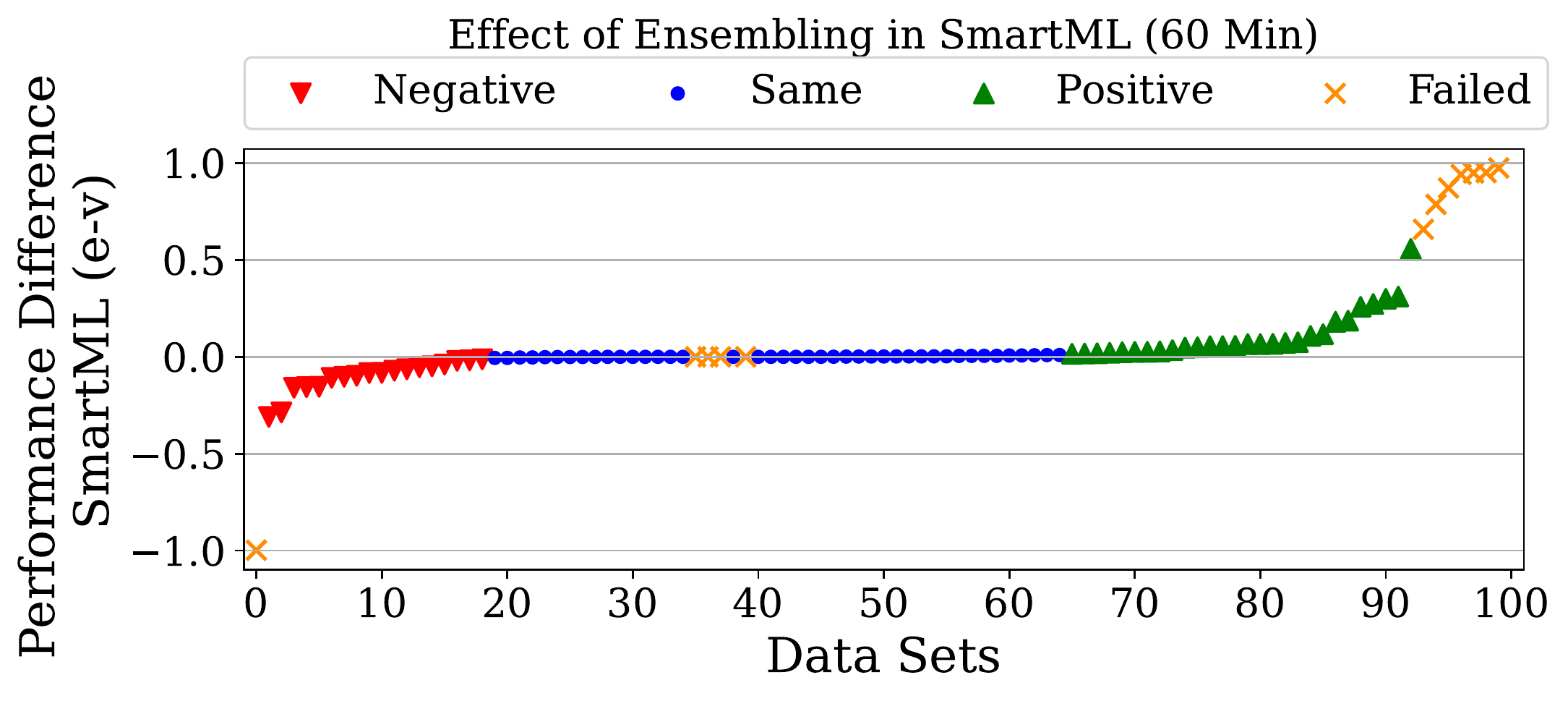}
}
\centering \subfigure[240 Min.] {
    \label{FIG:Ensembling240smart}
    \includegraphics[width=0.47\textwidth]{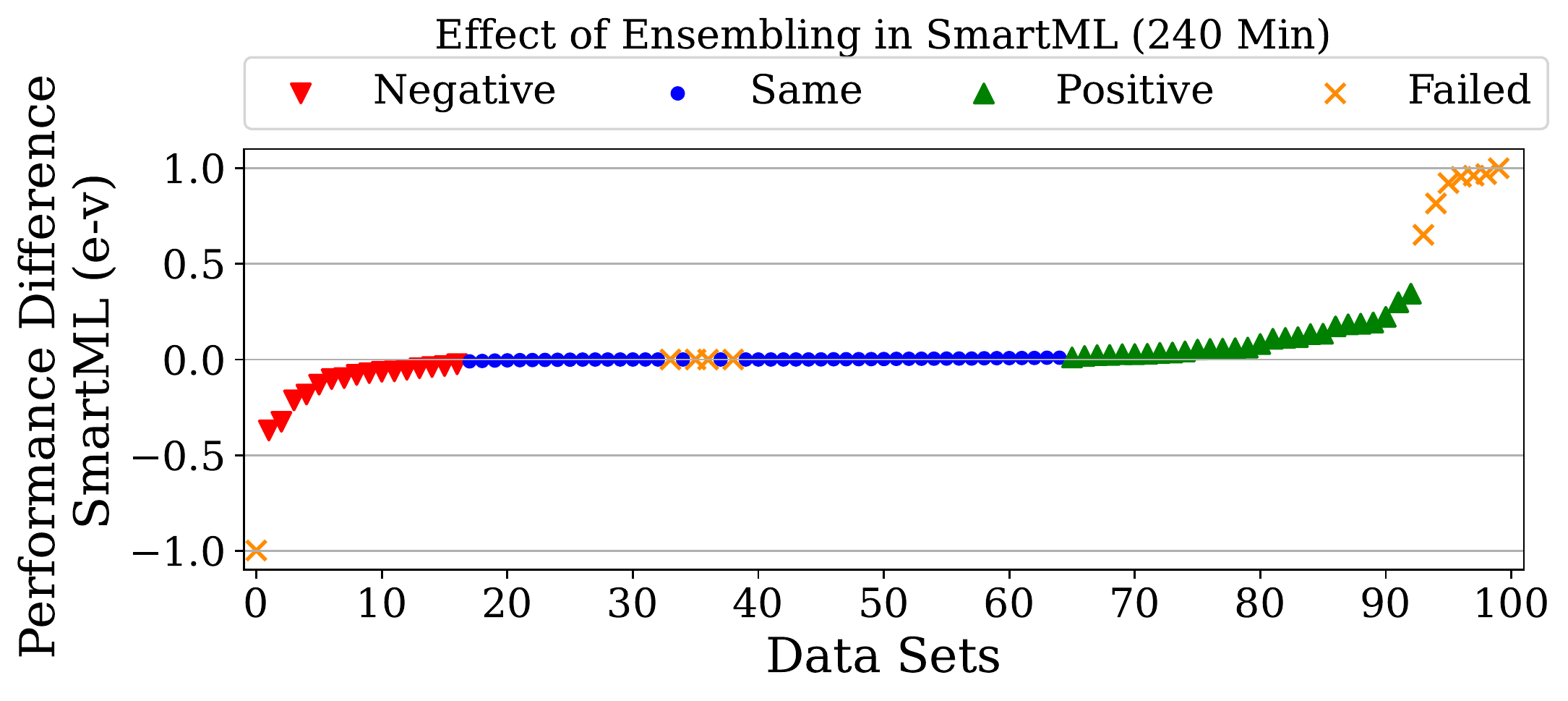}
}
\caption{\textcolor{red}{The performance difference between the \texttt{SmartML-m} and \texttt{SmartML-e}. Green markers represent better performance with \texttt{SmartML-e}, blue markers represent comparable performance with a difference
less than 1\%, red markers represent better performance with \texttt{SmartML}, yellow markers on the right represent failed runs
with \texttt{SmartML-m} but successful with \texttt{SmartML-e}, yellow markers on the left
represent failed runs with \texttt{SmartML-e} but successful with \texttt{SmartML-m} and yellow markers in the middle represent failed runs with both \texttt{SmartML-m}
and \texttt{SmartML-e}.}}
\label{FIG:Ensembling_smartML}
\end{figure*}

Figures~\ref{FIG:Ensembling} and ~\ref{FIG:Ensembling_smartML} shows the performance difference between the ensembling version and the vanilla/base version of \texttt{AutoSKlearn} and \texttt{SmartML}, respectively over 10, 30, 60 and 240 minutes time budgets.

\section{Impact of time budget}\label{App:ImpactTimeBudget}
\begin{figure*}[!ht]
\centering \subfigure[10-30 Min.] {
    \includegraphics[width=0.47\textwidth]{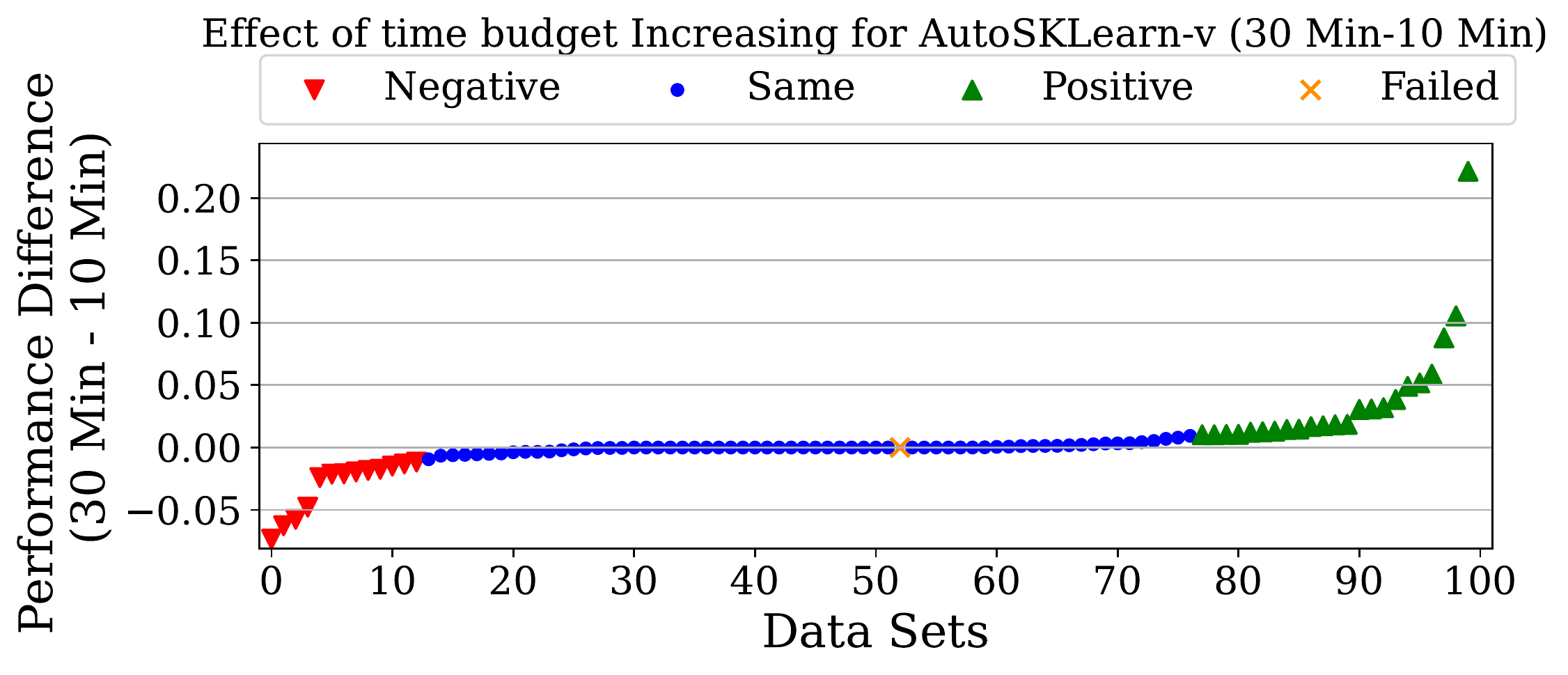}
}
\centering \subfigure[10-60 Min.] {
    \includegraphics[width=0.47\textwidth]{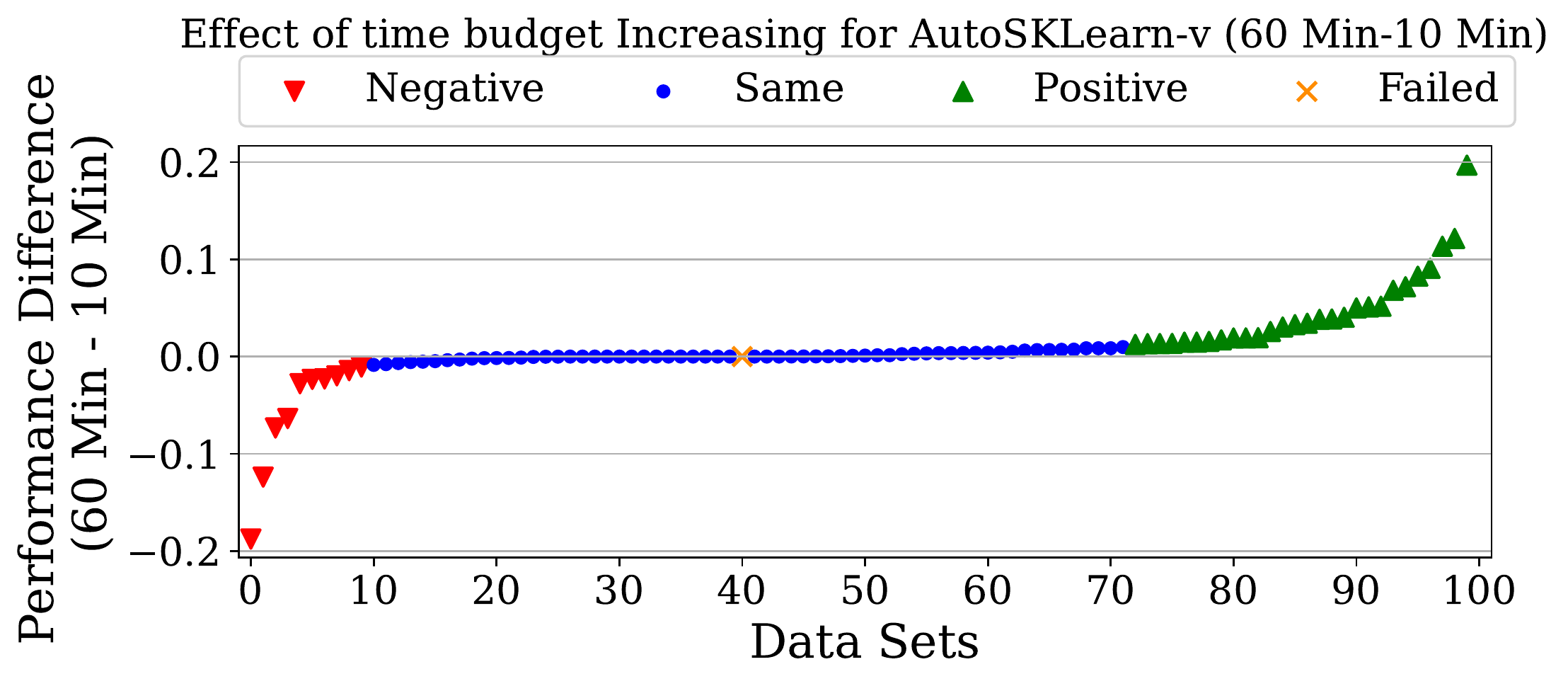}
}
\centering \subfigure[10-240 Min.] {
    \includegraphics[width=0.47\textwidth]{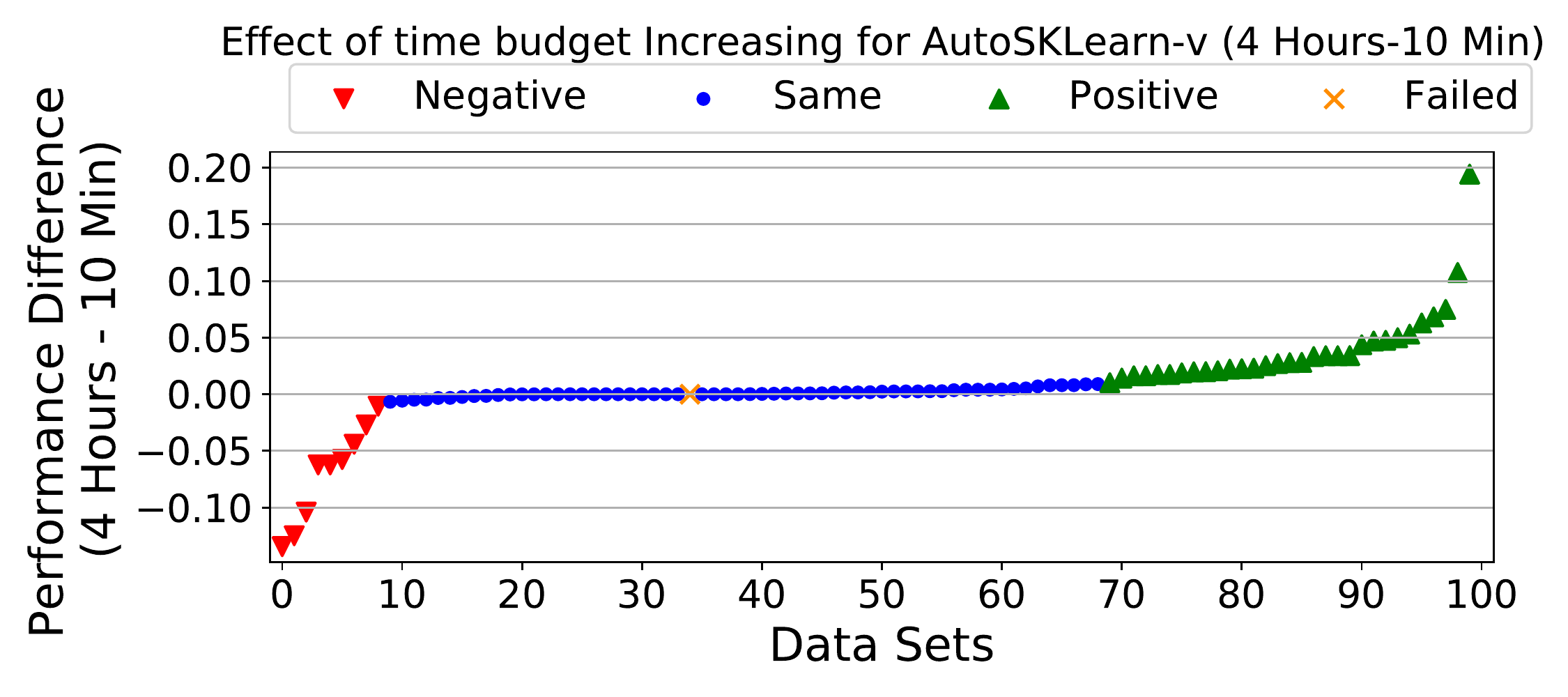}
}
\centering \subfigure[30-60 Min.] {
    \includegraphics[width=0.47\textwidth]{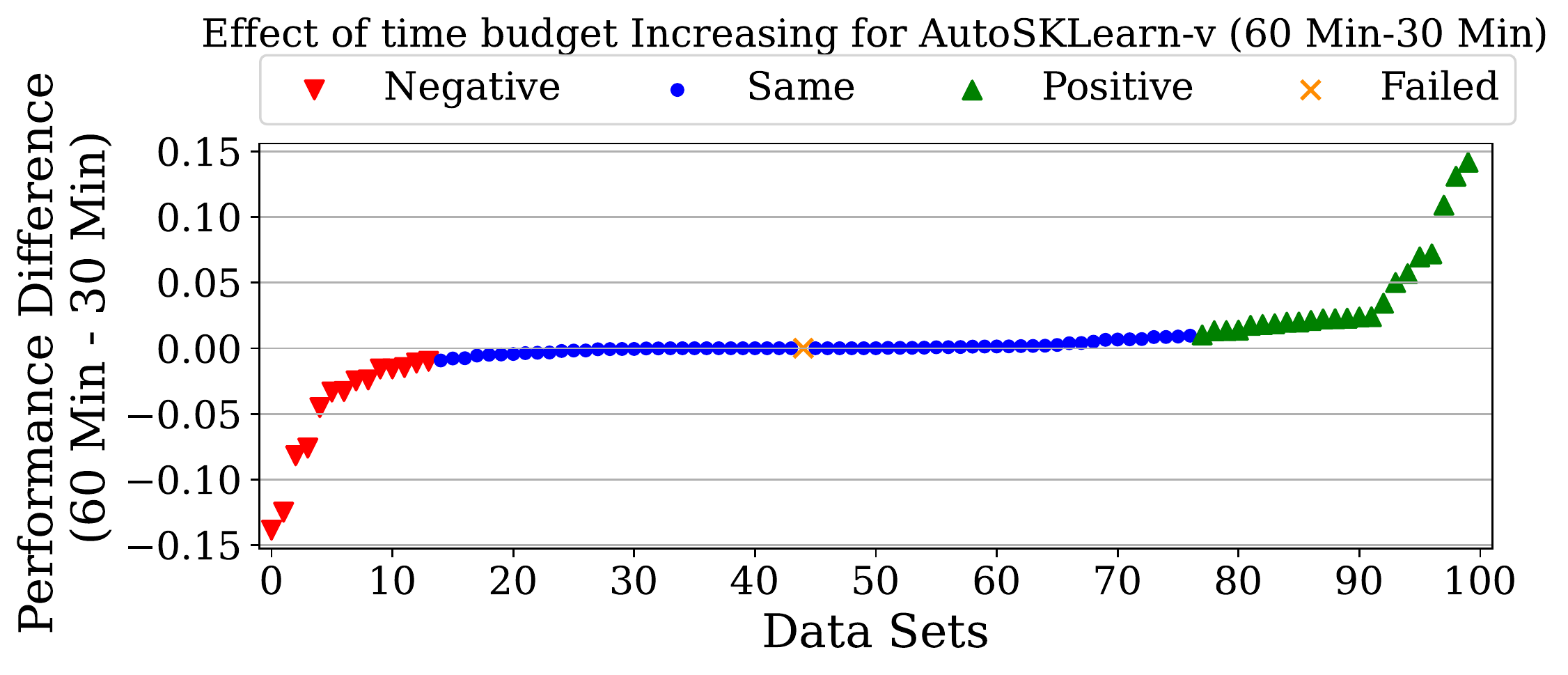}
}
\centering \subfigure[30-240 Min.] {
    \includegraphics[width=0.47\textwidth]{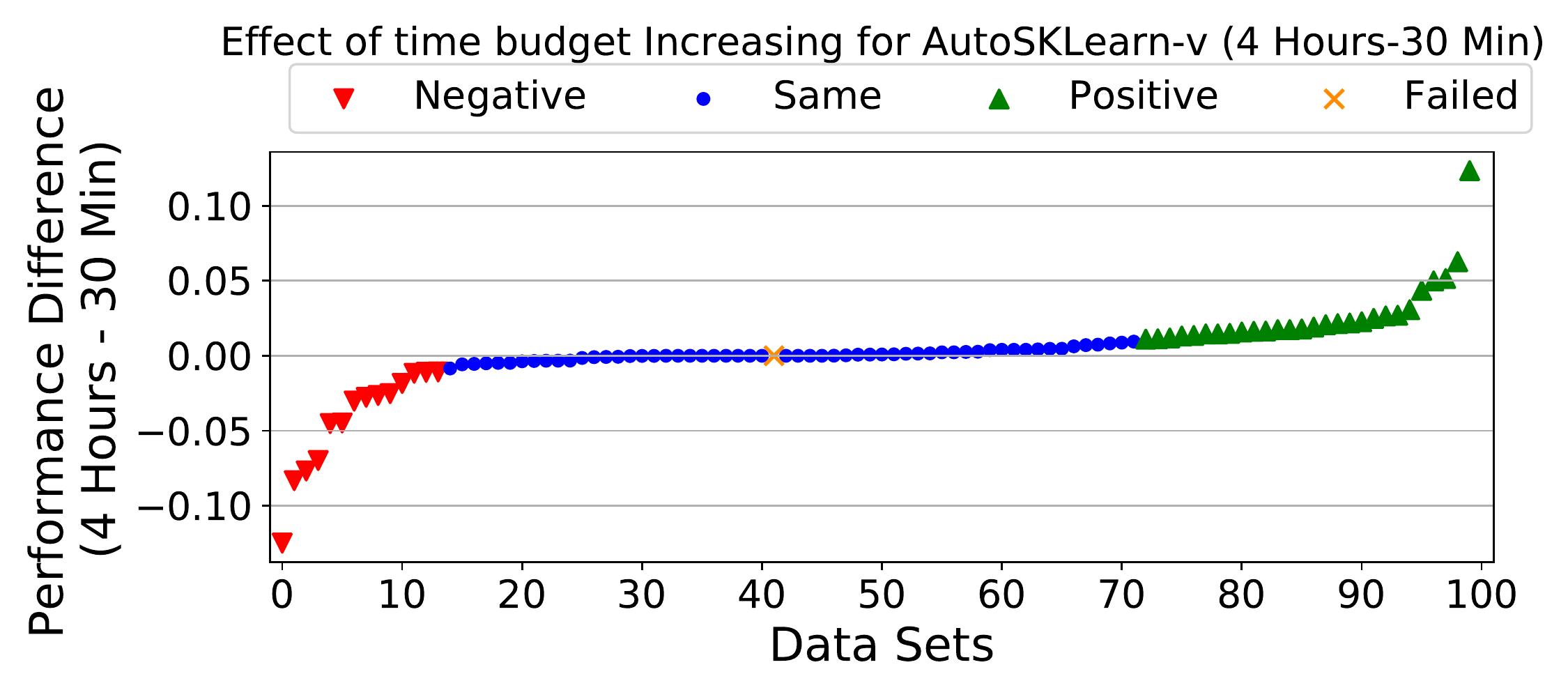}
}
\centering \subfigure[60-240 Min.] {
    \includegraphics[width=0.47\textwidth]{Figures/EffectoftimebudgetIncreasingforAutoSKlearnv4Hours30Min.pdf}
}

\caption{The impact of increasing the time budget on \texttt{AutoSKlearn-v} performance from $x$ to $y$ minutes (x-y). Green markers represent better performance with $y$ time budget, blue markers means that the difference between $x$ and $y$ is $<1$. Red markers represent better performance on $x$ time budget.}
\label{FIG:TimeBudgetSklearn-v}
\end{figure*}

\begin{figure*}[!ht]
\centering \subfigure[10-30 Min.] {
    \includegraphics[width=0.47\textwidth]{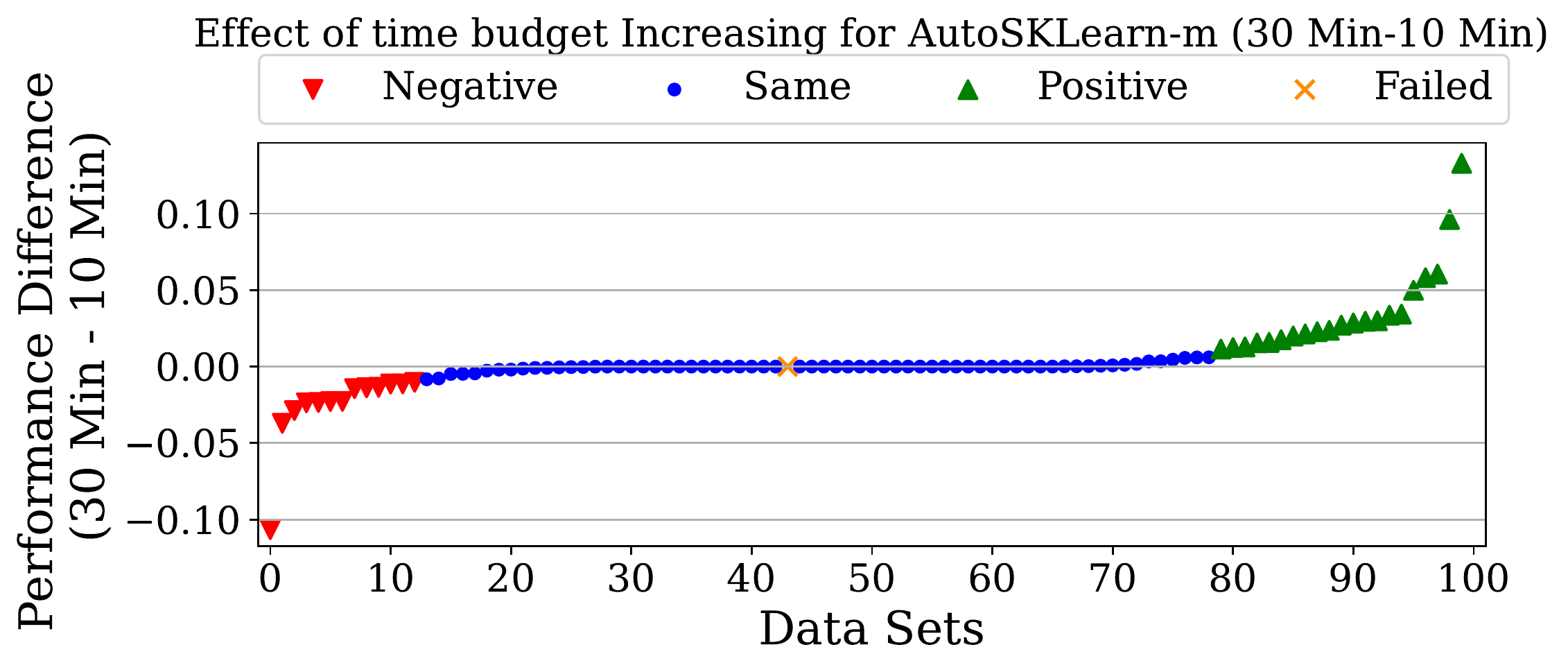}
}
\centering \subfigure[10-60 Min.] {
    \includegraphics[width=0.47\textwidth]{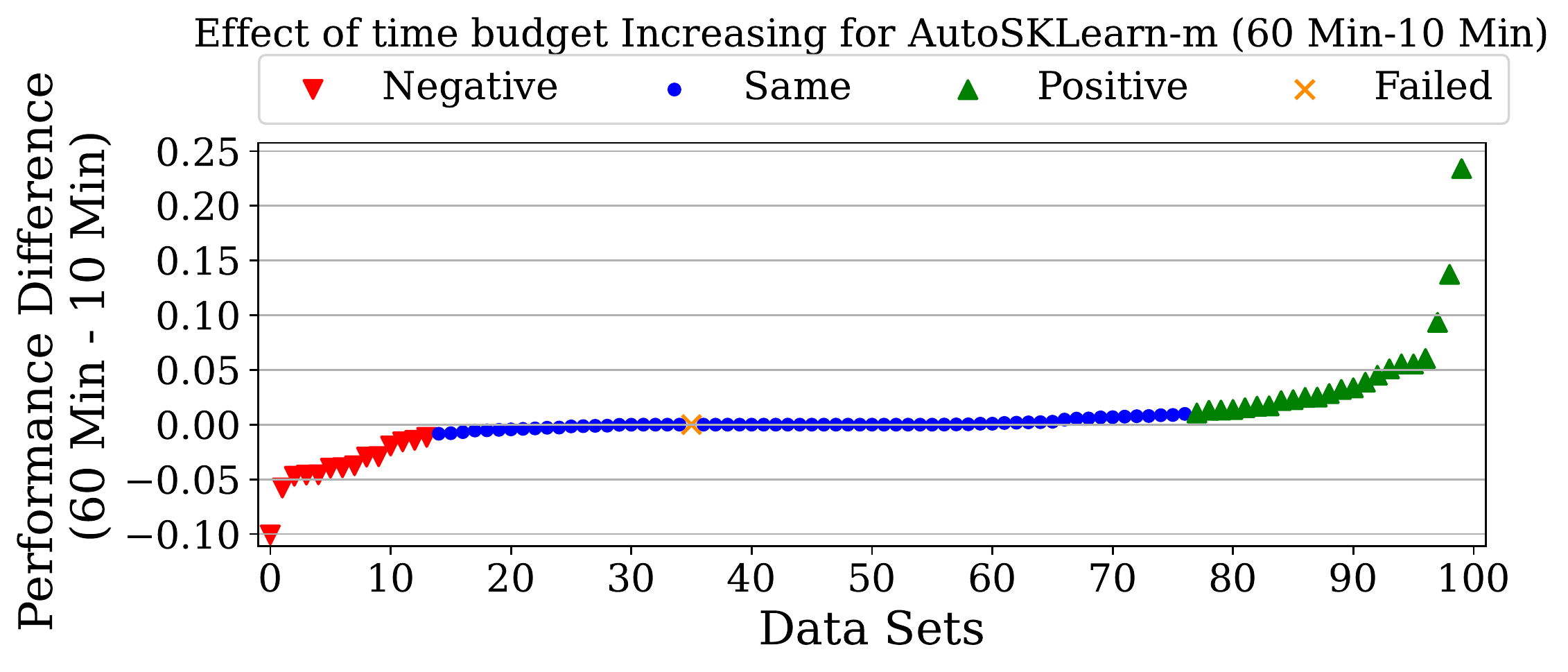}
}
\centering \subfigure[10-240 Min.] {
    \includegraphics[width=0.47\textwidth]{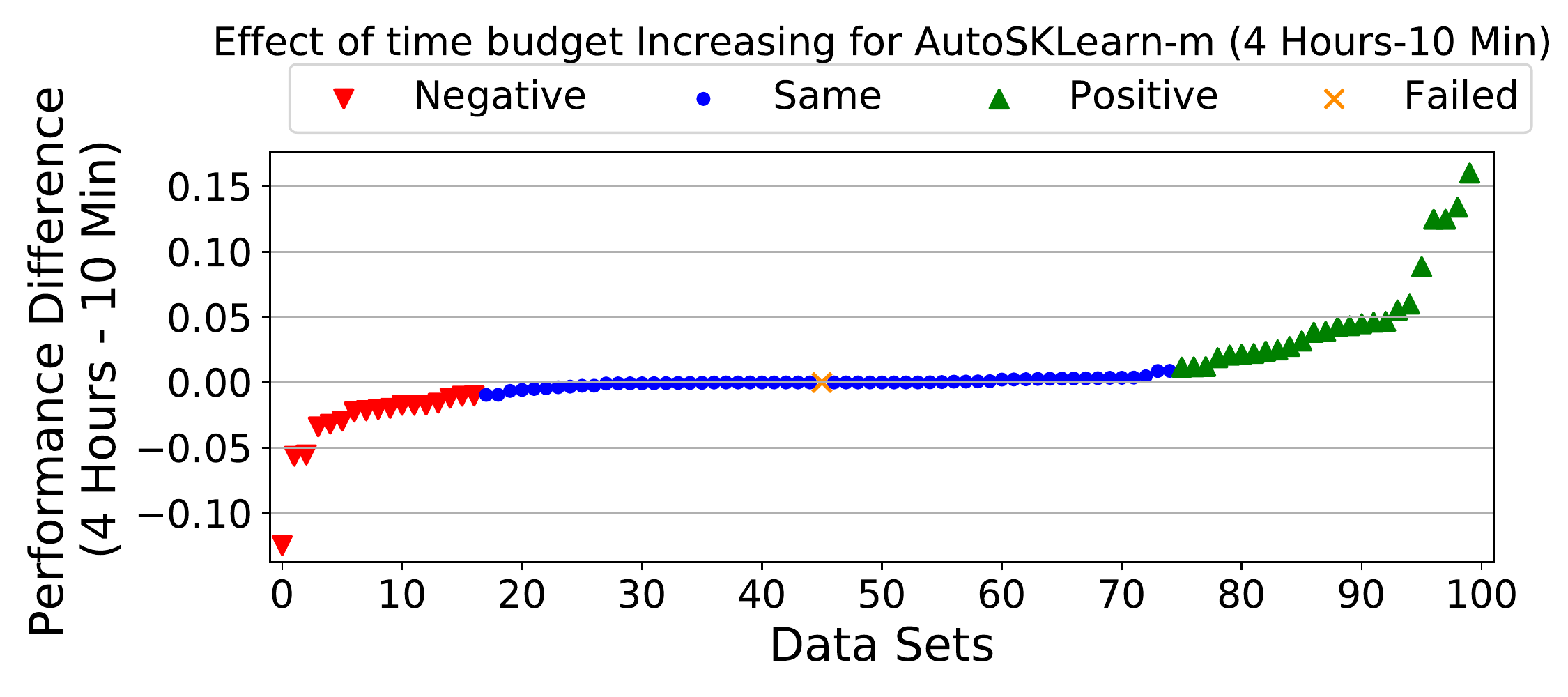}
}
\centering \subfigure[30-60 Min.] {
    \includegraphics[width=0.47\textwidth]{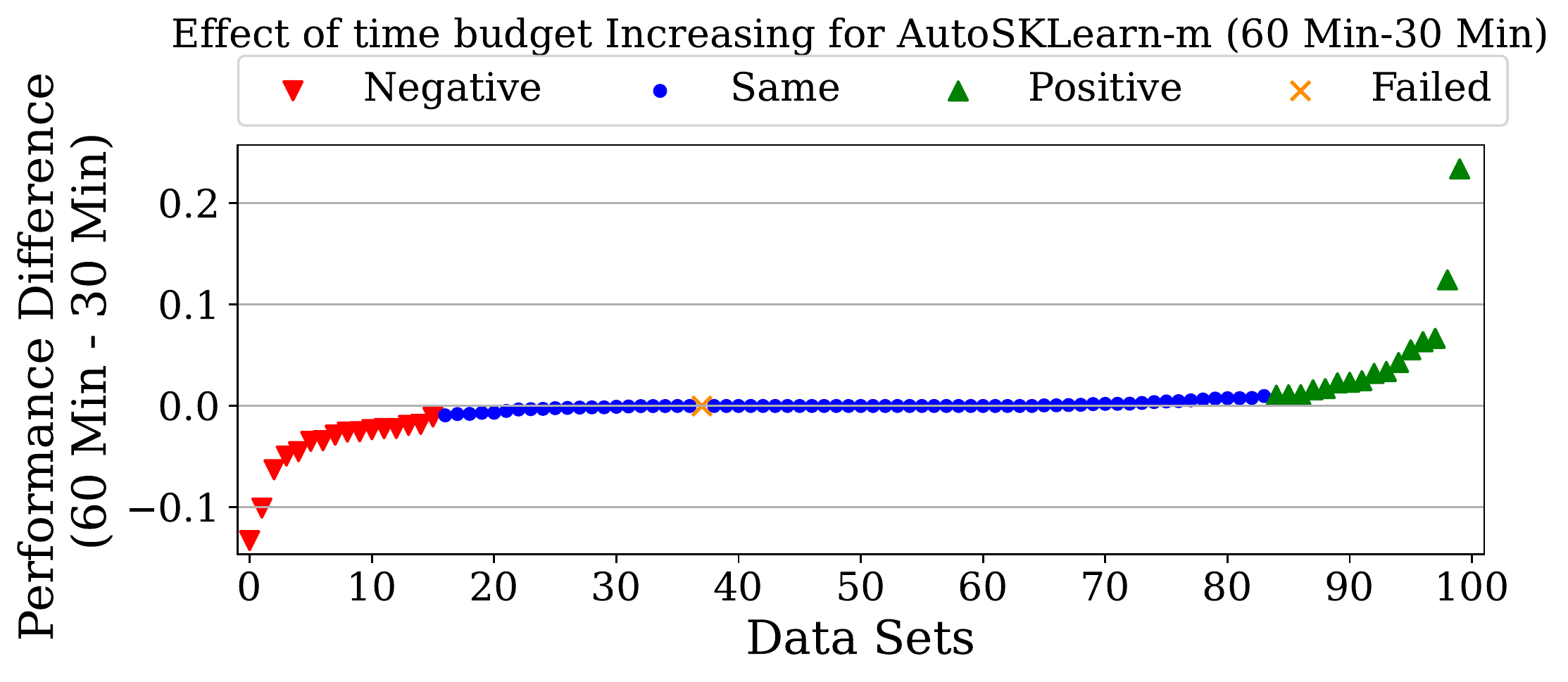}
}
\centering \subfigure[30-240 Min.] {
    \includegraphics[width=0.47\textwidth]{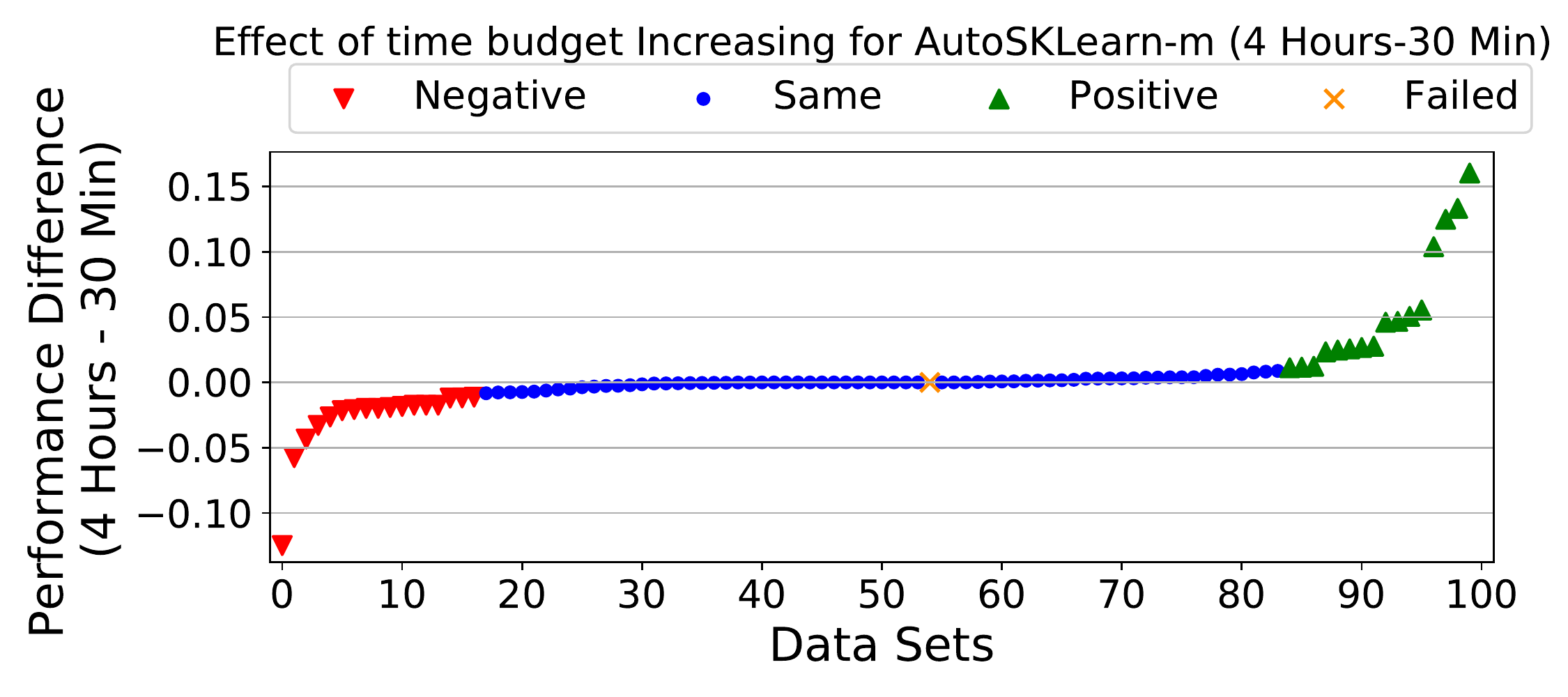}
}
\centering \subfigure[60-240 Min.] {
    \includegraphics[width=0.47\textwidth]{Figures/EffectoftimebudgetIncreasingforAutoSKlearnm4Hours30Min.pdf}
}

\caption{The impact of increasing the time budget on \texttt{AutoSKlearn-m} performance from $x$ to $y$ minutes (x-y). Green markers represent better performance with $y$ time budget, blue markers means that the difference between $x$ and $y$ is $<1$. Red markers represent better performance on $x$ time budget.}
\label{FIG:TimeBudgetSklearn-m}
\end{figure*}

\begin{figure*}[!ht]
\centering \subfigure[10-30 Min.] {
    \includegraphics[width=0.47\textwidth]{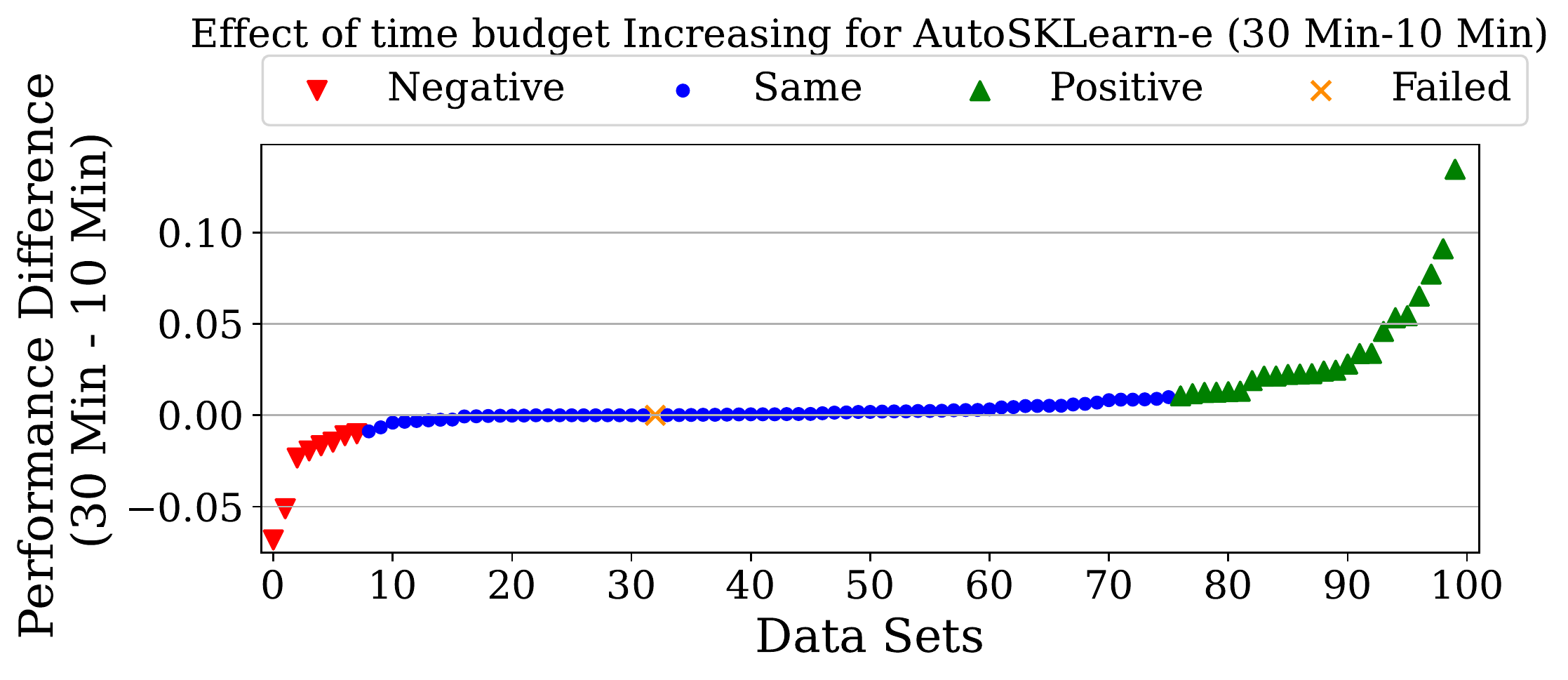}
}
\centering \subfigure[10-60 Min.] {
    \includegraphics[width=0.47\textwidth]{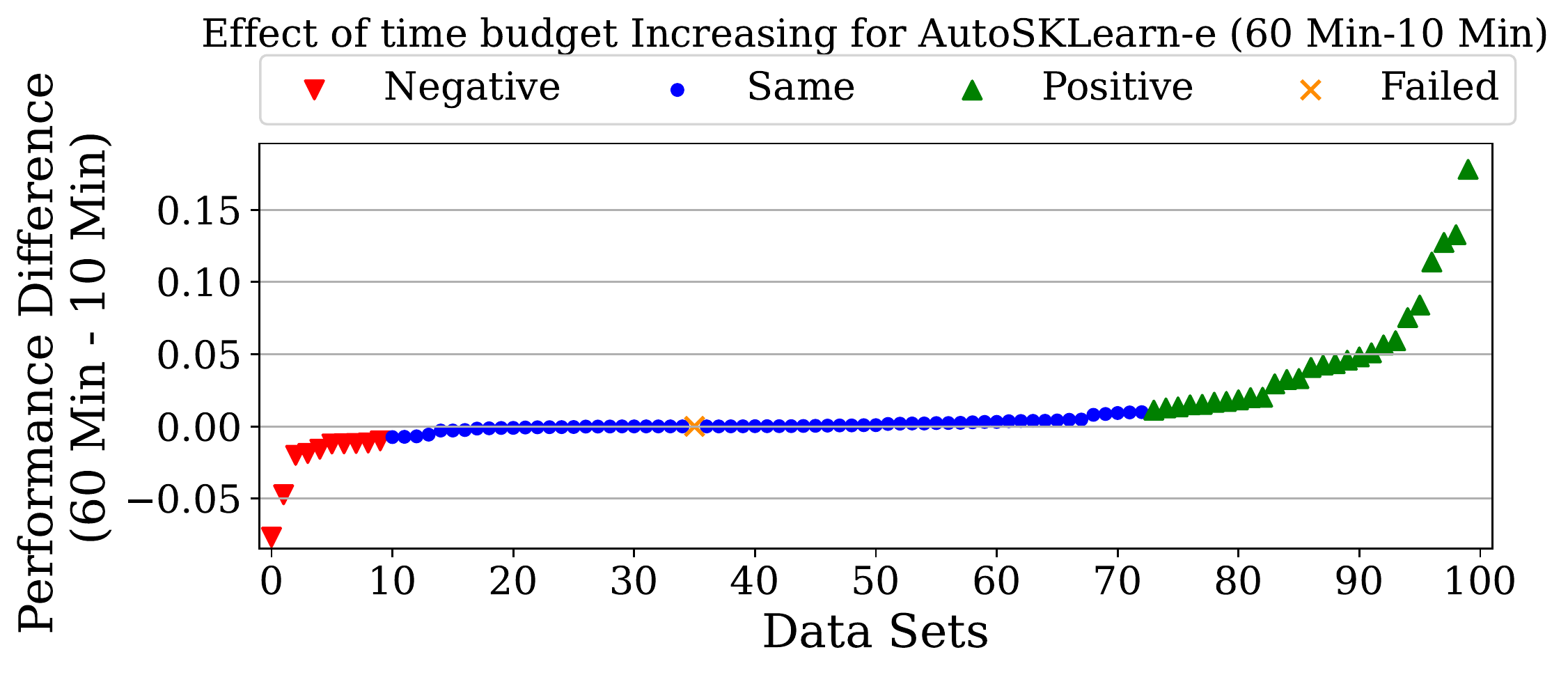}
}
\centering \subfigure[10-240 Min.] {
    \includegraphics[width=0.47\textwidth]{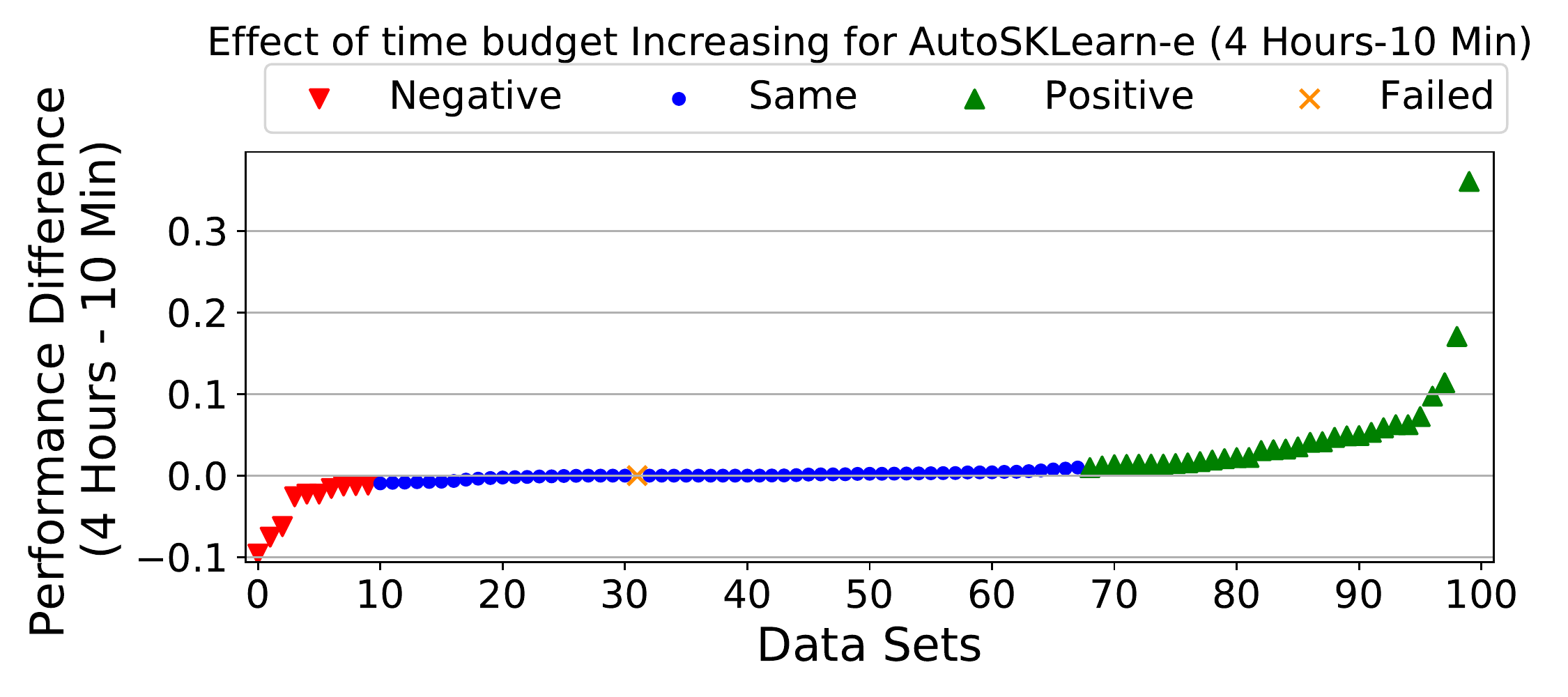}
}
\centering \subfigure[30-60 Min.] {
    \includegraphics[width=0.47\textwidth]{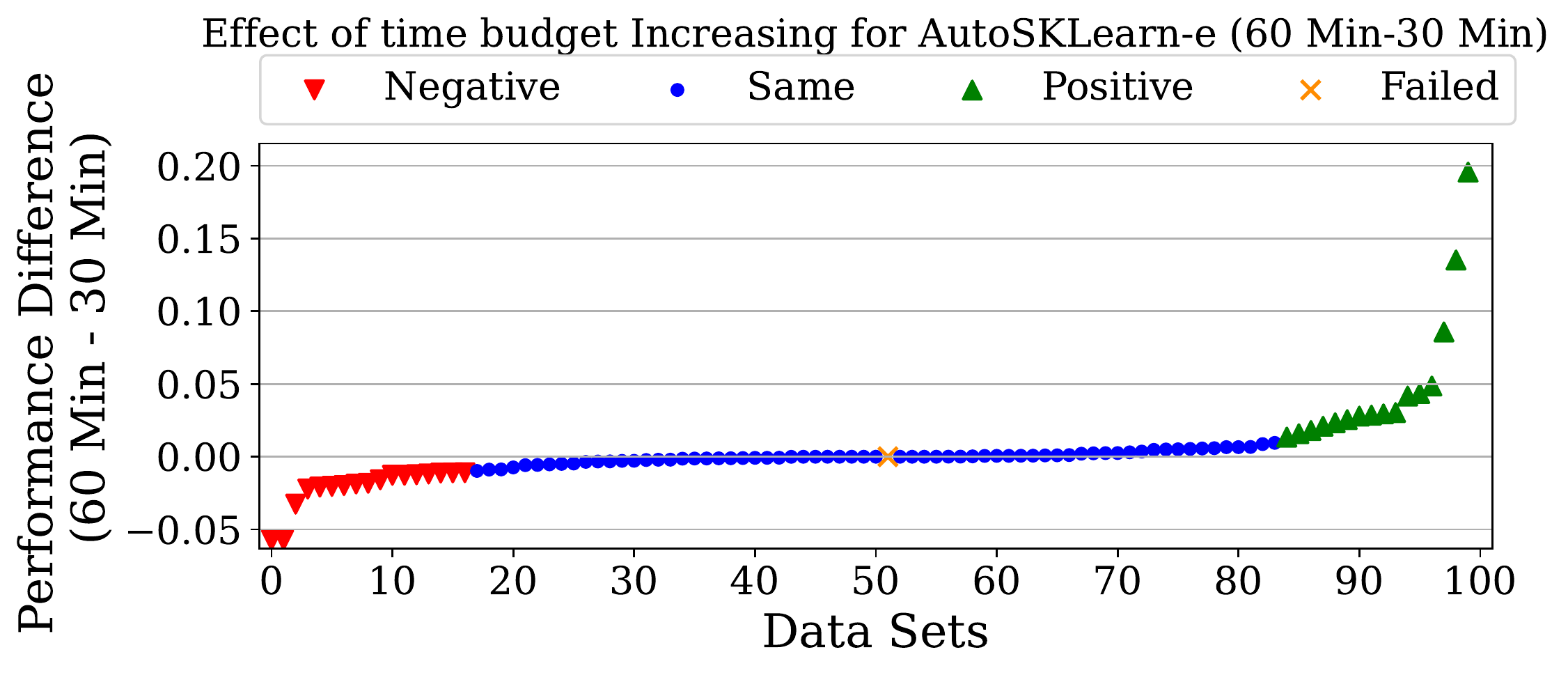}
}
\centering \subfigure[30-240 Min.] {
    \includegraphics[width=0.47\textwidth]{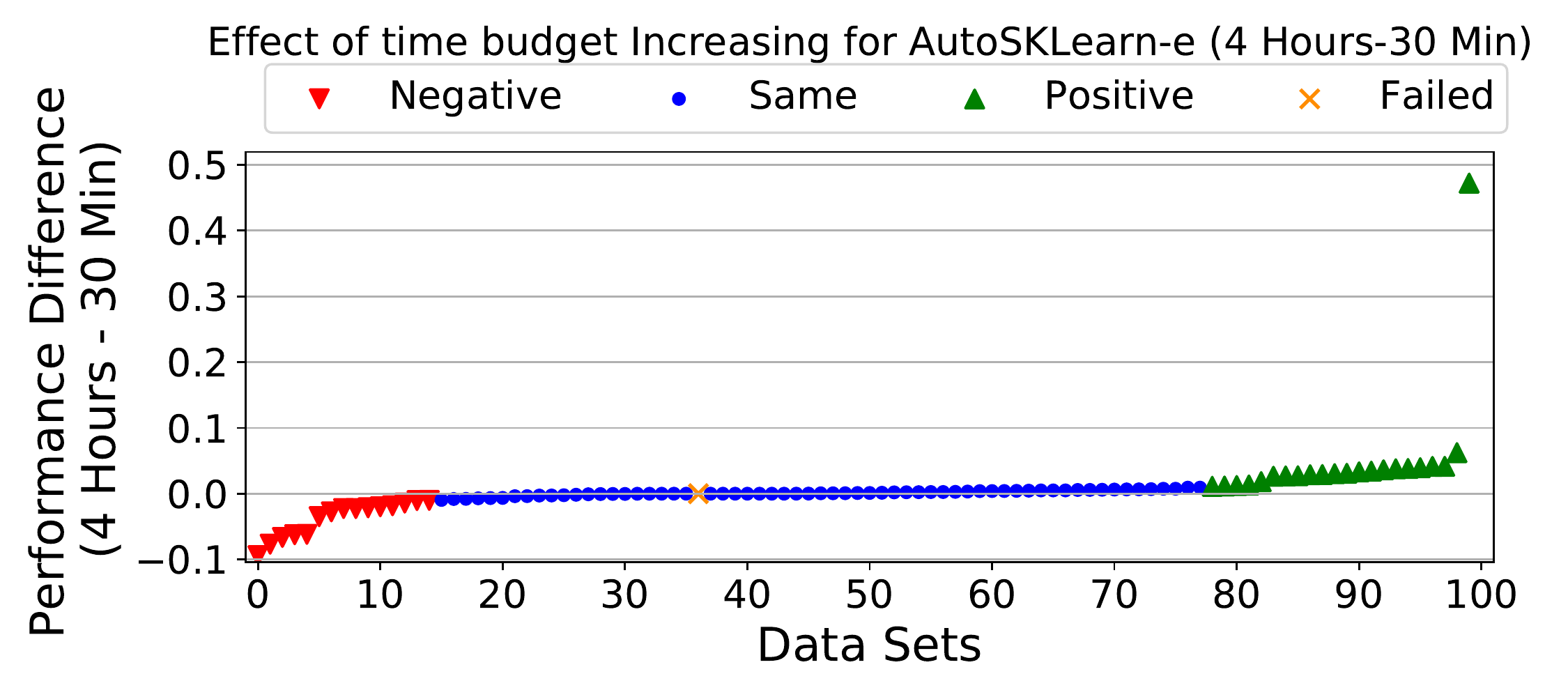}
}
\centering \subfigure[60-240 Min.] {
    \includegraphics[width=0.47\textwidth]{Figures/EffectoftimebudgetIncreasingforAutoSKlearne4Hours30Min.pdf}
}

\caption{The impact of increasing the time budget on \texttt{AutoSKlearn-e} performance from $x$ to $y$ minutes (x-y). Green markers represent better performance with $y$ time budget, blue markers means that the difference between $x$ and $y$ is $<1$. Red markers represent better performance on $x$ time budget.}
\label{FIG:TimeBudgetSklearn-e}
\end{figure*}

\begin{figure*}[!ht]
\centering \subfigure[10-30 Min.] {
    \includegraphics[width=0.47\textwidth]{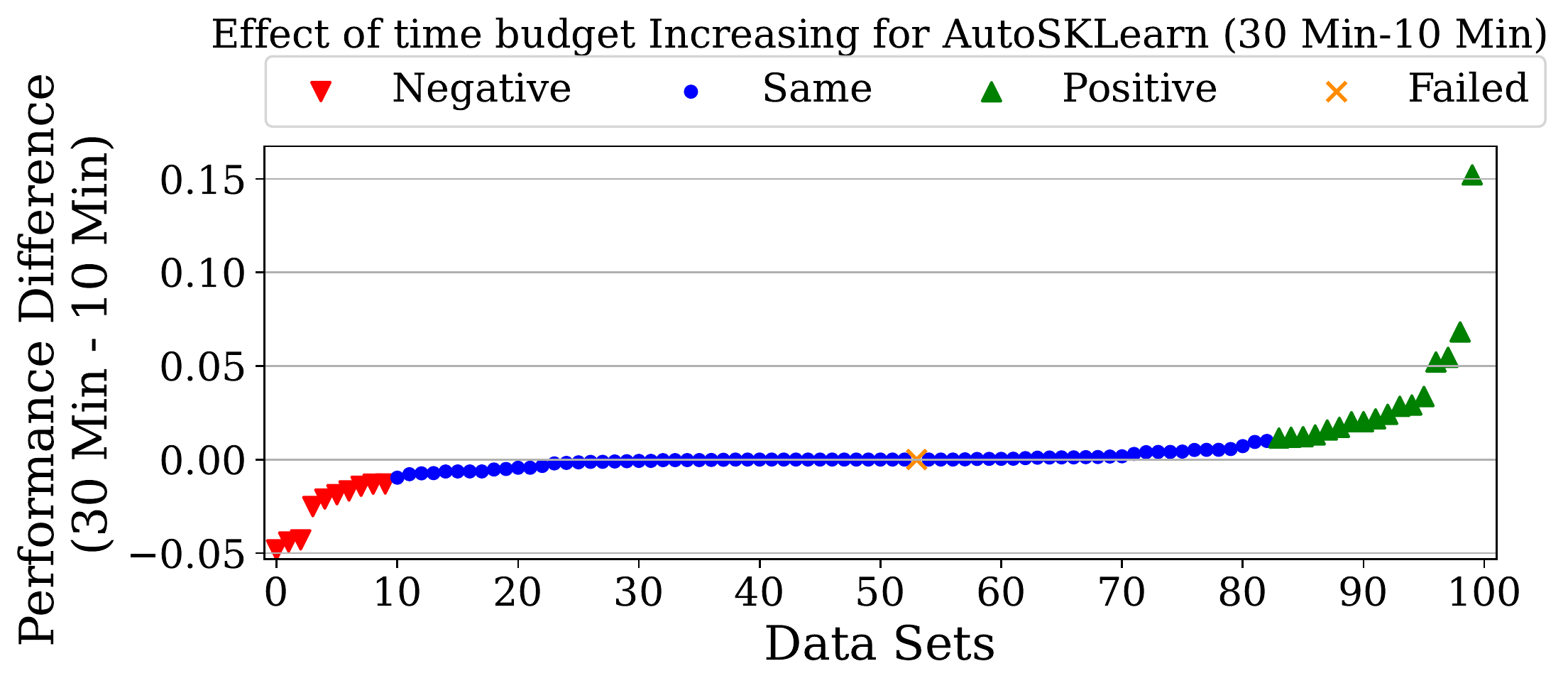}
}
\centering \subfigure[10-60 Min.] {
    \includegraphics[width=0.47\textwidth]{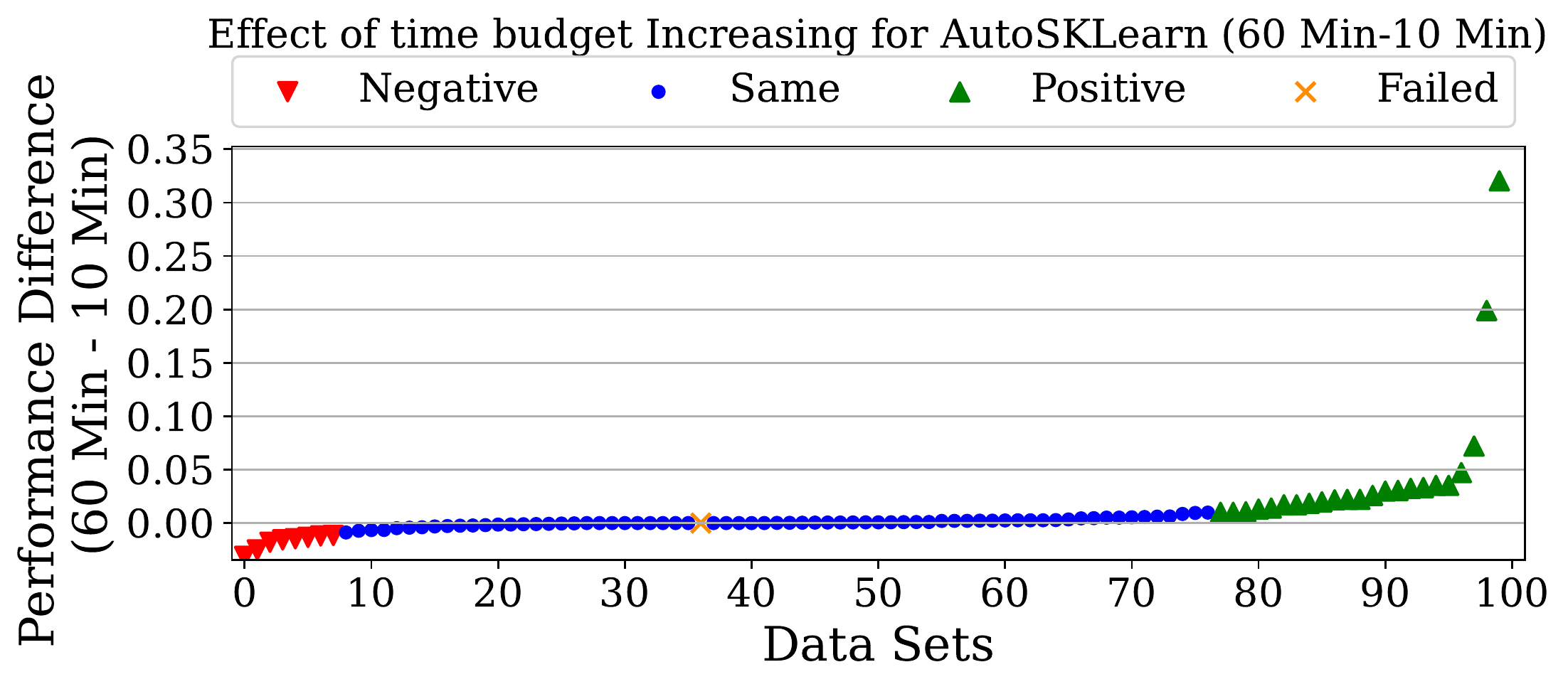}
}
\centering \subfigure[10-240 Min.] {
    \includegraphics[width=0.47\textwidth]{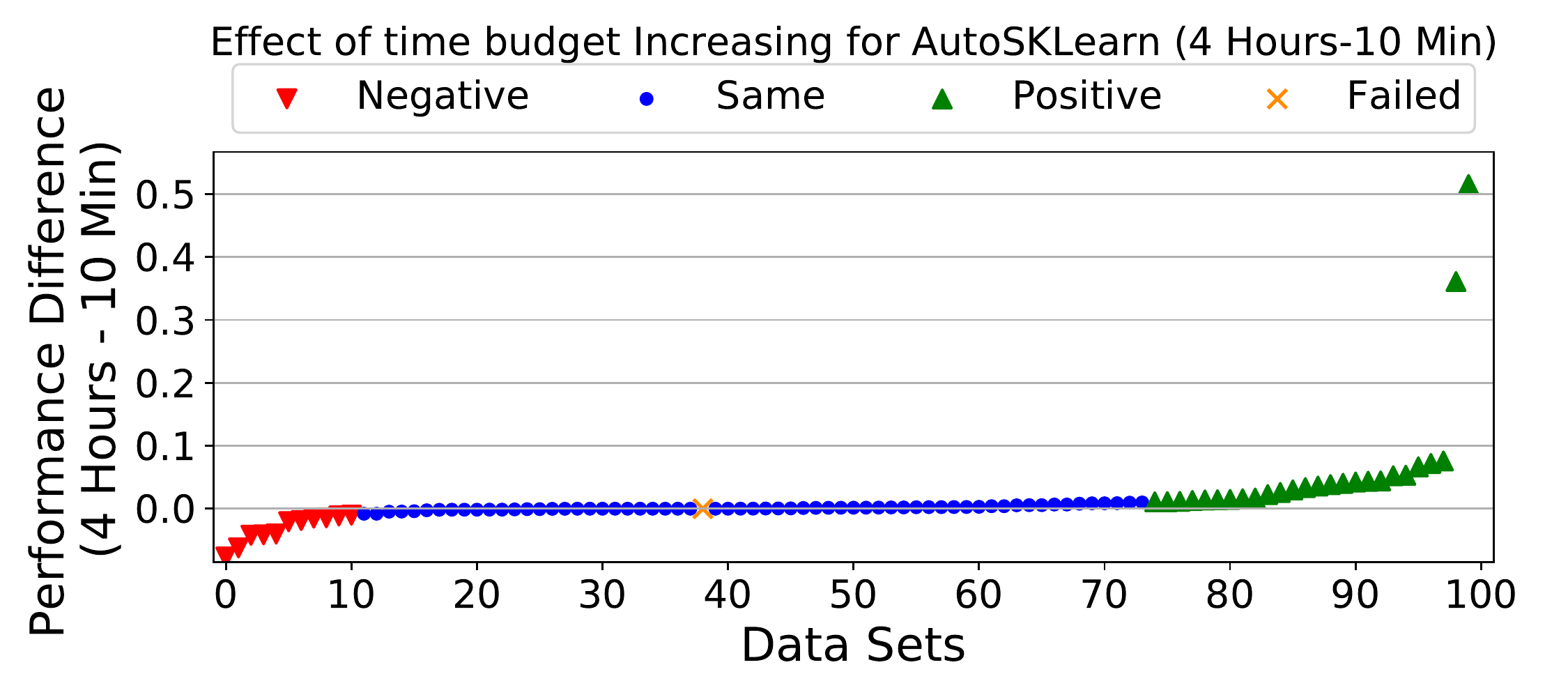}
}
\centering \subfigure[30-60 Min.] {
    \includegraphics[width=0.47\textwidth]{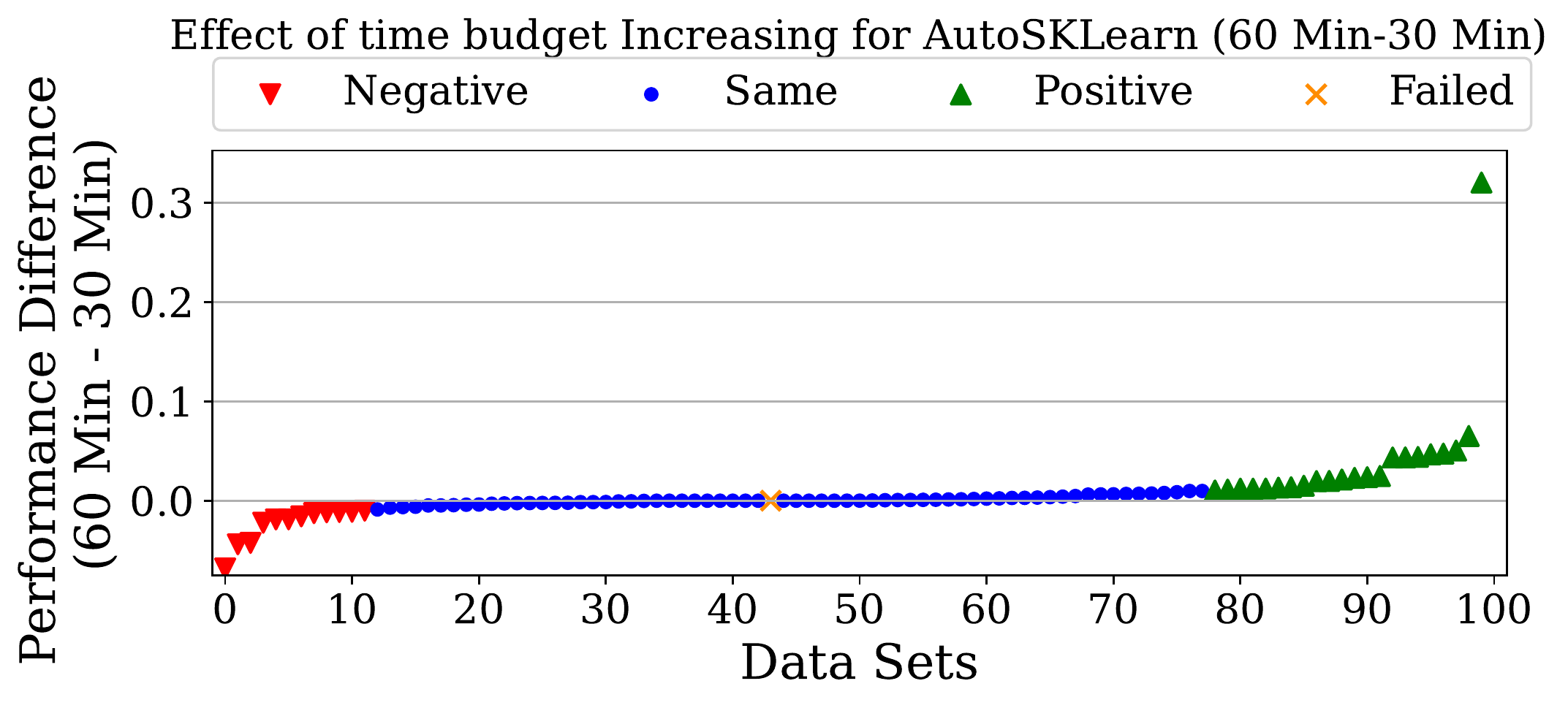}
}
\centering \subfigure[30-240 Min.] {
    \includegraphics[width=0.47\textwidth]{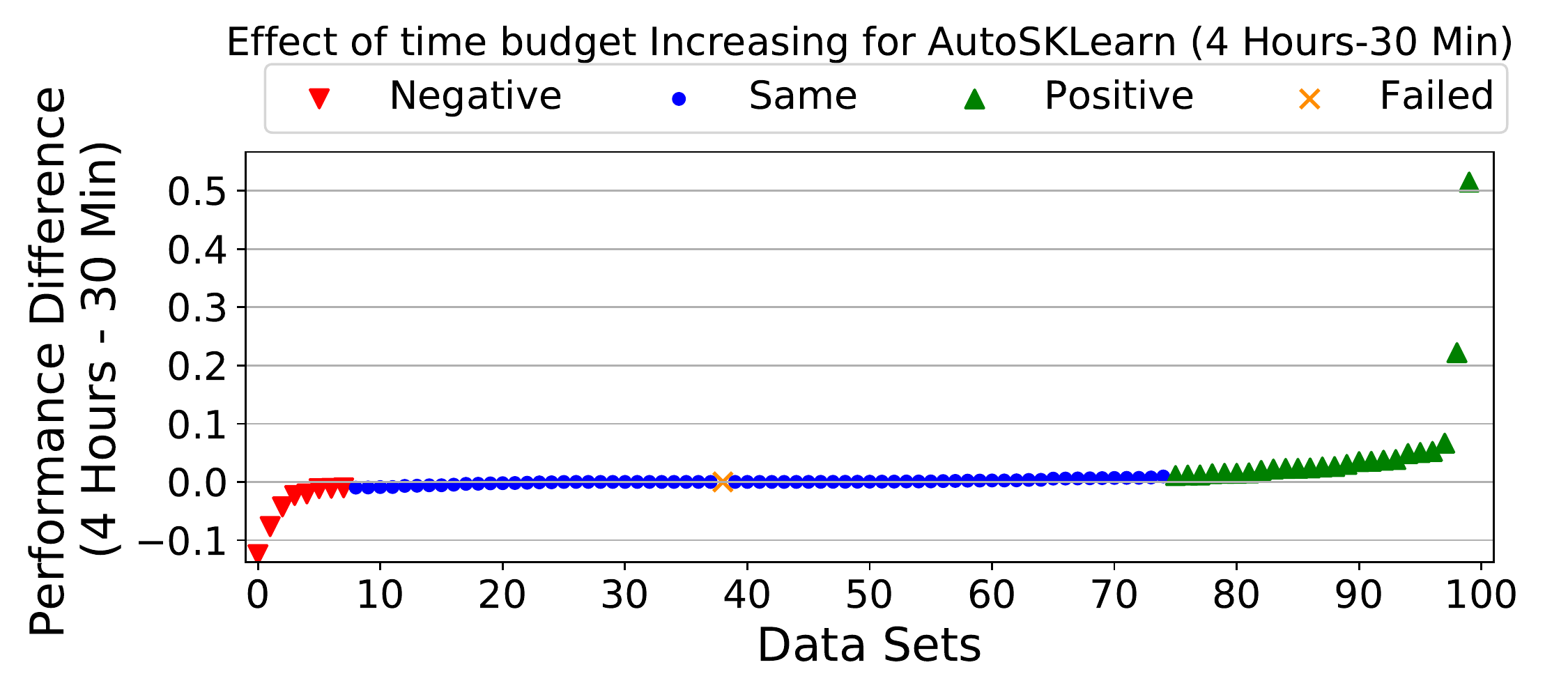}
}
\centering \subfigure[60-240 Min.] {
    \includegraphics[width=0.47\textwidth]{Figures/EffectoftimebudgetIncreasingforAutoSKlearn4Hours30Min.pdf}
}

\caption{The impact of increasing the time budget on \texttt{AutoSKlearn} performance from $x$ to $y$ minutes (x-y). Green markers represent better performance with $y$ time budget, blue markers means that the difference between $x$ and $y$ is $<1$. Red markers represent better performance on $x$ time budget.}
\label{FIG:TimeBudgetSklearn}
\end{figure*}

\begin{figure*}[!ht]
\centering \subfigure[10-30 Min.] {
    \includegraphics[width=0.47\textwidth]{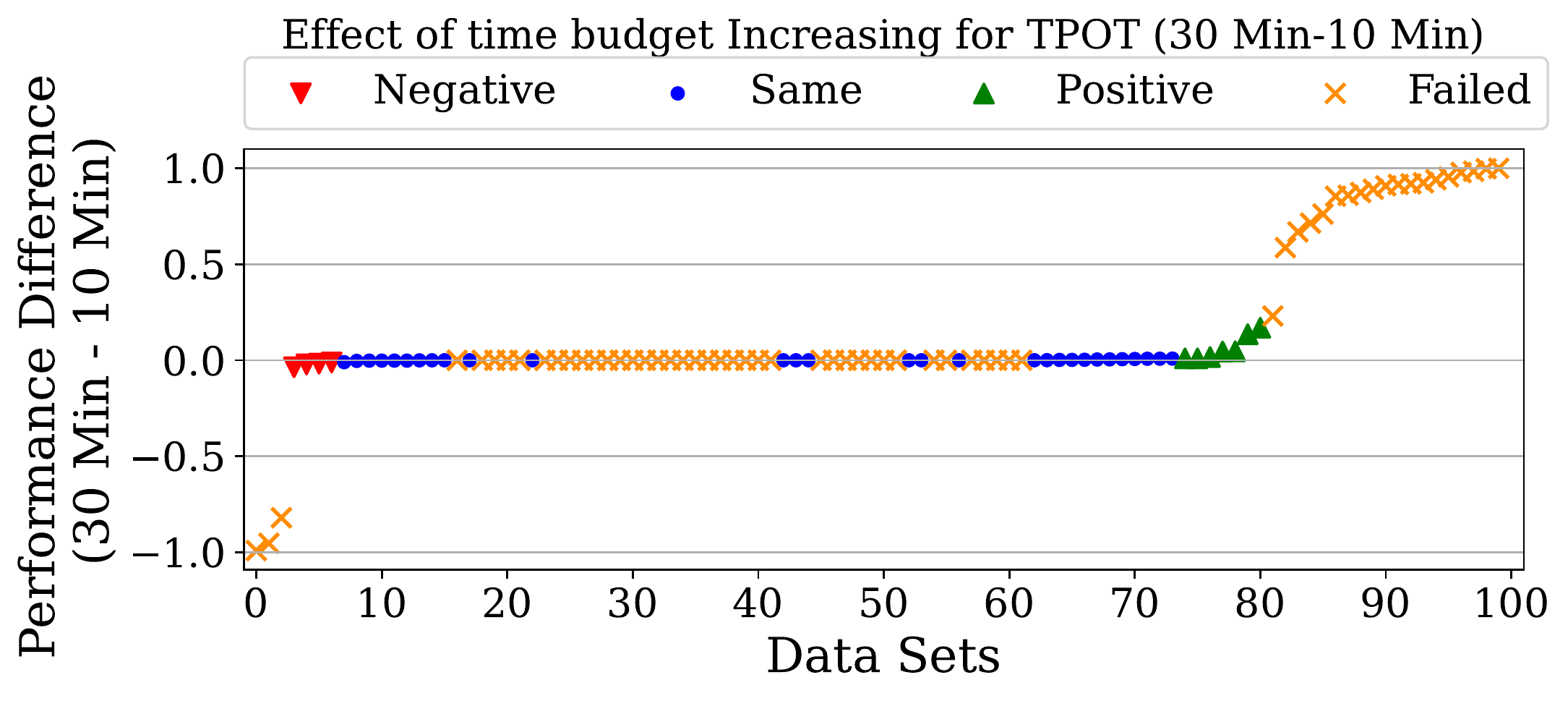}
}
\centering \subfigure[10-60 Min.] {
    \includegraphics[width=0.47\textwidth]{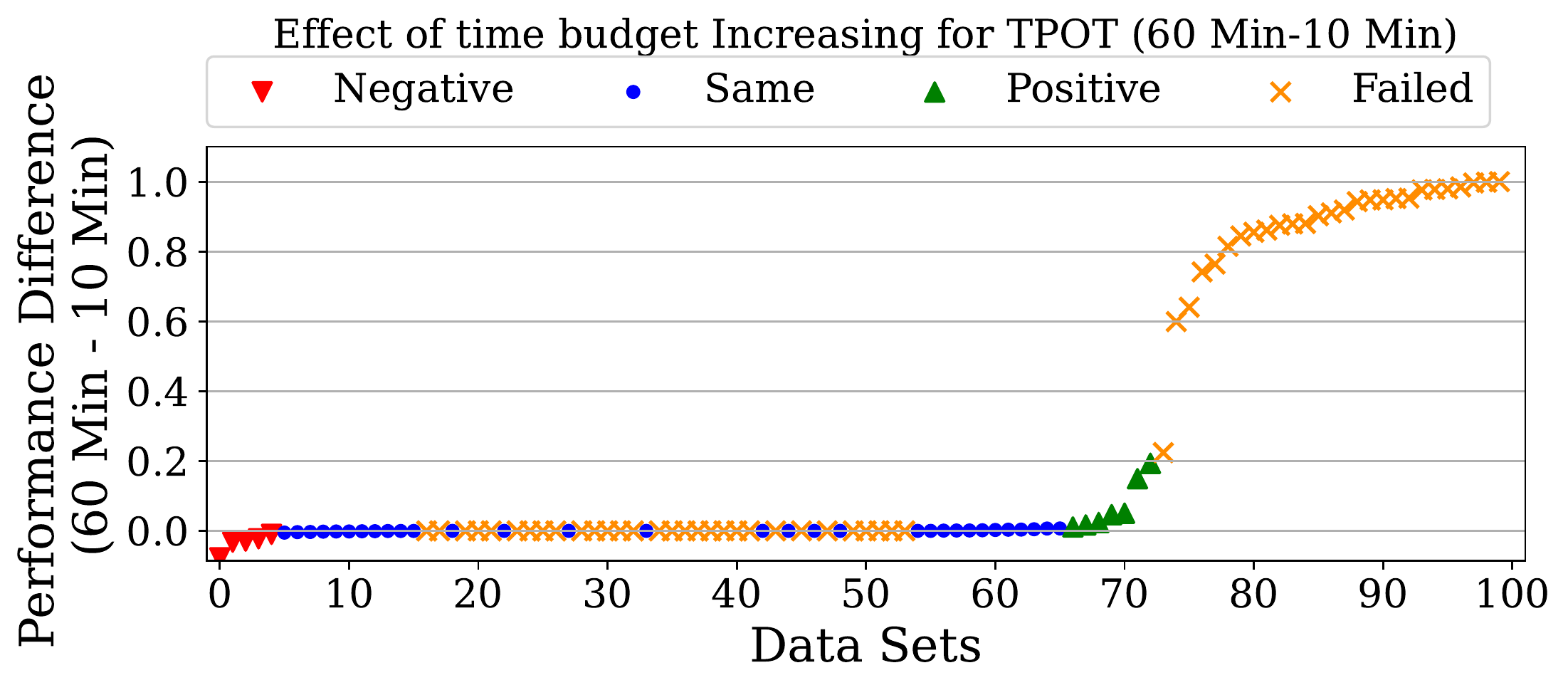}
}
\centering \subfigure[10-240 Min.] {
    \includegraphics[width=0.47\textwidth]{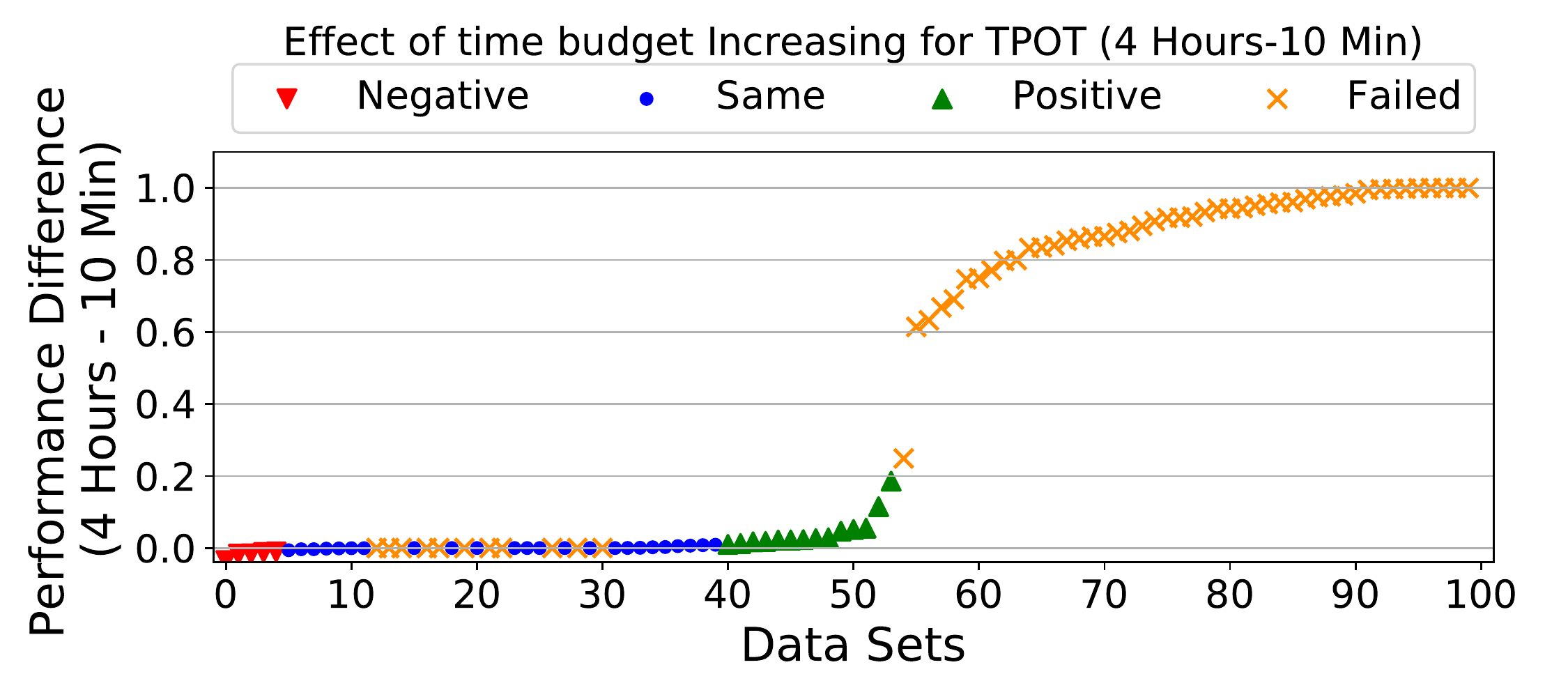}
}
\centering \subfigure[30-60 Min.] {
    \includegraphics[width=0.47\textwidth]{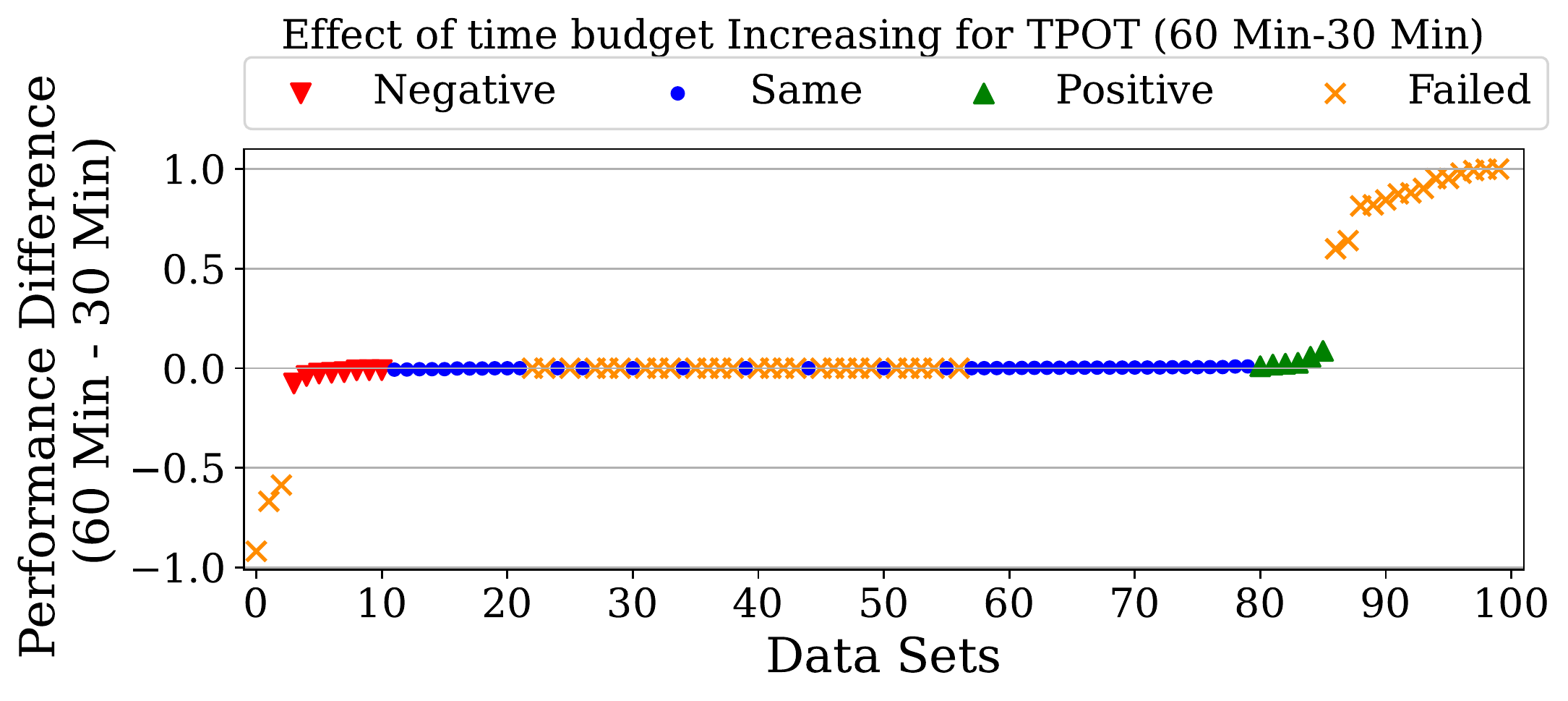}
}
\centering \subfigure[30-240 Min.] {
    \includegraphics[width=0.47\textwidth]{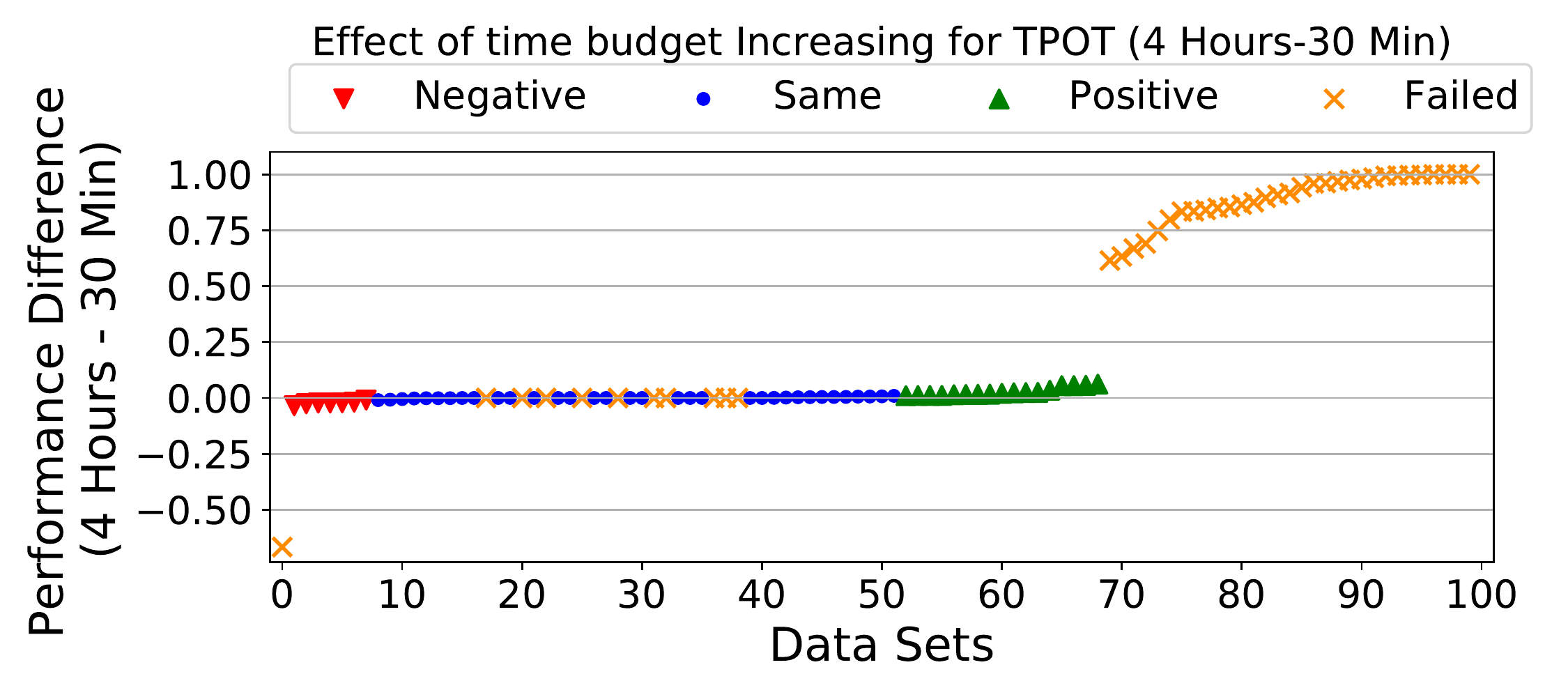}
}
\centering \subfigure[60-240 Min.] {
    \includegraphics[width=0.47\textwidth]{Figures/EffectoftimebudgetIncreasingforTPOT4Hours30Min.pdf}
}

\caption{The impact of increasing the time budget on \texttt{TPOT} performance from $x$ to $y$ minutes (x-y). Green markers represent better performance with $y$ time budget, blue markers means that the difference between $x$ and $y$ is $<1$. Red markers represent better performance on $x$ time budget.}
\label{FIG:TimeBudgetTPOT}
\end{figure*}

\begin{figure*}[!ht]
\centering \subfigure[10-30 Min.] {
    \includegraphics[width=0.47\textwidth]{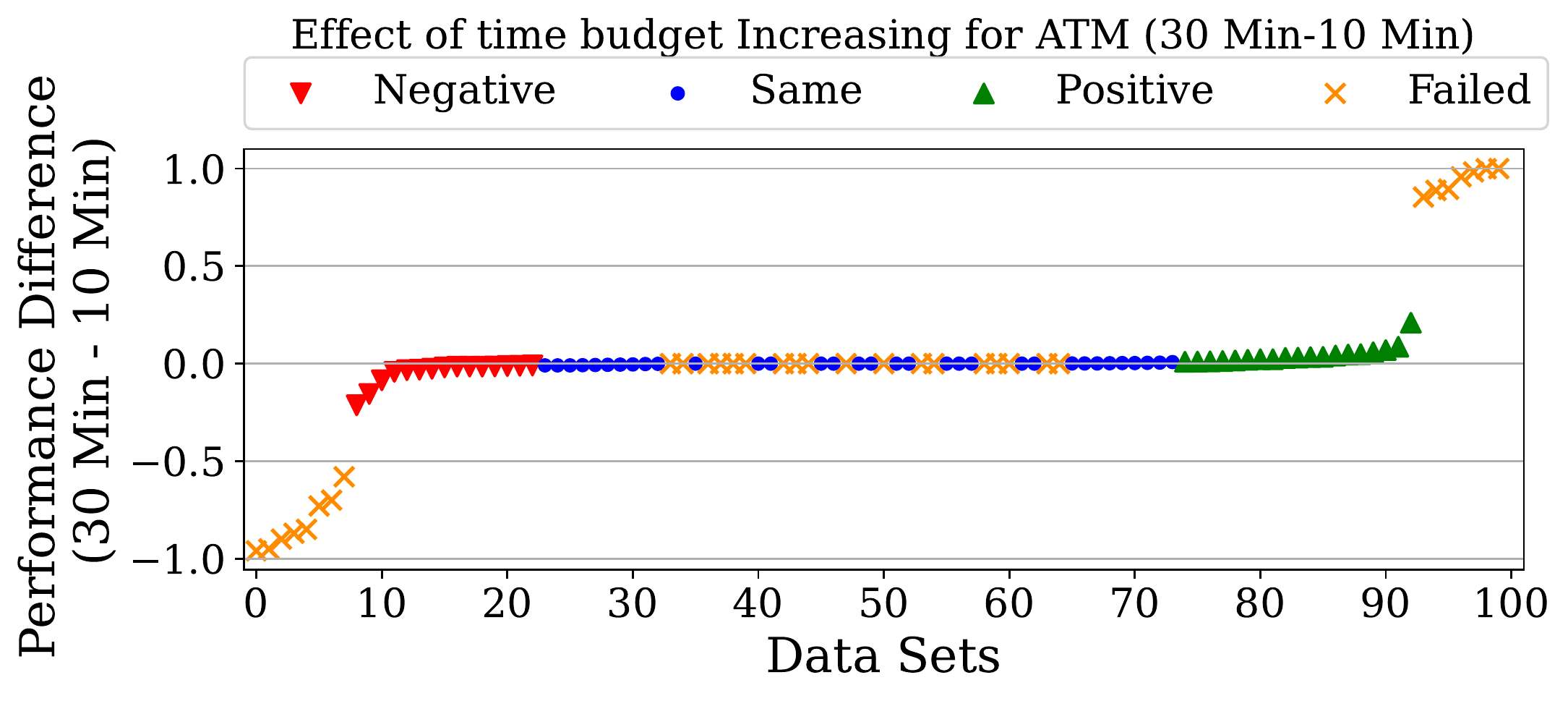}
}
\centering \subfigure[10-60 Min.] {
    \includegraphics[width=0.47\textwidth]{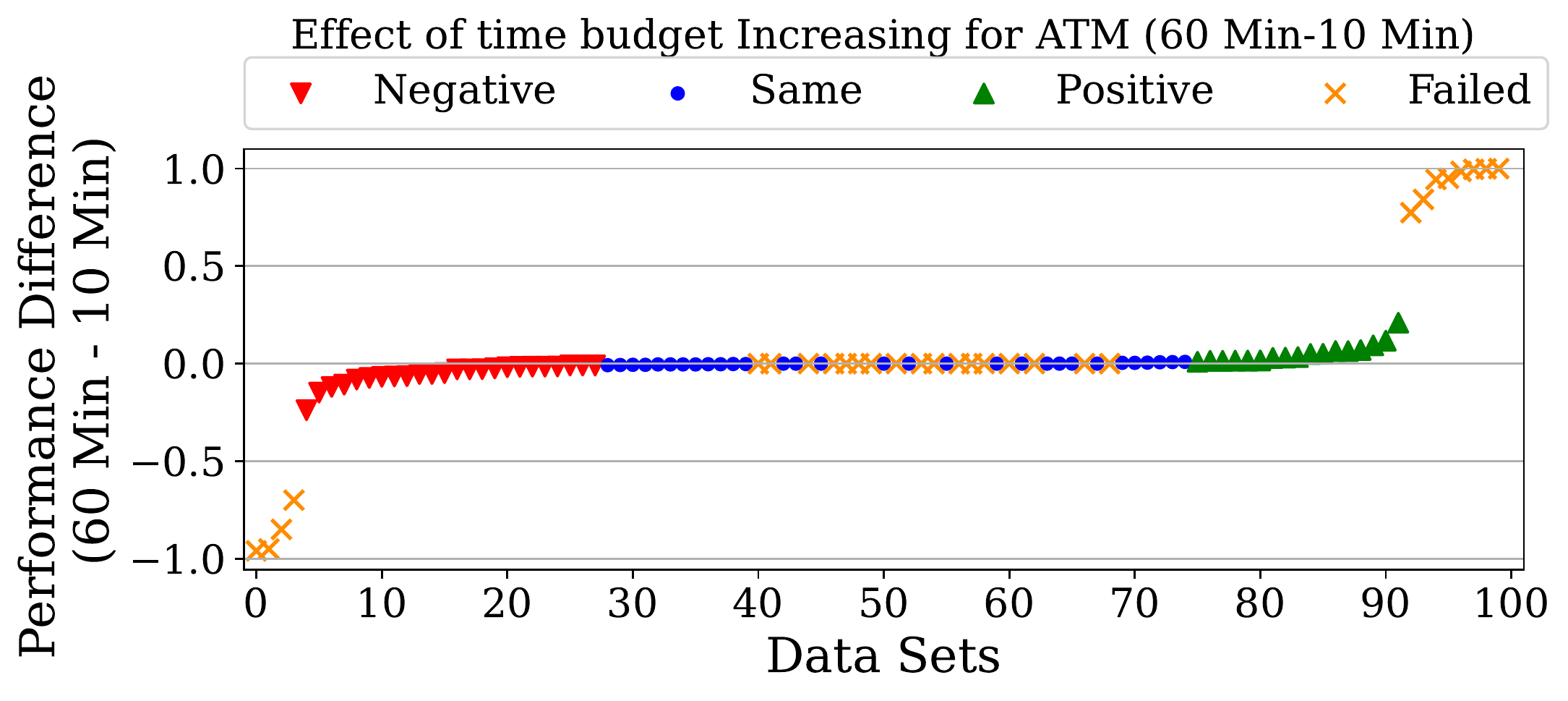}
}
\centering \subfigure[10-240 Min.] {
    \includegraphics[width=0.47\textwidth]{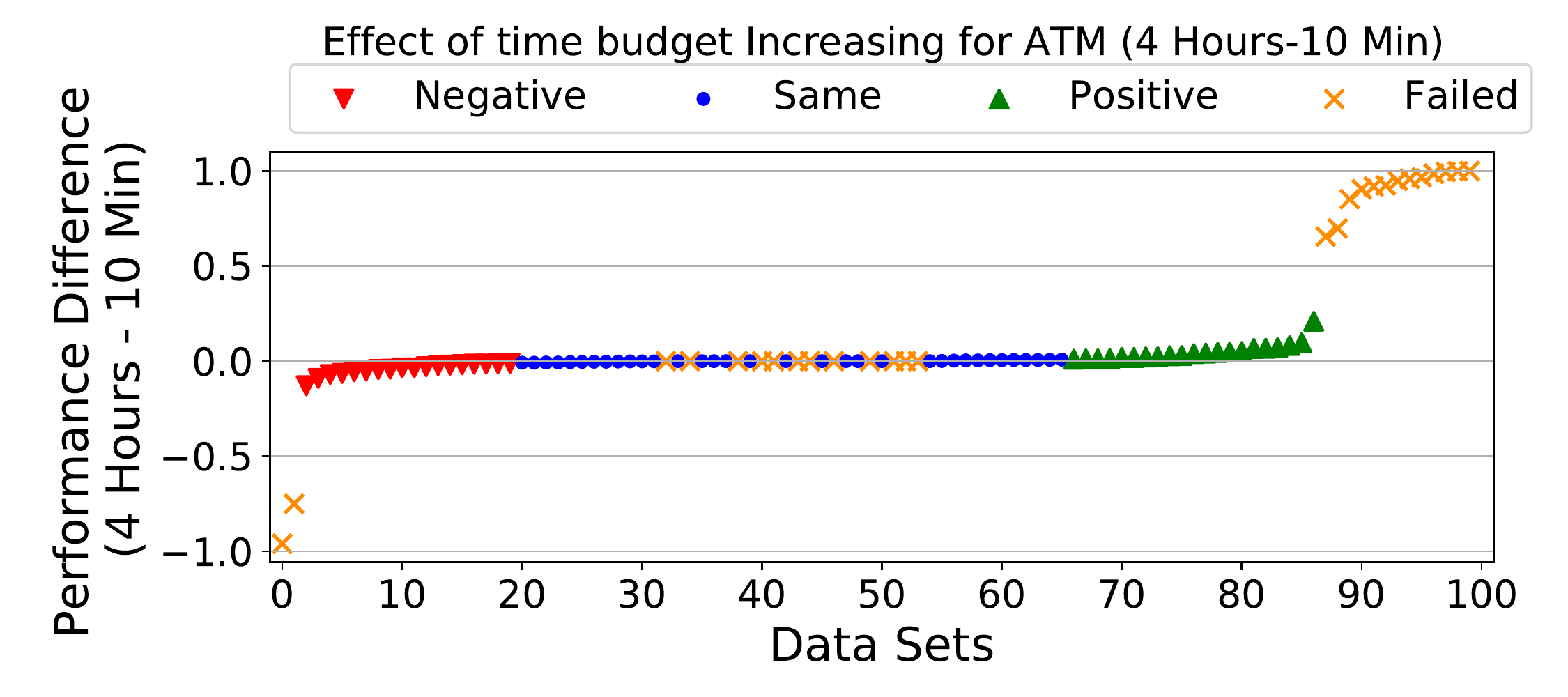}
}
\centering \subfigure[30-60 Min.] {
    \includegraphics[width=0.47\textwidth]{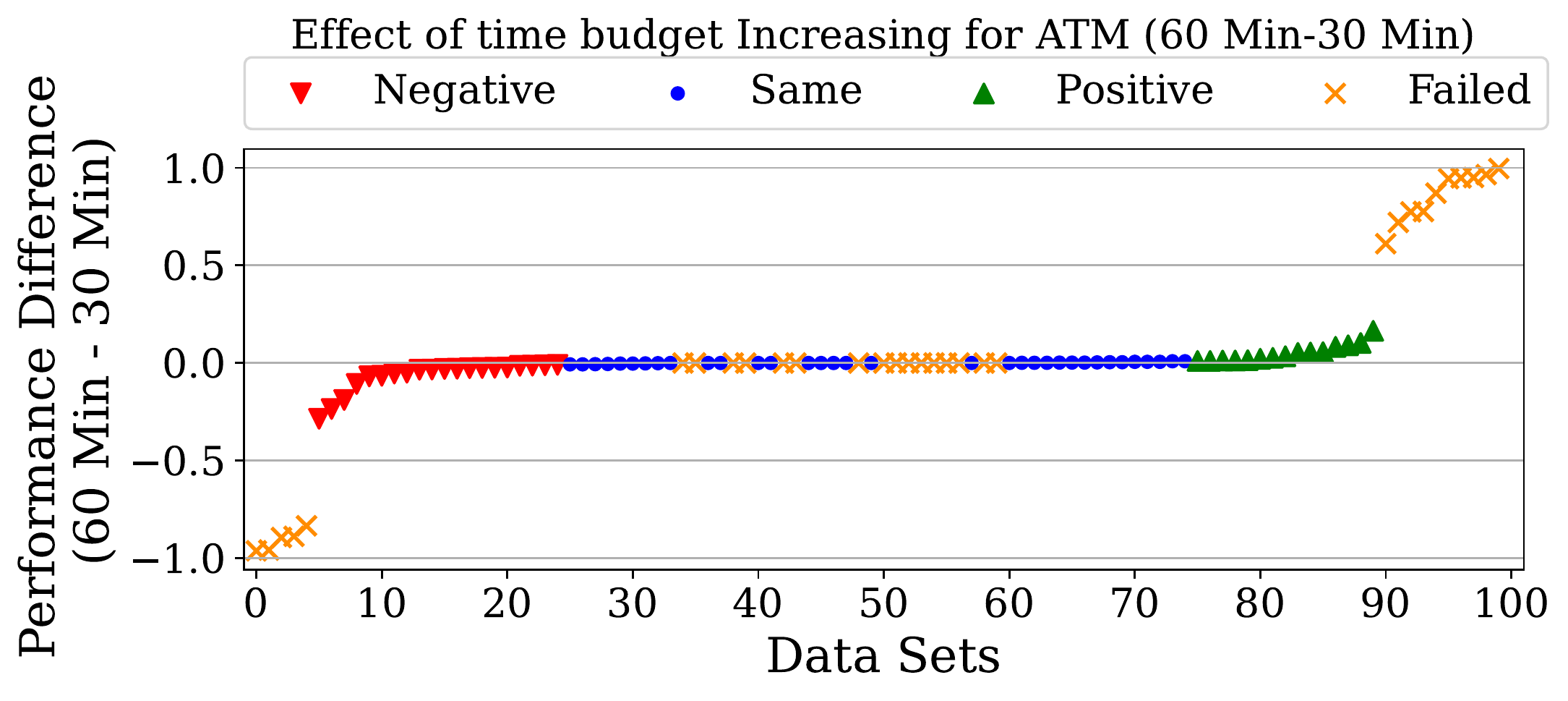}
}
\centering \subfigure[30-240 Min.] {
    \includegraphics[width=0.47\textwidth]{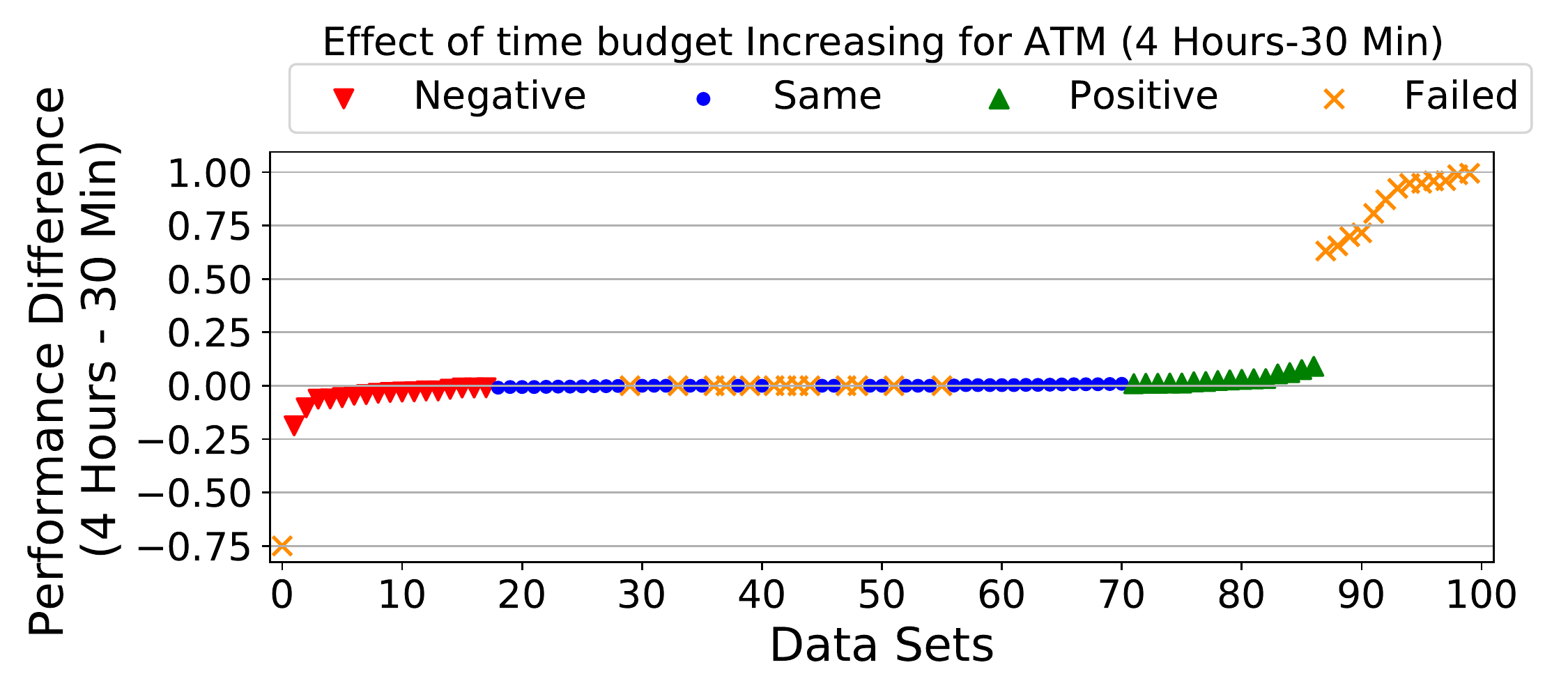}
}
\centering \subfigure[60-240 Min.] {
    \includegraphics[width=0.47\textwidth]{Figures/EffectoftimebudgetIncreasingforATM4Hours30Min.pdf}
}

\caption{The impact of increasing the time budget on \texttt{ATM} performance from $x$ to $y$ minutes (x-y). Green markers represent better performance with $y$ time budget, blue markers means that the difference between $x$ and $y$ is $<1$. Red markers represent better performance on $x$ time budget.}
\label{FIG:TimeBudgetATM}
\end{figure*}

\begin{figure*}[!ht]
\centering \subfigure[10-30 Min.] {
    \includegraphics[width=0.47\textwidth]{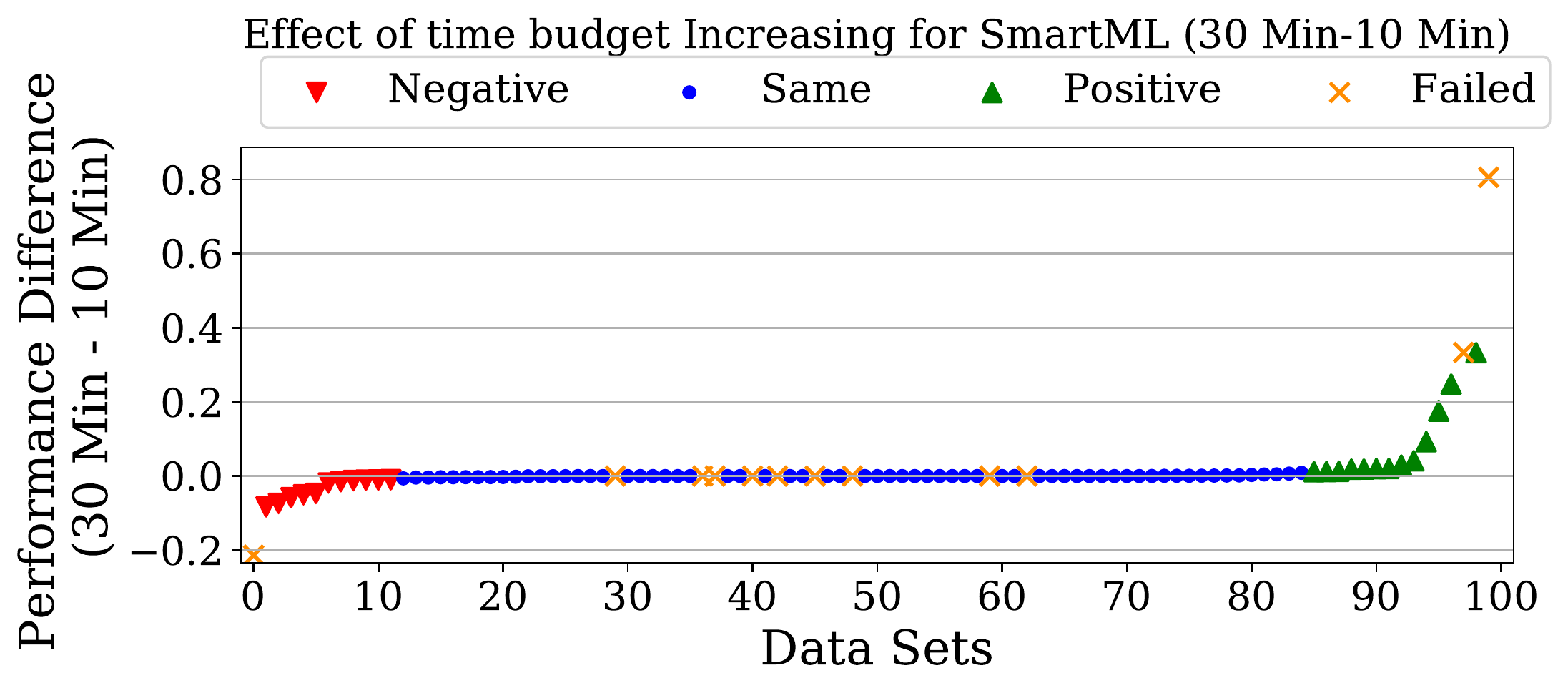}
}
\centering \subfigure[10-60 Min.] {
    \includegraphics[width=0.47\textwidth]{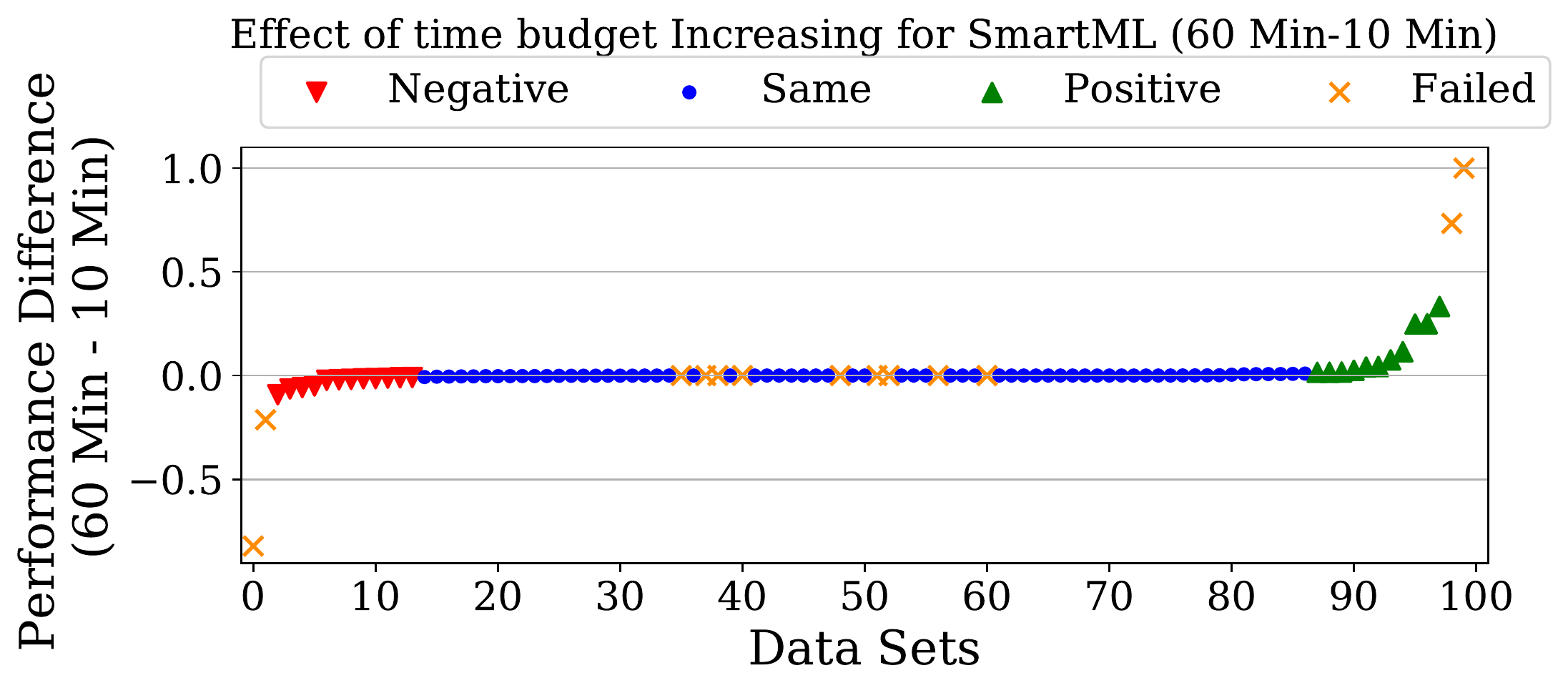}
}
\centering \subfigure[10-240 Min.] {
    \includegraphics[width=0.47\textwidth]{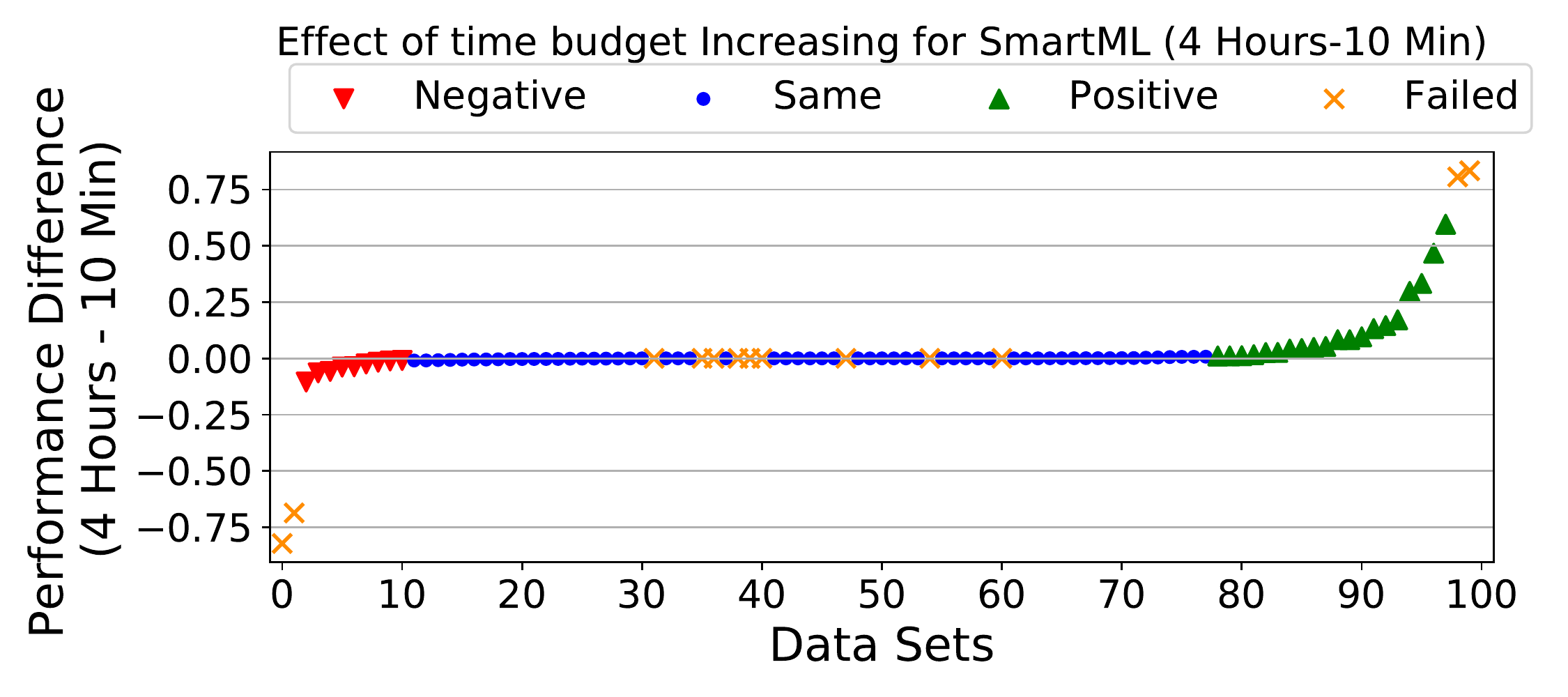}
}
\centering \subfigure[30-60 Min.] {
    \includegraphics[width=0.47\textwidth]{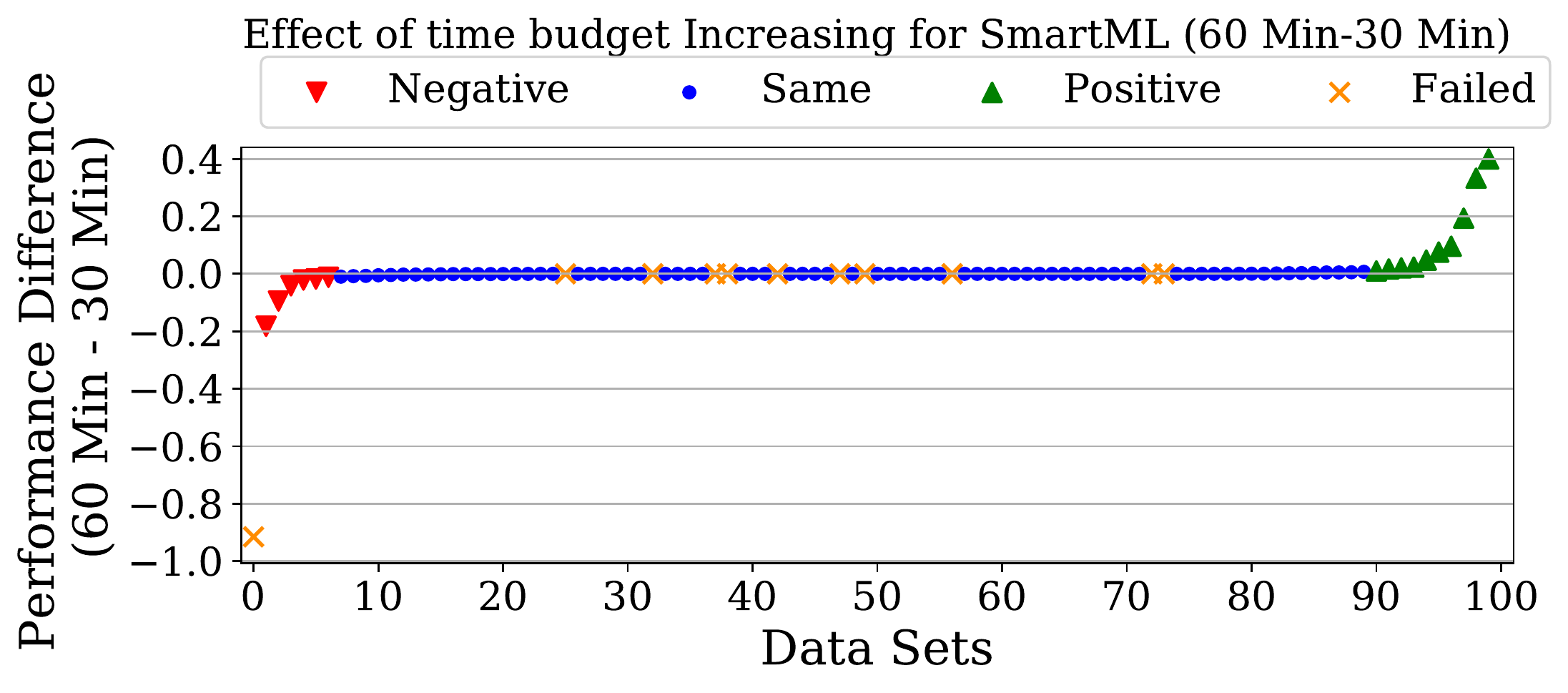}
}
\centering \subfigure[30-240 Min.] {
    \includegraphics[width=0.47\textwidth]{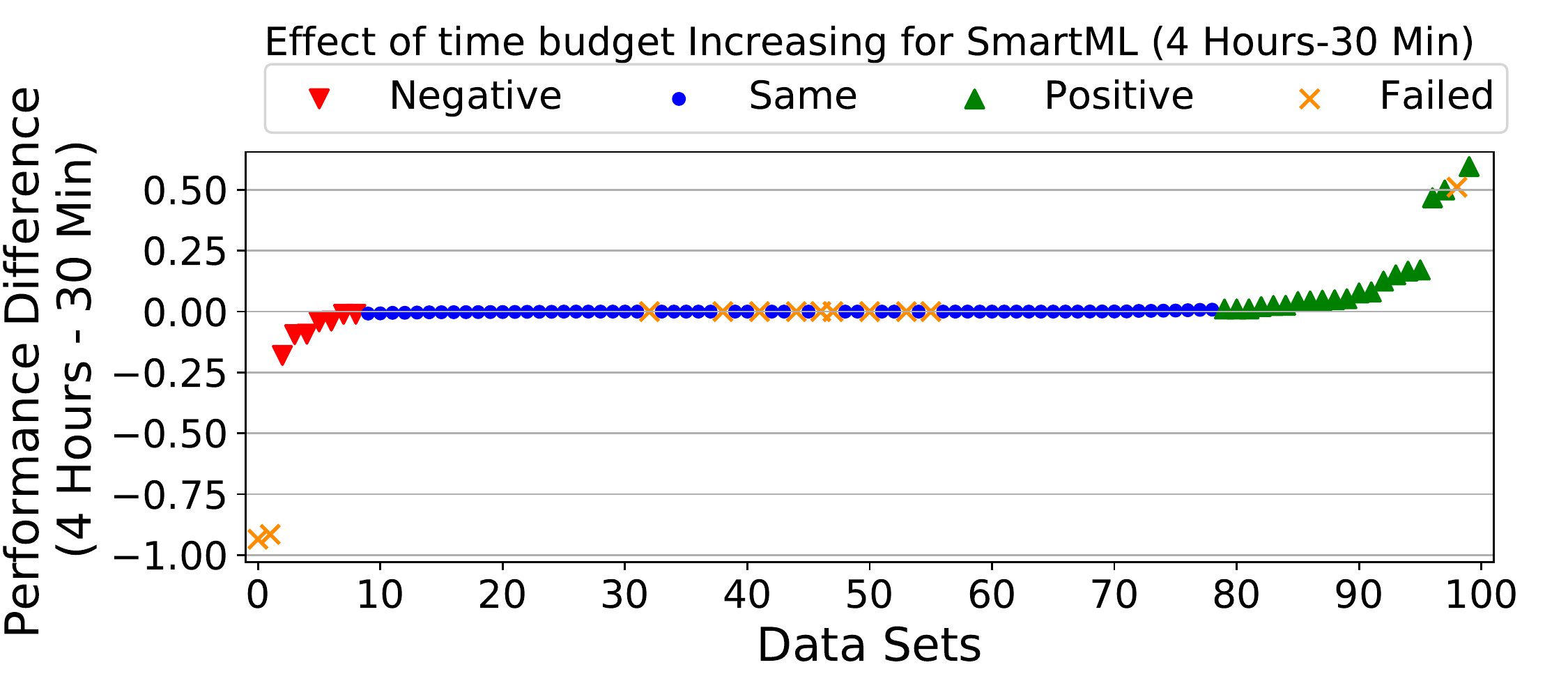}
}
\centering \subfigure[60-240 Min.] {
    \includegraphics[width=0.47\textwidth]{Figures/EffectoftimebudgetIncreasingforSmartML4Hours30Min.pdf}
}

\caption{The impact of increasing the time budget on \texttt{SmartML-m} performance from $x$ to $y$ minutes (x-y). Green markers represent better performance with $y$ time budget, blue markers means that the difference between $x$ and $y$ is $<1$. Red markers represent better performance on $x$ time budget.}
\label{FIG:TimeBudgetSmartML}
\end{figure*}

\begin{figure*}[!ht]
\centering \subfigure[10-30 Min.] {
    \includegraphics[width=0.47\textwidth]{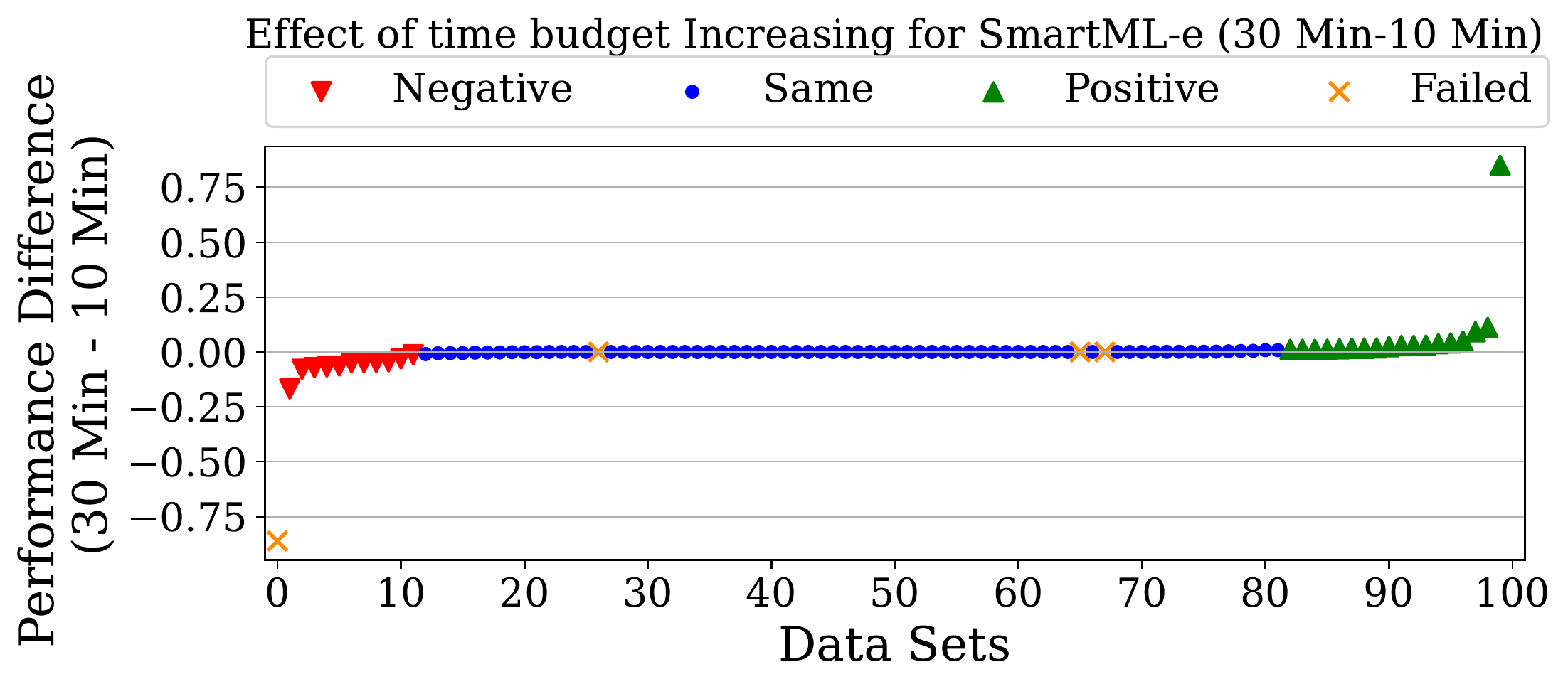}
}
\centering \subfigure[10-60 Min.] {
    \includegraphics[width=0.47\textwidth]{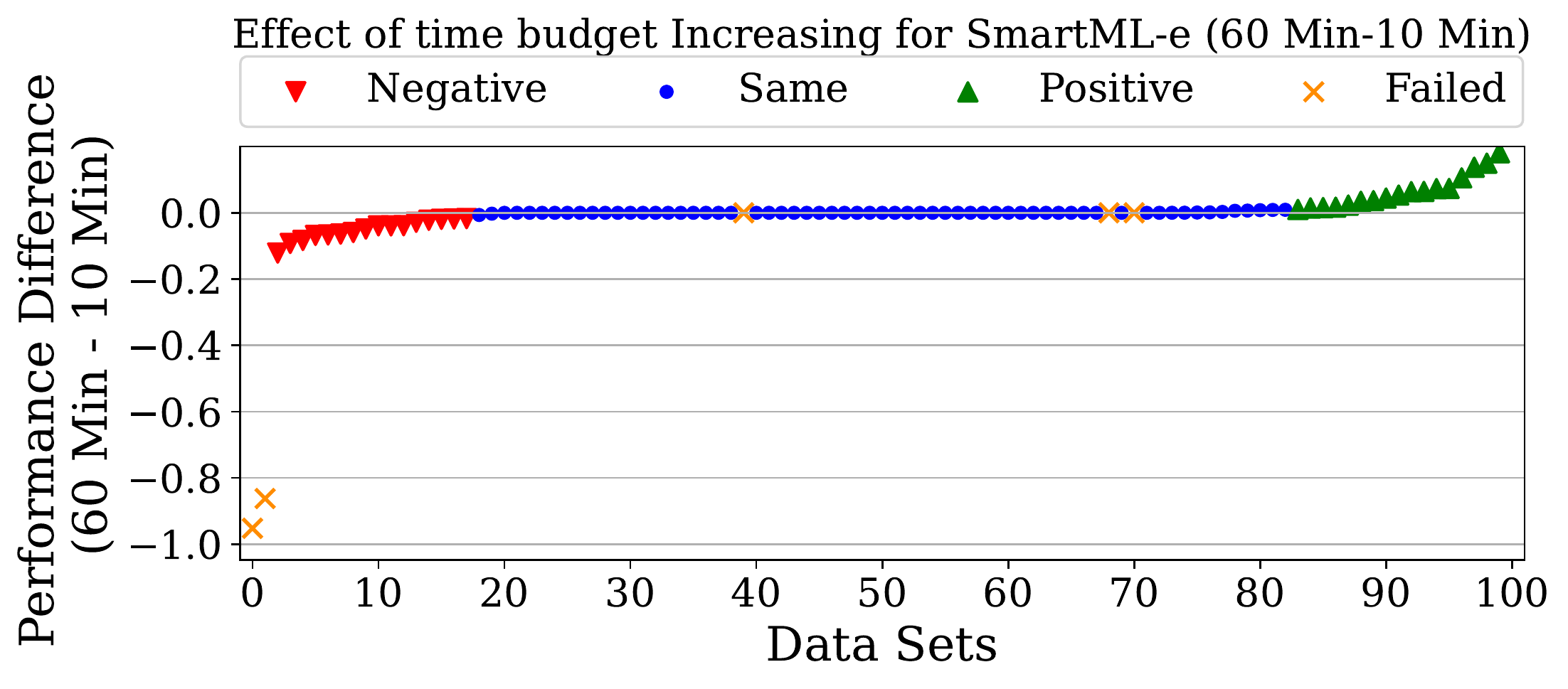}
}
\centering \subfigure[10-240 Min.] {
    \includegraphics[width=0.47\textwidth]{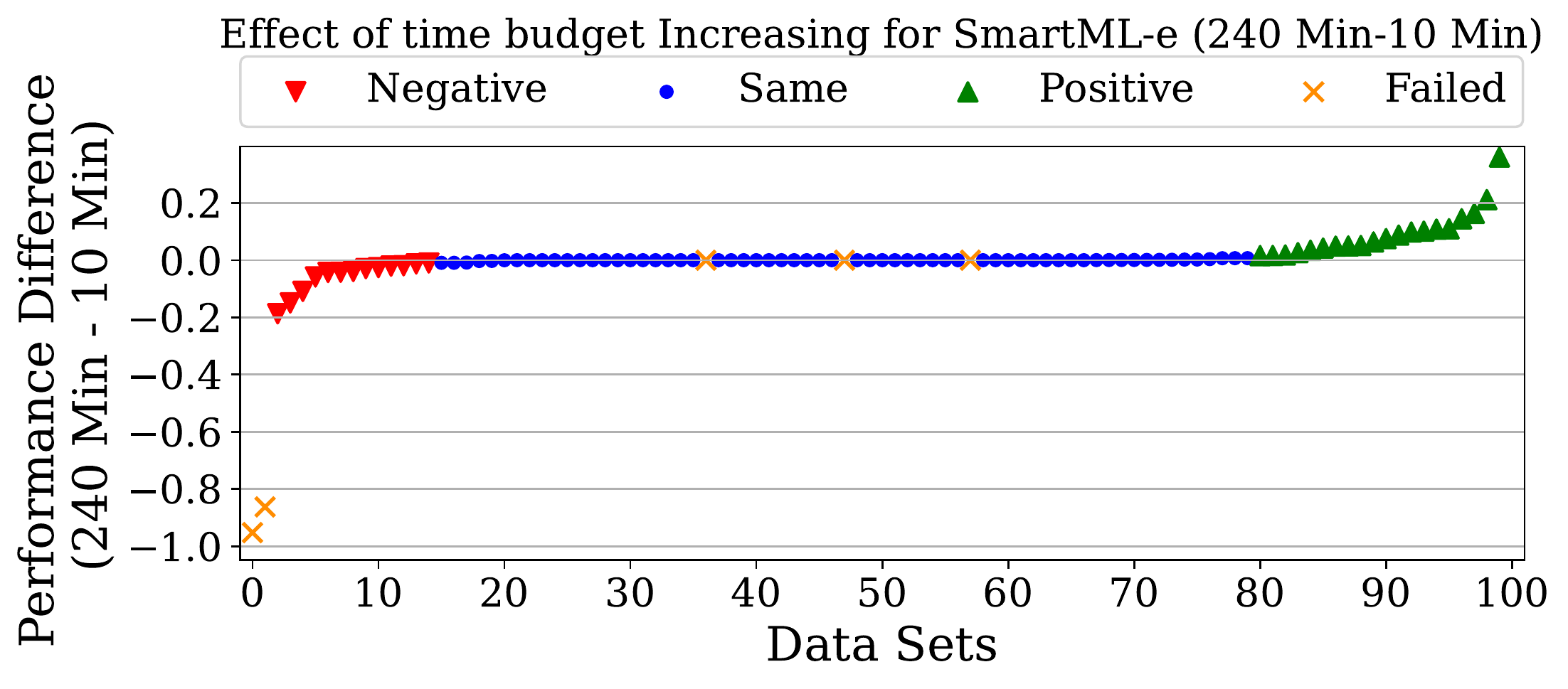}
}
\centering \subfigure[30-60 Min.] {
    \includegraphics[width=0.47\textwidth]{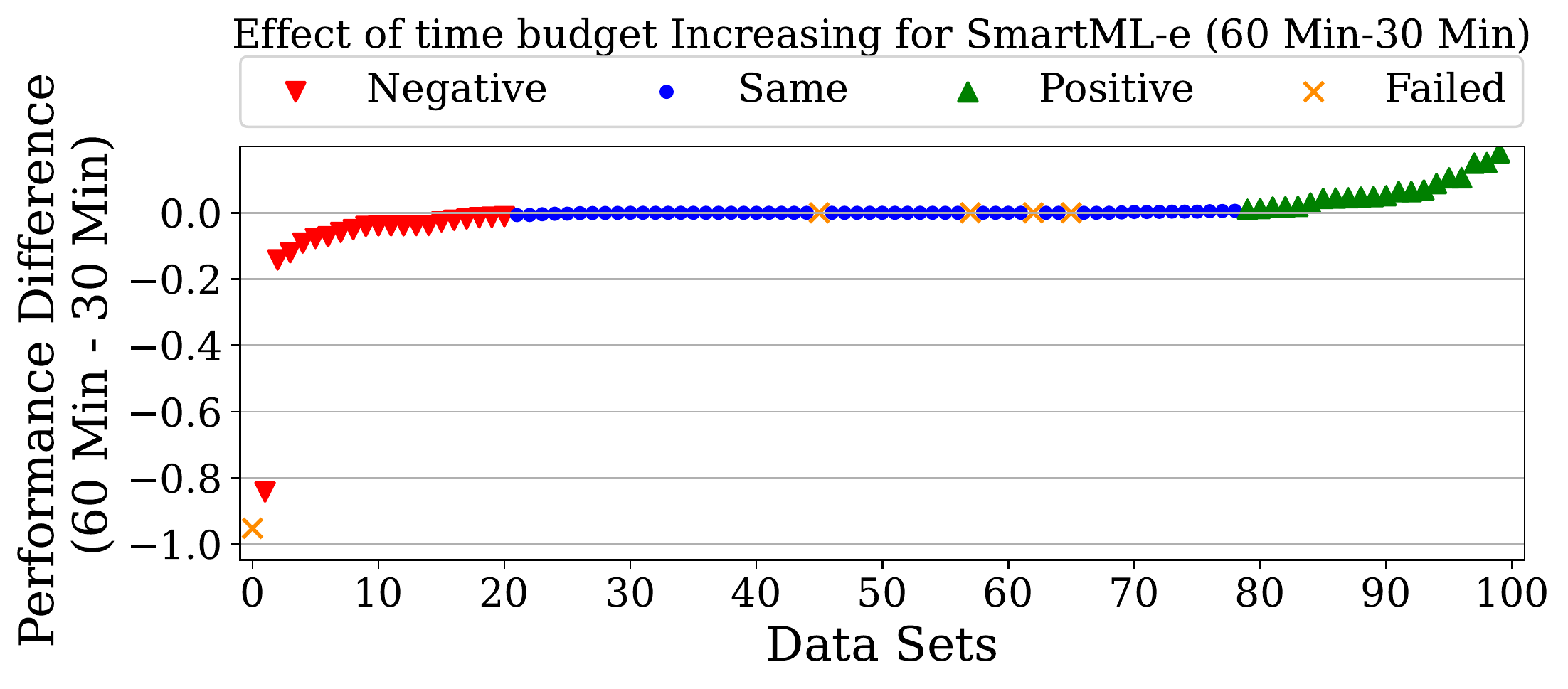}
}
\centering \subfigure[30-240 Min.] {
    \includegraphics[width=0.47\textwidth]{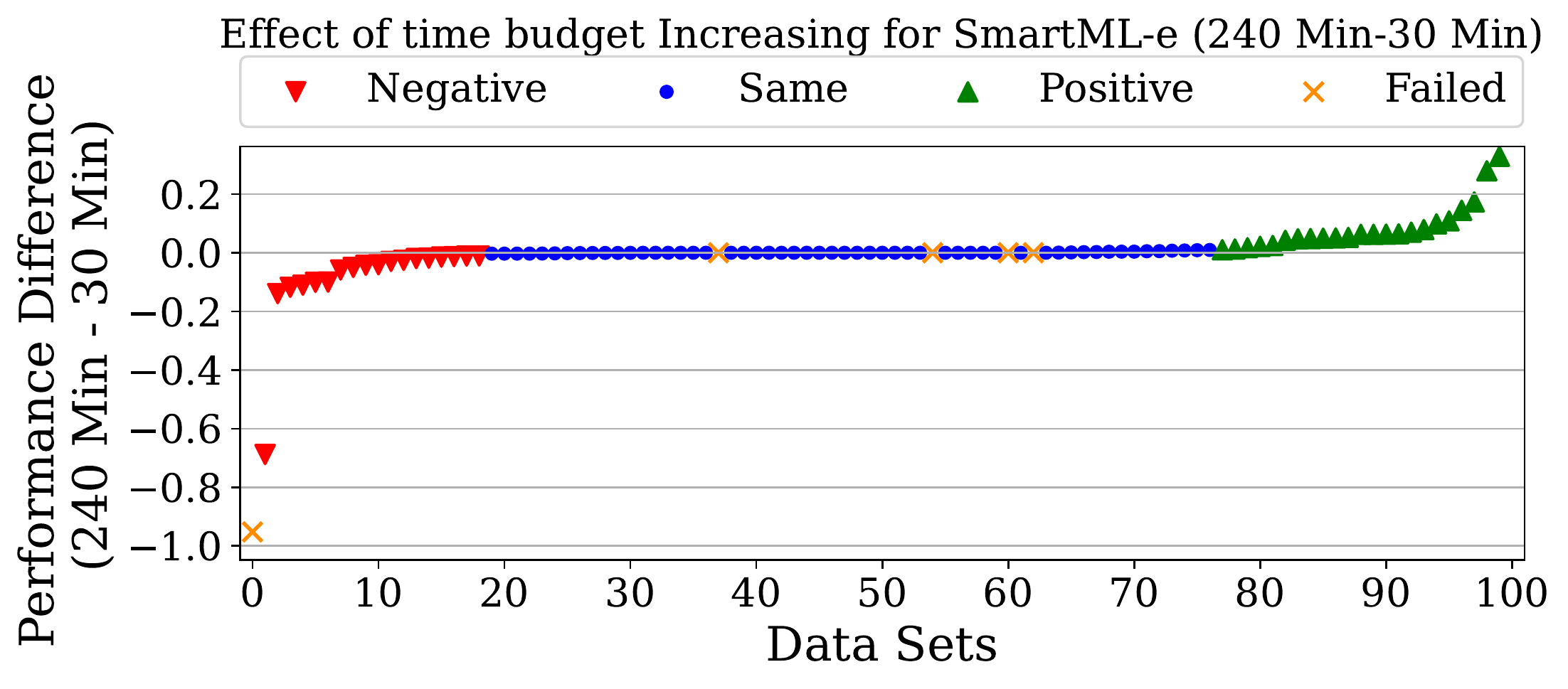}
}
\centering \subfigure[60-240 Min.] {
    \includegraphics[width=0.47\textwidth]{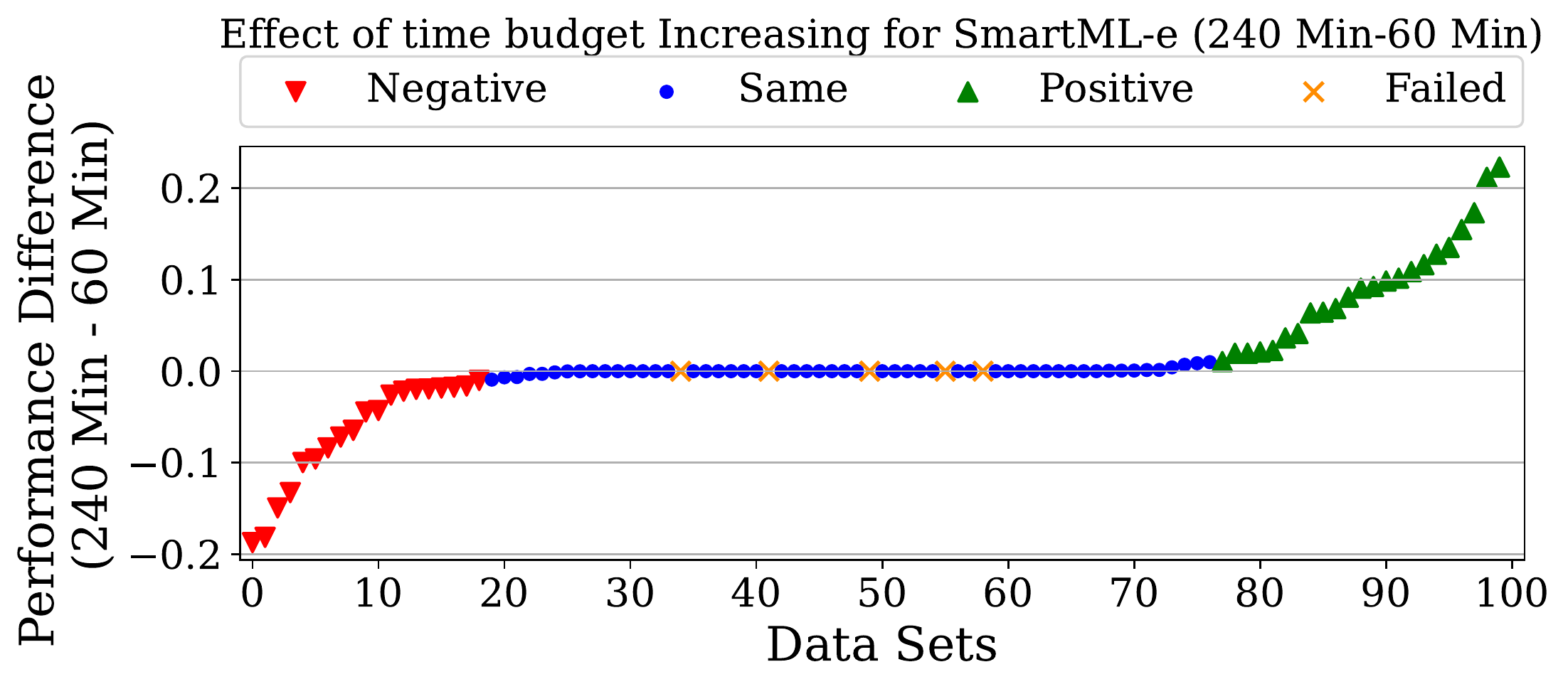}
}

\caption{The impact of increasing the time budget on \texttt{SmartML-e} performance from $x$ to $y$ minutes (x-y). Green markers represent better performance with $y$ time budget, blue markers means that the difference between $x$ and $y$ is $<1$. Red markers represent better performance on $x$ time budget.}
\label{FIG:TimeBudgetSmartML-e}
\end{figure*}

\begin{figure*}[!ht]
\centering \subfigure[10-30 Min.] {
    \includegraphics[width=0.47\textwidth]{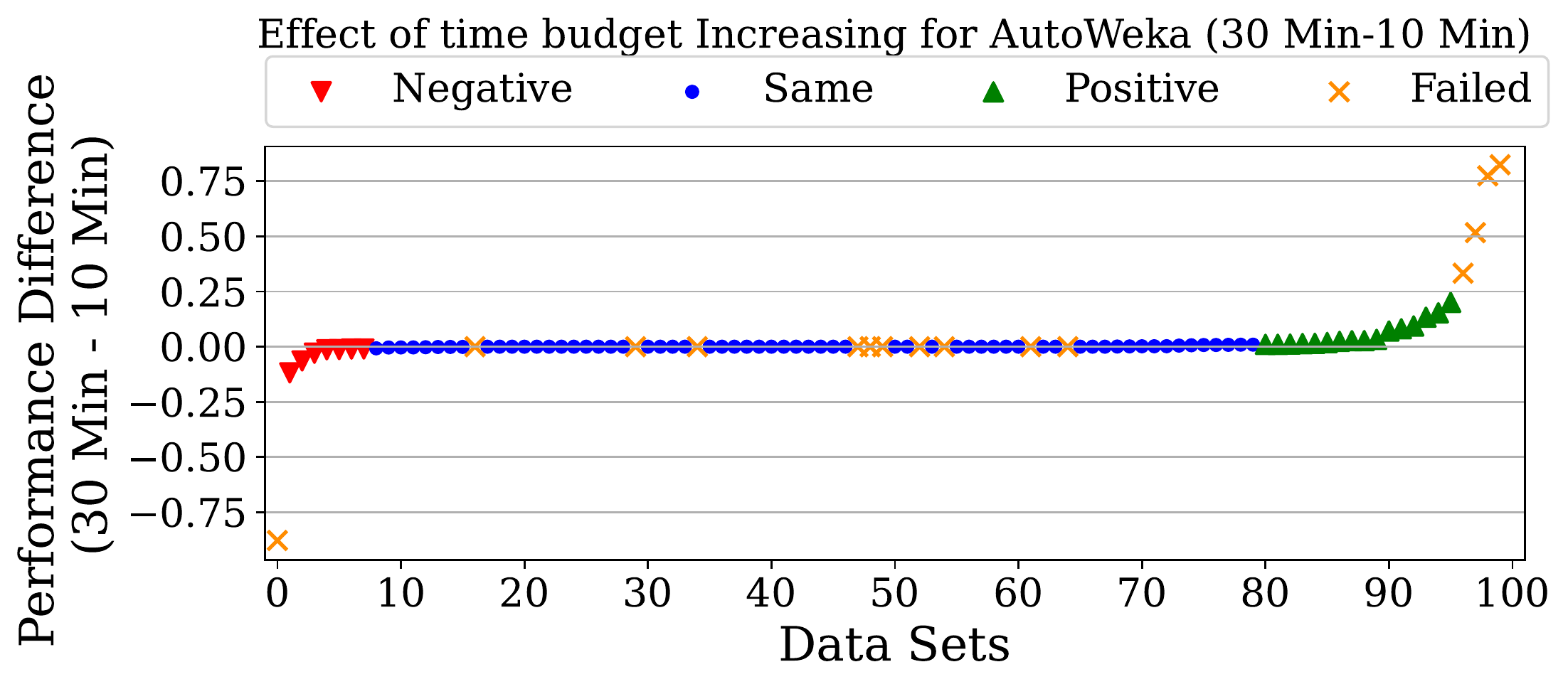}
}
\centering \subfigure[10-60 Min.] {
    \includegraphics[width=0.47\textwidth]{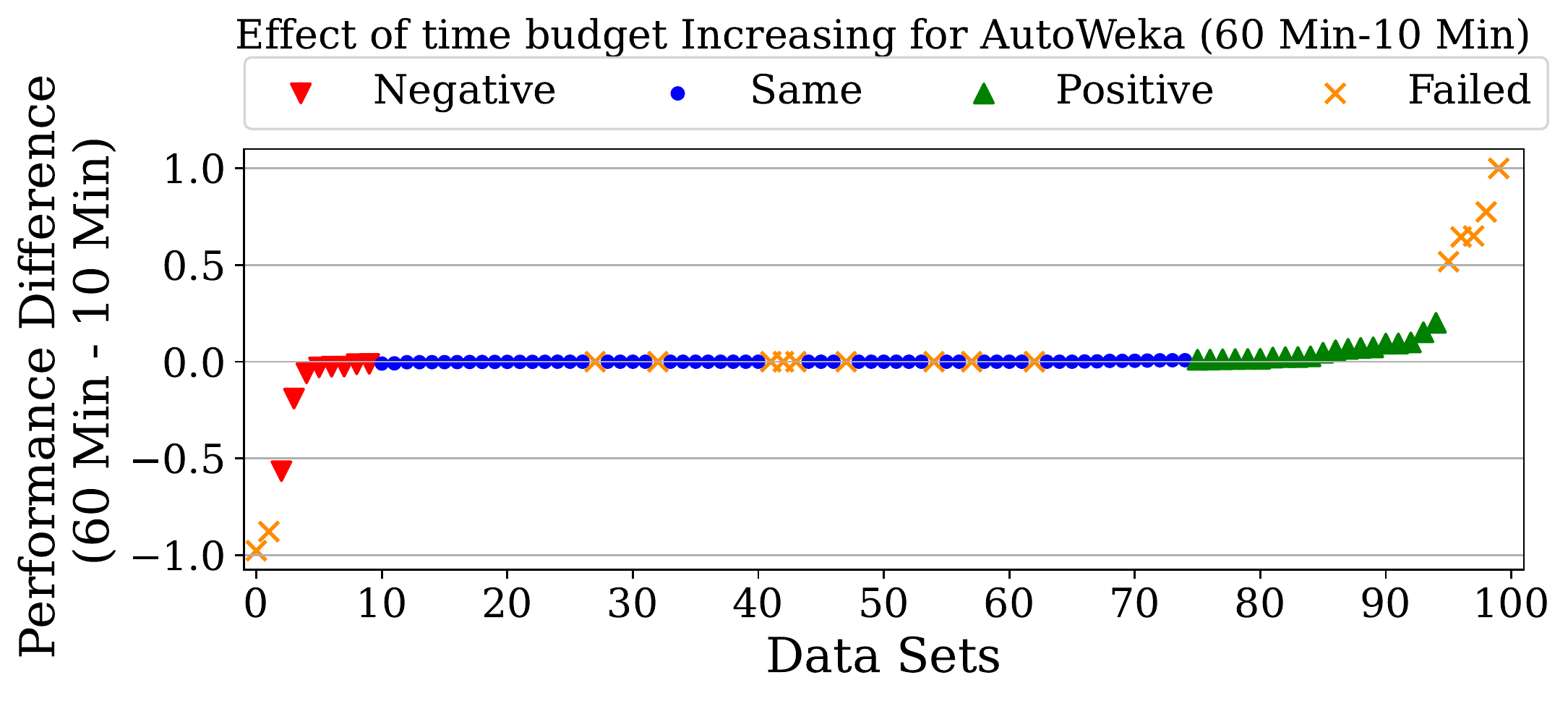}
}
\centering \subfigure[10-240 Min.] {
    \includegraphics[width=0.47\textwidth]{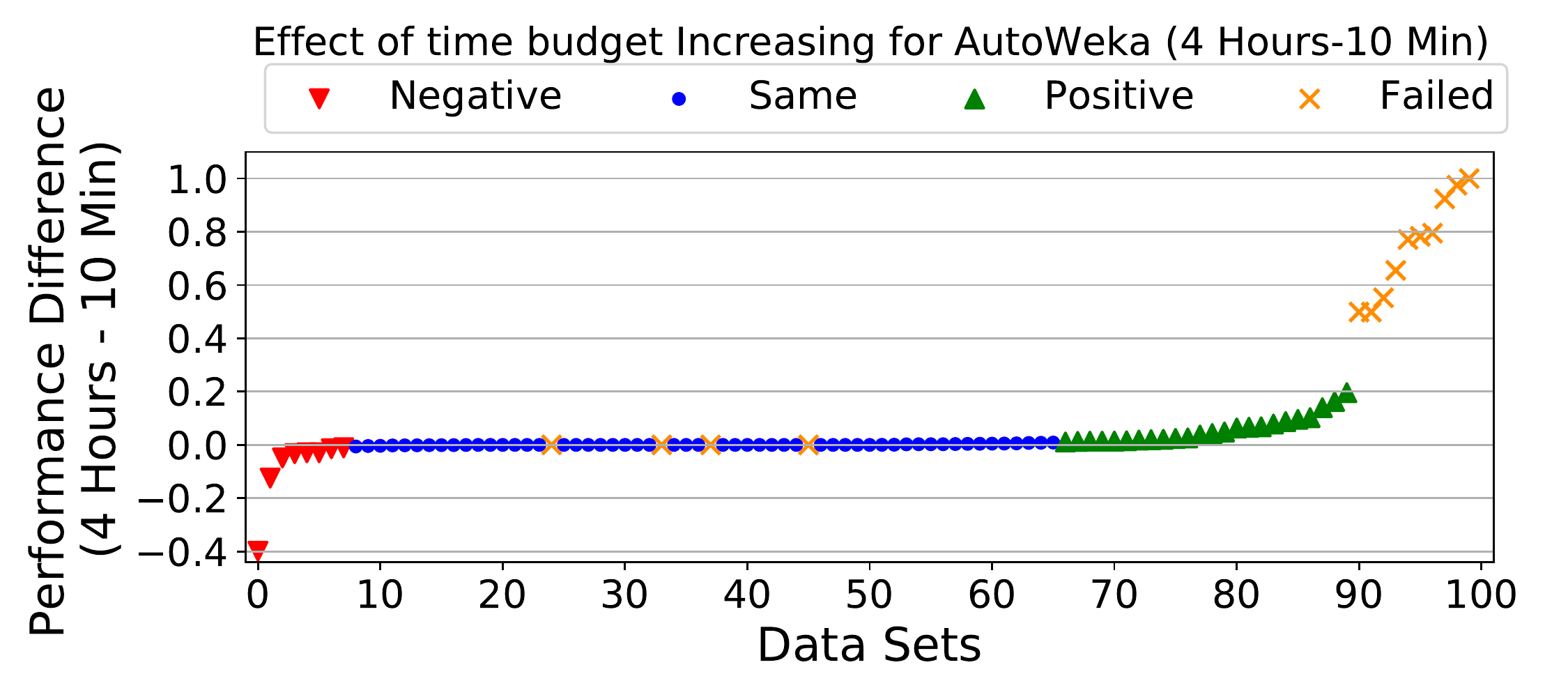}
}
\centering \subfigure[30-60 Min.] {
    \includegraphics[width=0.47\textwidth]{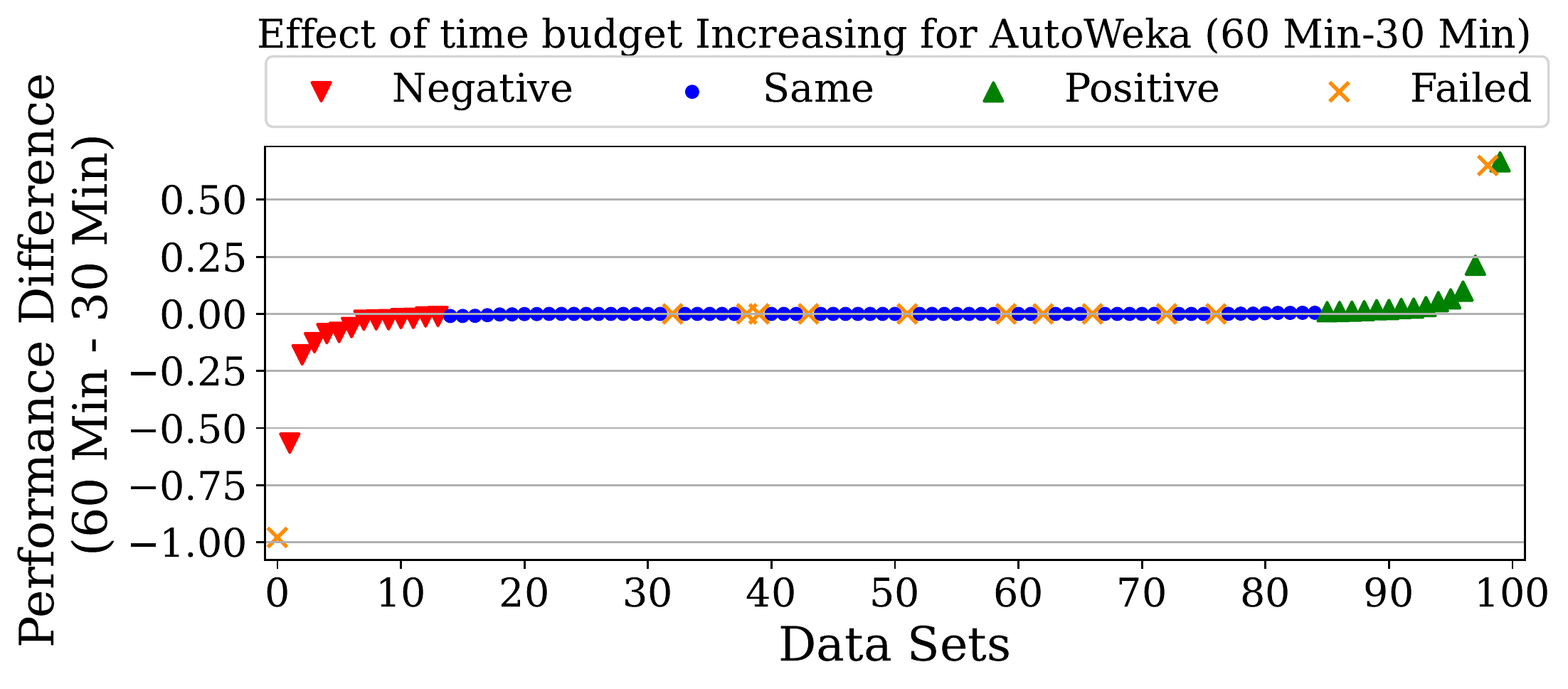}
}
\centering \subfigure[30-240 Min.] {
    \includegraphics[width=0.47\textwidth]{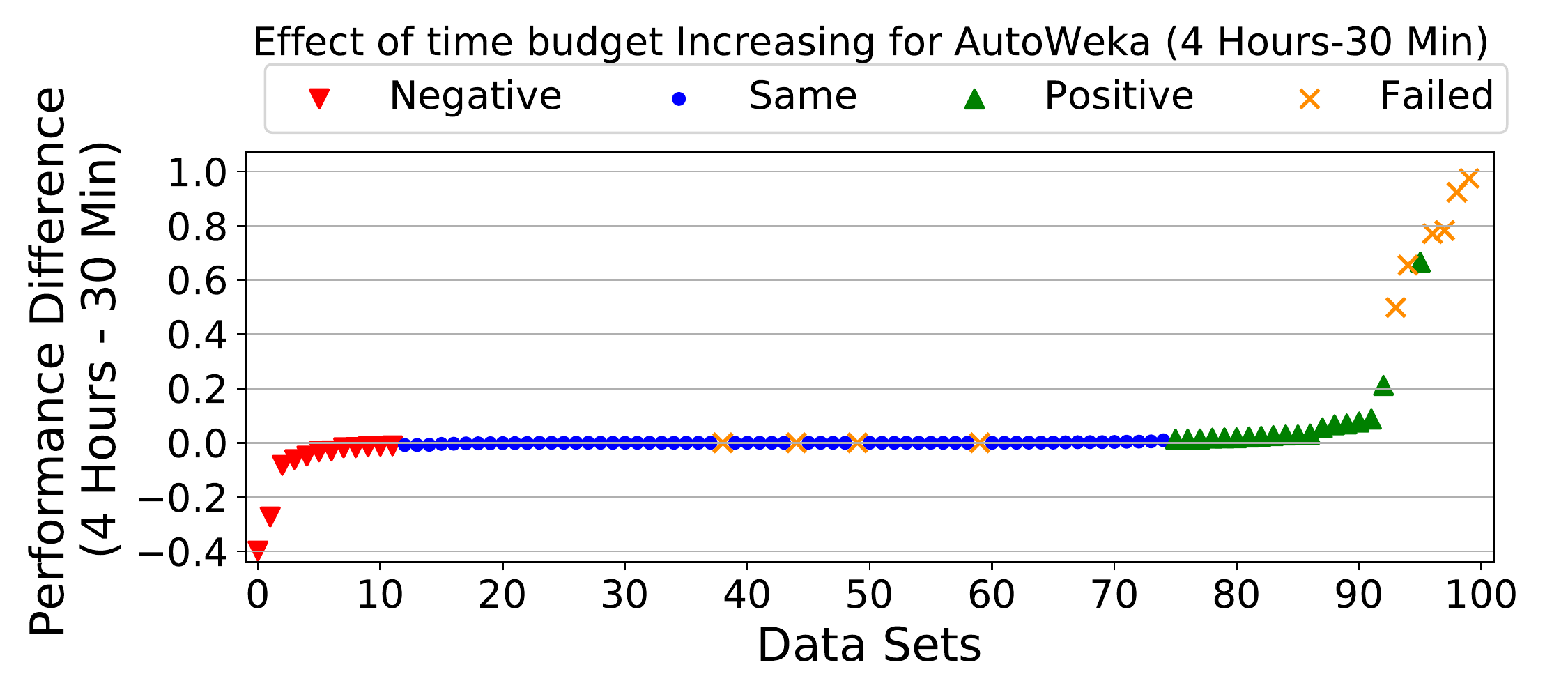}
}
\centering \subfigure[60-240 Min.] {
    \includegraphics[width=0.47\textwidth]{Figures/EffectoftimebudgetIncreasingforAutoWeka4Hours30Min.pdf}
}

\caption{The impact of increasing the time budget on \texttt{AutoWeka} performance from $x$ to $y$ minutes (x-y). Green markers represent better performance with $y$ time budget, blue markers means that the difference between $x$ and $y$ is $<1$. Red markers represent better performance on $x$ time budget.}
\label{FIG:TimeBudgetAutoWeka}
\end{figure*}

\begin{figure*}[!ht]
\centering \subfigure[10-30 Min.] {
    \includegraphics[width=0.47\textwidth]{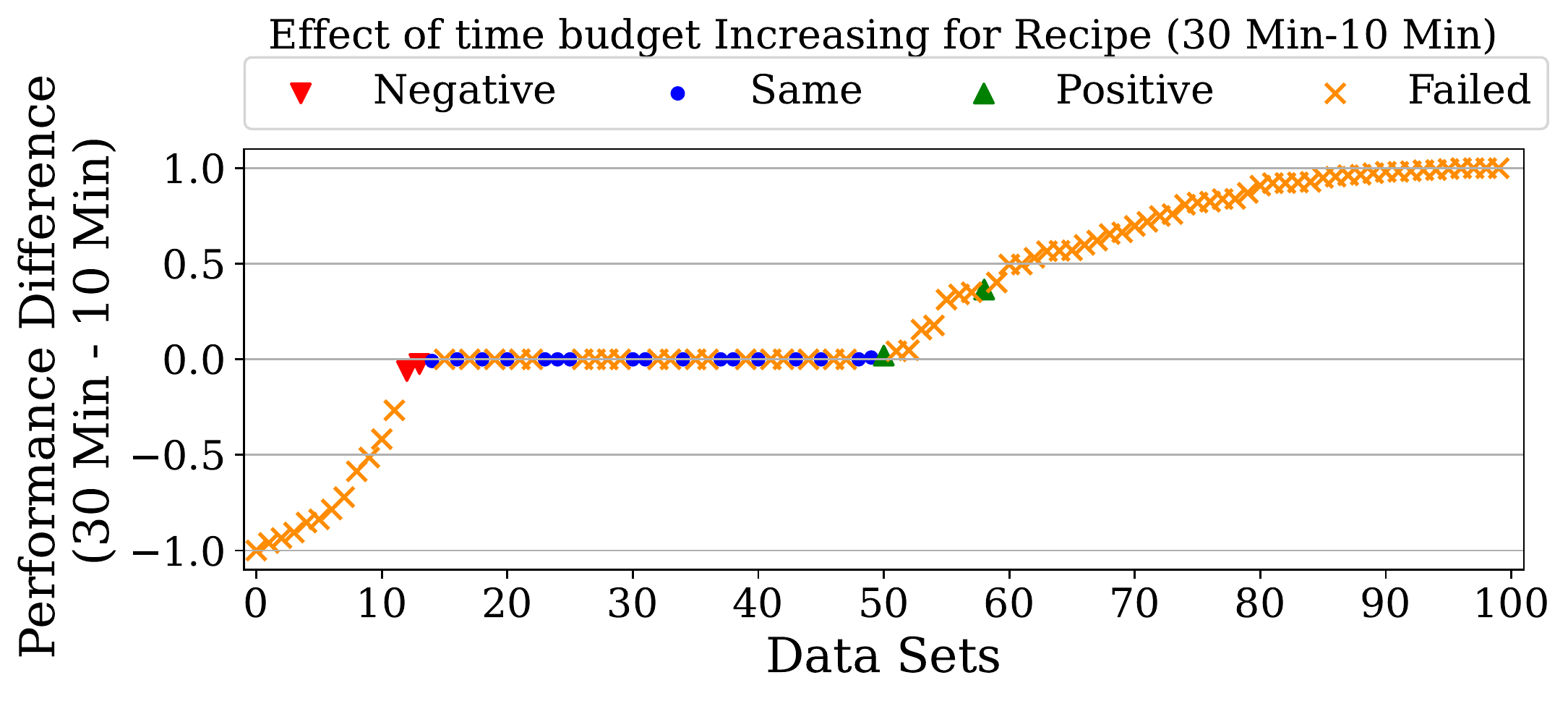}
}
\centering \subfigure[10-60 Min.] {
    \includegraphics[width=0.47\textwidth]{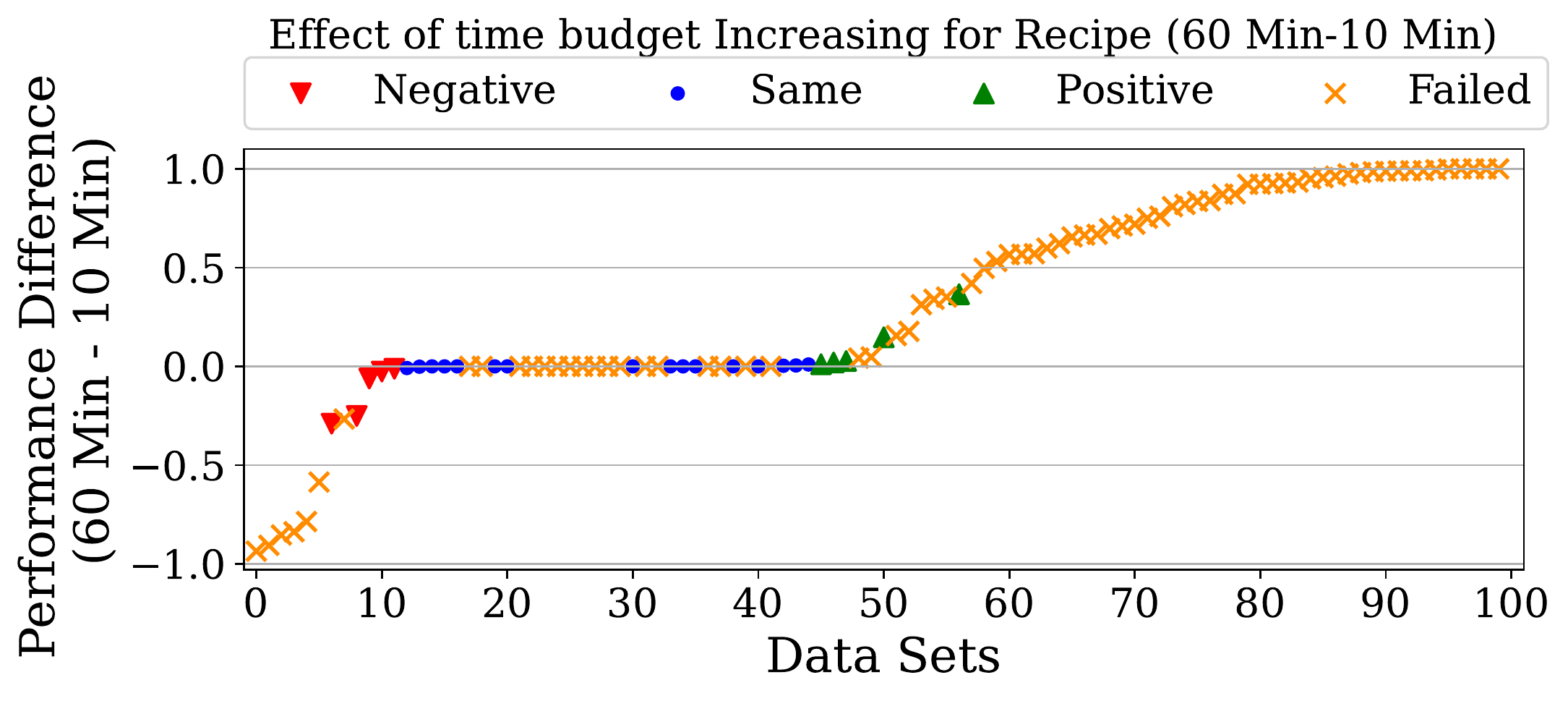}
}
\centering \subfigure[10-240 Min.] {
    \includegraphics[width=0.47\textwidth]{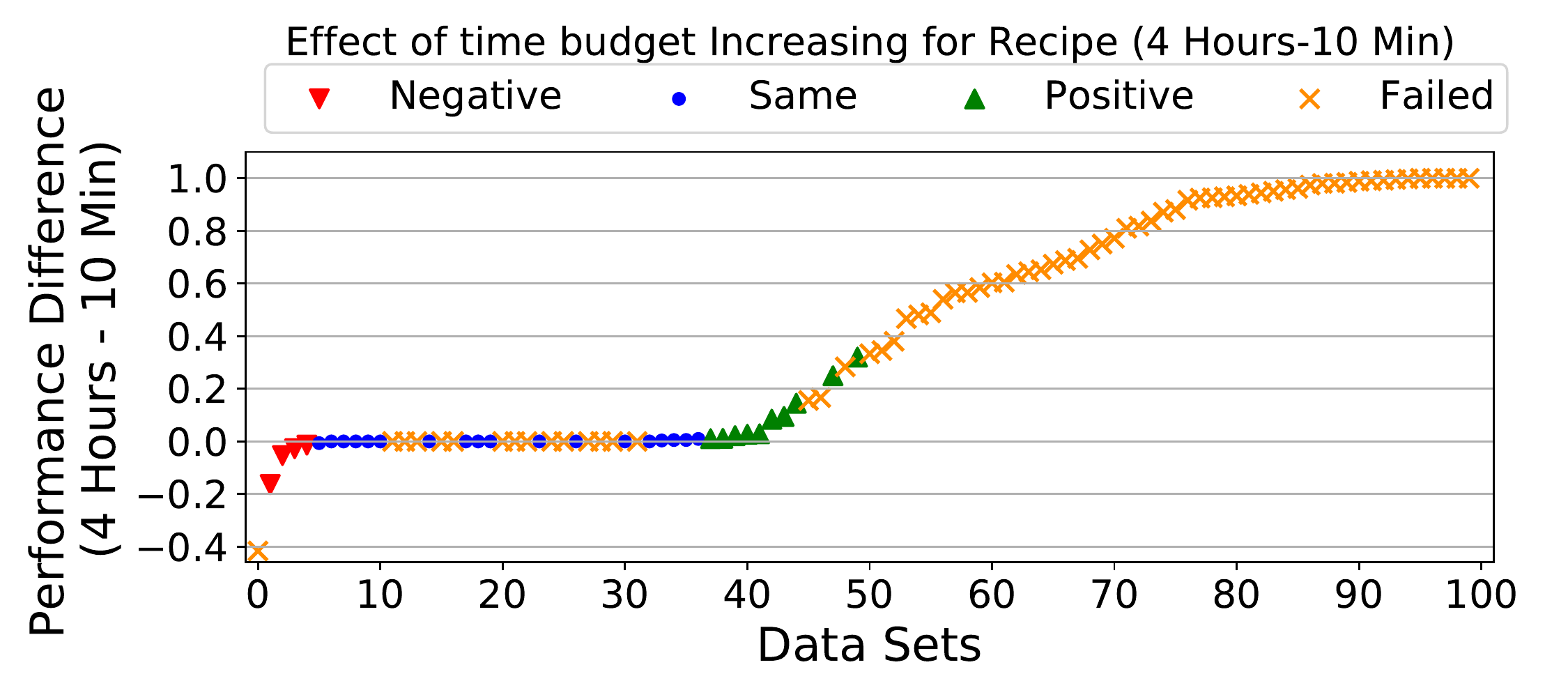}
}
\centering \subfigure[30-60 Min.] {
    \includegraphics[width=0.47\textwidth]{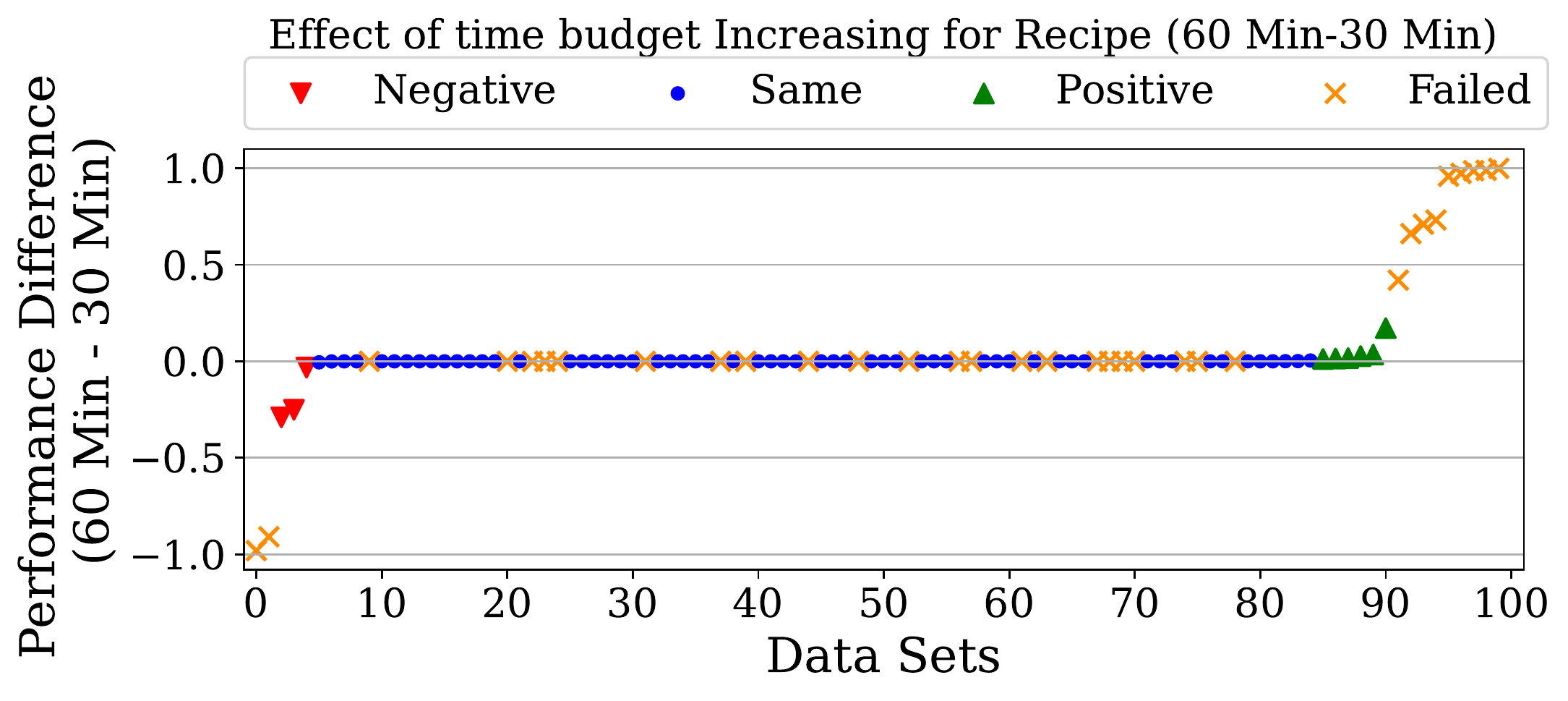}
}
\centering \subfigure[30-240 Min.] {
    \includegraphics[width=0.47\textwidth]{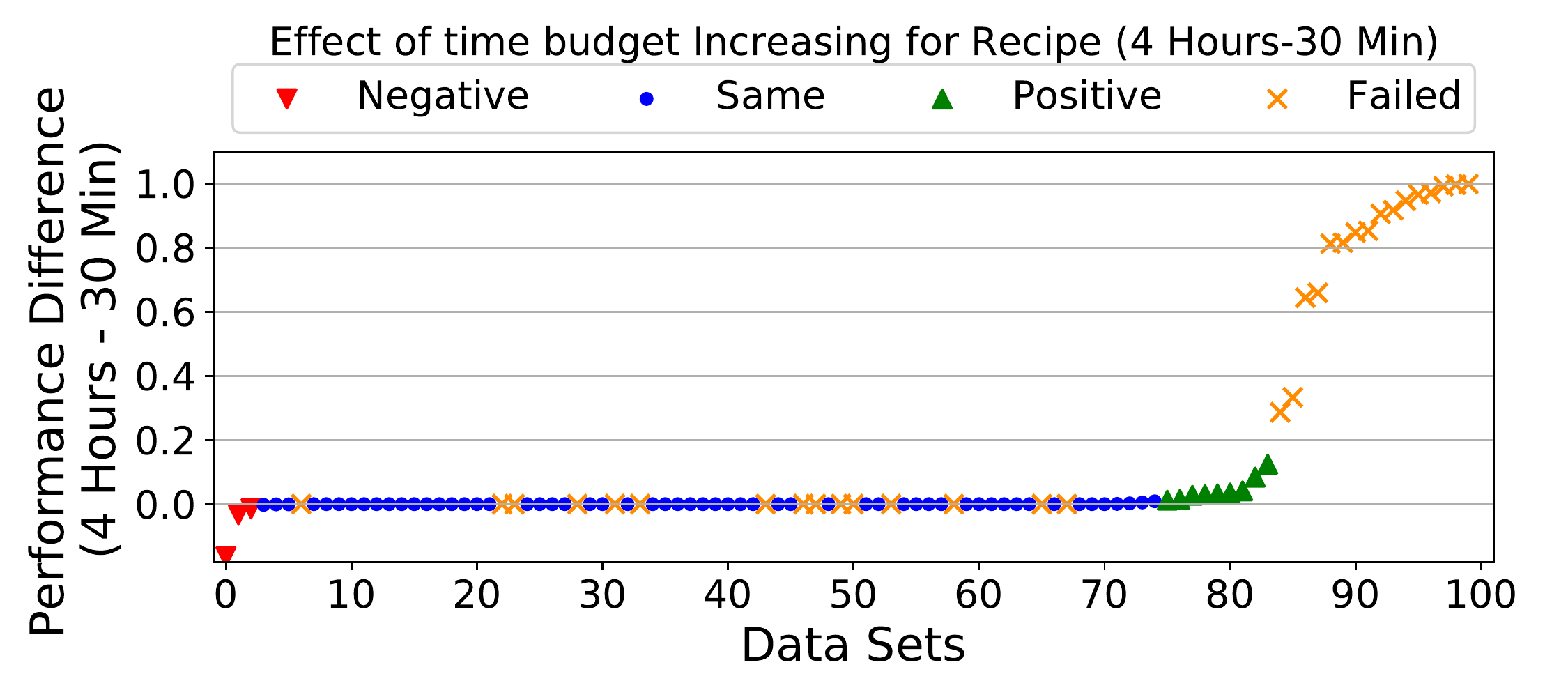}
}
\centering \subfigure[60-240 Min.] {
    \includegraphics[width=0.47\textwidth]{Figures/EffectoftimebudgetIncreasingforRecipe4Hours30Min.pdf}
}

\caption{The impact of increasing the time budget on \texttt{Recipe} performance from $x$ to $y$ minutes (x-y). Green markers represent better performance with $y$ time budget, blue markers means that the difference between $x$ and $y$ is $<1$. Red markers represent better performance on $x$ time budget.}
\label{FIG:TimeBudgetRecipe}
\end{figure*}

~\Cref{FIG:TimeBudgetSklearn-v,FIG:TimeBudgetSklearn-m,FIG:TimeBudgetSklearn-e,FIG:TimeBudgetSklearn,FIG:TimeBudgetTPOT,FIG:TimeBudgetATM,FIG:TimeBudgetSmartML,FIG:TimeBudgetSmartML-e,FIG:TimeBudgetAutoWeka,FIG:TimeBudgetRecipe} show the impact of increasing the time budget on the performance of the all the AutoML frameworks considered in this work.


\end{appendices}


\clearpage

\bibliography{sn-bibliography}


\end{document}